\documentclass[journal]{IEEEtran}
\ifCLASSINFOpdf
\else
\fi

\usepackage{lineno}
\usepackage{amsmath,amssymb,amsfonts}
\usepackage{amsthm}
\usepackage{enumitem}
\usepackage{algorithm}
\usepackage{algorithmic}
\usepackage{textcomp}
\usepackage{multirow}
\usepackage{multicol}
\usepackage{longtable}
\usepackage{lipsum}
\usepackage{float}
\usepackage{dblfloatfix}
\usepackage{mathtools}
\usepackage{xcolor}
\usepackage{array}
\usepackage{graphicx}
\usepackage{bbold}

\newtheorem{theorem}{Theorem}
\newtheorem{property}[theorem]{Property}

\ifCLASSOPTIONcompsoc
\usepackage[caption=false,font=normalsize,labelfon
t=sf,textfont=sf]{subfig}
\else
\usepackage[caption=false,font=footnotesize]{subfig}
\fi

\newcommand{\PreserveBackslash}[1]{\let\temp=\\#1\let\\=\temp}
\newcolumntype{C}[1]{>{\PreserveBackslash\centering}p{#1}}
\newcolumntype{R}[1]{>{\PreserveBackslash\raggedleft}p{#1}}
\newcolumntype{L}[1]{>{\PreserveBackslash\raggedright}p{#1}}

\usepackage{scalerel}
\usepackage{tikz}
\usetikzlibrary{svg.path}

\definecolor{orcidlogocol}{HTML}{A6CE39}
\tikzset{
	orcidlogo/.pic={
		\fill[orcidlogocol] svg{M256,128c0,70.7-57.3,128-128,128C57.3,256,0,198.7,0,128C0,57.3,57.3,0,128,0C198.7,0,256,57.3,256,128z};
		\fill[white] svg{M86.3,186.2H70.9V79.1h15.4v48.4V186.2z}
		svg{M108.9,79.1h41.6c39.6,0,57,28.3,57,53.6c0,27.5-21.5,53.6-56.8,53.6h-41.8V79.1z M124.3,172.4h24.5c34.9,0,42.9-26.5,42.9-39.7c0-21.5-13.7-39.7-43.7-39.7h-23.7V172.4z}
		svg{M88.7,56.8c0,5.5-4.5,10.1-10.1,10.1c-5.6,0-10.1-4.6-10.1-10.1c0-5.6,4.5-10.1,10.1-10.1C84.2,46.7,88.7,51.3,88.7,56.8z};
	}
}

\newcommand\orcidicon[1]{\href{https://orcid.org/#1}{\mbox{\scalerel*{
				\begin{tikzpicture}[yscale=-1,transform shape]
				\pic{orcidlogo};
				\end{tikzpicture}
			}{|}}}}
		
\usepackage[hidelinks]{hyperref}

\begin{document}
%
\title{An Online Learning Algorithm for a Neuro-Fuzzy Classifier with Mixed-Attribute Data}
%
%
%

\author{Thanh Tung Khuat$^{\textsuperscript{\orcidicon{0000-0002-6456-8530}}}$,~\IEEEmembership{Student Member,~IEEE},~and~Bogdan Gabrys$^{\textsuperscript{\orcidicon{0000-0002-0790-2846}}}$,~\IEEEmembership{Senior Member,~IEEE}

\thanks{T.T. Khuat (email: thanhtung.khuat@student.uts.edu.au) and B. Gabrys (email: Bogdan.Gabrys@uts.edu.au) are with Advanced Analytics Institute, Faculty of Engineering and Information Technology, University of Technology Sydney, Ultimo, NSW 2007, Australia.
}}

%
%

\markboth{}
{Khuat and Gabrys: An Online Learning Algorithm for a Neuro-Fuzzy Classifier with Mixed-Attribute Data}

%



\maketitle

\begin{abstract}
General fuzzy min-max neural network (GFMMNN) is one of the efficient neuro-fuzzy systems for data classification. However, one of the downsides of its original learning algorithms is the inability to handle and learn from the mixed-attribute data. While categorical features encoding methods can be used with the GFMMNN learning algorithms, they exhibit a lot of shortcomings. Other approaches proposed in the literature are not suitable for on-line learning as they require entire training data available in the learning phase. With the rapid change in the volume and velocity of streaming data in many application areas, it is increasingly required that the constructed models can learn and adapt to the continuous data changes in real-time without the need for their full retraining or access to the historical data. This paper proposes an extended online learning algorithm for the GFMMNN. The proposed method can handle the datasets with both continuous and categorical features. The extensive experiments confirmed superior and stable classification performance of the proposed approach in comparison to other relevant learning algorithms for the GFMM model.
\end{abstract}


%
\IEEEpeerreviewmaketitle

%
%
%
%

\section{Introduction}
\label{intro}
\IEEEPARstart{C}{lassical} batch learning algorithms usually require the complete availability of data at the training time. These algorithms do not constantly accommodate new information to the built models. Instead, we need to reconstruct the model from scratch when the underlying data changes. This operation is time-consuming, especially in the case of massive data, and the constructed models are more likely to be outdated in dynamically changing environments. Taking an advertising recommendation system as an example, this system constructs a customer preference model based on the tracking information about the shopping and browsing behaviors of the users. The buying activities and preferences are temporary and continuously changing. For example, the pandemic such as COVID-19 has dramatically changed the online shopping behaviors of customers where people tend to purchase things they have never bought before. Therefore, the learning models trained on consumer behavior data prior to the pandemic have been deteriorated or crashed. As a result, these models need to be retrained on new (normal?) behavior data. In this context, and many others characterised by streaming data in changing environments, it is desirable or even necessary to have online learning algorithms that can learn constantly new information without retraining from scratch. 

With the increase in the data volume and the rapid change of the environmental conditions nowadays, online learning algorithms are in high demand \cite{Lakshminarayanan2014, Lughofer2018}. These algorithms require smaller or no data storage as they only need one or few newest training samples at one time to rapidly update the constructed model. Hence, the online learning models are ideal candidates for the systems with frequently updating demands. General fuzzy min-max (GFMM) neural network \cite{Gabrys2000,Gabrys99} is such an incremental learning model, which can be effectively utilized for data classification problems. This type of learning model combines the artificial neural network with the fuzzy set theory to form a consolidated framework. The model creates new hyperboxes or adjusts the existing hyperboxes to cover new samples in its structure. Each hyperbox is defined by the minimum and maximum points in an n-dimensional space. The degree-of-fit of an input pattern to a hyperbox is identified by a membership function.

GFMM neural network \cite{Gabrys2000} is a significant enhancement of the FMNN \cite{Simpson1992}. Unlike the FMNN, the GFMM model can handle the uncertainty associated with the input data by accepting the input patterns not only as single points but also hyperboxes. In addition, it can handle both labeled and unlabeled data samples in a single model. The GFMMNN still maintains the online learning ability from the FMNN using a single-pass through training samples learning algorithms to expand or create new hyperboxes. To avoid the ambiguity in the classification phase, the original online learning algorithm proposed in \cite{Gabrys2000} does not allow the overlap between hyperboxes representing different classes. Therefore, after expanding a hyperbox to cover the input pattern, a hyperbox contraction procedure must be performed if there is an overlapping region between two hyperboxes belonging to different classes. However, the hyperbox contraction operation can lead to undesirable classification errors as shown in the original paper and subsequent publications \cite{Gabrys2000,Gabrys2002b, Gabrys04,Bargiela2004,Khuat2020iol}. As a result, in a recent study, we have proposed an improved online learning algorithm for the GFMM model (IOL-GFMM) \cite{Khuat2020iol} to overcome this limitation by not using the hyperbox contraction step during the learning process. This algorithm integrates the strong points of the batch learning algorithm proposed in \cite{Gabrys2002b} and the incremental learning ability of the original algorithm into a single algorithm.

However, both the original online learning algorithm \cite{Gabrys2000} and the IOL-GFMM algorithm \cite{Khuat2020iol} work well on the datasets with only numerical features. To perform classification for the datasets with mixed-type features, we would need to use the encoding methods to transform the categorical values into numerical values. As shown in a recent study \cite{Khuat2020enc}, each encoding method has its own drawbacks and except for the CatBoost \cite{Prokhorenkova18} and label encoding techniques, all of the remaining encoding approaches need to use the entire training set to encode the categorical features. Therefore, they are not appropriate for incremental learning algorithms, where the new values can appear during the operation time. In addition, according to the empirical results in \cite{Khuat2020enc}, the classification performance of the online learning algorithms using the CatBoost or label encoding method for the GFMM model is quite poor. It is because the label encoding method imposes an artificial distance metric for categorical groups, in which this distance is not correspondent to the correlation among original categorical values \cite{Brouwer02}. Not only this poses a serious problem but the CatBoost encoding method is sensitive to the order of training samples presentation and a shift in the encoded values between training and testing data as well as between training samples have been observed. For the same categorical value, its encoded value in the training data may be distinct from that in the testing data. Even in the training set, the same categorical value may be mapped into many different encoded values depending on the historical patterns prior to the current training pattern. Our proposed method in this paper avoids all of these issues by not using any encoding methods for discrete attributes in the first place. 

Many real-world datasets are in the form of mixed-type features. The mixed-attribute data contain both continuous and discrete (or categorical) features. Nowadays, the mixed-attribute data are more and more popular in a wide range of applications from the credit approval data to medical diagnostic data \cite{Huang19}. Hence, to apply the GFMMNN to such problems, we need to extend its current learning algorithms so that they can deal effectively with mixed-attribute data. Although there are a large number of improved algorithms of the FMNN model, only two existing studies have focused on expanding the learning algorithms for both categorical and numerical features as shown in a recent survey paper \cite{Khuat2019}. The first study was proposed in \cite{Castillo12} (denoted by Onln-GFMM-M1 in this paper) using the correlation between the occurrence frequency of categorical values and classes to determine the similarity degrees among categorical values for each categorical feature. After that, the authors proposed to extend the original online learning algorithm \cite{Gabrys2000} for mixed-attribute data. The second idea of expanding the original online learning algorithm of the FMNN model for both numerical and categorical features was introduced in \cite{Shinde16}, called Onln-GFMM-M2 in this paper. It uses the one-hot encoding method for the categorical features and logical operators such as AND and OR to operate on the categorical groups. However, the main weak point of both algorithms is the use of the entire training set to encode or compute the similarity degree between categorical values. If a new value occurs without being encountered during a training process before, these algorithms cannot handle such situation and produce a valid prediction. Different from these two approaches, this paper proposes a new incremental learning algorithm for both continuous and categorical features. The proposed method does not use any encoding methods for categorical values. Instead, it uses a union operator of a set to add new categorical value to the current set of values in each categorical feature of a hyperbox. The decision on expanding a selected hyperbox to accommodate a new input pattern is based on the change in the entropy for each categorical feature. We also modify the membership function to handle both categorical and numerical attributes. The membership degree for all categorical features is computed from the average probability of categorical values in the input sample with regard to all of the existing discrete values stored in discrete features of the hyperbox. In short, our main contribution in this paper can be summarized as follows:

\begin{itemize}
    \item We propose a novel online learning algorithm for the GFMMNN able to learn from mixed-attribute data. To the best of our knowledge, this is the first online learning algorithm for the family of fuzzy min-max neural networks which can handle both continuous and categorical features without using any encoding methods.
    \item We present and prove several properties of the proposed method with regard to the categorical/discrete attributes.
    \item We conduct extensive experiments to prove the effectiveness of the proposed method in comparison to other relevant methods.
    \item We assess the impact of hyper-parameters on the classification performance of the proposed method and propose a simple method for the parameter estimation.
\end{itemize}

The rest of this paper is structured as follows. Section \ref{preliminary} summarizes briefly the architecture of the GFMMNN and its improved online learning algorithm. Section \ref{proposed_method} is devoted to describing the proposed method and its properties. Experimental results and discussion are shown in Section \ref{experiment}. Section \ref{conclusion} concludes the key findings in this paper and informs potential research directions.

\section{Preliminaries} \label{preliminary}
\subsection{General fuzzy min-max neural network}
The GFMMNN \cite{Gabrys2000} are composed of three layers, i.e., input, hyperbox (hidden), and output layers. The input layer in the GFMM model can accept both real valued point and interval (hyperbox-typed) based input samples. If each input pattern has $n$ dimensions, there will be $2n$ nodes in the input layer, in which the first $n$ nodes are for the lower bounds and the remaining $n$ nodes represent the upper bounds. The hidden layer contains hyperboxes dynamically generated in the learning process. The connection weights between the lower bound nodes and a hyperbox $B_i$ form a vector $V_i$ storing the minimum coordinates for that hyperbox. Similarly, the connection weights from the upper bound input nodes to a hyperbox $B_i$ are represented by a vector $W_i$ containing the maximum coordinates of that hyperbox. The values of matrices \textbf{V} and \textbf{W} for all hyperboxes are tuned during the learning process. Each hyperbox $B_i$ in the hidden layer is fully connected to all output nodes. The connection weights between the hyperbox layer and output layer are kept in a matrix \textbf{U} and each of its element $u_{ij}$ is computed as follows:
\begin{equation}
u_{ij} = \begin{cases}
1, \mbox{if } class(B_i) = c_j \\
0, \mbox{otherwise}
\end{cases}
\end{equation}
where $c_j$ is the $j$-$th$ class node in the output layer.

Each hyperbox $B_i = [V_i, W_i]$, where $V_i = [v_{i1},\ldots,v_{in}]$ and $W_i = [w_{i1},\ldots,w_{in}]$ are the minimum and maximum points respectively, is associated with a membership function $b_i(X, B_i)$. This membership function is used to calculate the degree-of-fit for each input pattern $X = [X^l, X^u]$ to the hyperbox $B_i$, where $X^l = [x_1^l,\ldots,x_n^l]$ and $X^u = [x_1^u,\ldots,x_n^u]$ are the lower and upper bounds of an input pattern suitably normalised within an $n$-dimensional unit hypercube $[0, 1]^n$. The membership function is given as follows:
\begin{equation}
\label{eq_mem}
\begin{split}
b_i(X, B_i) = \min \limits_{j = 1}^{n} (\min(&[1 - f(x_{j}^u - w_{ij}, \gamma_j)], \\
&[1 - f(v_{ij} - x_{j}^l, \gamma_j)]))
\end{split}
\end{equation}
where $ f(\xi, \gamma) $ is a ramp function defined in \eqref{eq_ramp}:
\begin{equation}
\label{eq_ramp}
f(\xi, \gamma) = \begin{cases}
1, & \mbox{if } \xi \cdot \gamma > 1 \\
\xi \cdot \gamma, & \mbox{if } 0 \leq \xi \cdot \gamma \leq 1 \\
0, & \mbox{if } \xi \cdot \gamma < 0
\end{cases}
\end{equation}
with $ \gamma = (\gamma_1, \gamma_2,..., \gamma_n) $ a sensitivity parameter controlling the decreasing speed of the membership degree, and $ 0 \leq b_i \leq 1 $. If $b_i(X, B_i) = 1$, then $X$ is fully contained in the core of the hyperbox $B_i$.

\subsection{An improved online learning algorithm}\label{iol_gfmm}

The learning process in GFMM consists of creating and adjusting hyperbox fuzzy sets on the basis of the presented input patterns. There have been a number of fundamental GFMM learning algorithms proposed in the literature which fall into one of two key categories: (i) incremental/on-line learning algorithms where the hyperboxes are adjusted, if needed, after every presentation of a single pattern \cite{Gabrys2000,Khuat2020iol} and (ii) batch learning algorithms which assume the full training data is available from the beginning of the training process for the training algorithm to use \cite{Gabrys2002b,Gabrys2002a,Gabrys04}. The performance of the original online algorithm \cite{Gabrys2000} is sensitive to the order of training data presentation and the maximum hyperbox size hyper-parameter setting. When inappropriate maximum size of hyperbox is selected and combined with the existing hyperbox contraction process, it can lead to undesired classification errors as analysed and illustrated in \cite{Khuat2020iol}. Therefore, in a recent study, we proposed an improved version of the original online learning algorithm, in which, similarly to the agglomerative algorithms in \cite{Gabrys2002b}, the contraction process is not used during the learning process. The algorithm contains only two main steps, i.e., creation or expansion of hyperboxes and overlap test. In the original online learning algorithm, if a selected hyperbox candidate fulfills the expansion condition related to the maximum hyperbox size, it will be expanded. Then, if the overlap between the newly expanded hyperbox and the existing hyperboxes belonging to other classes occurs, the relevant hyperboxes are contracted. In contrast, in the IOL-GFMM algorithm, if the undesired overlap would happen after the expansion, the selected candidate will not be expanded.

For a training sample $X = [X^l, X^u, c_X]$, the algorithm first filters all existing hyperboxes with the same class as $c_X$. After that, the membership values between $X$ and these hyperboxes are computed and sorted in descending order. Hyperbox candidates will be then checked with regard to meeting the expansion conditions beginning from the hyperbox with the highest membership degree. If the maximum membership value is one, i.e. $X$ is contained in the hyperbox, the learning algorithm continues with the next training sample. Otherwise, the expansion condition checking process only terminates when there is a hyperbox candidate which can be expanded to cover $X$ or no further hyperbox candidates exist. If none of the existing candidates can be expanded, a new hyperbox is generated with the same coordinates as $X$. The first expansion condition is the maximum hyperbox size. For the hyperbox candidate $B_i$, first, the maximum hyperbox size condition given in \eqref{exp_num_cond} is checked:
\begin{equation} \label{exp_num_cond}
    \max(w_{ij}, x^u_j) - \min(v_{ij}, x^l_j) \leq \theta, \; \forall{j \in [1, n]}
\end{equation}
where $n$ is the number of features. If this condition is met, the hyperbox $B_i$ is temporarily expanded to new size as follows:
\begin{equation} \label{expand_num}
    \begin{split}
        w_{ij}^{new} &= \max(w_{ij}^{old}, x^u_j)\\
        v_{ij}^{new} &= \min(v_{ij}^{old}, x^l_j), \; \forall{j \in [1, n]}
    \end{split}
\end{equation}
Then, the newly expanded $B_i$ will be tested for undesired overlaps with all of the hyperboxes representing the other classes (i.e. different from the class associated with $B_i$). There are four overlap test cases shown in \cite{Gabrys2000}. If there is no overlapping area occurring, the new size of $B_i$ is kept. Otherwise, $B_i$ is reverted to the coordinates before expanding and the next hyperbox candidate is considered.

For an unseen pattern, its predicted class is the class of the hyperbox representing the highest membership value for that input pattern among all existing hyperboxes in the model. In the case when many hyperboxes representing $ K $ different classes have the same maximum membership degree ($ b_{win}$), an additional criterion is used to find the appropriate class for $X$. The final class of $X$ is the class $c_k$ with the highest score of $\mathcal{P}(c_k|X) $ given by:
\begin{equation}
    \label{probcard}
    \mathcal{P}(c_k|X) = \cfrac{\sum_{j \in \mathcal{I}_{win}^k} n_j \cdot b_j}{\sum_{i \in \mathcal{I}_{win}} n_i \cdot b_i }
\end{equation}
where $k \in [1, K]$ and $ \mathcal{I}_{win} = \{ i, \mbox{if } b_i = b_{win} \}$ comprises the indexes of all hyperboxes with the maximum membership value of $b_{win}$, $ I_{win}^k = \{ j, \mbox{if } class(B_j) = c_k \mbox{ and } b_j = b_{win} \}$ is a subset of $ I_{win} $ containing indexes of the $k$-$th$ class, and $n_i$ is the number of training samples covered by hyperbox $ B_i$. We would like to refer the interested readers to references \cite{Khuat2020iol} and \cite{Khuat2020acc} for the detailed algorithm as well as its time complexity.

\section{Proposed Method} \label{proposed_method}
\subsection{Formal Description}
Let $\mathcal{T}_N = \{(X_i^l, X_i^u, X_i^d, c_i)\}^{N}_{i=1}$ be $N$ training patterns, where $c_i$ is the class of the $i$-$th$ pattern, $X_i^l = (x_{i1}^l, \ldots, x_{in}^l)$ and $X_i^u = (x_{i1}^u, \ldots, x_{in}^u)$ are $n$ continuous attributes (determined in a unit hyper-cube $[0, 1]^n$) of lower bound $X_i^l$ and upper bound $X_i^u$ for the $i$-$th$ training sample, $X_i^d = (x_{i1}^d, \ldots, x_{ir}^d)$ represent $r$ discrete attributes for the $i$-$th$ training sample, $x_{ij}^d$ is a categorical value of the $j$-$th$ categorical feature ($A_j^d$) at the $i$-$th$ training sample, $x_{ij}^d \in \mbox{DOM}(A_j^d) = \{a_{1j}, a_{2j}, \ldots, a_{n_j j}\}$, where $\mbox{DOM}(A_j^d)$ is a domain of discrete values for the categorical attribute $A_j^d$ and $n_j$ is the number of symbolic values of $A_j^d$. This paper proposes an online learning algorithm to train an efficient GFMM classifier from $\mathcal{T}_N$.
\begin{figure}[!ht]
    \centering
    \includegraphics[width=0.32\textwidth]{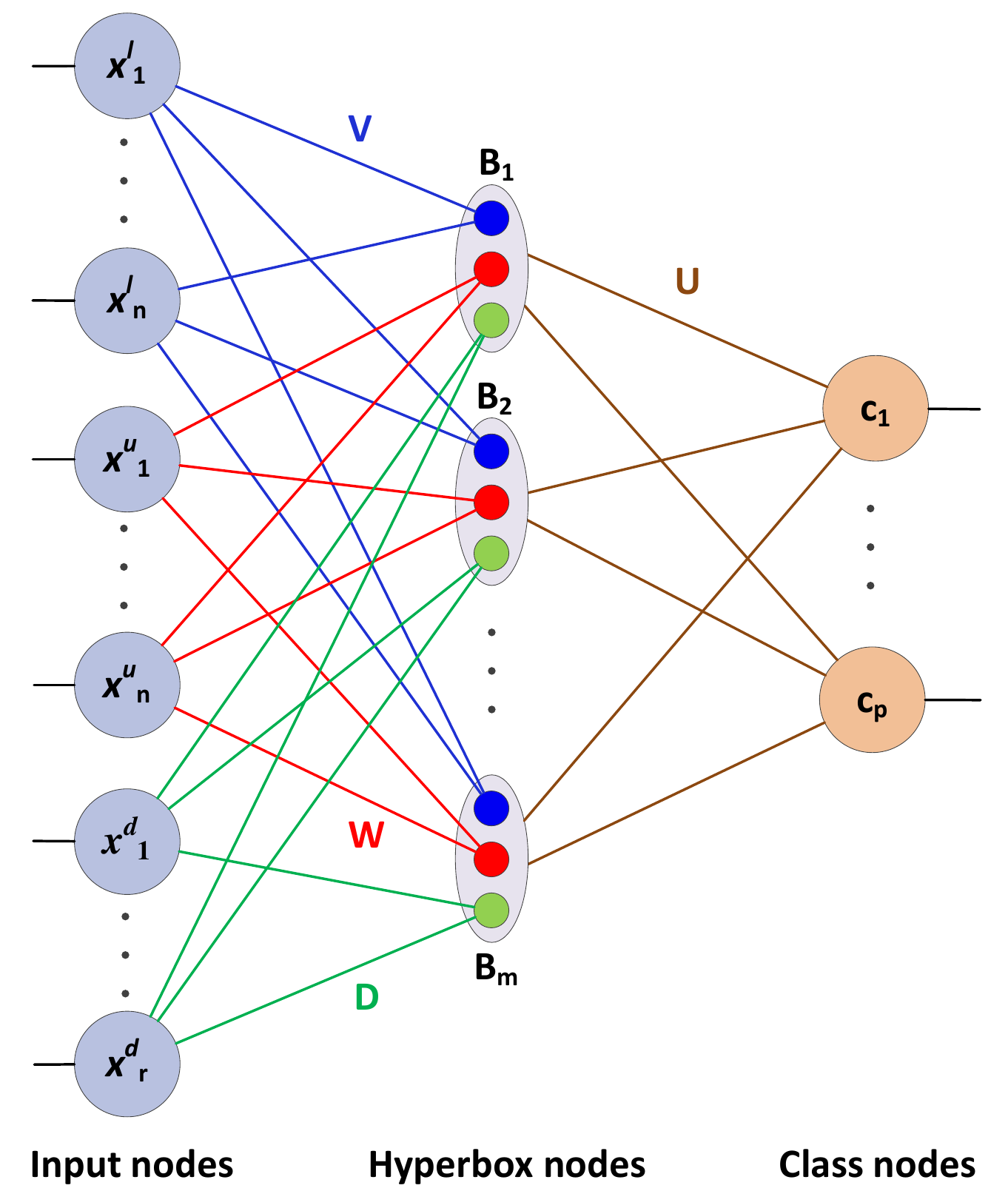}
    \caption{The extended architecture of GFMMNN for mixed-attribute data}
    \label{Fig_1}
\end{figure}
\subsubsection{Architecture of GFMMNN for Mixed-Attribute Data} \hfill

First of all, we need to expand the architecture of the GFMMNN for mixed-attribute data. Instead of using $2n$ input nodes as in the GFMM model for continuous data, we will need $2n + r$ nodes for the input layer. The first $2n$ nodes are lower bound and upper bound nodes for $n$ numerical features, respectively. The last $r$ remaining nodes correspond to $r$ categorical features in each input pattern. These $r$ input nodes are connected to $m$ hyperboxes by connection weights stored in a matrix \textbf{D}. New architecture of the GFMMNN is shown in Fig. \ref{Fig_1}. Beside the minimum points $V_i$ and the maximum points $W_i$, each hypebox $B_i$ in the hidden layer also contains a vector $D_i$ storing $r$ discrete-valued sets. Each element $d_{ij} \in D_i \; (1 \leq j \leq r)$ is a set of symbolic values with their cardinalities for the $j$-$th$ categorical dimension of the hyperbox $B_i$. For example, $d_{i1} = \{apple:5, orange:1\}$ means that the first categorical feature of the hyperbox $B_i$ contains 5 values of \textit{apple} and 1 value of \textit{orange}. The values of vectors $V_i$, $W_i$, and $D_i$ for each hyperbox $B_i$ are generated and adjusted during the learning process. The membership function of each hyperbox $B_i$ with regard to each input pattern with mixed-attribute $X = (X^l, X^u, X^d, c_X)$ is modified as follows:
\begin{equation}
\label{eq_new_mem}
\begin{split}
b_i(X, &B_i) = \alpha \cdot \min \limits_{j = 1}^{n} (\min([1 - f(x_{j}^u - w_{ij}, \gamma_j)], \\
&[1 - f(v_{ij} - x_{j}^l, \gamma_j)])) + \cfrac{1 - \alpha}{r} \cdot \sum\limits_{j = 1}^{r} \mathsf{P}_j(x_j^d \in d_{ij})
\end{split}
\end{equation}
where $\alpha \:(\alpha \in [0, 1])$ is a trade-off factor regulating the contribution level of numerical features part and categorical features part to the membership score, and $\mathsf{P}_j(x_j^d \in d_{ij})$ is a probability of encountering a symbolic value $x_j^d$ in the\break$j$-$th$ categorical attribute of the hyperbox $B_i$. This probability is formally defined as follows:
\begin{equation}
    \label{prob_cat}
    \mathsf{P}_j(x_j^d \in d_{ij}) = \cfrac{|\{a \in d_{ij} | a = x_j^d\}|}{|d_{ij}|}
\end{equation}
where $|\cdot|$ is the cardinality of a set. For the above example, we obtain $\mathsf{P}_j(\{orange\} \in d_{i1}) = 1/6$, $\mathsf{P}_j(\{apple\} \in d_{i1}) = 5/6$, and $\mathsf{P}_j(\{banana\} \in d_{i1}) = 0$. Unlike the numerical part in the membership function, we use an average operation for the categorical part to reduce sensitivity to the membership value. If we also use the $\min$ operator for the categorical part, the membership value for the categorical features will get the value of zero when there is only one discrete feature getting a new symbolic value.

\subsubsection{Extended Improved Online Learning Algorithm for Mixed-Attribute Data Classification} \hfill

To create new hyperboxes or adjust existing hyperboxes towards learning mixed-attribute training samples in the GFMM model, we need to expand the current improved learning algorithm presented in subsection \ref{iol_gfmm}, denoted by EIOL-GFMM in this paper. The proposed modifications include the expansion condition for categorical features, the way of accommodating a categorical value into the hyperbox, and the overlap test for categorical features.

For each training sample, $X = [X^l, X^u, X^d, c_X] \in \mathcal{T}_N$, the algorithm first filters all of the existing hyperboxes representing the same class as $c_X$. Then, the membership values of $X$ in these selected hyperboxes are calculated and sorted in descending order. After that, in turn, we select expandable hyperbox candidates starting from the hyperbox with the highest membership degree if the highest membership score is smaller than one. Assuming that $B_i$ is the currently considered hyperbox, the numerical features of $B_i$ are checked for the maximum hyperbox size condition as shown in \eqref{exp_num_cond}. If the expansion condition for continuous features is satisfied, the algorithm continues to check the constraint for discrete features.

Entropy-based measures can be used to assess the heterogeneity of data in clusters, and they are appropriate for clustering of categorical data due to the lack of explicit distance measures between discrete values \cite{Li04}. We propose to use the change in the entropy value of categorical values contained in the hyperbox to decide whether the current hyperbox can be expanded to accommodate the categorical values of a new training sample. Given a categorical attribute $j$, let $H_j(B_i)$ be the current entropy of hyperbox $B_i$ for the $j$-$th$ categorical feature, computed from the probability of all current categorical values stored in the $j$-$th$ attribute as follows:
\begin{equation} \label{entropy}
    H_j(B_i) = - \sum\limits_{l = 1}^{N_j} \mathsf{P}_j(a_l \in d_{ij}) \cdot \log_2 \mathsf{P}_j(a_l \in d_{ij})
\end{equation}
where $N_j$ is the number of different categorical values ($a_l$) in the $j$-$th$ attribute, and $\mathsf{P}_j$ is defined in \eqref{prob_cat}. It is clear that if we add a new sample to the hyperbox for which most of the sample's categorical values existed in the categorical attributes of the hyperbox, the change in the entropy of that hyperbox is small. In contrast, if we add a sample into the hyperbox for which most of the sample's categorical values are new symbolic values to the set of existing discrete attributes of the hyperbox, the homogeneity of this hyperbox is significantly changed, and so the entropy will increase. As a result, we can use the change in the entropy of the hypebox as an expansion condition for categorical attributes. This entropy changing value is defined in \eqref{entropy_change} for each discrete feature $j$ of each hyperbox candidate $B_i$.
\begin{equation}
    \label{entropy_change}
    Z_j = H_j(B_i \cup \{X\}) - \cfrac{n_i}{n_i + 1} H_j(B_i)
\end{equation}
where $H_j(B_i \cup \{X\})$ is the entropy of the hyperbox $B_i$ on the $j$-$th$ attribute after covering the input pattern $X$, computed using \eqref{entropy}, $n_i$ is the number of samples contained in the hyperbox $B_i$ ($n_i$ is also equal to the summation of cardinalities of categorical values on the dimension $j$).

Based on $Z_j$, we have two ways to construct the expansion condition for categorical attributes:
\begin{itemize}
    \item The first method is similar to the expansion condition for continuous features. The extended algorithm using this way is denoted by EIOL-GFMM-v1 in this paper. We require the change in the entropy for every categorical attribute smaller than a maximum entropy changing threshold $\delta$ for all categorical dimensions:
    \begin{equation}
        Z_j \leq \delta
    \end{equation}
    \item The second approach to build the expansion condition for categorical features uses a weaker condition compared to the first way. The proposed online learning algorithm adopting this condition is called EIOL-GFMM-v2 in this paper. This method requires the average change in the entropy of all of the $r$ categorical attributes smaller than a maximum average entropy changing threshold $\delta$:
    \begin{equation}
        \sum\limits_{j=1}^r Z_j \leq \delta
    \end{equation}
\end{itemize}

If both conditions for categorical and numerical features are met for the hyperbox $B_i$, it will be temporarily expanded to new coordinates. The expansion of numerical features is performed using \eqref{expand_num}. Each categorical feature $d_{ij}$ of $B_i$ is expanded as follows:
\begin{equation}
    d_{ij}^{new} = \begin{cases}
    d_{ij}^{old} \cup \{x_{j}^d: 1\}, &\mbox{if } \nexists a_j \in d_{ij}^{old}: a_j = x_j^d \\
    d_{ij}^{old}.\textbf{update}(a_j), &\mbox{if } \exists a_j \in d_{ij}^{old}: a_j = x_j^d
    \end{cases}
\end{equation}
where $\textbf{update}(a_j)$ is a function to increase the number of elements of the categorical value $a_j$ by 1. After that, an overlap checking procedure is performed for the newly expanded $B_i$ to examine whether $B_i$ overlaps with any hyperboxes belonging to other classes. In the improved online learning algorithm for numerical features, only four overlap test cases are used as in the original online learning algorithm. However, these four cases are not sufficient to identify all potential overlap cases between two hyperboxes. Therefore, in this extended version, we will deploy a similarity measure between two hyperboxes based on their smallest gap introduced in \cite{Gabrys2002b} to check the overlap for numerical features between $B_i$ and other hyperboxes $B_k$ representing different classes. This similarity measure $s_{ik}$ is defined in \eqref{sim_dist}.
\begin{equation}
    \label{sim_dist}
        s_{ik} = \min\limits_{j=1}^n [\min(1 - f(v_{kj} - w_{ij}, 1), 1 - f(v_{ij} - w_{kj}, 1))]
\end{equation}
where $n$ is the number of continuous features, $f$ is a ramp function given in \eqref{eq_ramp}. If $B_i$ and $B_k$ overlap with each other, $s_{ik} = 1$; otherwise, $s_{ik} < 1$. If $B_i$ does not overlap with any $B_k$ representing other classes on the numerical features, we do not need to check the overlap conditions for their discrete features. Otherwise, we have to verify the overlap for the discrete features between $B_i$ and hyperboxes $B_k$ overlapping with $B_i$ on the continuous features. Let $\Omega_{ij}$ and $\Omega_{kj}$ be the set of categorical values on the $j$-$th$ discrete attribute of two hyperboxes $B_i$ and $B_k$, respectively. $B_i$ overlaps with $B_k$ on the $j$-$th$ categorical feature if and only if:
\begin{equation} \label{overlap_cat}
    \begin{split}
        &\Omega_j = \Omega_{ij} \cap \Omega_{kj} \ne \varnothing \mbox{ and}\\
        & \exists a_j \in \Omega_j: \mathsf{P}_j(a_j \in d_{ij}) = \mathsf{P}_j(a_j \in d_{kj})
    \end{split}
\end{equation}
where $\mathsf{P}_j$ is defined in \eqref{prob_cat}. $B_i$ overlaps with $B_k$ on discrete attributes if the equation \eqref{overlap_cat} is true for all of the $r$ categorical features of these two hyperboxes.

If the hyperbox candidate $B_i$ does not overlap with any hyperboxes $B_k$ representing other classes on either categorical or continuous features, the new coordinates of $B_i$ remain unchanged and the algorithm continues with the next training sample. Otherwise, the coordinates of $B_i$ are reverted to the previous values and the hyperbox candidate with the next highest membership value is selected as an expandable hyperbox candidate, and the above steps are re-iterated.

If none of the hyperbox candidates can be extended to accommodate the input pattern $X$, a new hyperbox $B_i$ is generated as follows. For each numerical feature $j$, we set $v_{ij} = x_j^l, w_{ij} = x_j^u, \forall{j \in [1, n]}$, and for each categorical feature $j$, we assign $d_{ij} = \{x_j^d: 1\}, \forall{j \in [1, r]}$.

The classification phase of the EIOL-GFMM algorithm remains unchanged as in the original IOL-GFMM algorithm. We can see that the way of working of EIOL-GFMM algorithm itself can explain the reason leading to the classification results based on the selection of the hyperbox with the maximum membership degree.

\subsection{Properties of the Change in Entropy of Categorical Features when Accommodating New Training Samples} \label{Properties}
This section presents several interesting properties related to the change of the entropy on each categorical attribute of a hyperbox $B_i$ when accommodating a new training sample $X$.
\begin{property}\label{prop_1}
When covering an input pattern, the change of the entropy on each discrete attribute $j$ of $B_i$ obtains its maximum value if and only if that attribute $j$ includes a new categorical value which does not exist in the list of its current categorical values. Formally,
\begin{equation}
    d_{ij}^{new} = d_{ij}^{old} \cup \{x_j^d: 1\} \Rightarrow Z_j \mapsto \max
\end{equation}
\begin{proof}
See Section I in the supplemental material.
\end{proof}
\end{property}
\begin{property}\label{prop_2}
The upper bound of the change in the entropy for every categorical dimension $j$ of $B_i$ depends on the current number of samples included in $B_i$. That is:
\begin{equation}
    Z_j \leq \log_2(n_i + 1) - \cfrac{n_i}{n_i + 1} \log_2 n_i
\end{equation}
\begin{proof}
See Section II in the supplemental material.
\end{proof}
\end{property}
\begin{property}\label{prop3}
The change of the entropy for each categorical dimension $j$ always falls in the range of \textnormal{[0, 1]}: $$0 \leq Z_j \leq 1; \quad \forall{j \in [1, r]}$$
\begin{proof}
See Section III in the supplemental material.
\end{proof}
\end{property}
Property \ref{prop3} also confirms that $0 \leq \delta \leq 1$.
\begin{property}\label{prop4}
When the number of samples contained in $B_i$ approaches infinity, the change of the entropy for every categorical dimension will be limited at 0. Formally,
\begin{equation}
    \lim \limits_{n_i \rightarrow +\infty} Z_j = 0, \; \forall{j \in [1, r]}
\end{equation}
\begin{proof}
See Section IV in the supplemental material.
\end{proof}
\end{property}
Property \ref{prop4} indicates that when the number of samples included in each hyperbox $B_i$ increases, the expansion condition for categorical attributes of this hyperbox becomes easier to be satisfied.

\section{Experimental Results}\label{experiment}
The main purposes of the experiments in this section are to
\begin{itemize}
    \item Analyze the critical roles of parameters $\alpha$ and $\delta$ on classification accuracy for the proposed method
    \item Compare the performance of the proposed method to relevant approaches of GFMMNN for mixed-attribute data using fixed settings and tuning parameters
    \item Assess several different methods to estimate the values of $\alpha$ if we have sufficient samples at the training time.
\end{itemize}

These experiments were conducted on 14 datasets taken from the UCI machine learning repository \footnote{https://archive.ics.uci.edu/ml/datasets.php}. These datasets were used in \cite{Khuat2020enc} to evaluate different methods to handle mixed-attribute data using the GFMMNN. We used the same datasets to compare our proposed approach to the results presented in that research. The details of these datasets can also be found in subsection V.A in the supplemental document. All of the experimental datasets in this paper are class-imbalanced, so we will use the class balanced accuracy (CBA) metric to assess the performance of classification algorithms. The superior facets of the CBA in comparison to other metrics for the class-imbalanced datasets were shown in \cite{Khuat2020enc} and \cite{Mosley13}.

\subsection{Analyzing the Sensitivity of Parameters}
There are three important parameters affecting the classification performance of the proposed method, i.e., the maximum hyperbox size for continuous attributes ($\theta$), the trade-off factor ($\alpha$) regarding the contribution levels of continuous and discrete attributes to the membership function, and the maximum entropy changing threshold ($\delta$) for discrete attributes. The role of $\theta$ was analyzed in a recent study \cite{Khuat2020}, in which smaller values of $\theta$ usually result in better performance than the use of larger values of $\theta$ does. However, the smaller values of $\theta$ are, the more complex the final model is (i.e. the larger number of generated hyperboxes). In this section, we only study the influence of two new parameters introduced in our proposed method, i.e., $\delta$ and $\alpha$.

\subsubsection{Parameter $\alpha$} \label{eva_alpha} \hfill

\begin{figure}[!ht]
    \centering
    \includegraphics[width=0.26\textwidth]{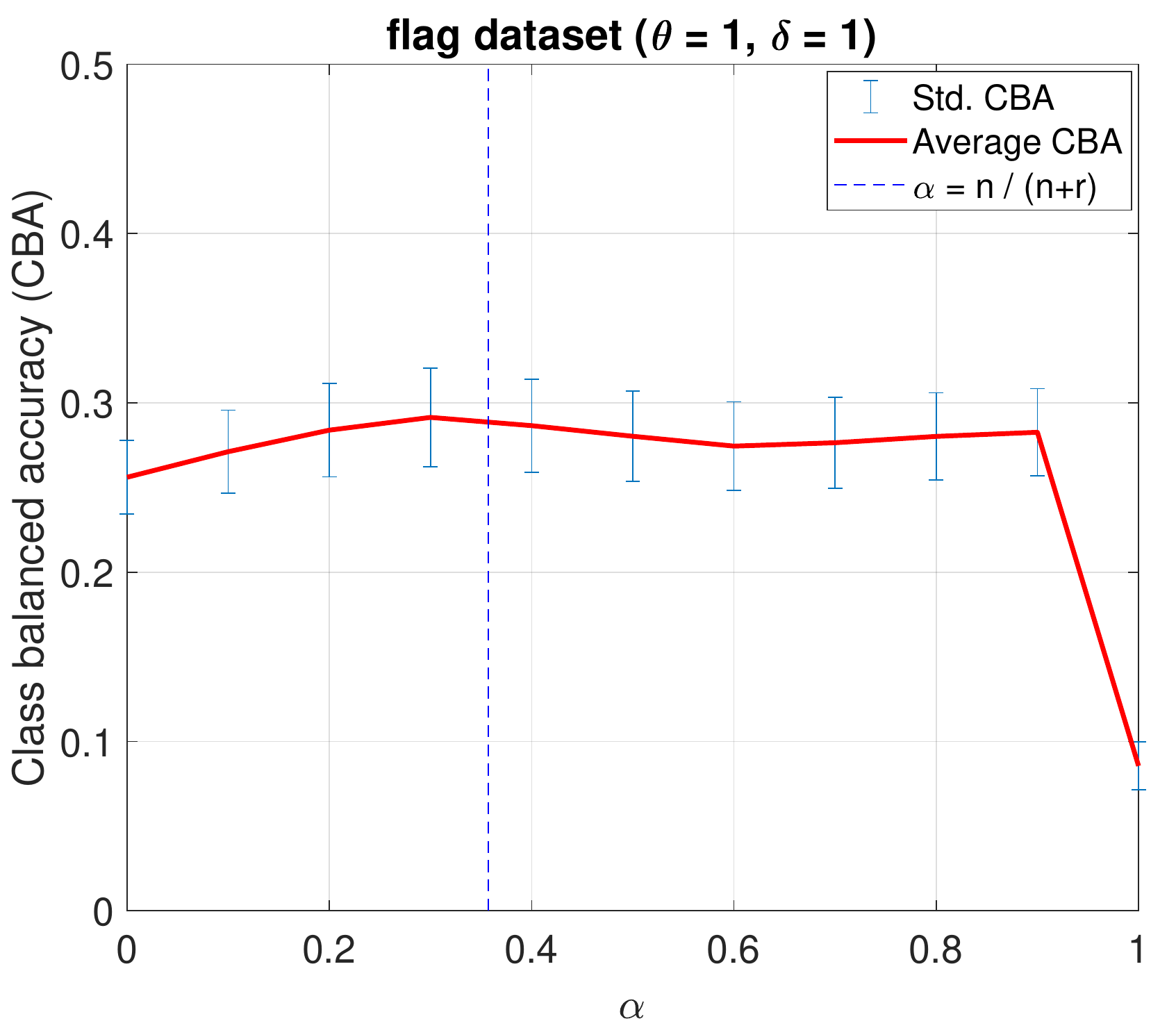}
    \caption{The change in the class balanced accuracy according to the different values of $\alpha$ for the \textit{flag} dataset ($\theta = 1, \delta = 1$).}
    \label{fig_2}
\end{figure}

\begin{figure}[!ht]
\centering
\begin{subfloat}[EIOL-GFMM-v1]{
			\includegraphics[width=0.23\textwidth]{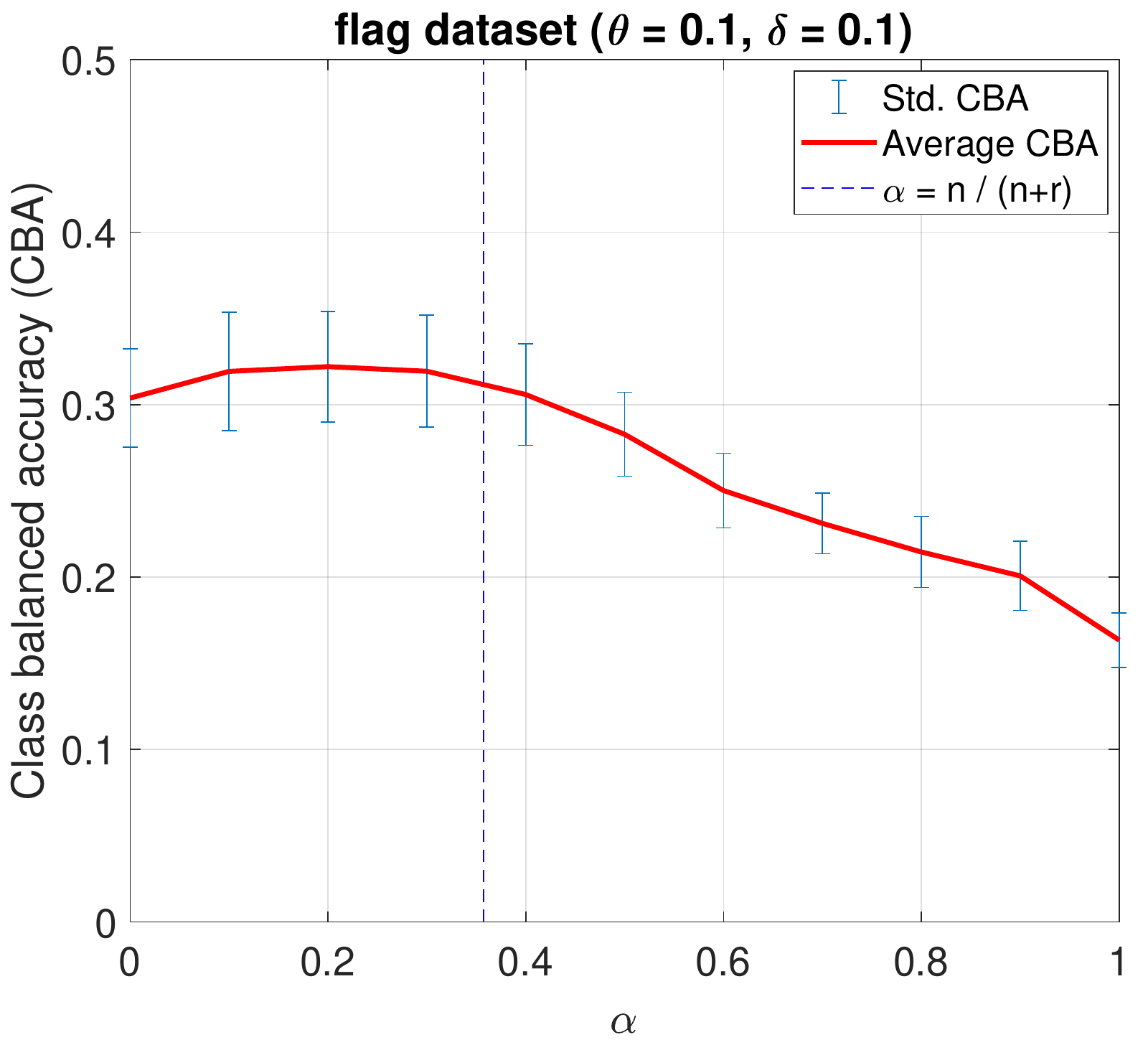}}
	\end{subfloat}
	\begin{subfloat}[EIOL-GFMM-v2]{
			\includegraphics[width=0.23\textwidth]{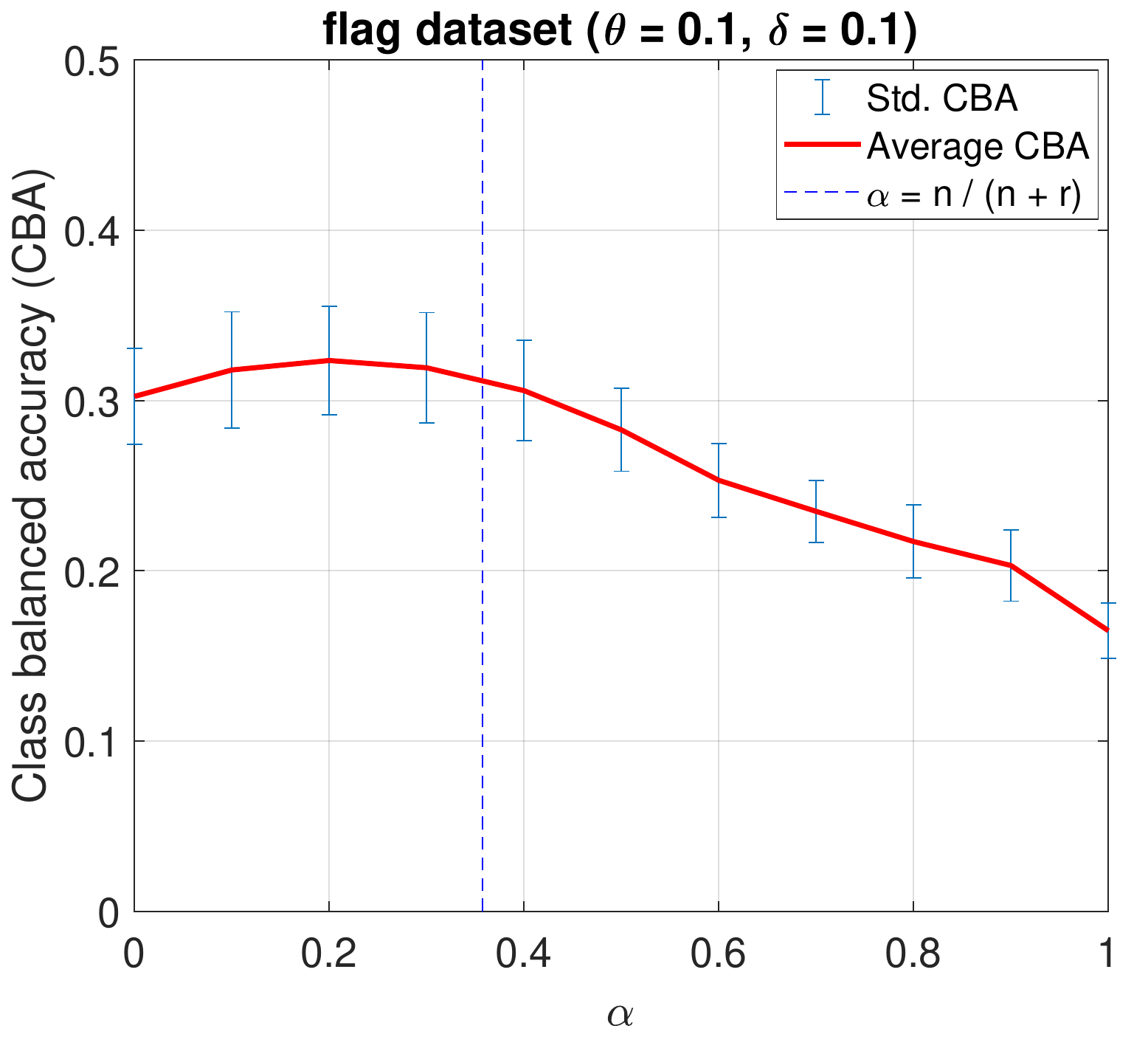}}
	\end{subfloat}
	\caption{The change in the class balanced accuracy according to the different values of $\alpha$ for the \textit{flag} dataset ($\theta = 0.1, \delta = 0.1$).}
	\label{Fig_3}
\end{figure}

To evaluate the impact of $\alpha$ on the performance of the EIOL-GFMM algorithms, we have changed the values of $\alpha$ from 0 to 1 with step 0.1 and recorded the average CBA scores using 10 times repeated stratified 4-fold cross-validation for 11 mixed-attribute datasets. The impact of $\alpha$ is studied for two cases, i.e., large-sized hyperboxes and small-sized hyperboxes. To obtain the large-sized hyperboxes, we established the parameters $\theta = \delta = 1$ so that the expansion process of hyperboxes is not constrained. To achieve small-sized resulting hyperboxes, we used the small values for both $\theta$ and $\delta$, i.e., $\theta = \delta = 0.1$ in this experiment. Fig. \ref{fig_2} shows the change in the CBA for different values of $\alpha$ in the case of large-sized hyperboxes. For $\delta = 1$, the behaviors of the EIOL-GFMM-v1 and EIOL-GFMM-v2 algorithms are identical. Fig. \ref{Fig_3} presents the change in the CBA results for different values of $\alpha$ in the case of small-sized hyperboxes for both EIOL-GFMM algorithms. We only present the results for a representative \textit{flag} dataset. The results of the remaining datasets are presented in Figs. S1, S2 and S3 in the supplemental document.


In general, the CBA values of both proposed learning algorithms at $\alpha = 0$ (using categorical features only) and $\alpha = 1$ (using numerical features only) are usually smaller than the results of using both types of features. The impact of $\alpha$ on both EIOL-GFMM-v1 and EIOL-GFMM-v2 algorithms are similar for many datasets. It can be observed that the influence of $\alpha$ on the GFMM models with small-sized hyperboxes is significantly higher than that with large-sized hyperboxes. This is demonstrated by the degree of oscillation in classification accuracy among different values of $\alpha$ in Fig. \ref{fig_2} and Fig. \ref{Fig_3}. It is because the value of $\alpha$ affects the results of the membership function, and the membership value in turn impacts the selection of the final hyperbox for each unseen pattern. In the case of small-sized hyperboxes, the number of hyperboxes is high and the small change in the membership value can lead to a significant change in the selected hyperbox.

As can be seen, the selection of $\alpha$ can result in the change in the classification performance, thus this parameter needs to be tuned in the learning process. However, for a large number of training samples, performing a hyper-parameter tuning step for $\alpha$ is time-consuming. For the online learning process, we also face another scenario, where we do not have sufficient samples at the training time. In this case, we cannot conduct the tuning process using a cross-validation technique to select an appropriate $\alpha$ for the learning algorithms. Therefore, we usually use a fixed setting for $\alpha$. From the empirical results in Figs. \ref{fig_2}, \ref{Fig_3} and Figs. S1, S2, and S3 in the supplemental document, it is interesting to observe that the highest CBA results are usually obtained for the value of $\alpha$ near the threshold $n/(n+r)$. Therefore, in the case of using a fixed setting for $\alpha$, we will set $\alpha = n/(n+r)$. With this setting, each feature is treated as equally important in decision making.

\subsubsection{Parameter $\delta$} \hfill

\begin{figure}[!ht]
\centering
\begin{subfloat}[EIOL-GFMM-v1]{
			\includegraphics[width=0.23\textwidth]{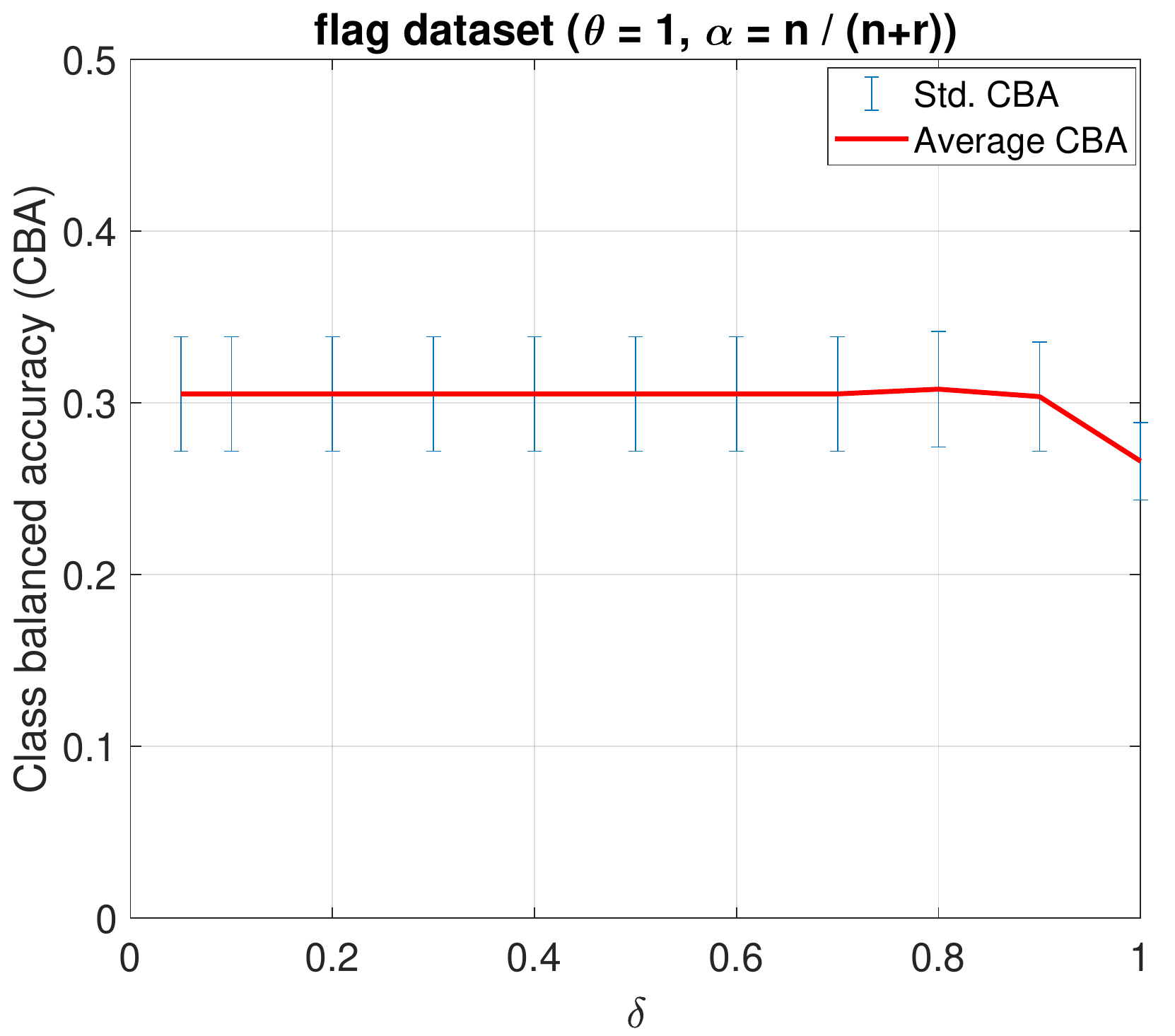}}
	\end{subfloat}
	\begin{subfloat}[EIOL-GFMM-v2]{
			\includegraphics[width=0.23\textwidth]{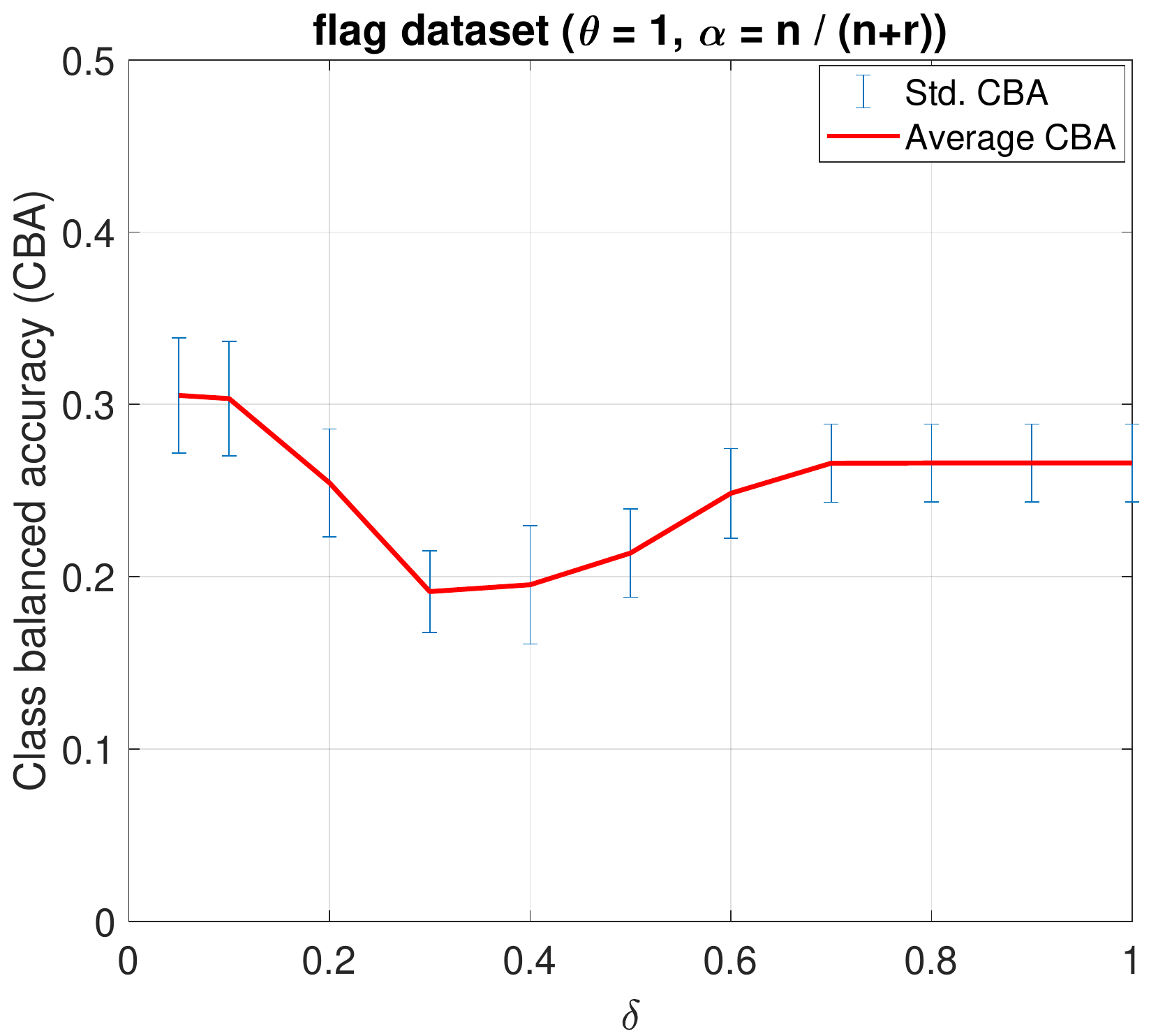}}
	\end{subfloat}
	\caption{The change in the class balanced accuracy according to the different values of $\delta$ for the $flag$ dataset ($\theta = 1, \alpha = n/(n+r)$).}
	\label{Fig_4}
\end{figure}

In this subsection, we will assess the impact of the maximum entropy changing threshold ($\delta$) for discrete attributes on the classification performance. To rule out the influence of $\theta$, we set $\theta = 1$ so that numerical features can be expanded without any limitation. From the above experimental results, we used $\alpha = n/(n+r)$. Therefore, the performance of the learning algorithms depends on the selection of $\delta$. We changed $\delta$ from 0.05 to 0.1 and kept the change step of 0.1 up to 1. The impact of $\delta$ on the IOL-GFMM-v1 and IOL-GFMM-v2 algorithms is illustrated in Fig. \ref{Fig_4} for the \textit{flag} dataset. The results for the remaining datasets can be found in Figs. S4 and S5 in the supplemental document.

From these results, it can be observed that the change in the CBA results for the EIOL-GFMM-v1 using $\delta < 0.7$ is very small. For $\alpha \geq 0.7$, its impact on the classification performance of the EIOL-GFMM-v1 on a number of datasets such as \textit{autralian}, \textit{heart}, and \textit{post operative} is significant, especially in the case of $\delta = 1$. In general, the classification error for $\delta \geq 0.7$ in the EIOL-GFMM-v1 is relatively high.

For the EIOL-GFMM-v2, however, the change in CBA values is small for $\delta < 0.2$ and $\delta \geq 0.7$. In contrast, the classification performance has significantly changed for the values of $\delta$ from 0.2 to 0.7. It can be seen that the impact of $\delta$ on the performance of the EIOL-GFMM-v2 algorithm is higher than that on the EIOL-GFMM-v1 algorithm. It is because the hyperbox expansion procedure for discrete features in the EIOL-GFMM-v2 algorithm can be performed much easier than that in the EIOL-GFMM-v1 algorithm. As a result, the number of generated hyperboxes in the EIOL-GFMM-v1 algorithm is higher than that of hyperboxes in the EIOL-GFMM-v2 algorithm, and so it can capture better the underlying data distribution. For the EIOL-GFMM-v1 algorithm, each discrete feature can only accommodate a new categorical value if the number of samples for the current categorical values is sufficiently large according to Properties \ref{prop_1} and \ref{prop_2}. As a result the homogeneity for each categorical feature of hyperboxes in the EIOL-GFMM-v1 is higher compared to that in the EIOL-GFMM-v2. Therefore, the change in the classification performance among the different values of $\delta$ in the first version of the proposed method is smaller in comparison to the second version.

\subsection{Comparing the Performance of the EIOL-GFMM Algorithms with Other Methods using the Fixed-Parameter Settings}
As we assume that there will not be sufficient number of training samples up front in the considered online learning scenarios, as discussed earlier, we will use a fixed setting for certain hyper-parameters which cannot be reliably tuned/optimised using available data. This section is to assess the proposed method in comparison to other solutions to deal with mixed-attribute data for the GFMMNN shown in \cite{Khuat2020enc} using previously evaluated fixed values of hyper-parameters. In particular, the proposed method will be compared with two learning algorithms with the mixed-attribute handling ability for the GFMM model including the Onln-GFMM-M1 \cite{Castillo12} and the Onln-GFMM-M2 \cite{Shinde16}. We will also compare the proposed method to the use of the original IOL-GFMM algorithm along with various encoding methods for categorical features.

\subsubsection{Algorithms with Mixed-Attribute learning ability}\label{fix_params} \hfill

In \cite{Khuat2020enc}, the different learning methods were compared to each other using three different settings for $\theta$, i.e., a small size $\theta = 0.1$, a large size $\theta = 0.7$, and an extreme case $\theta = 1$. To compare the proposed method to the previous solutions, we also used the same settings for the $\theta$ parameter. For the $\gamma$ parameter, we set $\gamma = 1$ as recommended in \cite{Abe01}. In addition to $\theta$ and $\gamma$, the existing learning algorithms for the GFMMNN with mixed-feature handling ability have their own hyper-parameters. The Onln-GFMM-M1 algorithm depends on the $\eta$ parameter, which represents the maximum hyperbox size for discrete features. We used $\eta \in \{0.1, 0.7, 1\}$ as shown in \cite{Khuat2020enc}. The Onln-GFMM-M2 algorithm has the $\beta$ parameter to control the minimum number of categorical features matched between the selected hyperbox and the input pattern so that hyperbox can be expanded to cover the input pattern. Similarly to \cite{Khuat2020enc}, we used $\beta \in \{25\%, 50\%, 75\%\}$ of the total number of features for each dataset. To be fair in the comparison, we used the parameter $\delta \in \{0.1, 0.7, 1\}$ and $\alpha = n/(n+r)$ for the proposed learning algorithms.

The average CBA values over 10 times repeated stratified 4-fold cross-validation with different parameter settings are shown in Table S.II in the supplemental document. It can be easily observed that in extreme cases (the largest values of parameters), the classification performance of our proposed method significantly outperforms the Onln-GFMM-M1 and Onln-GFMM-M2 algorithms. To facilitate the comparison of results, for each value of $\theta$, we will use the best results of the remaining parameter to rank four algorithms over 14 datasets. For example, for $\theta = 0.7$, the Onln-GFMM-M1 usually obtains the best performance using $\eta = 0.1$, the Onln-GFMM-M2 achieve its best results with $\beta = 0.75r$, and two proposed methods attain their best results using $\delta = 0.1$. The average ranking of algorithms using their best settings is shown in Table \ref{table1}.

\begin{table}[!ht]
\centering
\caption{The Average Rank for the Algorithms Using Their Best Settings}\label{table1}
\begin{scriptsize}
\begin{tabular}{cccc}
\hline
$\theta$ & Method       & Other parameters & Average rank \\ \hline
\multirow{4}{*}{0.1}  & Onln-GFMM-M1  & $\eta = 0.1$     & 3            \\ 
                      & Onln-GFMM-M2  & $\beta = 0.75r$  & 3.214        \\  
                      & EIOL-GFMM-v1 & $\delta = 0.1$   & 1.893        \\ 
                      & EIOL-GFMM-v2 & $\delta = 0.1$   & 1.893        \\ \hline
\multirow{4}{*}{0.7}  & Onln-GFMM-M1  & $\eta = 0.1$     & 2.429        \\ 
                      & Onln-GFMM-M2  & $\beta = 0.75r$  & 3.857        \\ 
                      & EIOL-GFMM-v1 & $\delta = 0.1$   & 1.607        \\ 
                      & EIOL-GFMM-v2 & $\delta = 0.1$   & 2.107        \\ \hline
\multirow{4}{*}{1}    & Onln-GFMM-M1  & $\eta = 0.1$     & 2.429        \\  
                      & Onln-GFMM-M2  & $\beta = 0.75r$  & 3.857        \\  
                      & EIOL-GFMM-v1 & $\delta = 0.1$   & 1.679        \\ 
                      & EIOL-GFMM-v2 & $\delta = 0.1$   & 2.036        \\ \hline
\end{tabular}
\end{scriptsize}
\end{table}

It can be seen that the classification performance of our proposed methods is better than that of two existing algorithms with the mixed-attribute learning ability for three different thresholds of $\theta$. To conclude if there are statistically significant differences among algorithms, we will carry out a non-parametric
test procedure as recommended in \cite{Demsar2006} employing the Friedman rank-sum test with a confidence level of 95\% (a significance level $\epsilon = 0.05$). The null hypothesis is ``there are no statistical differences between learning algorithms", and if this hypothesis is rejected, then we perform the Nemenyi post-hoc test to determine the particular differences. For $Z$ datasets and $M$ algorithms, the Friedman statistic distribution is computed using average ranks $R_j (j \in [1, M])$ of each algorithm $j$ as follows:
\begin{equation}
    \chi_F^2 = \cfrac{12 Z}{M \cdot (M + 1)} \left[ \sum_{j=1}^M R^2_j - \cfrac{M \cdot (M + 1)^2}{4}\right]
\end{equation}

From $\chi_F^2$, a F-distribution with $M - 1$ and $(M - 1) \cdot (N - 1)$ degrees of freedom can be calculated using \eqref{f_dis}.
\begin{equation}\label{f_dis}
    F_F = \cfrac{(Z - 1) \cdot \chi_F^2}{Z \cdot (M - 1) - \chi_F^2}
\end{equation}
The rejection of the null hypothesis occurs with the significance level $\epsilon$ if $F_F$ is smaller than a critical value of $F(M - 1, (M-1)\cdot(N - 1), \epsilon)$. In this experiment, we used 14 datasets and four learning algorithms, so $F_F$ is distributed according to the F-distribution with $4 - 1 = 3$ and $(4 - 1) \cdot (14 - 1) = 39$ degrees of freedom. The critical value of $F(3, 39)$ for $\epsilon = 0.05$ is 2.845.
\begin{figure}[!ht]
    \centering
    \includegraphics[width=0.4\textwidth]{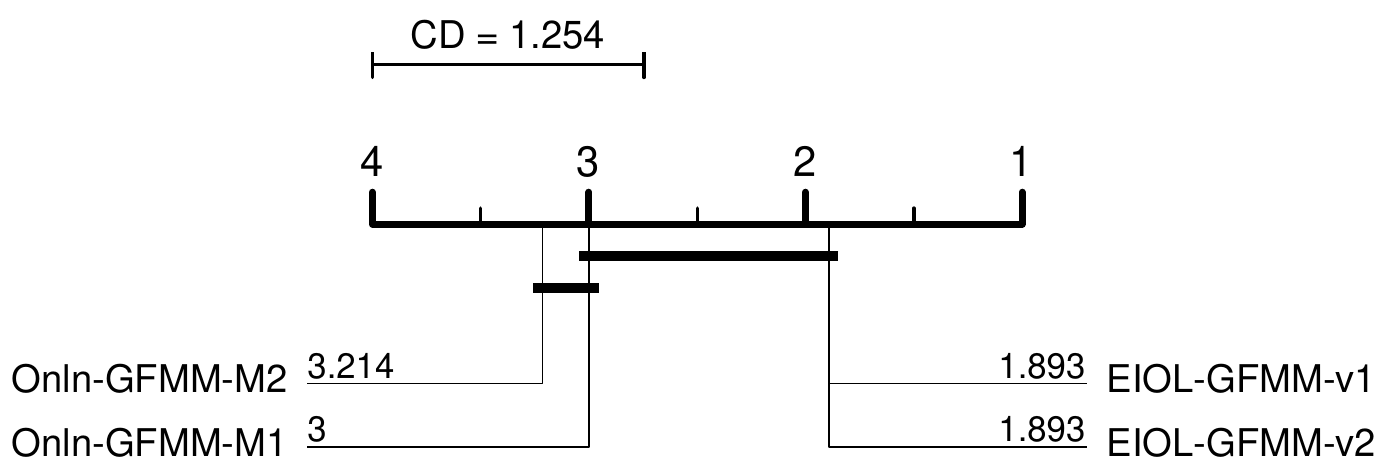}
    \caption{Critical difference diagram for four learning algorithms ($\theta = 0.1$).}
    \label{Fig_5}
\end{figure}

For $\theta = 0.1$, we obtain $F_F = 5.5539 > 2.845$, and so the null hypothesis is rejected. This means that there are significant differences between the results of learning algorithms. Using the Nemenyi post-hoc test, we obtain a critical difference (CD) diagram in Fig. \ref{Fig_5}. The groups of algorithms that are not significantly different from each other are connected by a solid line. We can see that our proposed methods are statistically better compared to the Onln-GFMM-M2 algorithm with the selected settings. However, there is no statistically significant difference in the classification performance between the proposed methods and the Onln-GFMM-M1 algorithm.
\begin{figure}[!ht]
    \centering
    \includegraphics[width=0.4\textwidth]{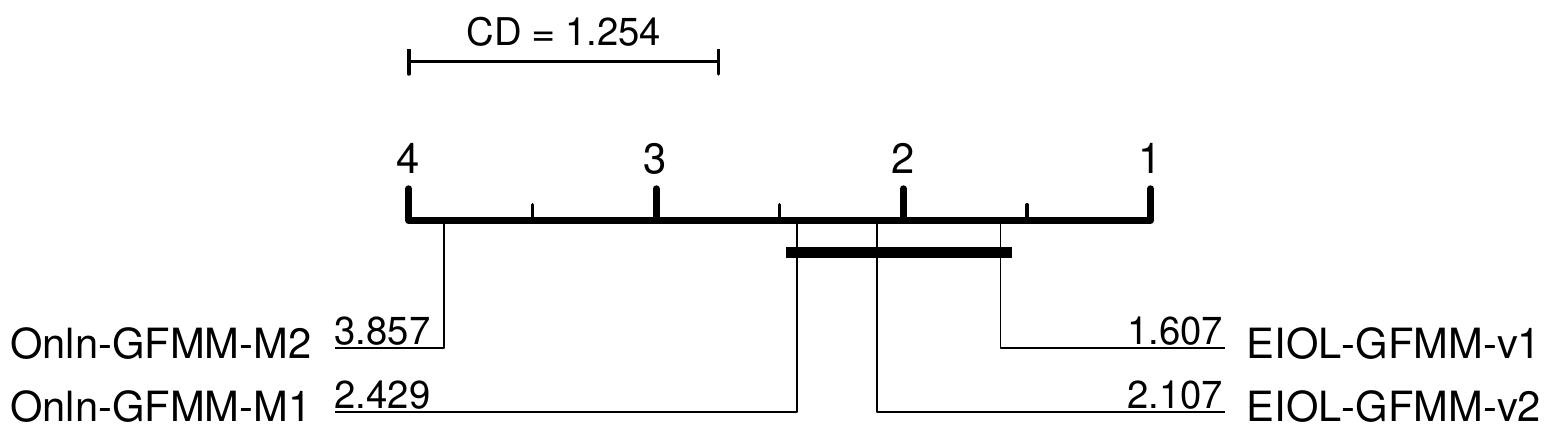}
    \caption{Critical difference diagram for four learning algorithms ($\theta = 0.7$).}
    \label{Fig_6}
\end{figure}

For $\theta = 0.7$, we have $F_F = 16.5238 > 2.845$, and so the null hypothesis is also rejected. Applying the Nemenyi post-hoc test, we have a CD diagram in Fig. \ref{Fig_6}. We can observe that there are no statistically significant differences among the obtained empirical results in the groups of two proposed learning algorithms and the Onln-GFMM-M1 algorithm. However, the algorithms in this group significantly outperform the Onln-GFMM-M2 algorithm.
\begin{figure}[!ht]
    \centering
    \includegraphics[width=0.4\textwidth]{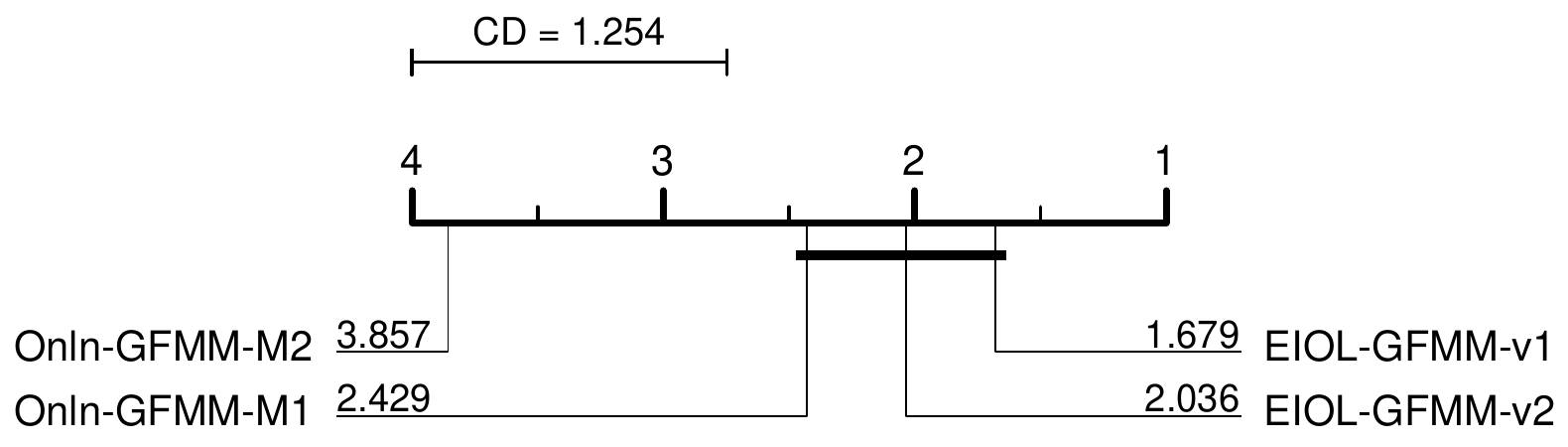}
    \caption{Critical difference diagram for four learning algorithms ($\theta = 1$).}
    \label{Fig_7}
\end{figure}

Similarly, for $\theta = 1$, we obtain $F_F = 15.7716 > 2.845$, and so the null hypothesis is rejected as well. Fig. \ref{Fig_7} shows the CD diagram using the Nemenyi post-hoc test. In this case, the statistical difference in the classification performance among the four methods is the same as in the case of $\theta = 0.7$.

For the complexity of the resulting GFMM models using these learning algorithms, the average number of generated hyperboxes for each method is shown in Table S.III in the supplemental document. We can see that in most of the cases, the complexity of the model using the Onln-GFMM-M2 algorithm is lowest, while the complexity of the GFMMNN using the EIOL-GFMM-v1 is highest. The number of generated hyperboxes using the EIOL-GFMM-v2 algorithm is usually smaller than that using the Onln-GFMM-M1 algorithm.

\subsubsection{Comparing the Proposed Method to the Original Learning Algorithm using Encoding Methods} \hfill

In this subsection, we will compare the EIOL-GFMM algorithms to the original IOL-GFMM algorithm using different encoding techniques. In \cite{Khuat2020enc}, there are eight encoding methods used to transform the categorical features into numerical features, i.e., Leave-One-Out (LOO), CatBoost, label, one-hot, target, James-Stein, Helmert, and Sum encoding techniques. Similarly to the above experiment, we will consider three different thresholds for $\theta$ including 0.1, 0.7, and 1. To be fair in the comparison, we established the value of $\delta$ equal to $\theta$.

The average CBA values over 10 times repeated stratified 4-fold cross-validation of these approaches are shown in Table S.IV, and their ranks are presented in Table S.V in the supplemental document. The average rank for these methods for different thresholds of $\theta$ is shown in Table \ref{table2}, in which the best results are highlighted in bold. In general, we can see that the average performance of our proposed method is better than that using the original IOL-GFMM algorithm along with encoding techniques. One of the strong points of our proposed method is that it does not use any encoding method for discrete attributes.

\begin{table}[!ht]
\centering
\caption{The Average Ranks for the Proposed Method and the Original IOL-GFMM using Different Encoding Techniques}\label{table2}
\begin{scriptsize}
\begin{tabular}{lccc}
\hline
\multirow{2}{*}{Method} & \multicolumn{3}{c}{$\theta (= \delta)$}         \\ \cline{2-4} 
                        & 0.1            & 0.7            & 1              \\ \hline
IOL-GFMM + CatBoost     & 5.357          & 5.071           & 5          \\ 
IOL-GFMM + One-hot      & 8.179          & 8.036          & 7.536            \\ 
IOL-GFMM + LOO          & 4.857          & 4.786            & 5.643              \\ 
IOL-GFMM + Label        & 5.107          & 5.607          & 5.464         \\ 
IOL-GFMM + Target       & 5.464          & 4.464          & 5.429          \\ 
IOL-GFMM + James-Stein  & 5.250          & 4.536          & 5.214          \\ 
IOL-GFMM + Helmert      & 7.893          & 7.750          & 7.179          \\ 
IOL-GFMM + Sum          & 5.607          & 5.750          & 7.821          \\
EIOL-GFMM-v1            & 3.679          & \textbf{3.536} & \textbf{2.857} \\ 
EIOL-GFMM-v2            & \textbf{3.607} & 5.464          & \textbf{2.857} \\ \hline
\end{tabular}
\end{scriptsize}
\end{table}

Similarly to the above experiments, we will use a statistical test procedure to analyze the statistical difference among the methods. The critical value of $F(9, 117)$ for 10 methods and 14 datasets at $\epsilon = 0.05$ is 1.9608.

\begin{figure}[!ht]
    \centering
    \includegraphics[width=0.48\textwidth]{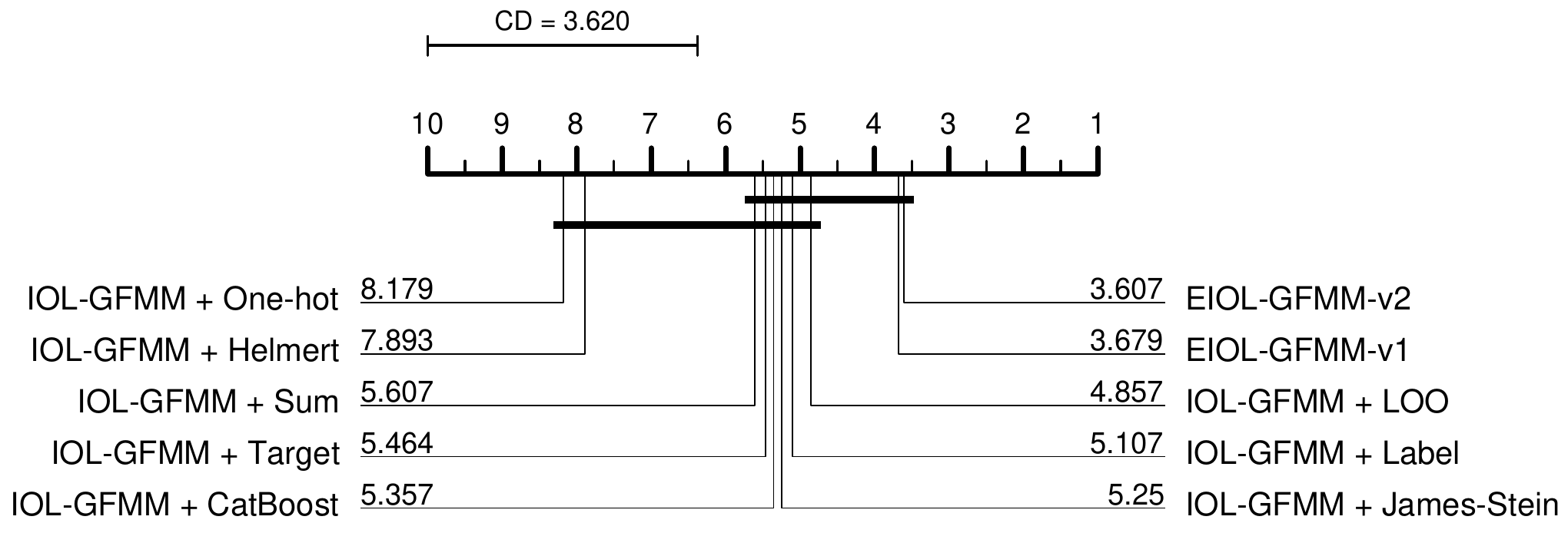}
    \caption{Critical difference diagram for the proposed method and the original algorithm using encoding methods ($\theta = 0.1$).}
    \label{Fig_8}
\end{figure}

For $\theta (=\delta) = 0.1$, we obtain $F_F = 4.289 > 1.9608$, and so there are statistically significant differences among methods. Using the Nemenyi post-hoc test, we obtain a CD diagram in Fig. \ref{Fig_8}. We can see that the proposed method is significantly better than the original IOL-GFMM algorithm using the one-hot or Helmert encoding method. However, there are no statistical differences between the proposed methods and the IOL-GFMM algorithm using the remaining encoding techniques.

\begin{figure}[!ht]
    \centering
    \includegraphics[width=0.47\textwidth]{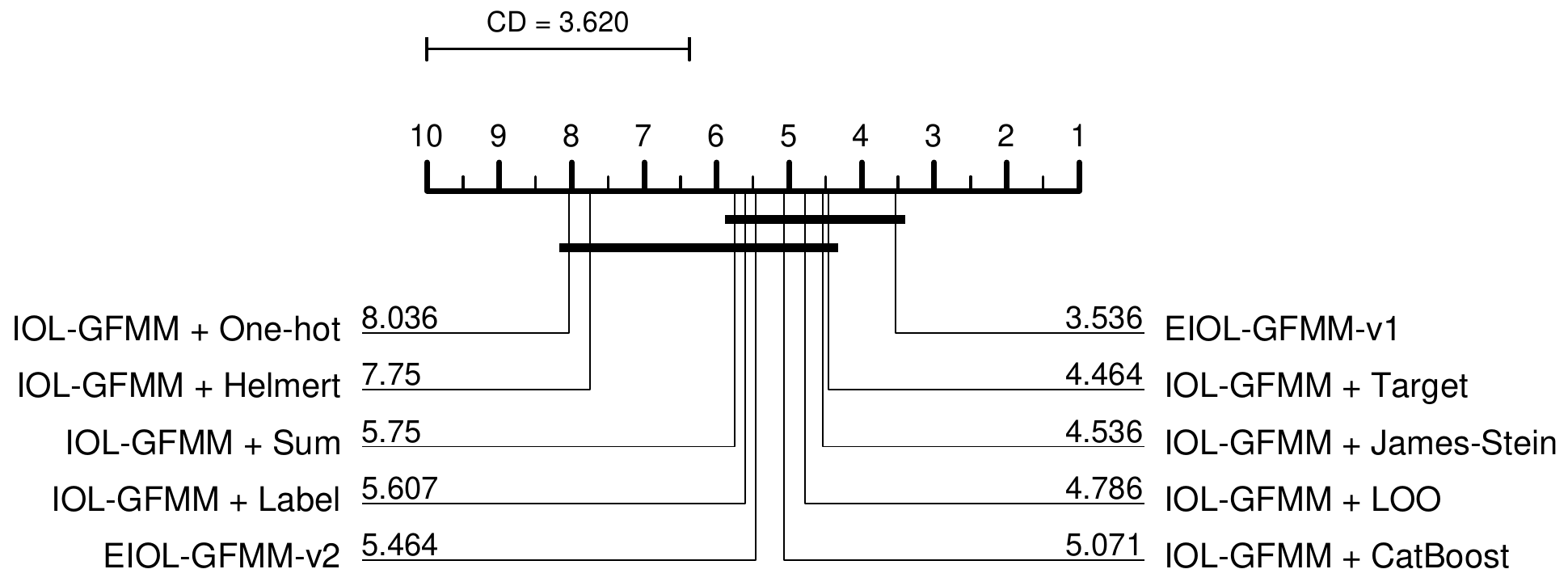}
    \caption{Critical difference diagram for the proposed method and the original algorithm using encoding methods ($\theta = 0.7$).}
    \label{Fig_9}
\end{figure}

For $\theta (=\delta) = 0.7$, we have $F_F = 3.6596 > 1.9608$, and so there are also statistically significant differences among methods in this case. Employing the Nemenyi post-hoc test, we obtain a CD diagram in Fig. \ref{Fig_9}. In this case, the EIOL-GFMM-v1 is significantly better than the original IOL-GFMM algorithm using the one-hot or Helmert encoding method as well, but it does not statistically outperform the original algorithm using the remaining encoding approaches. Moreover, in this case, there is not sufficient evidence to conclude that the EIOL-GFMM-v2 is statistically better than the original algorithm employing encoding methods.

\begin{figure}[!ht]
    \centering
    \includegraphics[width=0.47\textwidth]{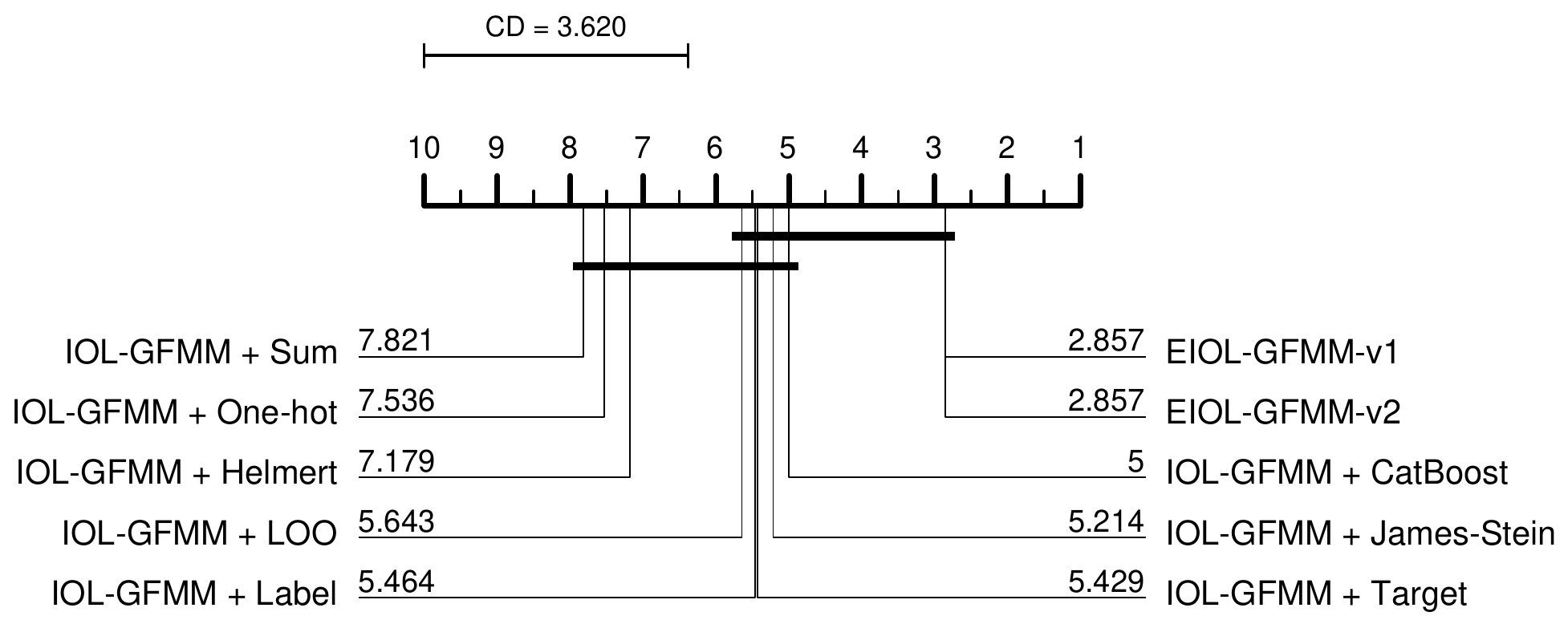}
    \caption{Critical difference diagram for the proposed method and the original algorithm using encoding methods ($\theta = 1$).}
    \label{Fig_10}
\end{figure}

For the extreme case $\theta (=\delta) = 1$, we obtain $F_F = 6.2138 > 1.9608$, and so the null hypothesis is also rejected. Fig. \ref{Fig_10} shows the CD diagram, in this case, using the Nemenyi post-hoc test. It can be observed that there are no statistical differences among the original algorithms using encoding methods. Our proposed methods significantly outperform the original learning algorithm using the one-hot, sum, or Helmert encoding method. However, there are no statistical differences between our proposed methods and the original IOL-GFMM algorithm using the remaining encoding approaches.

\subsection{Evaluating the Role of the Hyper-Parameter Tuning on the Performance of the EIOL-GFMM Algorithms}
\subsubsection{Hyper-Parameter Tuning and the Estimation of $\alpha$} \hfill

In the case that we are given a large number of samples at the training time to build an initial model, we can perform a hyperparameter tuning step for $\alpha$ but this process is time-consuming. Meanwhile, the empirical results in subsection \ref{eva_alpha} indicate the relation between suitable values of $\alpha$ and the ratio of the number of continuous features over the total number of features. In this subsection, we propose a simple way to estimate the appropriate value of $\alpha$ in a data-driven manner. The estimation method does not loop through predefined values of $\alpha$ as in the tuning process, and so they will run faster than the hyper-parameter tuning step for $\alpha$.

For each training fold $T_i$, we split it into three inner folds. To estimate the value of $\alpha$ for each training fold $T_i$, we will repeat the learning process three times. Each time two inner folds are used to build the GFMM model and the remaining inner fold is used as a validation set. For each inner training fold $T_{ij} (j\in[1, 3])$, we split it into two separate parts, in which each part contains either continuous attributes or discrete attributes. Then, we will construct two separate GFMM models using the EIOL-GFMM algorithm from these two training parts. After that, we compute the CBA value for each trained model using the inner validation fold $V_{ij}$. Let $CBA_{ij1}$ and $CBA_{ij2}$ be the CBA scores for the GFMM models trained on continuous features only and discrete features only, respectively. We have two ways to estimate the value of $\alpha$ for each training fold $T_i$. The first way uses both the CBA values and the number of features, denoted Est-$\alpha$-v1 in this paper, as follows:
\begin{equation}
\hat{\alpha} = \cfrac{\sum\limits_{j=1}^3 CBA_{ij1} \cdot n}{\sum\limits_{j=1}^3 CBA_{ij1} \cdot n + \sum\limits_{j=1}^3 CBA_{ij2} \cdot r}
\end{equation}
The second estimation way of $\alpha$ uses only the obtained CBA values, called Est-$\alpha$-v2, as follows:
\begin{equation}
    \hat{\alpha} = \cfrac{\sum\limits_{j=1}^3 CBA_{ij1}}{\sum\limits_{j=1}^3 CBA_{ij1} + \sum\limits_{j=1}^3 CBA_{ij2}}
\end{equation}
Two ways of estimating $\alpha$ are summarized in Fig. S6 in the supplemental material. After obtaining the value of $\alpha$, we use it to train a final model using the whole mixed-attribute training fold $T_i$ and evaluate its performance using the $i$-$th$ testing fold. The above process is repeated 40 times (10 times repeated stratified 4-fold cross-validation) to compare the average CBA values among different methods. 

This section will compare the effectiveness of two above estimation methods to the fixed setting of $\alpha = n/(n+r)$ and the parameter tuning method for $\alpha$. In the parameter tuning method, for each training fold $T_i$, we split $T_i$ into three inner training folds. Two inner training folds are used to build the GFMM model using the proposed EIOL-GFMM algorithm and the remaining fold is used as a validation fold. We will iterate this process three times to obtain three CBA values from three validation folds for each value $\alpha \in \{0, 0.1, \ldots, 0.9, 1\}$. The value of $\alpha$ resulting in the highest average CBA value over three inner validation folds is used to build the final GFMM model on the whole training fold $T_i$, and the trained model is assessed by the CBA value on the $i$-$th$ testing fold. The whole process is repeated 40 times for different training folds $T_i$.

The average CBA results of 40 GFMM models trained using the proposed algorithms with two estimation methods of $\alpha$, the parameter tuning method and the fixed setting of $\alpha$ for 11 datasets are shown in Table S.VI in the supplemental material. The rank for these methods is presented in Table S.VII in the supplemental document. It is noted that the results are reported over 11 out of 14 experimental datasets because these datasets contain both continuous and discrete features while three remaining datasets consist of only discrete features. The average rank over 11 datasets for different methods of finding the value of $\alpha$ for the GFMM model trained using the proposed algorithm is shown in Table \ref{table3}. Similarly to subsection \ref{eva_alpha}, we compare the methods of finding $\alpha$ in two cases, i.e., small-sized hyperboxes ($\theta = \delta = 0.1$) and large-sized hyperboxes ($\theta = \delta = 1$). The best rank in each row is highlighted in bold. In the case of $\theta = \delta = 1$, the behavior of both EIOL-GFMM-v1 and EIOL-GFMM-v2 is the same, and so they lead to the same results.

\begin{table}[!ht]
\centering
\caption{Average Rank for Different Methods used to Find the Values for Parameter $\alpha$}\label{table3}
\begin{scriptsize}
\begin{tabular}{lccccc}
\hline
\multicolumn{1}{c}{Algorithm} & $\theta = \delta$ & Tuning $\alpha$ & Est-$\alpha$-v1 & Est-$\alpha$-v2 & $\alpha = n/(n+r)$ \\ \hline
EIOL-GFMM-v1                    & 0.1               & 2.727          & \textbf{2}               & 3           & 2.273              \\ 
EIOL-GFMM-v2                    & 0.1               & 2.545          & \textbf{2.318}           & 2.727           & 2.409              \\ 
Both                            & 1                 & \textbf{2.273}          & 2.773           & \textbf{2.273}           & 2.682              \\ \hline
\end{tabular}
\end{scriptsize}
\end{table}

We can observe that for small values of $\theta$ and $\delta$, the estimation method using the CBA values from two separate models along with the number of features usually results in the best average CBA values in comparison to the second estimation method, the parameter tuning approach, and the fixed setting of $\alpha$ for both learning algorithms. However, the second estimation method without using the number of features often leads to the worst results. Interestingly, in this case, the fixed value of $\alpha = n/(n+r)$ shows slightly better results than the hyper-parameter tuning method. In the case of generating the largest hyperbox sizes, the best predictive results belong to the models using the hyper-parameter tuning method and the Est-$\alpha$-v2 method. Meanwhile, the first estimation method usually leads to the worst classification performance.

\begin{figure}[!ht]
    \centering
    \includegraphics[width=0.26\textwidth]{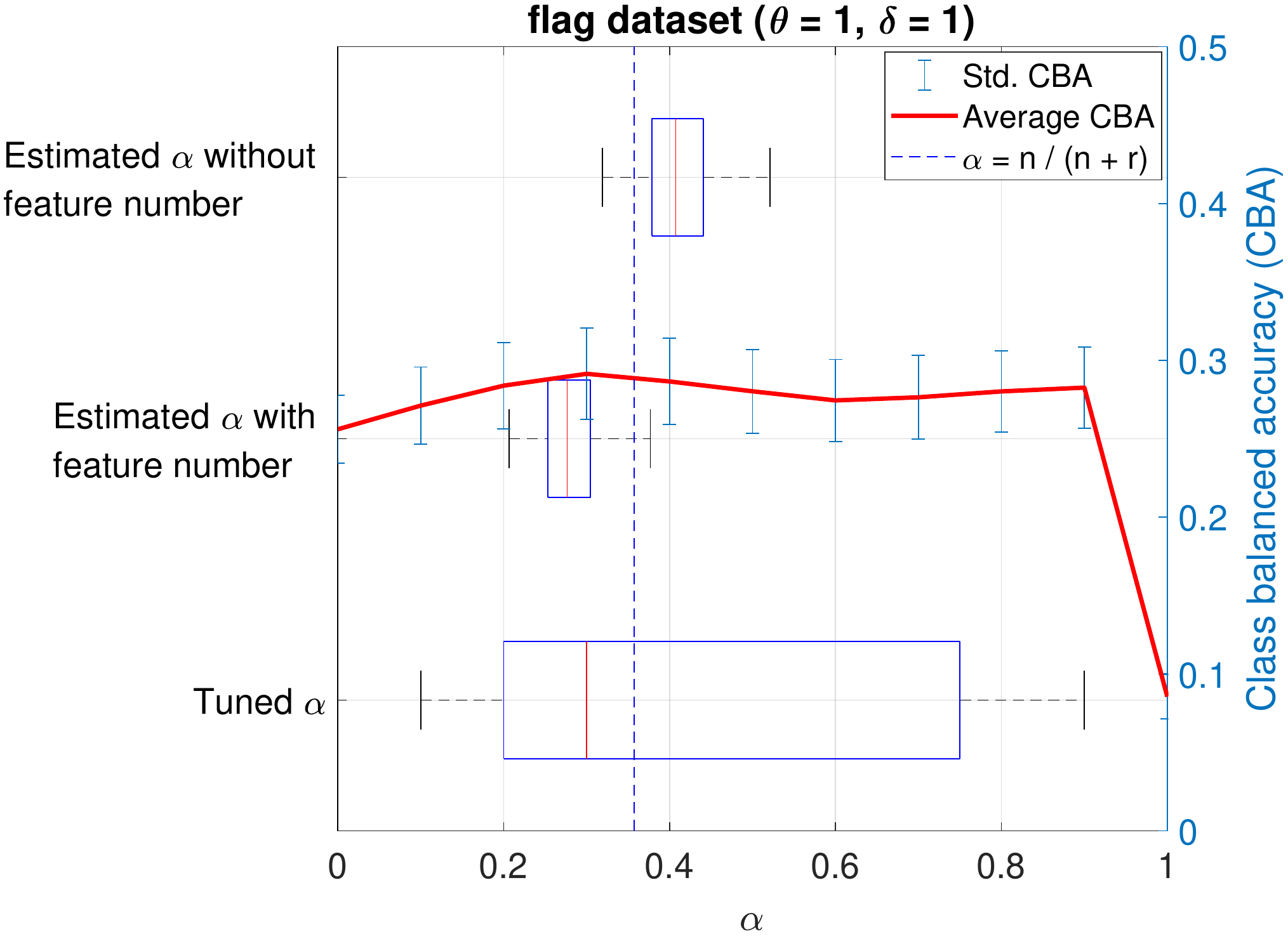}
    \caption{The distribution of the obtained $\alpha$ values for different methods used to find $\alpha$ and the CBA values for the \textit{flag} dataset ($\theta = 1, \delta = 1$).}
    \label{Fig_11}
\end{figure}

To explain these facts, we will examine the distribution of the obtained values of $\alpha$ through 40 iterations and the change in the corresponding CBA values. Fig. \ref{Fig_11} shows the distribution of the obtained $\alpha$ values for different methods in the case of largest-sized hyperboxes for the \textit{flag} dataset. The results in the case of $\theta = \delta = 0.1$ are presented in Fig. \ref{Fig_12}. Similar results for all of the remaining datasets can be found in Figs. S7, S8, and S9 in the supplemental material.

\begin{figure}[!ht]
\centering
\begin{subfloat}[EIOL-GFMM-v1]{
			\includegraphics[width=0.23\textwidth]{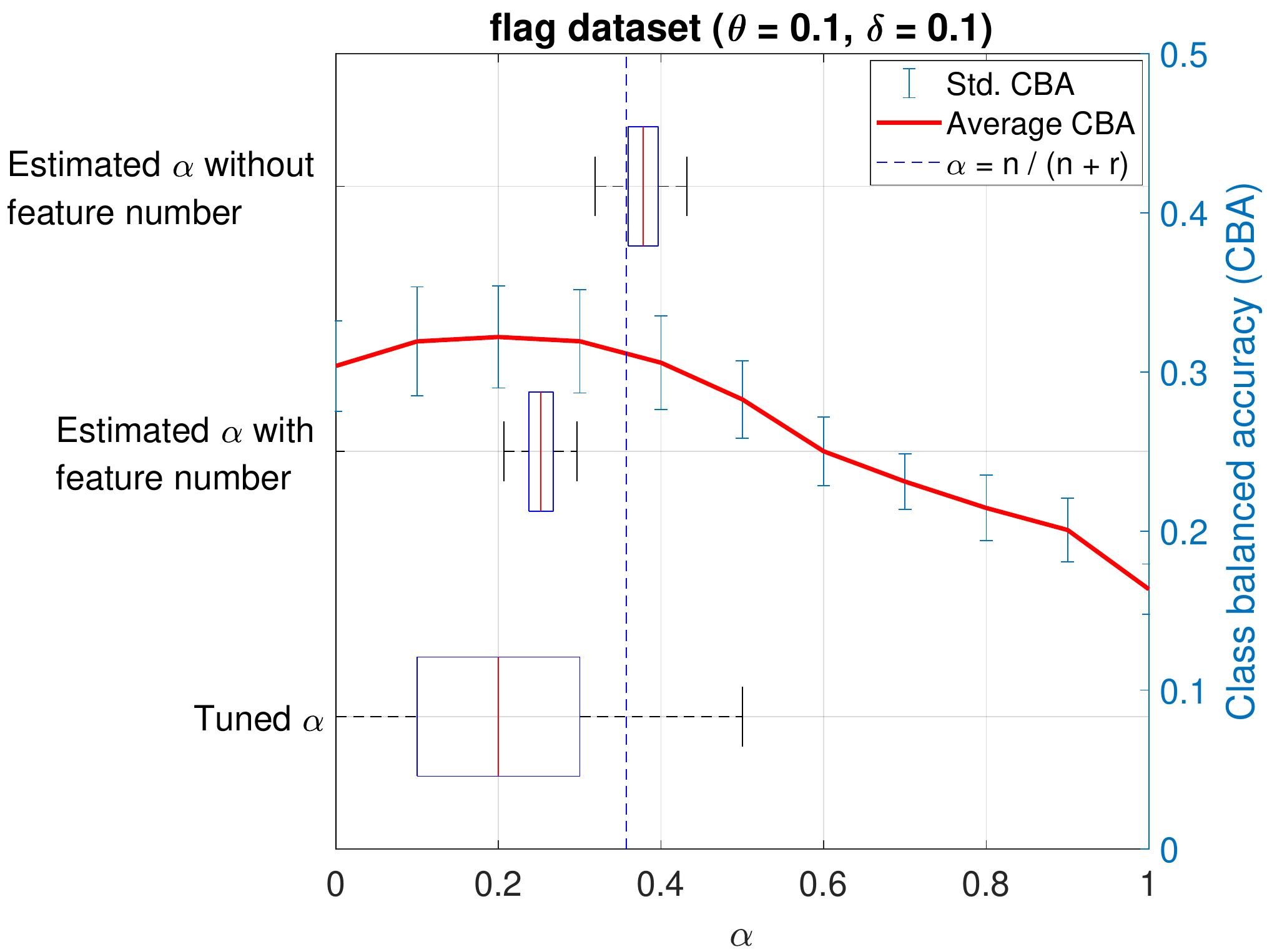}}
	\end{subfloat}
	\begin{subfloat}[EIOL-GFMM-v2]{
			\includegraphics[width=0.23\textwidth]{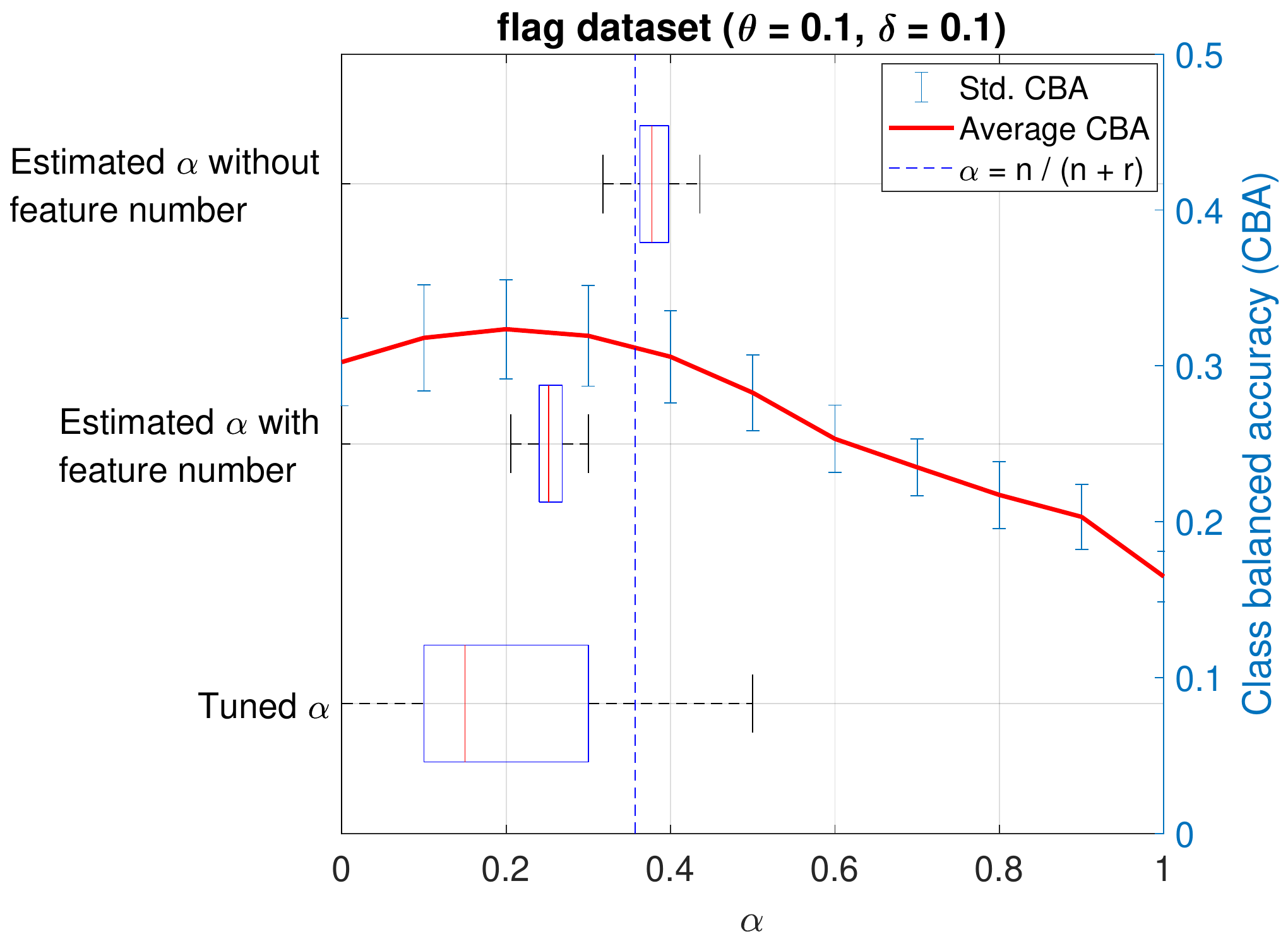}}
	\end{subfloat}
	\caption{The distribution of the obtained $\alpha$ values for different methods used to find $\alpha$ and the CBA values for the \textit{flag} dataset ($\theta = 0.1, \delta = 0.1$).}
	\label{Fig_12}
\end{figure}

We can see that, in both cases, the use of the hyper-parameter tuning method returns a wide range of values for $\alpha$, in which the obtained median value of $\alpha$ locates near the $\alpha$ value resulting in the best classification result. In the case of small-sized hyperboxes, it can be seen that the deviation in the classification results among adjacent values of $\alpha$ is high. Therefore, a wide range of $\alpha$ values usually leads to a low average classification result compared to the use of a narrow range of $\alpha$ values near the best results. We can see from Fig. \ref{Fig_12}, the obtained $\alpha$ values employing two estimation methods are distributed in a narrower area than that using the hyper-parameter tuning approach. Also, the range of the obtained $\alpha$ values of the Est-$\alpha$-v2 is wider than that using the Est-$\alpha$-v1 method. However, the range of the obtained $\alpha$ values using the Est-$\alpha$-v1 is nearer the $\alpha$ value leading to the best classification performance than one using the Est-$\alpha$-v2. Hence, in this case, the Est-$\alpha$-v1 method usually gives the best classification results among the four methods.

In the case of largest-sized hyperboxes, the difference in the performance among different values of $\alpha$ is small. Therefore, the wide range of the obtained $\alpha$ values using the hyper-parameter tuning method regularly leads to a better average classification result compared to the outcomes employing other methods. As can be seen from Fig. S7 in the supplemental document that two estimation methods return a narrower range of the obtained $\alpha$ values in comparison to the use of the hyper-parameter tuning approach. However, in this case, the obtained $\alpha$ values using the Est-$\alpha$-v2 usually locates nearer the $\alpha$ values leading to much better performance than those in the case of using the Est-$\alpha$-v1 method. Therefore, the performance of the GFMM model using the Est-$\alpha$-v2 outperforms that adopting the Est-$\alpha$-v1 method.

In short, the second estimation method is appropriate for the model having a small number of hyperboxes, while the first estimation method should be used in the case when the resulting model has a large number of hyperboxes.

\subsubsection{Comparing the EIOL-GFMM Algorithms to Other Algorithms with the Mixed-Attribute Learning Ability} \hfill

In this experiment, we will compare the classification performance of our proposed method and two existing algorithms with the mixed-attribute learning ability using the hyper-parameter tuning procedure for important parameters in each learning algorithms. For the $\theta$ value in all learning algorithms, we will find its best parameter value in the range of $\{0.1, 0.2, \ldots, 0.9, 1\}$ for each training fold. The $\eta$ parameter for the Onln-GFMM-M1 algorithm is searched in the range of $\{0.1, 0.3, 0.5, 0.9, 1\}$. The searching range of the $\beta$ parameter for the Onln-GFMM-M2 is $\{10\%, 30\%, 50\%, 70\%, 90\%, 100\%\}$ of the total number of categorical features. For the two proposed algorithms in this paper, the $\delta$ parameter is searched in the range of $\{0.1, 0.3, 0.5, 0.9, 1\}$, while the $\alpha$ value is sought in the range of $\{0, 0.1,\ldots, 0.9, 1\}$.

Each training fold $T_i$ is split into three inner folds, in which two inner folds are used for training a GFMM model using learning algorithms. Then, we use the remaining fold to obtain the CBA value. This process is repeated three times for every inner validation fold. The combination of parameters resulting in the best average CBA values through three validation folds is used to train the final GFMM model using the whole training fold $T_i$. After that, this model is evaluated using the corresponding testing fold. This process is iterated 40 times (10 times repeated stratified 4-fold cross-validation) for each dataset. The average CBA results for four learning methods using the above hyper-parameter tuning approach are shown in Table S.VIII and their ranks are presented in Table S.IX in the supplemental document. The average rank for each method over 11 mixed-attribute datasets is shown in Table \ref{table4}.

\begin{table}[!ht]
\centering
\caption{Average Ranks for the Learning Algorithms using the Hyper-Parameter Tuning Approach} \label{table4}
\begin{scriptsize}
\begin{tabular}{lc}
\hline
Algorithm          & Average rank \\ \hline
Onln-GFMM-M1 & 3.182 \\
Onln-GFMM-M2 & 2.909\\
EIOL-GFMM-v1 & \textbf{1.5}\\
EIOL-GFMM-v2 & 2.409 \\ \hline
\end{tabular}
\end{scriptsize}
\end{table}

We can observe that the two proposed learning algorithms outperform two existing learning algorithms with the mixed-attribute handling ability, in which the best performance belongs to the EIOL-GFMM-v1 algorithm. For the experimental results in subsection \ref{fix_params}, we can see that the Onln-GFMM-M1 algorithm is better than the Onln-GFMM-M2 algorithm using the fixed-parameter settings. However, by using the hyper-parameter tuning method, the Onln-GFMM-M2 algorithm overcomes the Onln-GFMM-M1. This is because of the difference in the distribution between the inner training set used to find the best combination of parameters and the training fold used to build the final model. The Onln-GFMM-M1 needs to use the entire training data to find the distance between categorical values based on the relationship between the occurrence frequency of discrete values and classes. These distance values are deployed to build membership functions. Therefore, when the training data change, the best combination of parameters on the inner training folds no longer maintains the superior classification performance when used on the training fold $T_i$. Our proposed methods do not use the training samples to build the similarity measure among categorical features, and so they still achieve the best performance as in the case of using the fixed parameter settings.

Interestingly, the classification performance of learning algorithms using the hyper-parameter tuning method in several datasets such as \textit{cmc}, \textit{cmc}, \textit{zoo}, \textit{australian}, and \textit{japanese credit} is worse than those using fixed parameter settings presented in subsection \ref{fix_params}. This is because the representativeness and distribution of the inner validation sets used to find the best combination of parameters are different from the training and testing folds. Therefore, the best parameters obtained from the inner validation folds may not lead to the best classification accuracy on the testing set. As a result, the hyper-parameter tuning method does not always result in better performance than the use of fixed parameters.

\begin{figure}[!ht]
    \centering
    \includegraphics[width=0.45\textwidth]{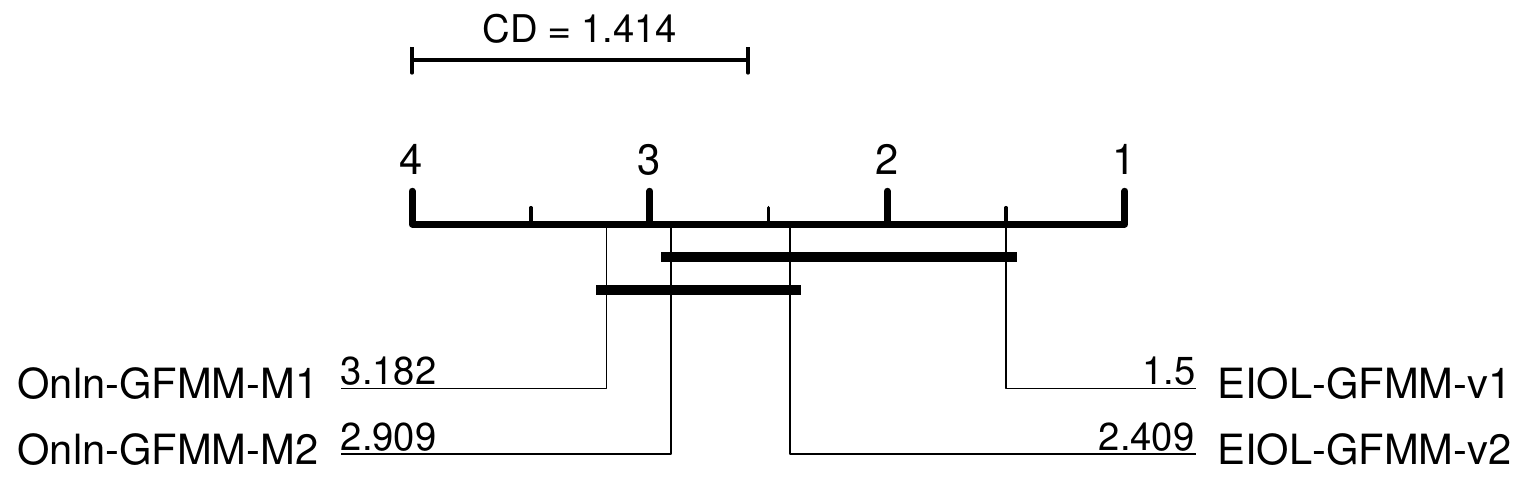}
    \caption{A CD diagram of four learning algorithms using the hyper-parameter tuning method.}
    \label{Fig_13}
\end{figure}

To verify the statistical difference in the performance among the learning algorithms, we will use the above Friedman rank-sum test. For 11 datasets and 4 learning algorithms, $F_F$ is distributed according to the F-distribution with $4 - 1 = 3$ and $(4 - 1) \cdot (11 - 1) = 30$ degrees of freedom. The critical value of $F(3, 30)$ at a significant level $\epsilon = 0.05$ is 2.9223. In this case, we obtain $F_F = 4.884 > 2.9223$. Therefore, there are statistically significant differences among the four considering algorithms. Using the Nemenyi post-hoc test, we achieve a CD diagram in Fig. \ref{Fig_13}.

We can see that there is a statistically significant difference in the classification performance between the EIOL-GFMM-v1 and the Onln-GFMM-M1 algorithms in this case. For $CD = 1.414$, we can also conclude that the EIOL-GFMM-v1 algorithm significantly better than the Onln-GFMM-M2 algorithm. However, the EIOL-GFMM-v2 does not statistically outperform both existing learning algorithms with the mixed-attribute learning ability.

\section{Conclusion and Future Work} \label{conclusion}
This paper presented a new online learning algorithm for the GFMMNN with mixed-attribute data. The proposed method expands the current membership function for both continuous and discrete features. We also extend the current architecture of the GFMM model for mixed-attribute data and introduce a new way of learning for categorical dimensions based on the change in the entropy when accommodating new discrete values without using any encoding methods. The experimental results confirmed the superior classification performance of our proposed method in comparison to the current solutions to handle the mixed-type datasets for the GFMMNN.

Although the GFMMNN for mixed-attribute data itself can explain the predicted results using the membership function to select the appropriate hyperbox, to make it friendly and easy-to-read for users, it is necessary to extract and optimize \textit{if-then} rule sets from the resulting hyperboxes for both continuous and discrete features in the future studies. The interpretability of predictive models is a critical factor when applying the machine learning algorithms for high-stakes applications such as medicine, finance, or criminal justice \cite{Rudin2019}. Furthermore, the classification accuracy depends on the selection of parameters, thus the next research should assess the use of optimization algorithms such as genetic algorithms \cite{Khuat2017} to evolve the hyperboxes and optimize their hyperparameters simultaneously. When applying the online learning algorithms for applications in dynamic changing environments, these learning algorithms need to detect and adapt to the change of the underlying data distribution \cite{Gabrys99,Gabrys04,Sahel2007,Kadlec2009,Salvador2016}. Therefore, one of the potential research directions is to integrate the adaptation ability into the proposed algorithm.

\section*{Acknowledgment}
T.T. Khuat acknowledges FEIT-UTS for awarding his PhD scholarships (IRS and FEIT scholarships).

\ifCLASSOPTIONcaptionsoff
  \newpage
\fi



\bibliographystyle{IEEEtran}
\bibliography{reference}
%

%

\vfill








\end{document}


%
\title{Supplemental Material for the Paper:\\ An Online Learning Algorithm for a Neuro-Fuzzy Classifier with Mixed-Attribute Data}
%
%
%

\author{Thanh Tung Khuat$^{\textsuperscript{\orcidicon{0000-0002-6456-8530}}}$,~\IEEEmembership{Student Member,~IEEE},~and~Bogdan Gabrys$^{\textsuperscript{\orcidicon{0000-0002-0790-2846}}}$,~\IEEEmembership{Senior Member,~IEEE}

}

\markboth{}
{Khuat and Gabrys: An Online Learning Algorithm for a Neuro-Fuzzy Classifier with Mixed-Attribute Data}

\maketitle

%
\IEEEpeerreviewmaketitle

%
%
%
%

\section{Proof of Property 1 from the Main Paper}
\begin{property}
When covering an input pattern, the change of the entropy on each discrete attribute $j$ of $B_i$ obtains its maximum value if and only if that attribute $j$ includes a new categorical value which does not exist in the list of its current categorical values. Formally,
\begin{equation*}
    d_{ij}^{new} = d_{ij}^{old} \cup \{x_j^d: 1\} \Rightarrow Z_j \mapsto \max
\end{equation*}
\end{property}
\begin{proof}
From the main paper, the current entropy of categorical values for the discrete feature $j$ is computed as follows:
\begin{equation*}
    H_j(B_i) = - \sum\limits_{l = 1}^{N_j} \mathsf{P}_j(a_l \in d_{ij}) \cdot \log_2 \mathsf{P}_j(a_l \in d_{ij})
\end{equation*}
We have:
\begin{equation*}
    \mathsf{P}_j(a_l \in d_{ij}) = \cfrac{|\{a \in d_{ij} | a = a_l\}|}{|d_{ij}|} = \cfrac{\varphi(a_l)}{n_i}
\end{equation*}
where $\varphi(a_l)$ is a function returning the number of elements of $a_l$ on the discrete dimension $d_{ij}$ of $B_i$ and $n_i$ is the number of samples included in the hyperbox $B_i$. Therefore,
\begin{equation*}
    H_j(B_i) = - \sum\limits_{l = 1}^{N_j} \cfrac{\varphi(a_l)}{n_i} \cdot \log_2 \cfrac{\varphi(a_l)}{n_i}
\end{equation*}
\textbf{Case 1}: The $j$-$th$ discrete feature includes a new categorical value $x_j^d$ which does not exist in the current list of categorical values of $d_{ij}$.

In this case, the number of categorical values on the discrete attribute $j$ after including $x_j^d$ is $N_j + 1$ and $\varphi(a_l) (l \in [1, N_j]$ is unchanged. The number of samples contained in $B_i$ is $n_i + 1$. As a result, we obtain: $\mathsf{P}_j(x_j^d \in d_{ij}) = \cfrac{1}{n_i + 1}$

Now, the new entropy on the  discrete feature $j$ when including $x_j^d$ is:
\begin{equation*}
\begin{split}
    H_j^{(1)}(B_i \cup \{X\}) &= - \sum\limits_{l = 1}^{N_j} \mathsf{P}_j(a_l \in d_{ij}) \cdot \log_2 \mathsf{P}_j(a_l \in d_{ij}) - \mathsf{P}_j(x_j^d \in d_{ij}) \cdot \log_2 \mathsf{P}_j(x_j^d \in d_{ij}) \\
    &= - \sum\limits_{l = 1}^{N_j} \cfrac{\varphi(a_l)}{n_i + 1} \cdot \log_2 \cfrac{\varphi(a_l)}{n_i + 1} - \cfrac{1}{n_i + 1} \cdot \log_2 \cfrac{1}{n_i + 1}
\end{split}
\end{equation*}
\textbf{Case 2}: The $j$-$th$ discrete feature includes a categorical value $x_j^d$ existed in the current list of categorical values of $d_{ij}$.

In this case, the number of distinct categorical values, $N_j$, storied on each of the $j$-$th$ discrete dimension is unchanged, while the number of samples included in $B_i$ increases by 1. Without loss of generality, we assume that $x_j^d = a_{N_j}$, then $\varphi(a_l) \; (l \in [1, N_j -1]$ is unchanged, while $\varphi(x_j^d) = \varphi^{new}(a_{N_j}) = \varphi^{old}(a_{N_j}) + 1$. Therefore, $\mathsf{P}_j(x_j^d \in d_{ij}) = \cfrac{\varphi^{old}(a_{N_j}) + 1}{n_i + 1}$

The new entropy on the  discrete feature $j$ when including $x_j^d$ is:
\begin{equation*}
\begin{split}
    H_j^{(2)}(B_i \cup \{X\}) &= - \sum\limits_{l = 1}^{N_j - 1} \mathsf{P}_j(a_l \in d_{ij}) \cdot \log_2 \mathsf{P}_j(a_l \in d_{ij}) - \mathsf{P}_j(x_j^d \in d_{ij}) \cdot \log_2 \mathsf{P}_j(x_j^d \in d_{ij}) \\
    &= - \sum\limits_{l = 1}^{N_j - 1} \cfrac{\varphi(a_l)}{n_i + 1} \cdot \log_2 \cfrac{\varphi(a_l)}{n_i + 1} - \cfrac{\varphi^{old}(a_{N_j}) + 1}{n_i + 1} \cdot \log_2 \cfrac{\varphi^{old}(a_{N_j}) + 1}{n_i + 1} \\
    &= - \sum\limits_{l = 1}^{N_j - 1} \cfrac{\varphi(a_l)}{n_i + 1} \cdot \log_2 \cfrac{\varphi(a_l)}{n_i + 1} - \cfrac{\varphi^{old}(a_{N_j})}{n_i + 1} \cdot \log_2 \cfrac{\varphi^{old}(a_{N_j}) + 1}{n_i + 1} - \cfrac{1}{n_i + 1} \cdot \log_2 \cfrac{\varphi^{old}(a_{N_j}) + 1}{n_i + 1}
\end{split}
\end{equation*}
Because $1 \leq \varphi^{old}(a_{N_j})$, we have $-\log_2 \cfrac{\varphi^{old}(a_{N_j}) + 1}{n_i + 1} < -\log_2 \cfrac{\varphi^{old}(a_{N_j})}{n_i + 1}$ and $-\log_2 \cfrac{\varphi^{old}(a_{N_j}) + 1}{n_i + 1} < -\log_2 \cfrac{1}{n_i + 1}$
Hence,
\begin{equation*}
\begin{split}
    H_j^{(2)}(B_j \cup \{X\}) < &-\sum\limits_{l = 1}^{N_j - 1} \cfrac{\varphi(a_l)}{n_i + 1} \cdot \log_2 \cfrac{\varphi(a_l)}{n_i + 1} - \cfrac{\varphi^{old}(a_{N_j})}{n_i + 1} \cdot \log_2 \cfrac{\varphi^{old}(a_{N_j})}{n_i + 1} - \cfrac{1}{n_i + 1} \cdot \log_2 \cfrac{1}{n_i + 1}\\
    &= -\sum\limits_{l = 1}^{N_j} \cfrac{\varphi(a_l)}{n_i + 1} \cdot \log_2 \cfrac{\varphi(a_l)}{n_i + 1} - \cfrac{1}{n_i + 1} \cdot \log_2 \cfrac{1}{n_i + 1} = H_j^{(1)}(B_j \cup \{X\})
\end{split}
\end{equation*}
As a result, the change in the entropy for the hyperbox $B_i$ on the $j$-$th$ discrete attribute:
\begin{equation*}
    Z_j^{(2)} = H_j^{(2)}(B_j \cup \{X\}) - H_j(B_j) < H_j^{(1)}(B_j \cup \{X\}) - H_j(B_j) = Z_j^{(1)}
\end{equation*}
Therefore, the change in the entropy for each discrete dimension of the hyperbox $B_i$ will takes its maximum value if the categorical value on that categorical dimension of the input pattern does not appear in the current list of categorical values stored on that discrete dimension. This property is proved.
\end{proof}

\section{Proof of Property 2 from the Main Paper}
\begin{property}
The upper bound of the change in the entropy for every categorical dimension $j$ of $B_i$ depends on the current number of samples included in $B_i$. That is:
\begin{equation*}
    Z_j \leq \log_2(n_i + 1) - \cfrac{n_i}{n_i + 1} \log_2 n_i
\end{equation*}
\end{property}
\begin{proof}
Based on the Property 1, we know that the maximum entropy change occurs on the $j$-$th$ categorical dimension when it accommodates a new categorical value which does not exist in its current values. Hence, we obtain:
\begin{equation*}
\begin{split}
    Z_j \leq Z_j^{(1)} &= H_j^{(1)}(B_i \cup \{X\}) - \cfrac{n_i}{n_i + 1} H_j(B_i) \\
    &= - \sum\limits_{l = 1}^{N_j} \cfrac{\varphi(a_l)}{n_i + 1} \cdot \log_2 \cfrac{\varphi(a_l)}{n_i + 1} - \cfrac{1}{n_i + 1} \cdot \log_2 \cfrac{1}{n_i + 1} + \cfrac{n_i}{n_i + 1} \cdot \sum\limits_{l = 1}^{N_j} \cfrac{\varphi(a_l)}{n_i} \cdot \log_2 \cfrac{\varphi(a_l)}{n_i}\\
    &= -\cfrac{1}{n_i + 1} \sum\limits_{l = 1}^{N_j} \left[ \varphi(a_l) \cdot [\log_2 \varphi(a_l) - \log_2 (n_i + 1)]\right] + \cfrac{\log_2 (n_i + 1)}{n_i + 1}  + \cfrac{1}{n_i + 1} \cdot \sum\limits_{l = 1}^{N_j} \varphi(a_l) \cdot [\log_2 \varphi(a_l) - \log_2 n_i] \\
    &= \cfrac{\log_2 (n_i + 1)}{n_i + 1} \sum\limits_{l = 1}^{N_j} \varphi(a_l) + \cfrac{\log_2 (n_i + 1)}{n_i + 1} - \cfrac{\log_2 n_i}{n_i + 1} \cdot \sum\limits_{l = 1}^{N_j} \varphi(a_l) \\
    &= \cfrac{\log_2 (n_i + 1)}{n_i + 1}\left[\sum\limits_{l = 1}^{N_j} \varphi(a_l) + 1\right] - \cfrac{\log_2 n_i}{n_i + 1}\cdot\sum\limits_{l = 1}^{N_j} \varphi(a_l)
\end{split}
\end{equation*}

We have: $\sum\limits_{l = 1}^{N_j} \varphi(a_l) = n_i$. Hence,
\begin{equation*}
    Z_j^{(1)} = \cfrac{\log_2 (n_i + 1)}{n_i + 1}(n_i + 1) - \cfrac{\log_2 n_i}{n_i + 1}\cdot n_i = \log_2 (n_i + 1) - \cfrac{n_i}{n_i + 1} \cdot \log_2 n_i
\end{equation*}
As a result,
\begin{equation*}
    Z_j \leq \log_2 (n_i + 1) - \cfrac{n_i}{n_i + 1} \cdot \log_2 n_i
\end{equation*}
The property is proved.
\end{proof}
\section{Proof of Property 3 from the Main Paper}
\begin{property}
The change of the entropy for each categorical dimension $j$ always falls in the range of \textnormal{[0, 1]}: $$0 \leq Z_j \leq 1; \quad \forall{j \in [1, r]}$$
\end{property}
\begin{proof}
First of all, we need to prove that $Z_j \leq 1; \; \forall{j \in [1, r]}$\\
Let $f(n) = \log_2 (n + 1) - \cfrac{n}{n + 1} \cdot \log_2 n; \; (n \geq 1)$. We have the first order derivative of $f(n)$:
\begin{equation*}
    \begin{split}
        f^{'}(n) = \cfrac{1}{(n+1) \ln{2}} - \cfrac{\log_2 n}{(n+1)^2} - \cfrac{1}{(n+1) \ln{2}} = -\cfrac{\log_2 n}{(n+1)^2}
    \end{split}
\end{equation*}
Therefore, $f^{'}(n) \leq 0, \; \forall{n \geq 1}$. As a result, $f(n)$ is decreasing on the interval $[1, +\infty]$. This leads to $f(n) \leq f(1) = 1; \; \forall{n \geq 1}$.
Hence, $\log_2 (n_i + 1) - \cfrac{n_i}{n_i + 1} \cdot \log_2 n_i \leq 1; \; \forall{n_i \geq 1}$. From the Property 2, we obtain $$Z_j \leq \log_2 (n_i + 1) - \cfrac{n_i}{n_i + 1} \cdot \log_2 n_i; \; \forall{j \in [1, r]}$$
As a result, $Z_j \leq 1; \; \forall{j \in [1, r]}$ is proved.

Next, we need to prove that $Z_j \geq 0; \; \forall{j \in [1, r]}$.
From the Property 1, we can see that the change in the entropy on each discrete dimension $j$ takes a smaller value if the included value $x_j^d$ has already been in the current categorical values of $d_{ij}$ in the hyperbox $B_i$. Hence, we only need to prove $Z_j \geq 0$ for this case. We obtain:
\begin{equation*}
    \begin{split}
        Z_j &= H_j^{(2)}(B_i \cup \{X\}) - \cfrac{n_i}{n_i + 1} \cdot H_j(B_i) \\
        &= -\sum\limits_{l=1}^{N_j} \mathsf{P}_j(a_l \in d_{ij} | B_i \cup \{X\}) \cdot \log_2 \mathsf{P}_j(a_l \in d_{ij} | B_i \cup \{X\}) + \cfrac{n_i}{n_i + 1} \cdot \sum\limits_{l=1}^{N_j} \mathsf{P}_j(a_l \in d_{ij} | B_i) \cdot \log_2 \mathsf{P}_j(a_l \in d_{ij} | B_i) \\
        &= -\sum\limits_{l=1}^{N_j} \cfrac{\varphi(a_l | B_i \cup \{X\})}{n_i + 1} \cdot \log_2 \cfrac{\varphi(a_l | B_i \cup \{X\})}{n_i + 1} + \cfrac{n_i}{n_i + 1} \cdot \sum\limits_{l=1}^{N_j} \cfrac{\varphi(a_l | B_i)}{n_i} \cdot \log_2 \cfrac{\varphi(a_l | B_i)}{n_i}
    \end{split}
\end{equation*}
To prove $Z_j \geq 0$, we need to prove that:
\begin{equation}\label{proof_zj}
\begin{split}
    &n_i \cdot \cfrac{\varphi(a_l | B_i)}{n_i} \cdot \log_2 \cfrac{\varphi(a_l | B_i)}{n_i} \geq (n_i + 1) \cdot \cfrac{\varphi(a_l | B_i \cup \{X\})}{n_i + 1} \cdot \log_2 \cfrac{\varphi(a_l | B_i \cup \{X\} )}{n_i + 1}, \forall{l \in [1, N_j]} \\
    &\Leftrightarrow \varphi(a_l | B_i) \cdot \log_2 \cfrac{\varphi(a_l | B_i)}{n_i} \geq \varphi(a_l | B_i \cup \{X\}) \cdot \log_2 \cfrac{\varphi(a_l | B_i \cup \{X\} )}{n_i + 1}, \forall{l \in [1, N_j]}\\
\end{split}
\end{equation}
Let $x = \varphi(a_l | B_i) \geq 1$, then $\varphi(a_l | B_i \cup \{X\}) = x + 1$ or $\varphi(a_l | B_i \cup \{X\}) = x$ because the number of elements of $a_l$ on the $j$-$th$ discrete feature can remain unchanged or increase by 1 when $B_i$ includes $X$. In the case that $\varphi(a_l | B_i \cup \{X\}) = x$, we have:
$$ \eqref{proof_zj} \Leftrightarrow x \cdot \log_2 \cfrac{x}{n_i} \geq x \cdot \log_2 \cfrac{x}{n_i + 1}$$
This inequality is obviously true because $x \geq 1$ and $\cfrac{x}{n_i} \geq \cfrac{x}{n_i + 1} \Leftrightarrow \log_2 \cfrac{x}{n_i} \geq \log_2 \cfrac{x}{n_i + 1} $.\\
In the case that $\varphi(a_l | B_i \cup \{X\}) = x + 1$, we need to prove that:
\begin{equation} \label{proof_zj_2}
    x \cdot \log_2 \cfrac{x}{n_i} \geq (x+1) \cdot \log_2 \cfrac{x+1}{n_i + 1}
\end{equation}
Let $a_1,\ldots, a_m$ and $b_1,\ldots, b_m$ be non-negative numbers, and $a = \sum\limits_{k = 1}^m a_k$, $b = \sum\limits_{k = 1}^m b_k$, the log-sum inequality states that
\begin{equation*}
    \sum\limits_{k = 1}^m a_i \cdot \log_2 \cfrac{a_i}{b_i} \geq a \cdot \log_2 \cfrac{a}{b}
\end{equation*}
Based on this log-sum inequality, we obtain:
\begin{equation*}
    x \cdot \log_2 \cfrac{x}{n_i} + 1 \cdot \log_2 \cfrac{1}{1} \geq (x+1) \cdot \log_2 \cfrac{x+1}{n_i + 1}
\end{equation*}
We have $1 \cdot \log_2 \cfrac{1}{1} = 0$, thus the inequality \eqref{proof_zj_2} is true.

From all above proofs, we obtain: $0 \leq Z_j \leq 1; \; \forall{j \in [1, r]}$. The property is proved.
\end{proof}

\section{Proof of Property 4 from the Main Paper}
\begin{property}\label{prop4}
When the number of samples contained in $B_i$ approaches infinity, the change of the entropy for every categorical dimension will be limited at 0. Formally,
\begin{equation}
    \lim \limits_{n_i \rightarrow +\infty} Z_j = 0, \; \forall{j \in [1, r]}
\end{equation}
\end{property}
\begin{proof}
From the properties 2 and 3, we have:
$$0 \leq Z_j \leq \log_2 (n_i + 1) - \cfrac{n_i}{n_i + 1} \cdot \log_2 n_i$$
To prove this property, therefore, we need to show that:
$$\lim \limits_{n_i \rightarrow +\infty} \log_2 (n_i + 1) - \cfrac{n_i}{n_i + 1} \cdot \log_2 n_i = 0$$
Let $f(n) = \log_2 (n + 1) - \cfrac{n}{n + 1} \cdot \log_2 n = \cfrac{(n+1) \cdot \log_2 (n+1) - n \cdot \log_2 n}{n + 1}$, then $\lim \limits_{n \rightarrow +\infty} f(n)$ is the indeterminate form $\cfrac{\infty}{\infty}$. Let's apply L'Hospital's rule:
\begin{equation*}
\begin{split}
    \lim \limits_{n \rightarrow +\infty} f(n) &= \lim \limits_{n \rightarrow +\infty} \cfrac{(n+1) \cdot \log_2 (n+1) - n \cdot \log_2 n}{n + 1} \\
    &= \lim \limits_{n \rightarrow +\infty} \left[\log_2 (n+1) + \cfrac{1}{\ln{2}} - \log_2 n - \cfrac{1}{\ln{2}} \right] = \lim \limits_{n \rightarrow +\infty} \log_2 (1 + \cfrac{1}{n}) = 0
\end{split}
\end{equation*}
The property is proved.
\end{proof}

\section{Additional Results}
\subsection{Datasets}
Table \ref{table_S1} shows the detailed information about the datasets used for the experiments in this study. These datasets were used in the previous research regarding different solutions to learn from mixed-attribute data for the general fuzzy min-max neural network. These datasets show the diversity in the numbers of samples, classes, and features. All datasets are class-imbalanced.

\begin{table}[!ht]
\scriptsize{
\centering
\caption{Experimental Dataset} \label{table_S1}
\begin{tabular}{cL{2cm}C{1.1cm}C{1.1cm}C{2.5cm}C{2.2cm}L{6.6cm}}
\hline
ID & Dataset            & \# Patterns & \# Classes & \# Continuous attributes & \# Discrete attributes & Summary of  discrete attributes                                                                                               \\ \hline
1  & abalone            & 4177       & 28         & 7                     & 1                       & 1 feature: 3 values\\                                                  
2  & australian         & 690        & 2          & 6                     & 8                       & 4 features: 2 values, 2 features: 3 values, 1 feature: 15   values, 1 feature: 8 values \\                                         
3  & cmc                & 1473       & 3          & 2                     & 7                       & 2 features: 4 values, 3 features: 2 values, 2 features: 4   values\\
4  & dermatology        & 358        & 6          & 1                     & 33                      & 1 feature: 2 values, 1 feature: 3 values, 31 features: 4   values \\     
5  & flag               & 194        & 8          & 10                    & 18                      & 1 feature: 4 values, 1 feature: 6 values, 1 feature: 7 values,   1 feature: 10 values, 2 features: 8 values, 12 features: 2 values\\
6  & german             & 1000       & 2          & 7                     & 13                      & 4 features: 4 values, 3 features: 5 values, 1 feature: 10   values, 3 features: 3 values, 2 features: 2 values   \\                
7  & heart              & 270        & 2          & 7                     & 6                       & 3 features: 2 values, 1 feature: 4 values, 2 features: 3   values\\
8  & japanese credit   & 653        & 2          & 6                     & 9                       & 4 features: 2 values, 3 features: 3 values, 1 feature: 14   values, 1 feature: 9 values\\
9  & molecular biology & 3190       & 3          & 0                     & 60                      & All features: 4 values\\
10 & nursery            & 12960      & 5          & 0                     & 8                       & 4 features: 3 values, 1 feature: 5 values, 2 features: 4   values, 1 feature: 2 values\\
11 & post operative    & 87         & 3          & 1                     & 7                       & 5 features: 3 values, 2 features: 2 values\\
12 & tae                & 151        & 3          & 1                     & 4                       & 1 feature: 2 values, 1 feature: 25 values, 1 feature: 26   values, 1 feature: 2 values\\
13 & tic tac toe      & 958        & 2          & 0                     & 9                       & All features: 3 values\\
14 & zoo                & 101        & 7          & 1                     & 15                      & All features: 2 values\\ \hline
\end{tabular}
}
\end{table}

\subsection{Additional Experimental Results for the Main Paper}
This section provides the interested readers with the detailed results for the experiments presented in the main paper.

\begin{figure}[!ht]
	\begin{subfloat}[abalone]{
			\includegraphics[width=0.32\textwidth, height=0.18\textheight]{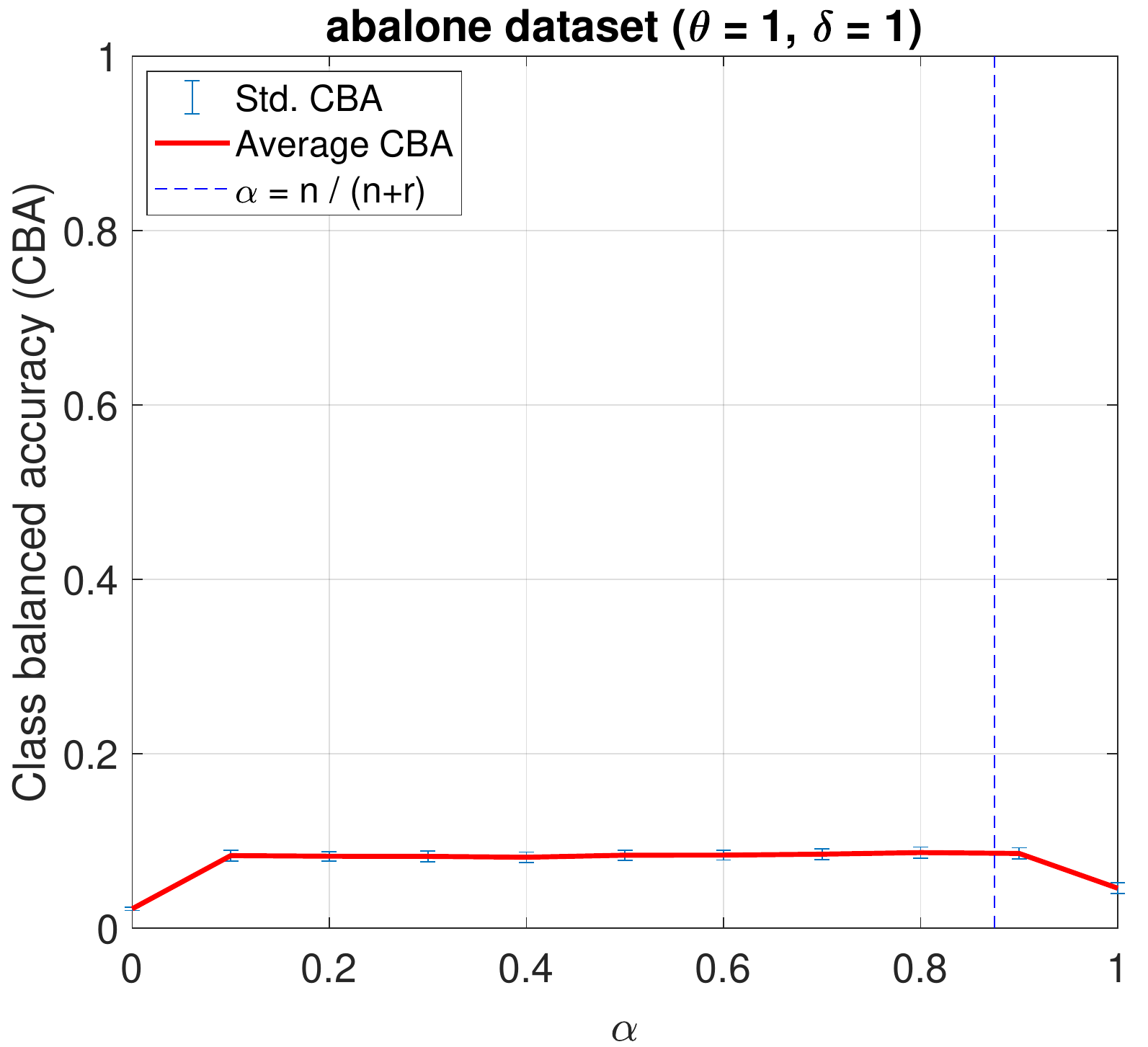}}
	\end{subfloat}
	\hfill%
	\begin{subfloat}[australian]{
			\includegraphics[width=0.32\textwidth,height=0.18\textheight]{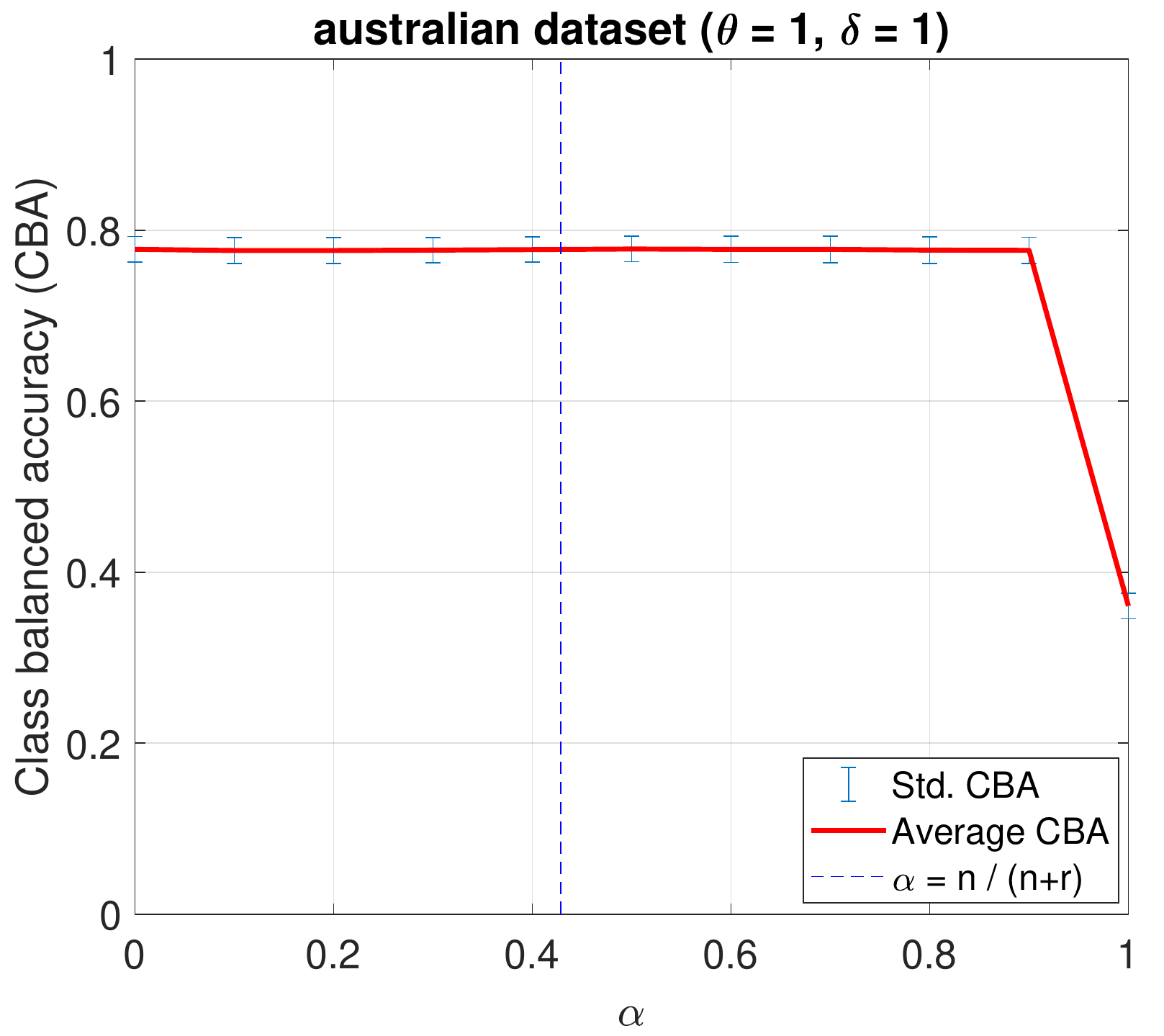}}
	\end{subfloat}
	\hfill%
	\begin{subfloat}[cmc]{
			\includegraphics[width=0.32\textwidth,height=0.18\textheight]{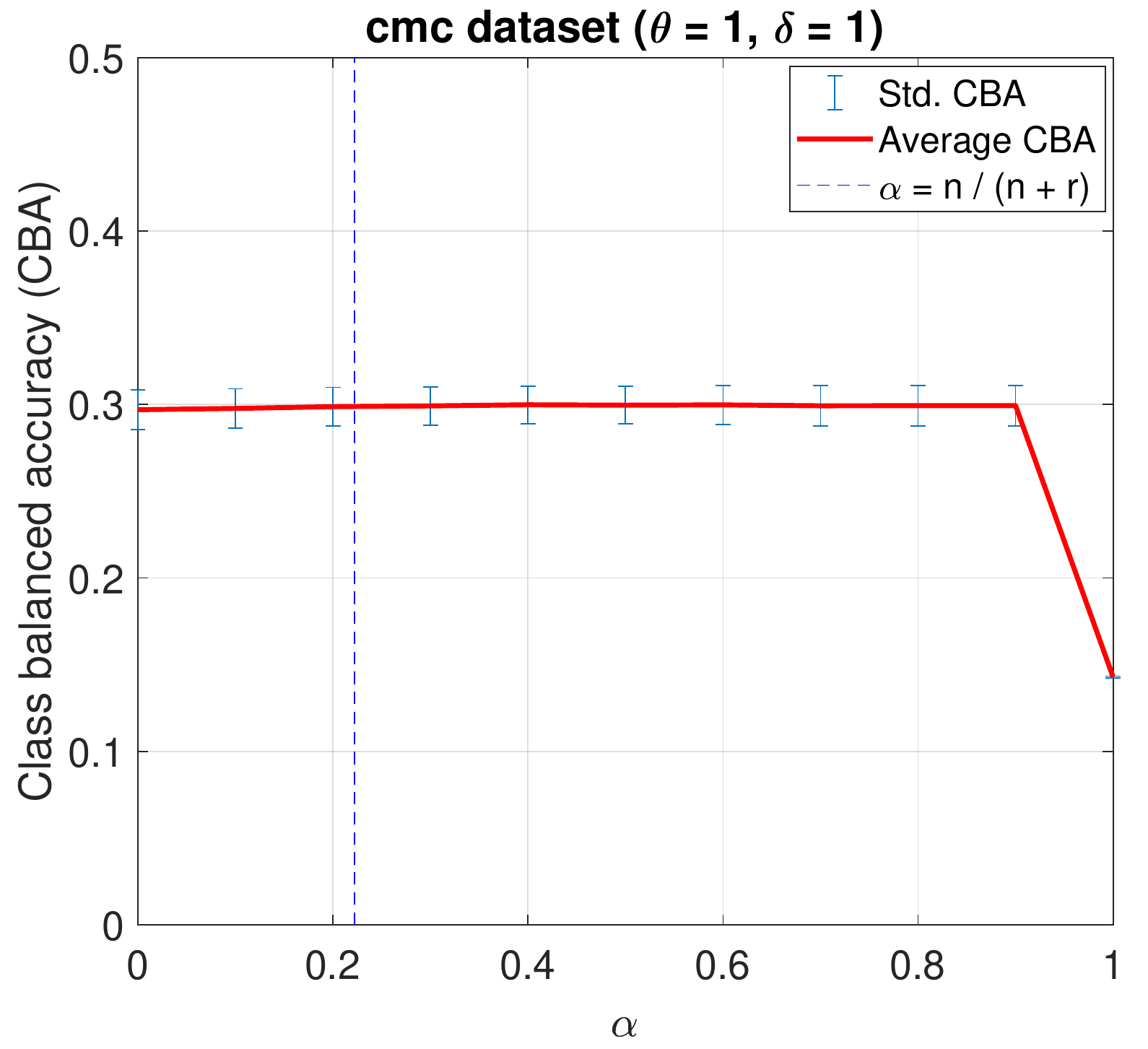}}
	\end{subfloat}
	\hfill%
	\begin{subfloat}[dermatology]{
			\includegraphics[width=0.32\textwidth,height=0.18\textheight]{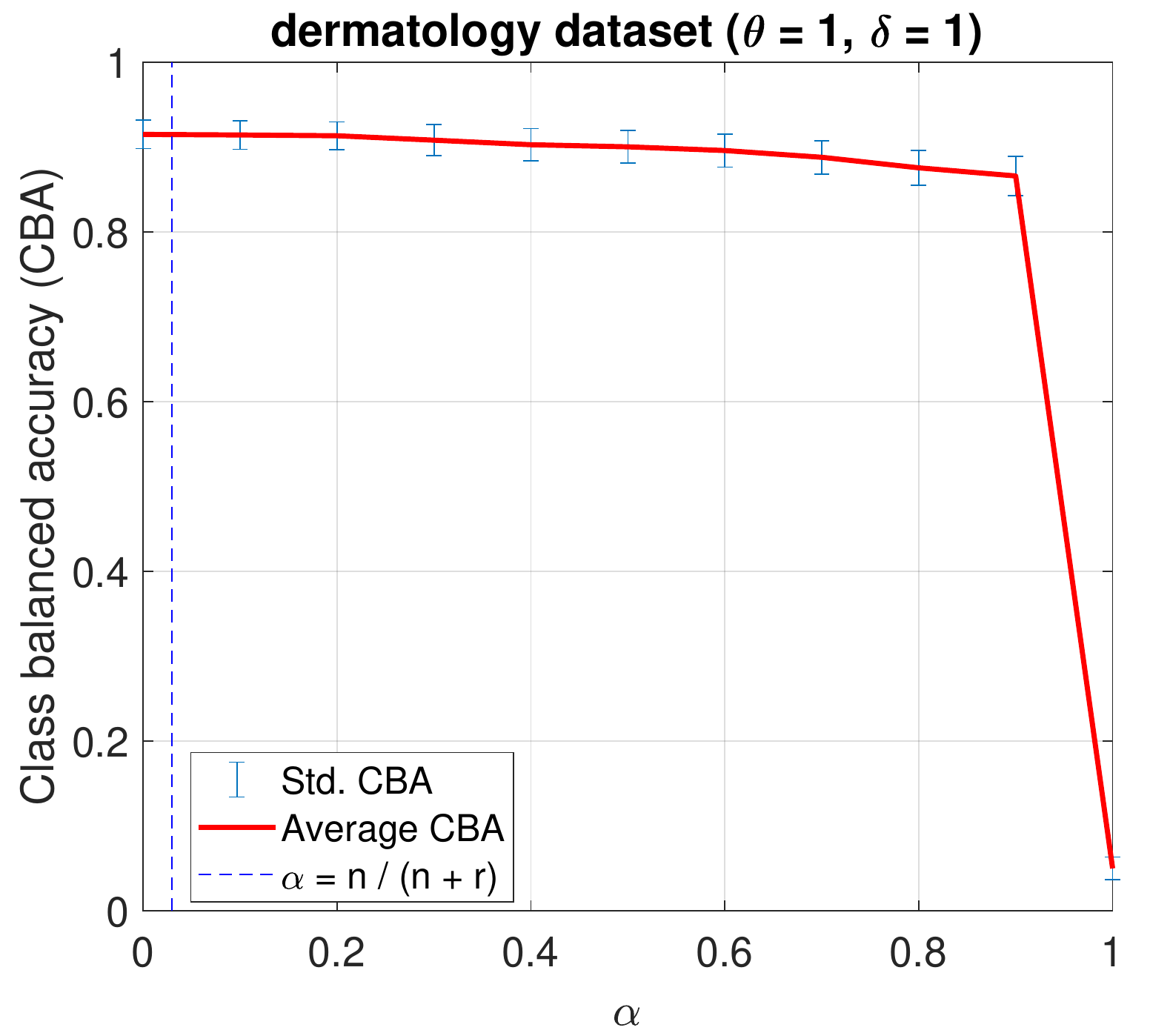}}
	\end{subfloat}
	\hfill%
	\begin{subfloat}[flag]{
			\includegraphics[width=0.32\textwidth,height=0.18\textheight]{flag_changed_alpha_11.pdf}}
	\end{subfloat}
	\hfill%
	\begin{subfloat}[german]{
			\includegraphics[width=0.32\textwidth,height=0.18\textheight]{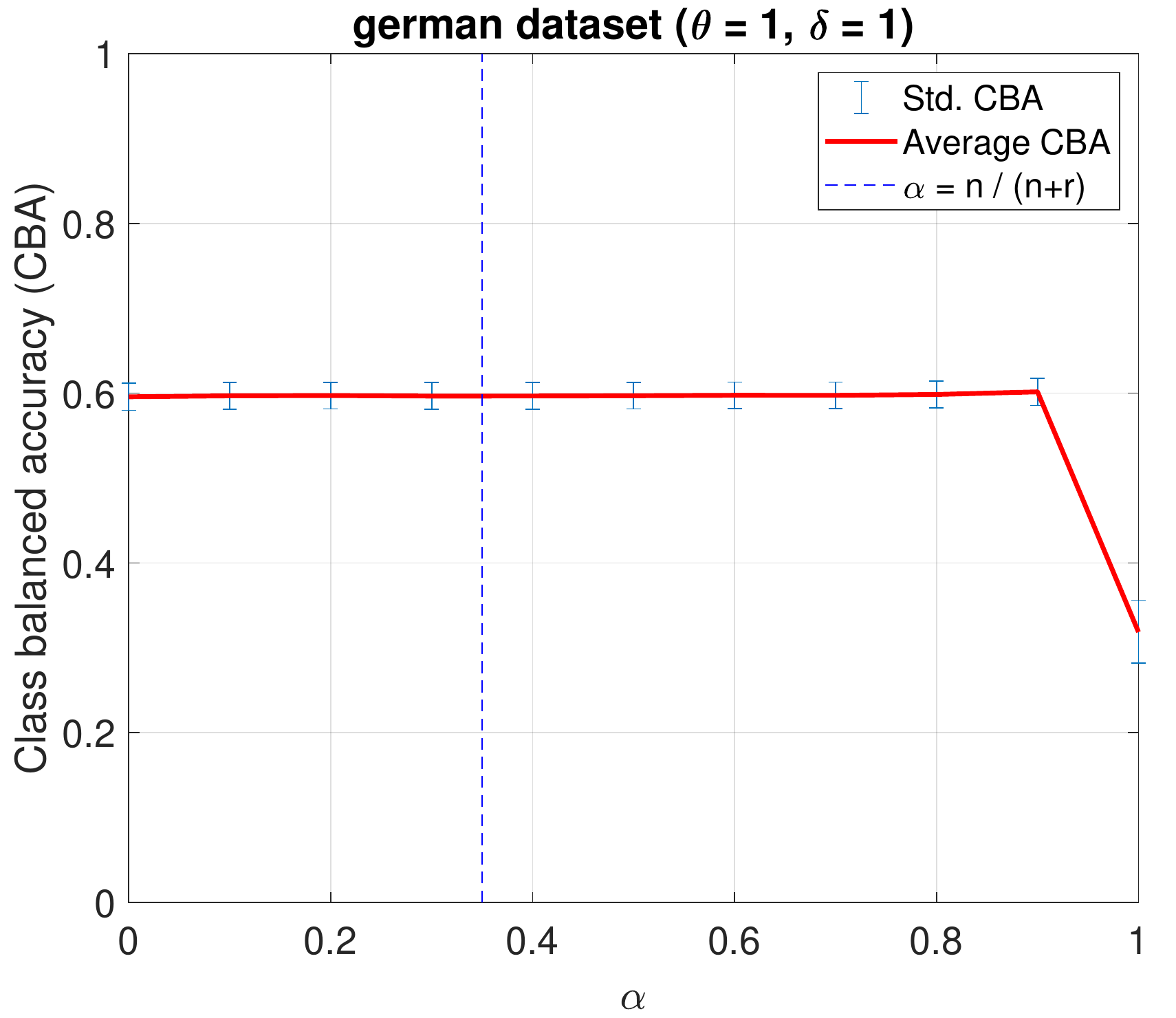}}
	\end{subfloat}
	\hfill
	\begin{subfloat}[heart]{
			\includegraphics[width=0.32\textwidth,height=0.18\textheight]{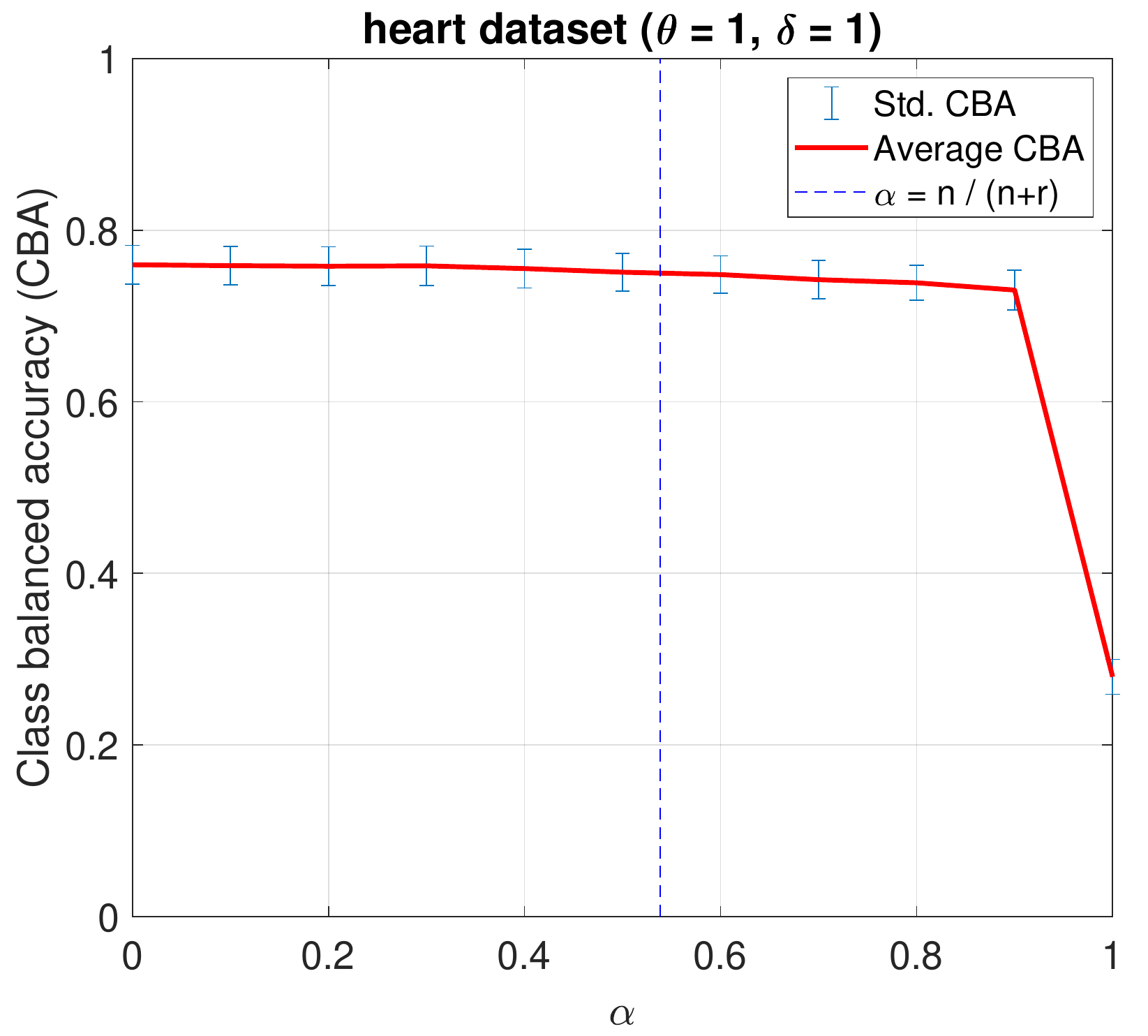}}
	\end{subfloat}
	\hfill
	\begin{subfloat}[japanese credit]{
			\includegraphics[width=0.32\textwidth,height=0.18\textheight]{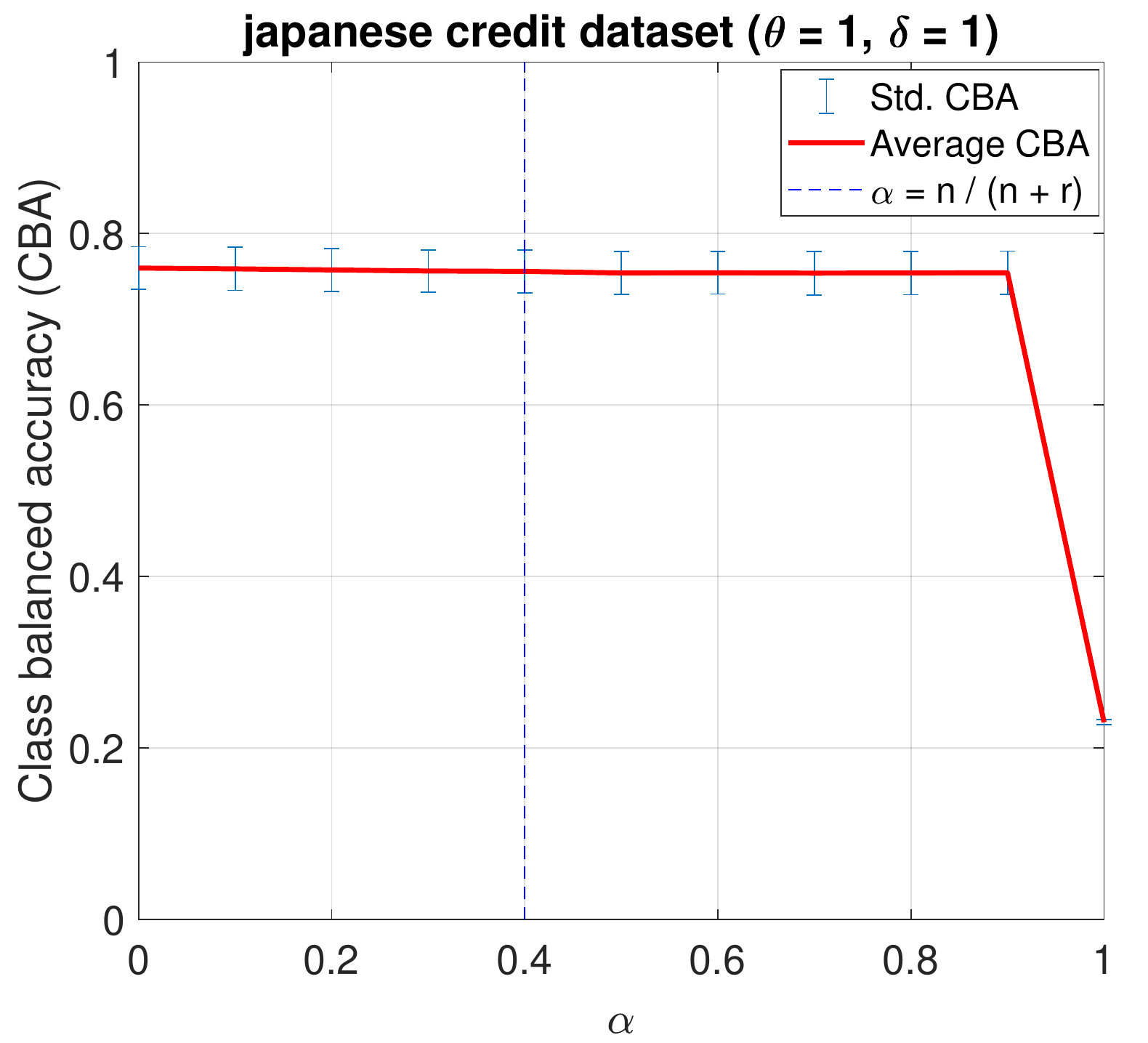}}
	\end{subfloat}
	\hfill
	\begin{subfloat}[post operative]{
			\includegraphics[width=0.32\textwidth,height=0.18\textheight]{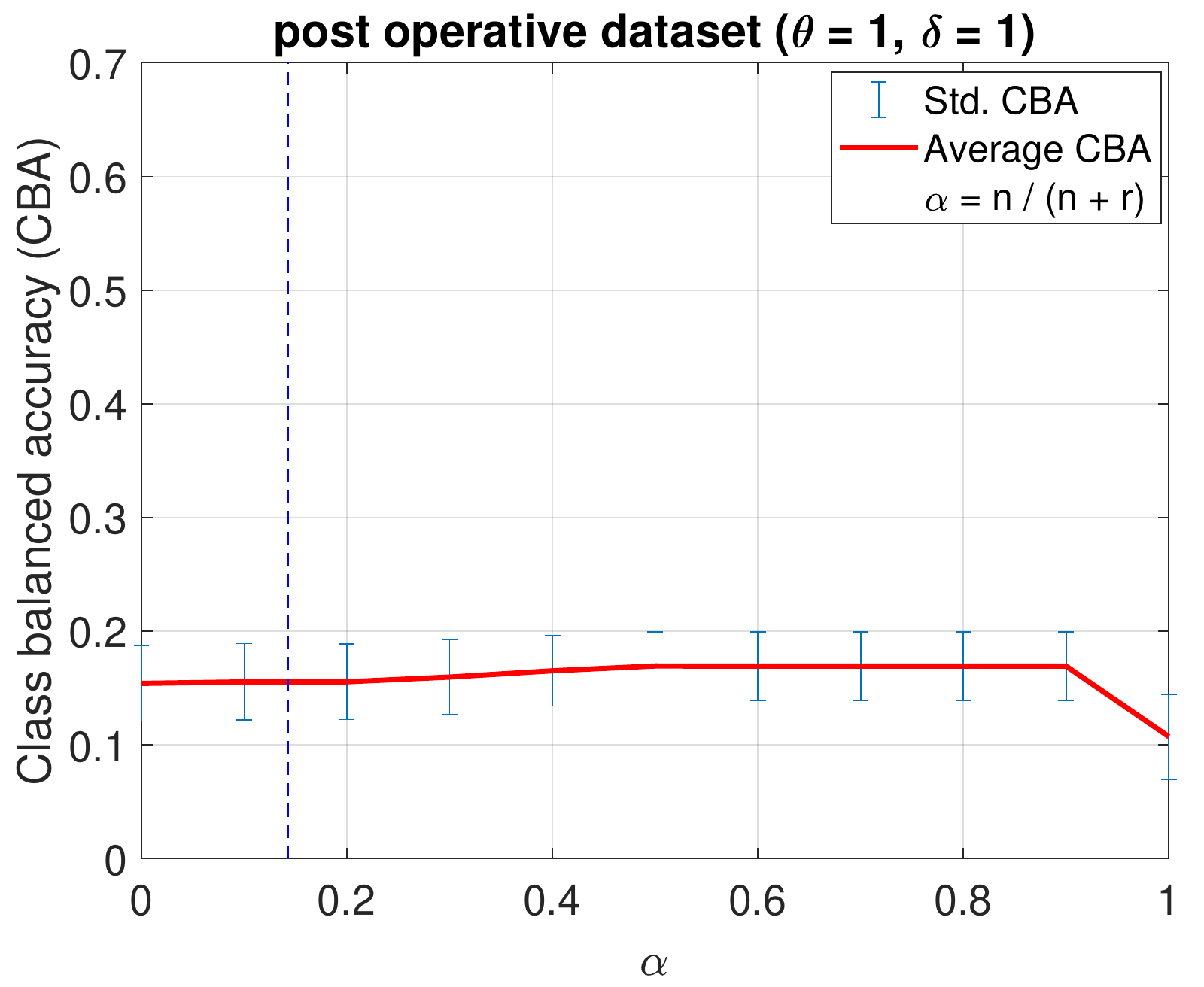}}
	\end{subfloat}
	\hfill
	\begin{subfloat}[tae]{
			\includegraphics[width=0.32\textwidth,height=0.18\textheight]{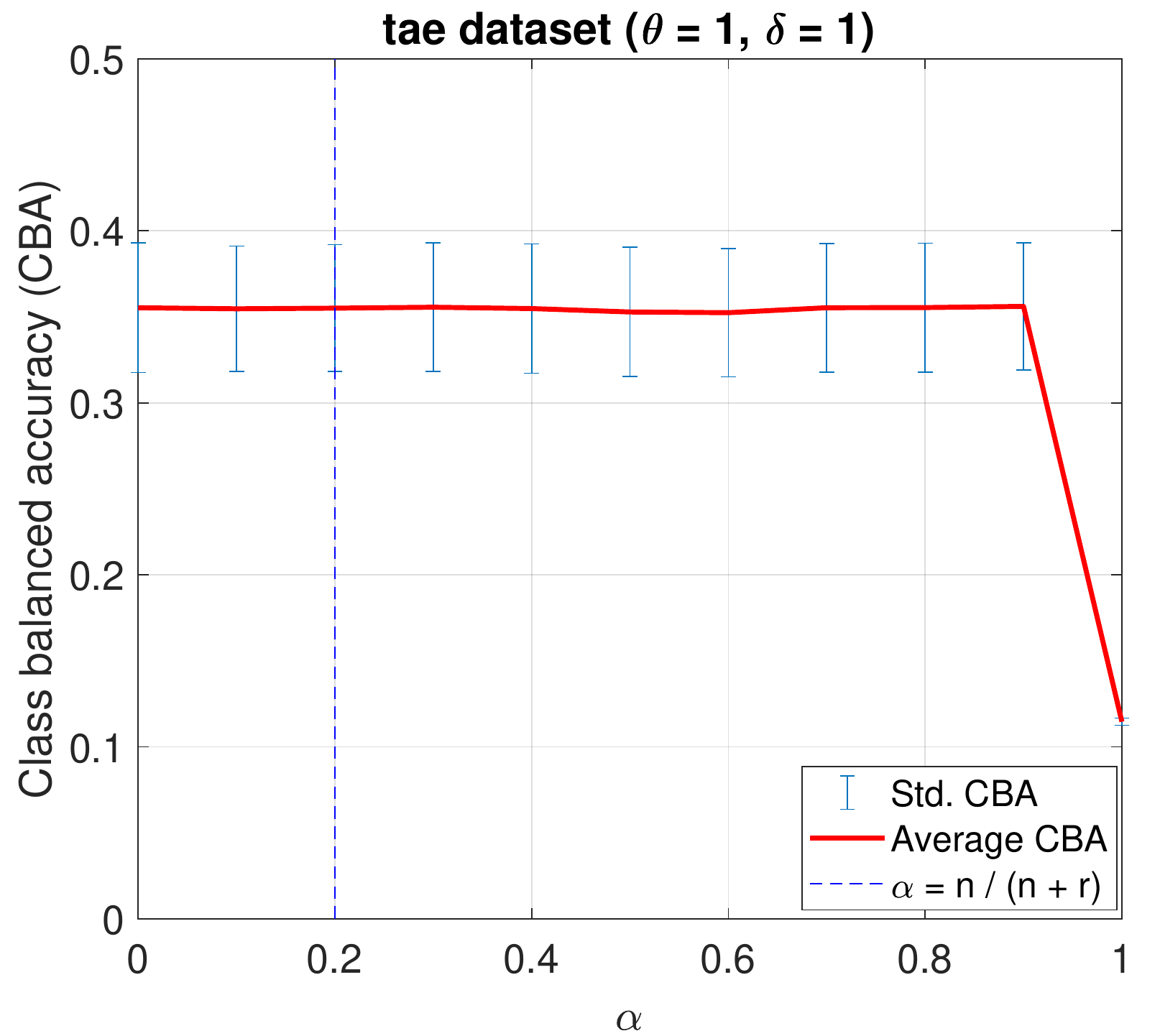}}
	\end{subfloat}
	\begin{subfloat}[zoo]{
			\includegraphics[width=0.32\textwidth,height=0.18\textheight]{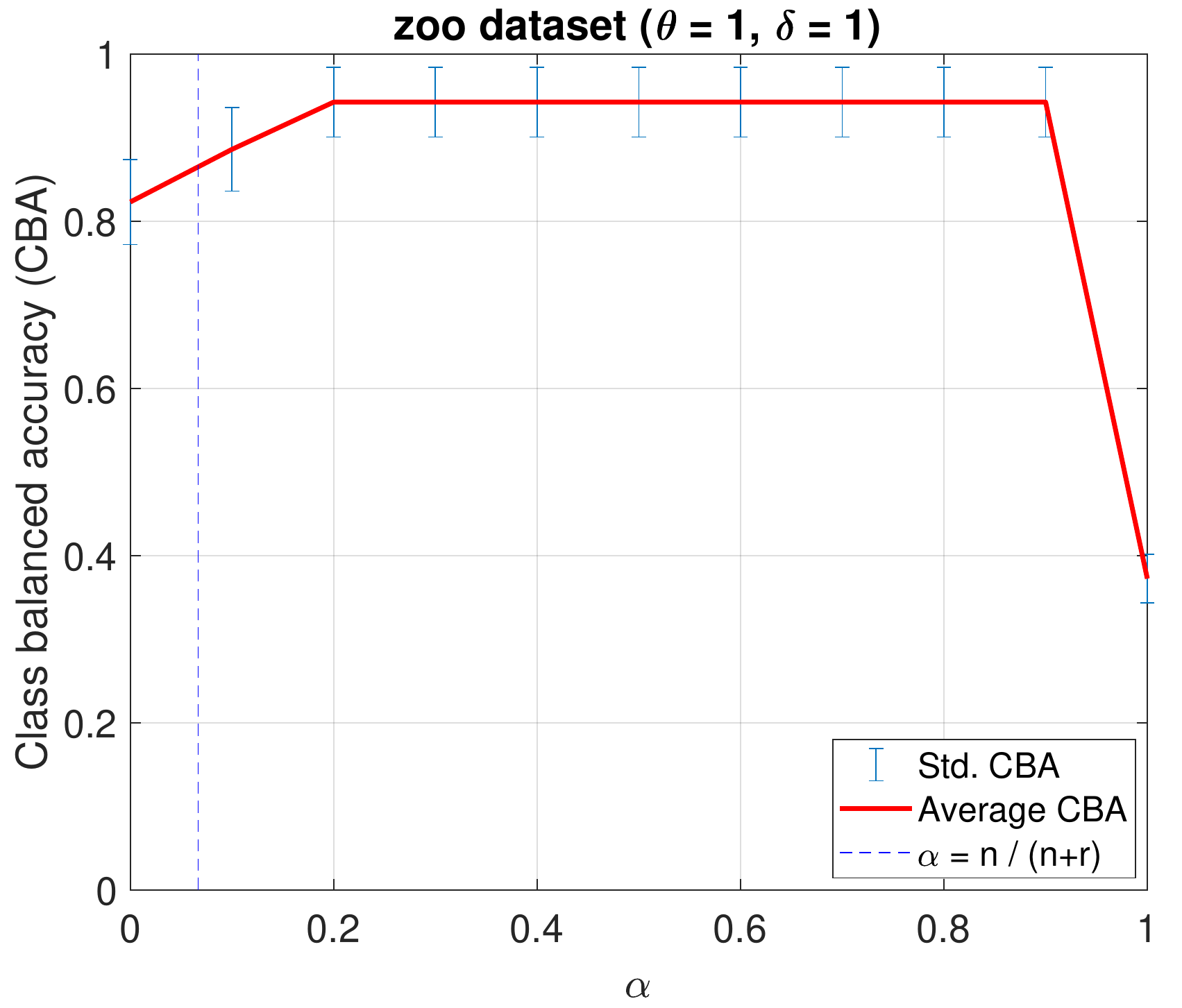}}
	\end{subfloat}
	\caption{Class balanced accuracy values for different values of $\alpha$ ($\theta = 1, \delta = 1$).}
	\label{Fig_S1}
\end{figure}

\begin{figure}[!ht]
	\begin{subfloat}[abalone]{
			\includegraphics[width=0.32\textwidth, height=0.18\textheight]{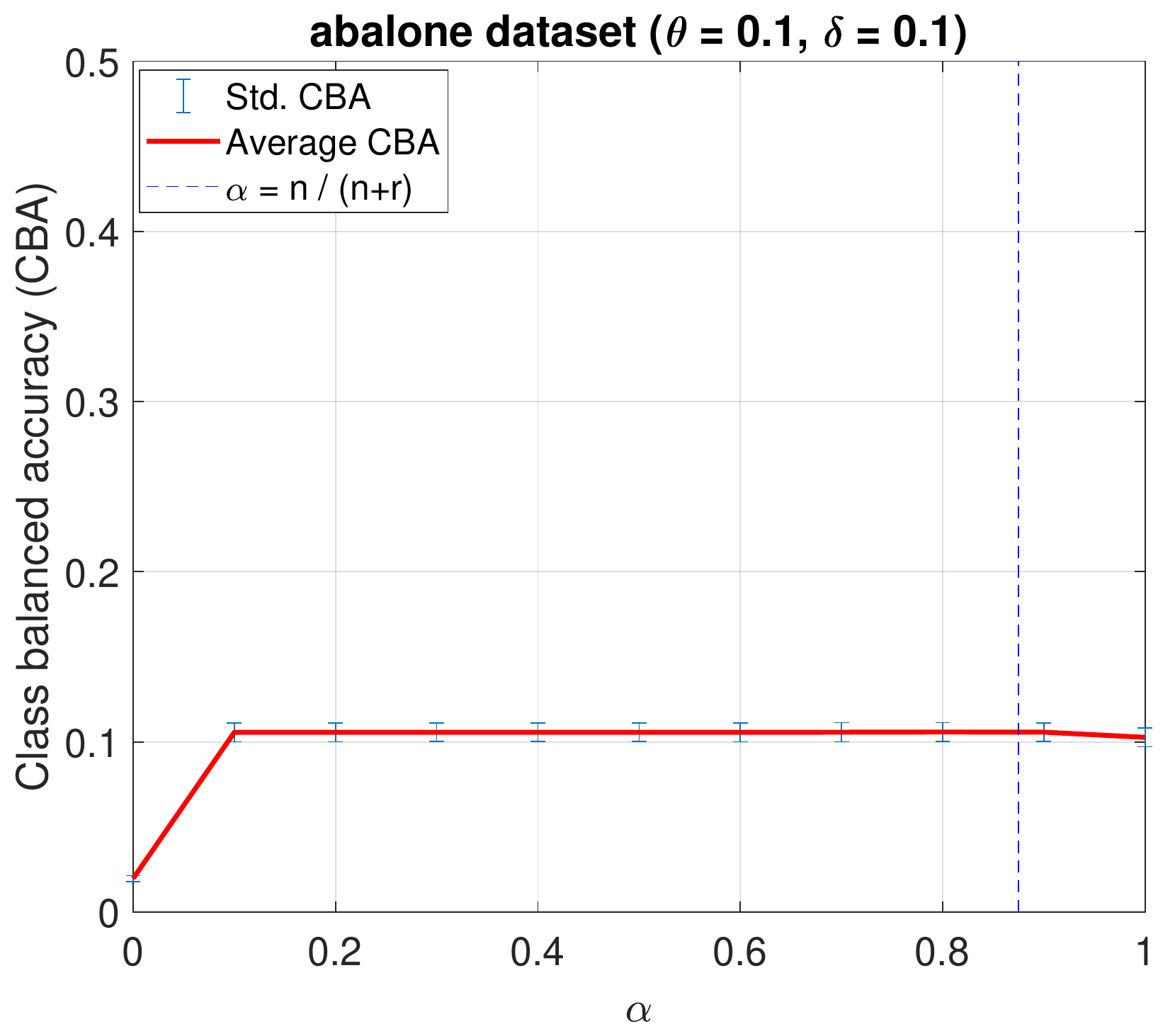}}
	\end{subfloat}
	\hfill%
	\begin{subfloat}[australian]{
			\includegraphics[width=0.32\textwidth,height=0.18\textheight]{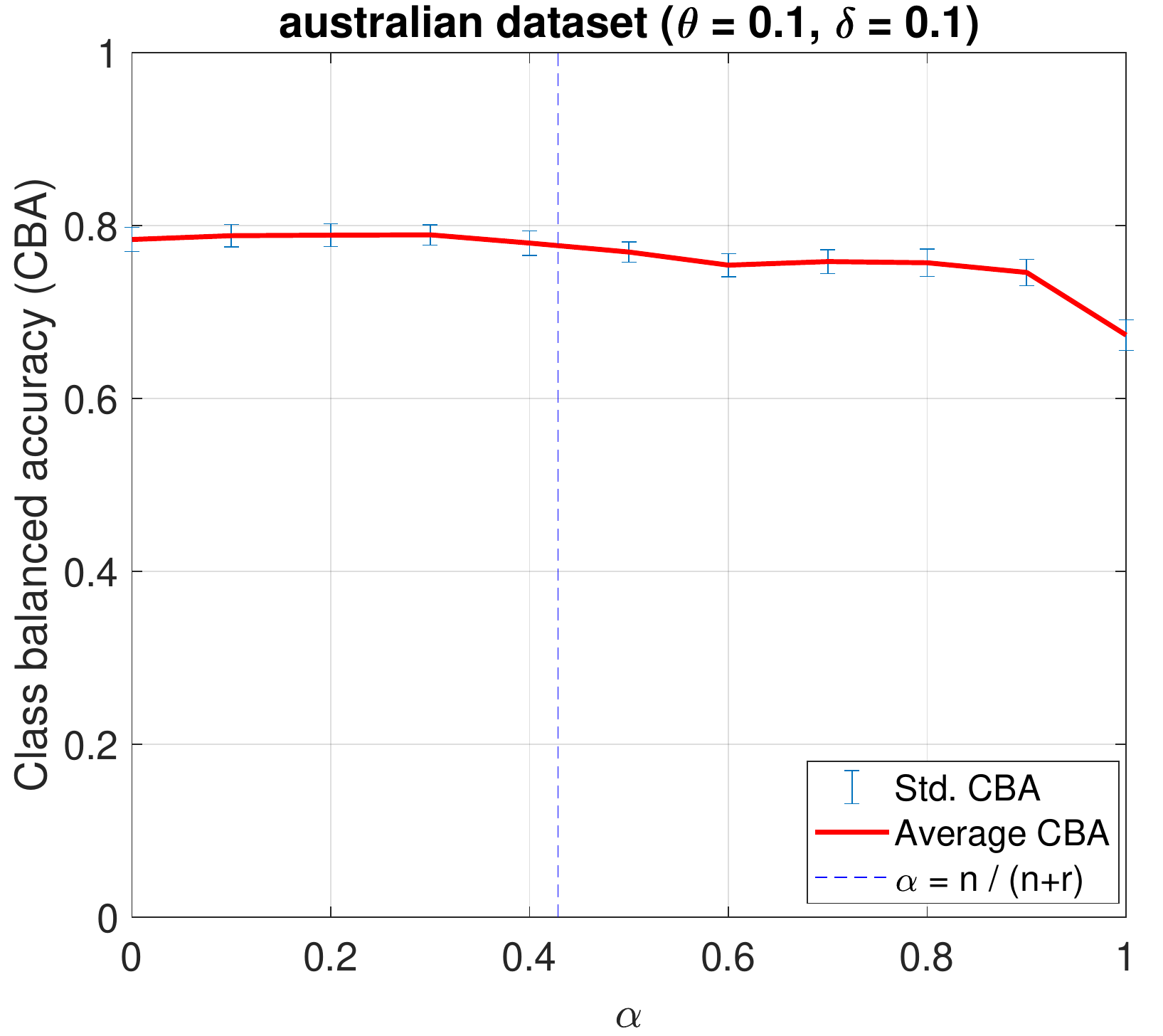}}
	\end{subfloat}
	\hfill%
	\begin{subfloat}[cmc]{
			\includegraphics[width=0.32\textwidth,height=0.18\textheight]{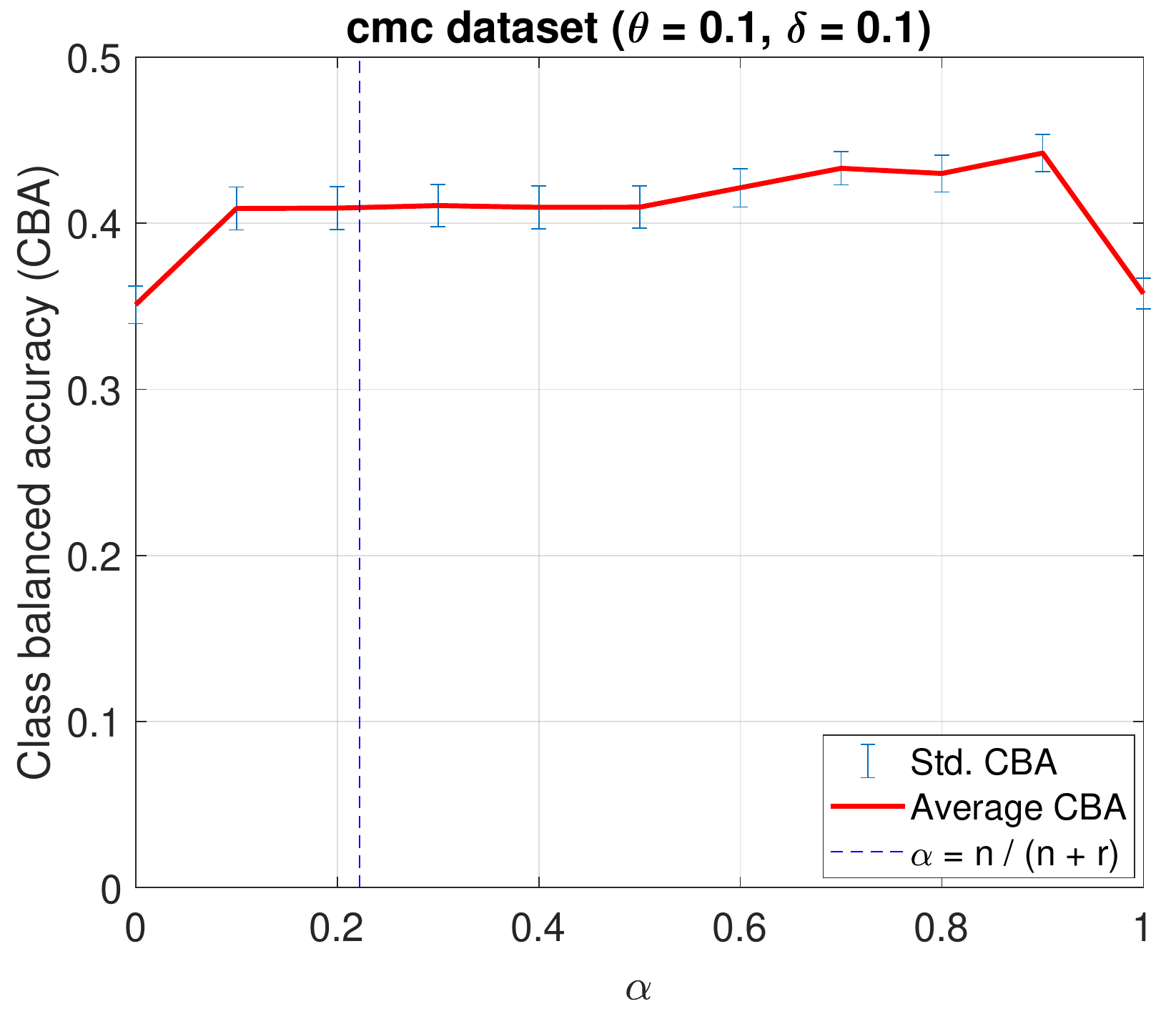}}
	\end{subfloat}
	\hfill%
	\begin{subfloat}[dermatology]{
			\includegraphics[width=0.32\textwidth,height=0.18\textheight]{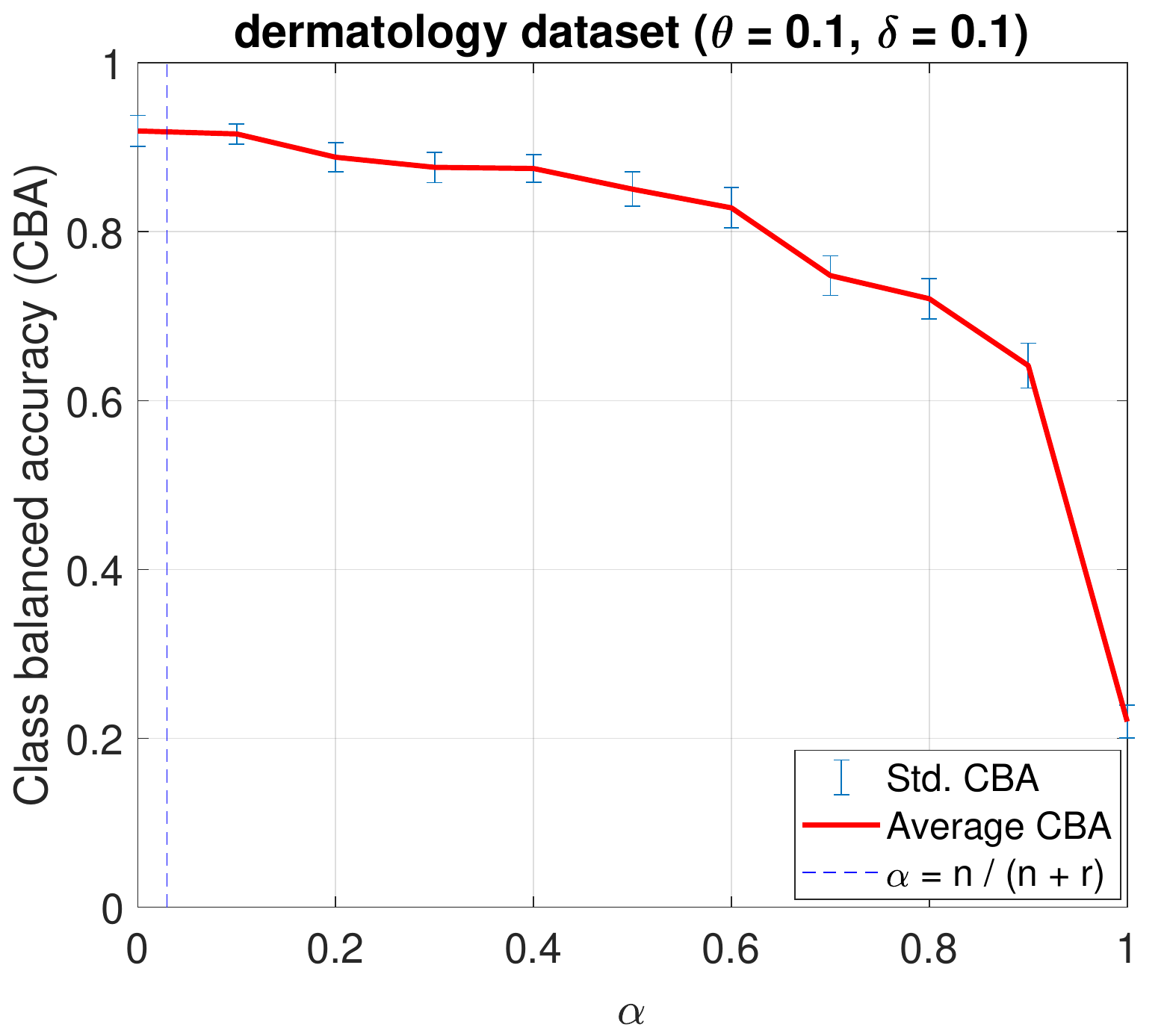}}
	\end{subfloat}
	\hfill%
	\begin{subfloat}[flag]{
			\includegraphics[width=0.32\textwidth,height=0.18\textheight]{flag_changed_alpha_0101_v1.pdf}}
	\end{subfloat}
	\hfill%
	\begin{subfloat}[german]{
			\includegraphics[width=0.32\textwidth,height=0.18\textheight]{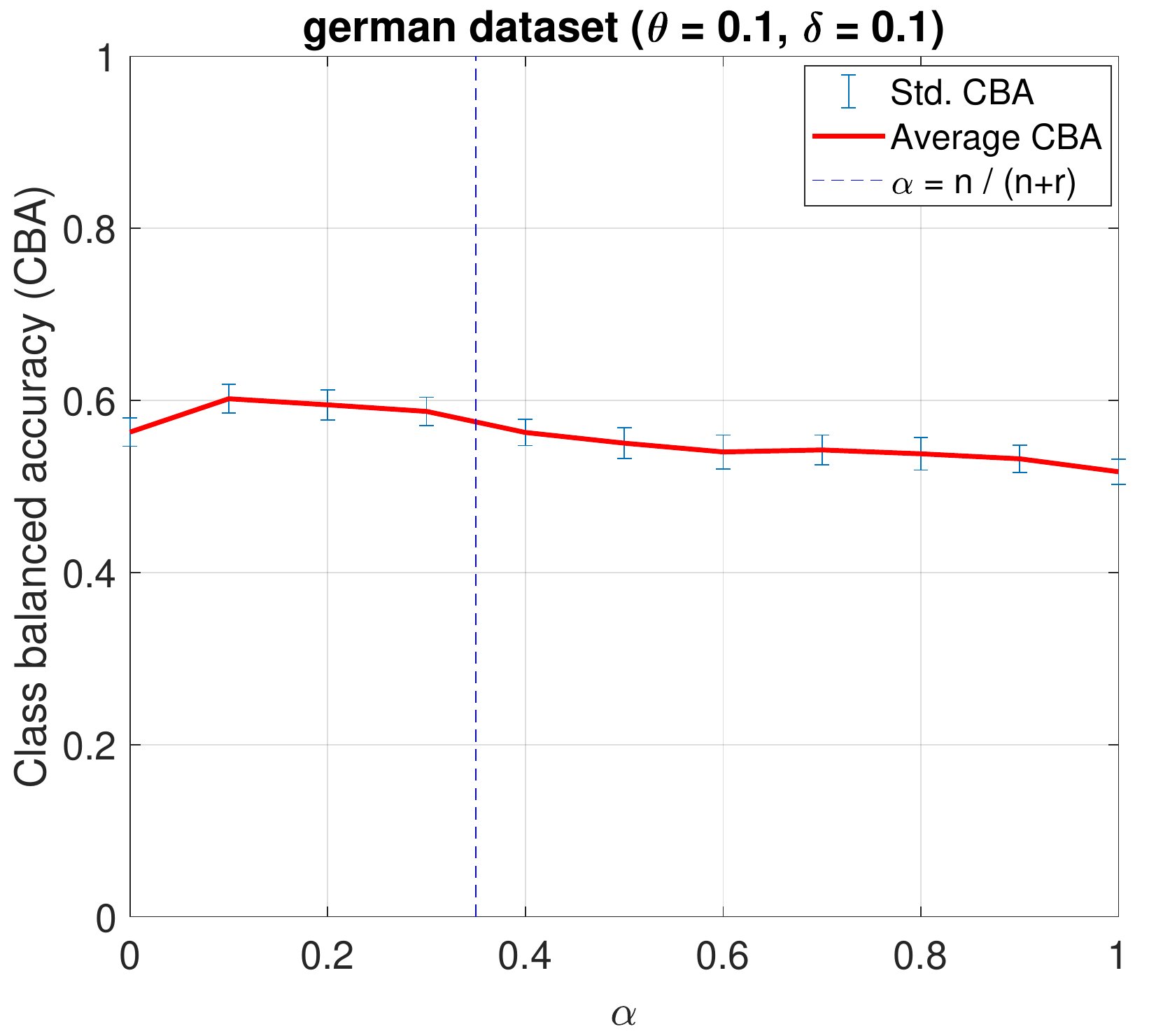}}
	\end{subfloat}
	\hfill
	\begin{subfloat}[heart]{
			\includegraphics[width=0.32\textwidth,height=0.18\textheight]{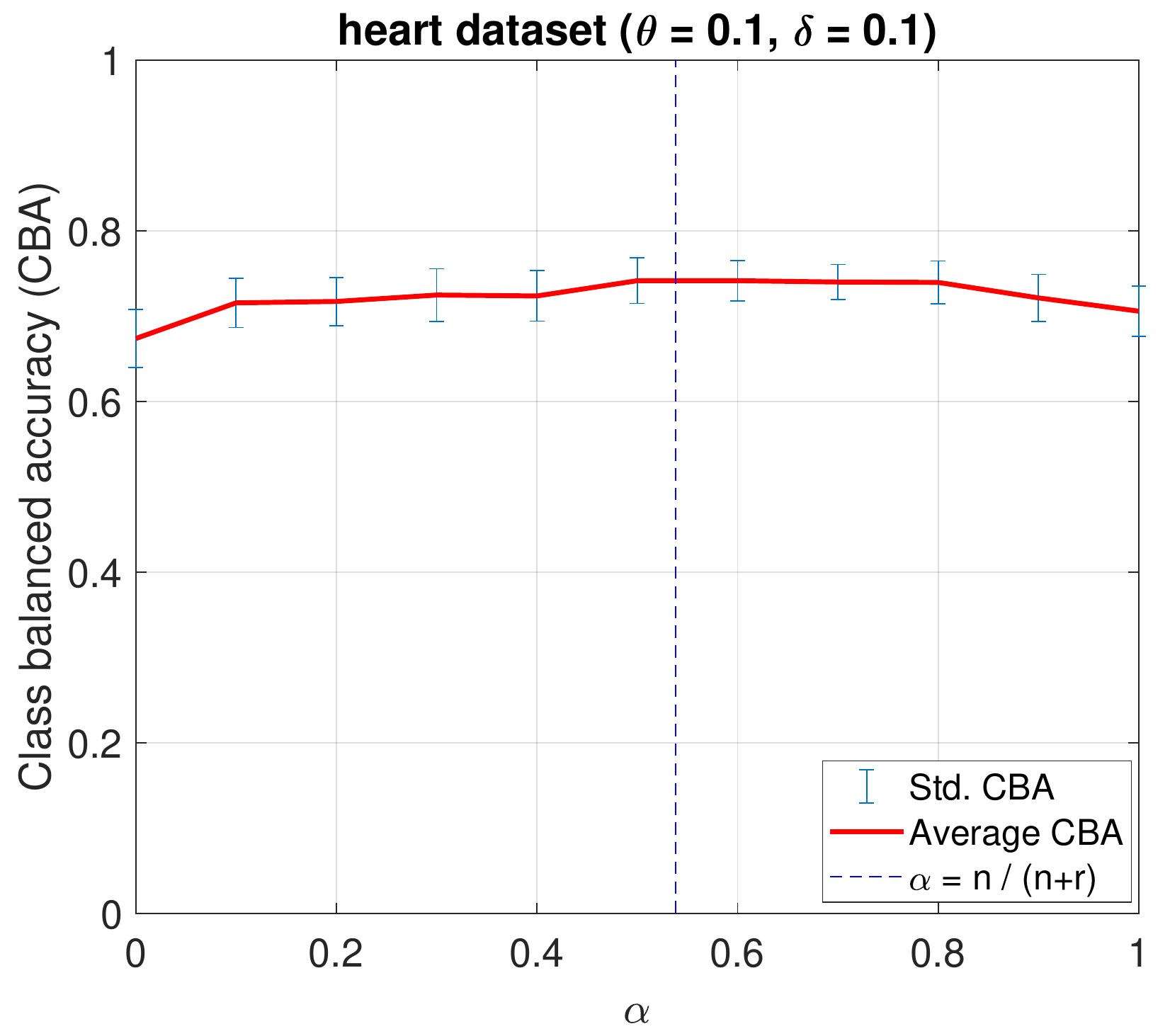}}
	\end{subfloat}
	\hfill
	\begin{subfloat}[japanese credit]{
			\includegraphics[width=0.32\textwidth,height=0.18\textheight]{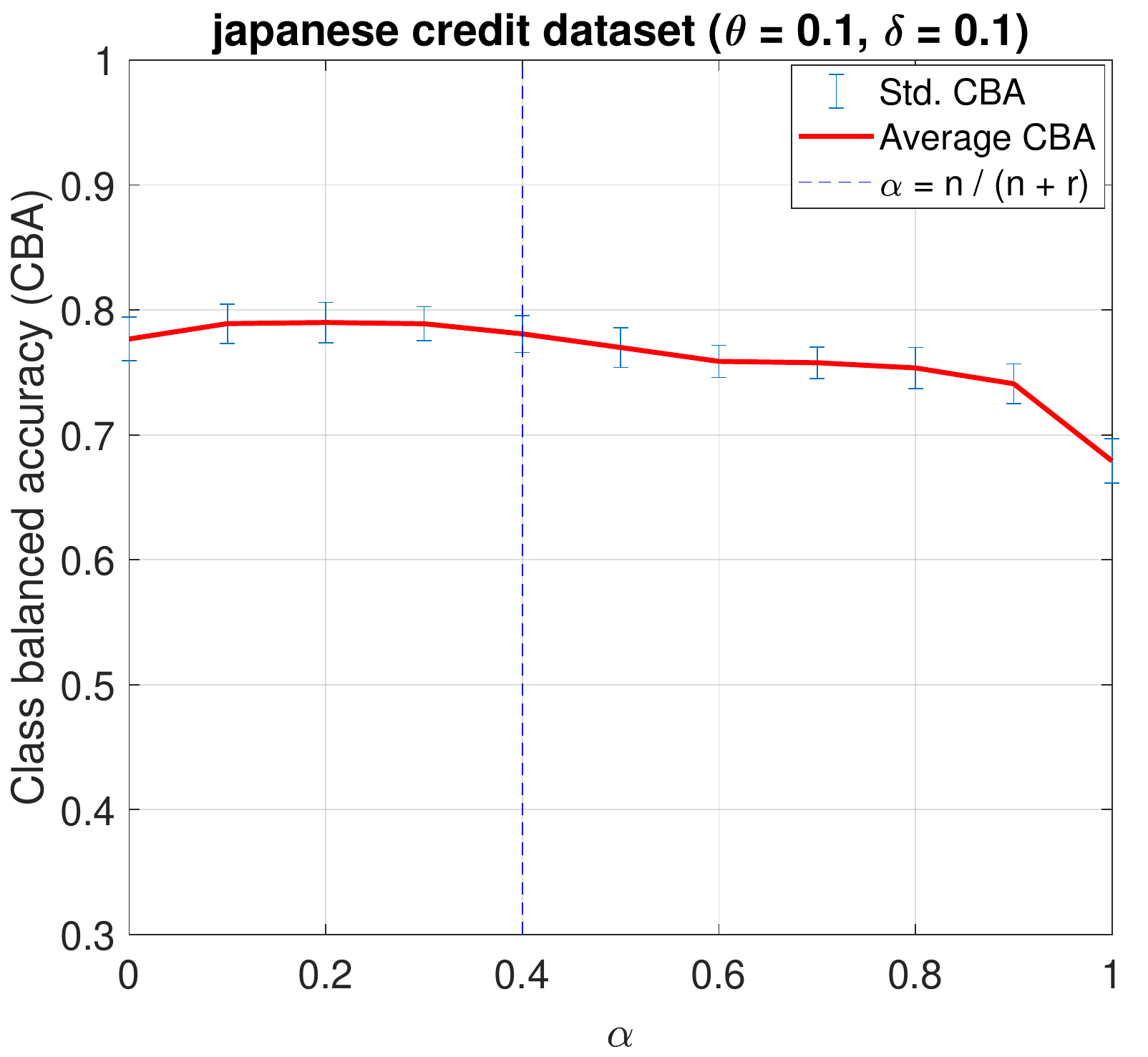}}
	\end{subfloat}
	\hfill
	\begin{subfloat}[post operative]{
			\includegraphics[width=0.32\textwidth,height=0.18\textheight]{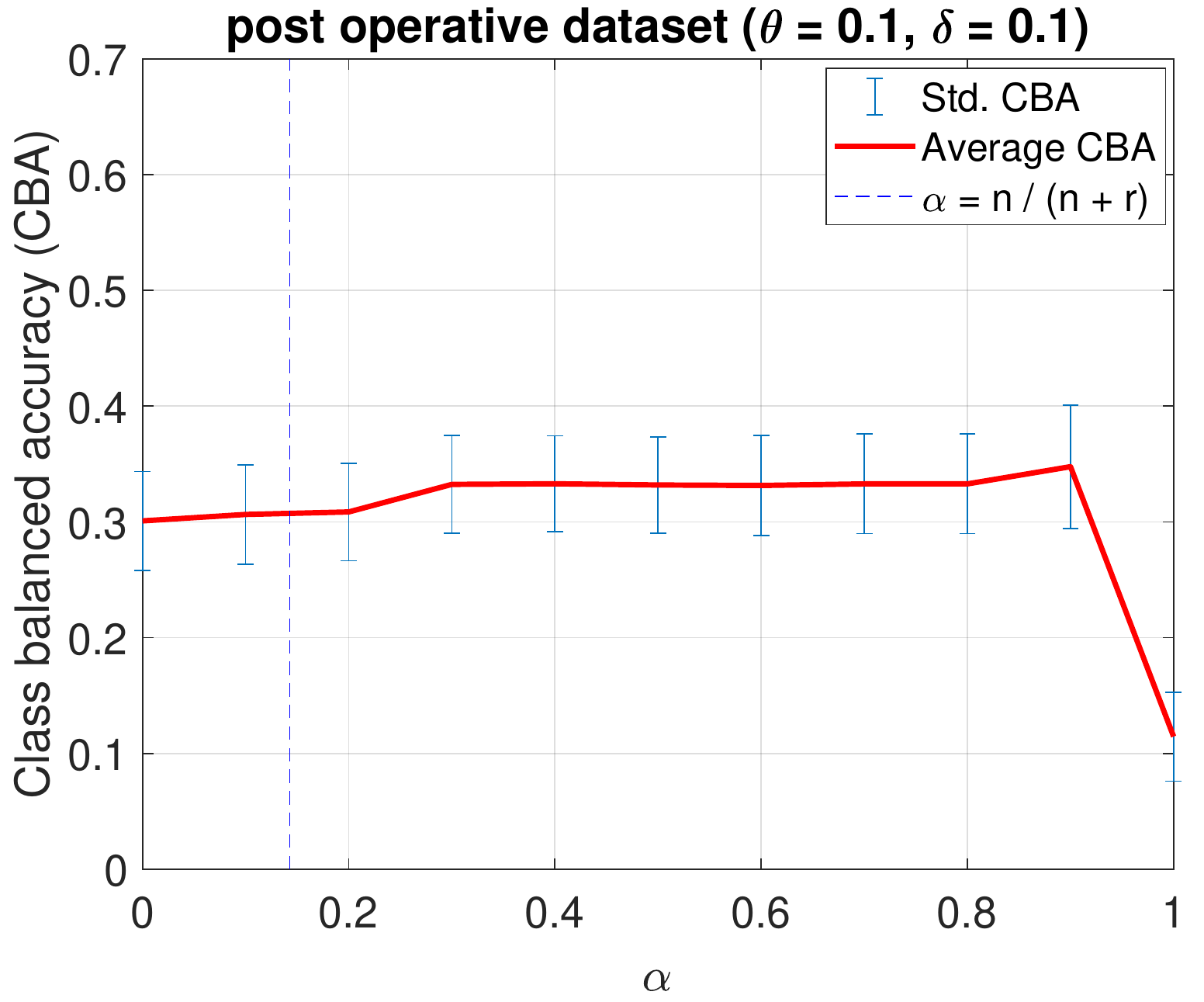}}
	\end{subfloat}
	\hfill
	\begin{subfloat}[tae]{
			\includegraphics[width=0.32\textwidth,height=0.18\textheight]{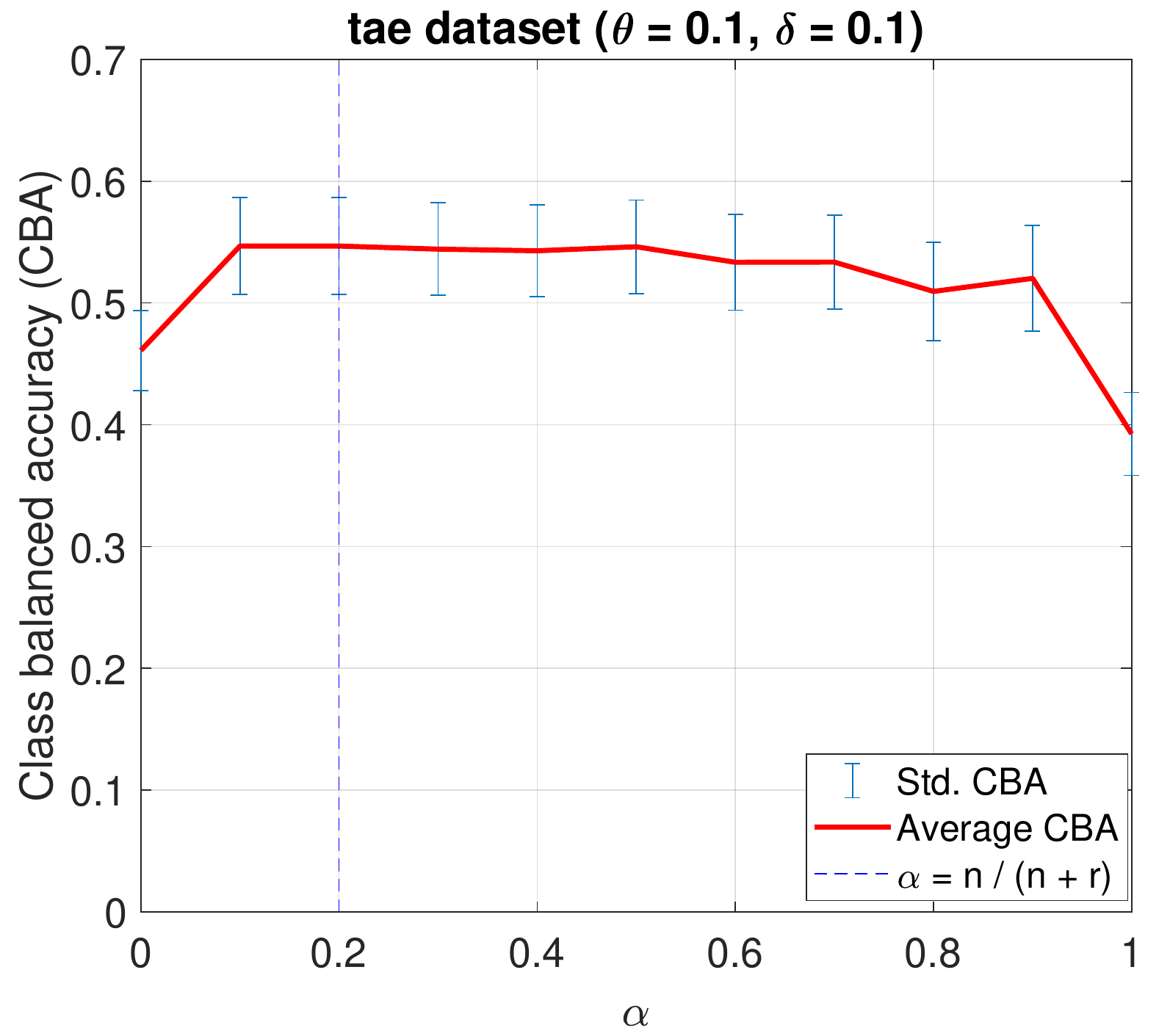}}
	\end{subfloat}
	\begin{subfloat}[zoo]{
			\includegraphics[width=0.32\textwidth,height=0.18\textheight]{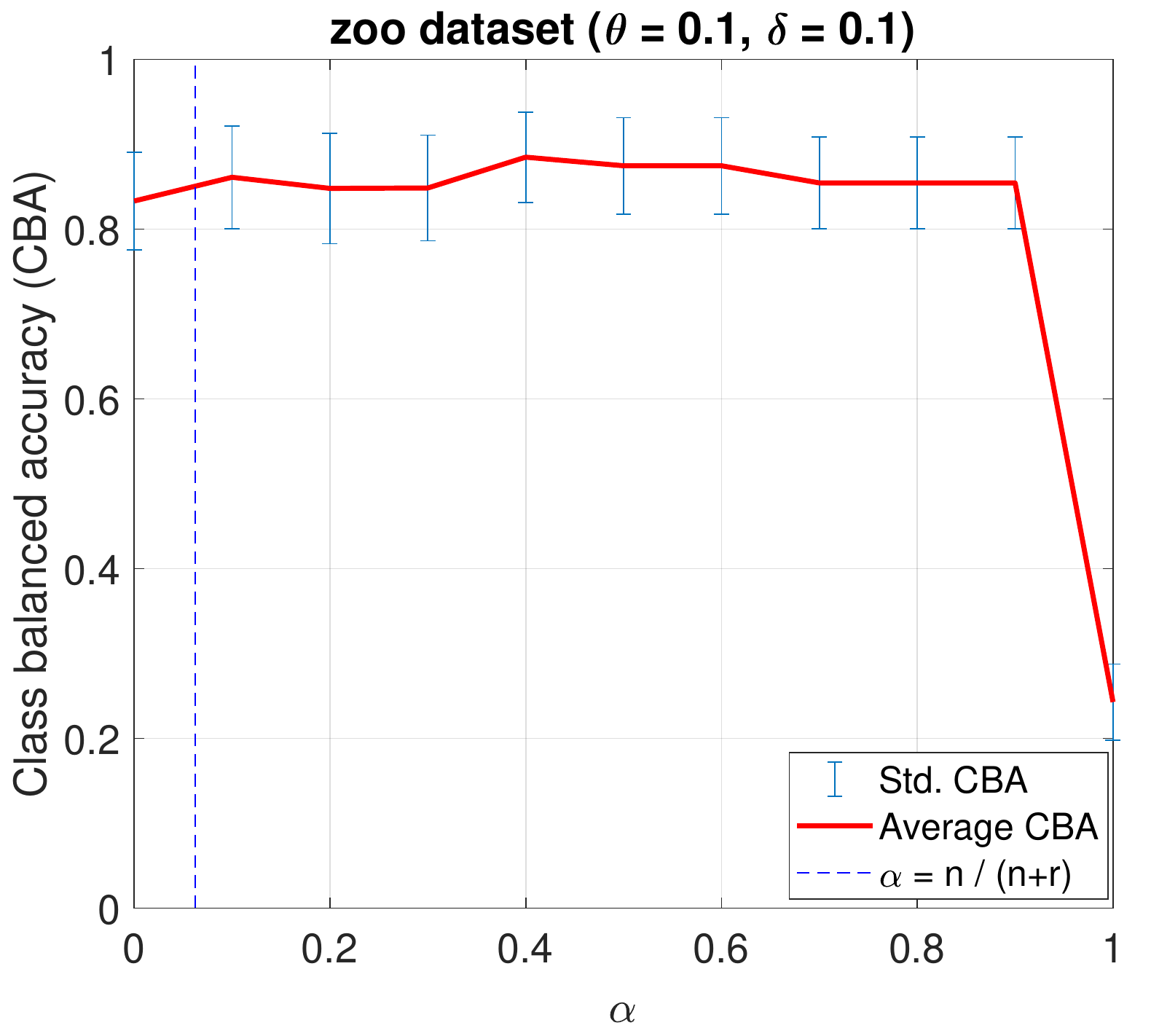}}
	\end{subfloat}
	\caption{Class balanced accuracy values for different values of $\alpha$ using the IOL-GFMM-v1 algorithm ($\theta = 0.1, \delta = 0.1$).}
	\label{Fig_S2}
\end{figure}

\begin{figure}[!ht]
	\begin{subfloat}[abalone]{
			\includegraphics[width=0.32\textwidth, height=0.18\textheight]{abalone_changed_alpha_0101.pdf}}
	\end{subfloat}
	\hfill%
	\begin{subfloat}[australian]{
			\includegraphics[width=0.32\textwidth,height=0.18\textheight]{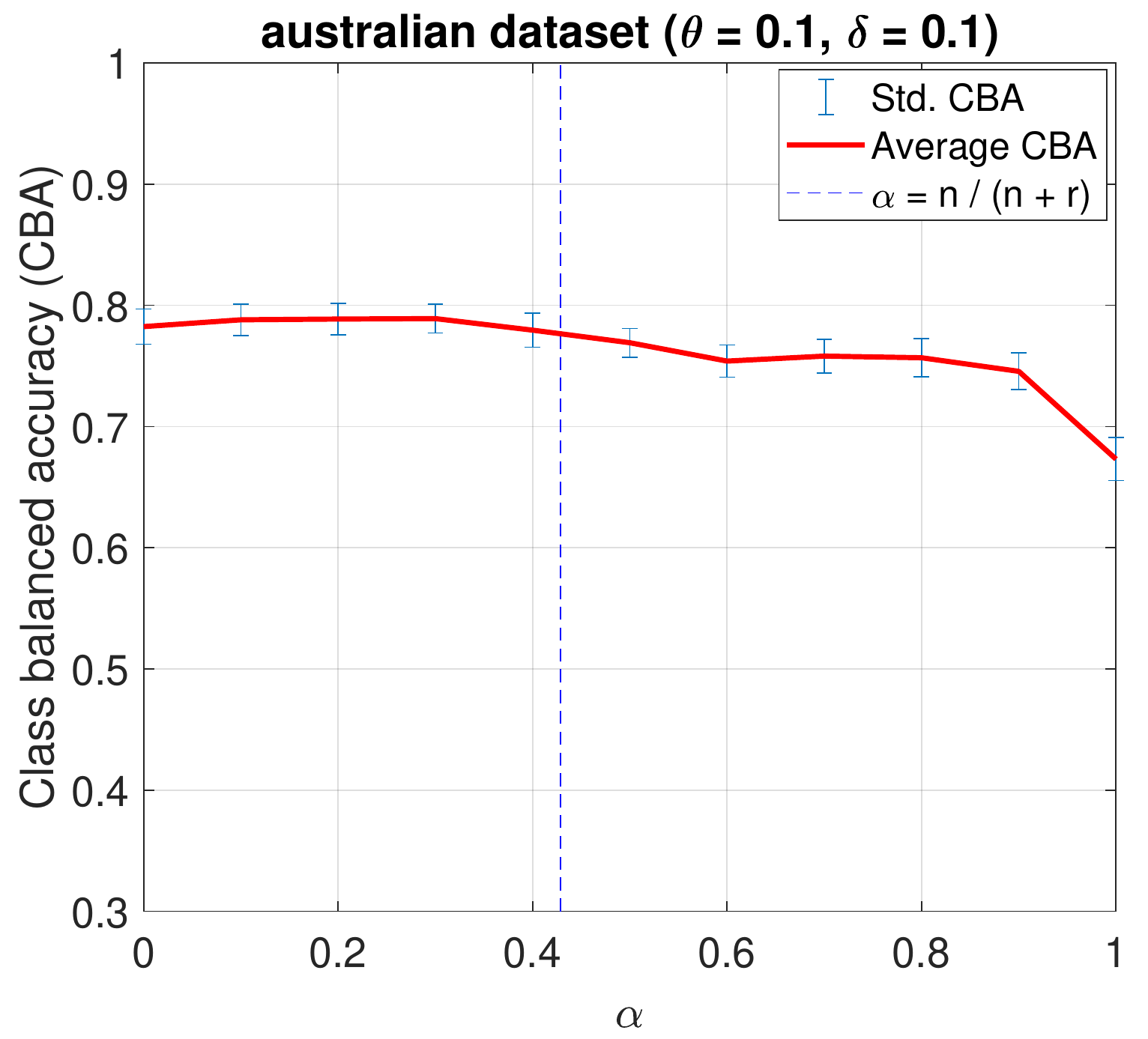}}
	\end{subfloat}
	\hfill%
	\begin{subfloat}[cmc]{
			\includegraphics[width=0.32\textwidth,height=0.18\textheight]{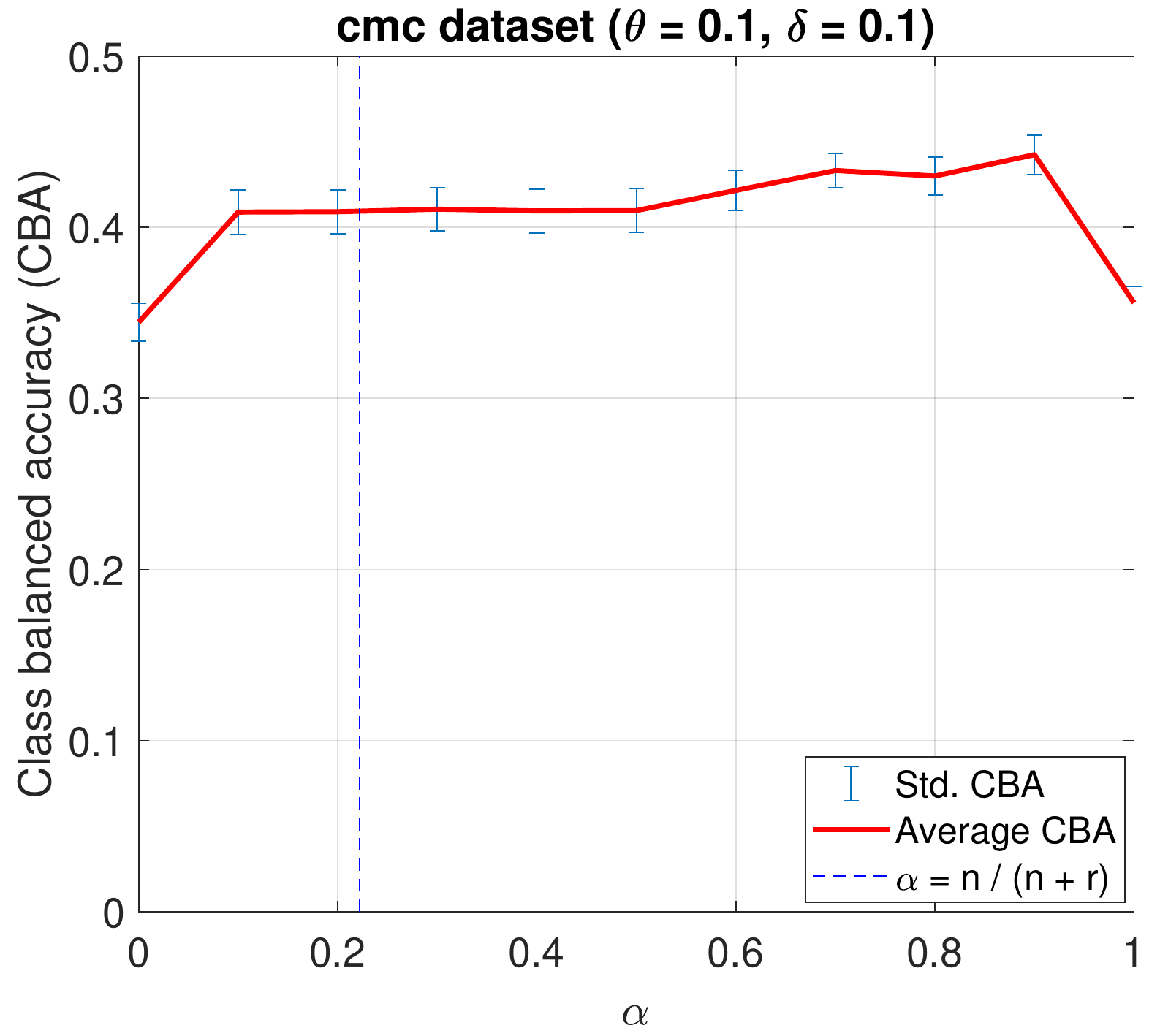}}
	\end{subfloat}
	\hfill%
	\begin{subfloat}[dermatology]{
			\includegraphics[width=0.32\textwidth,height=0.18\textheight]{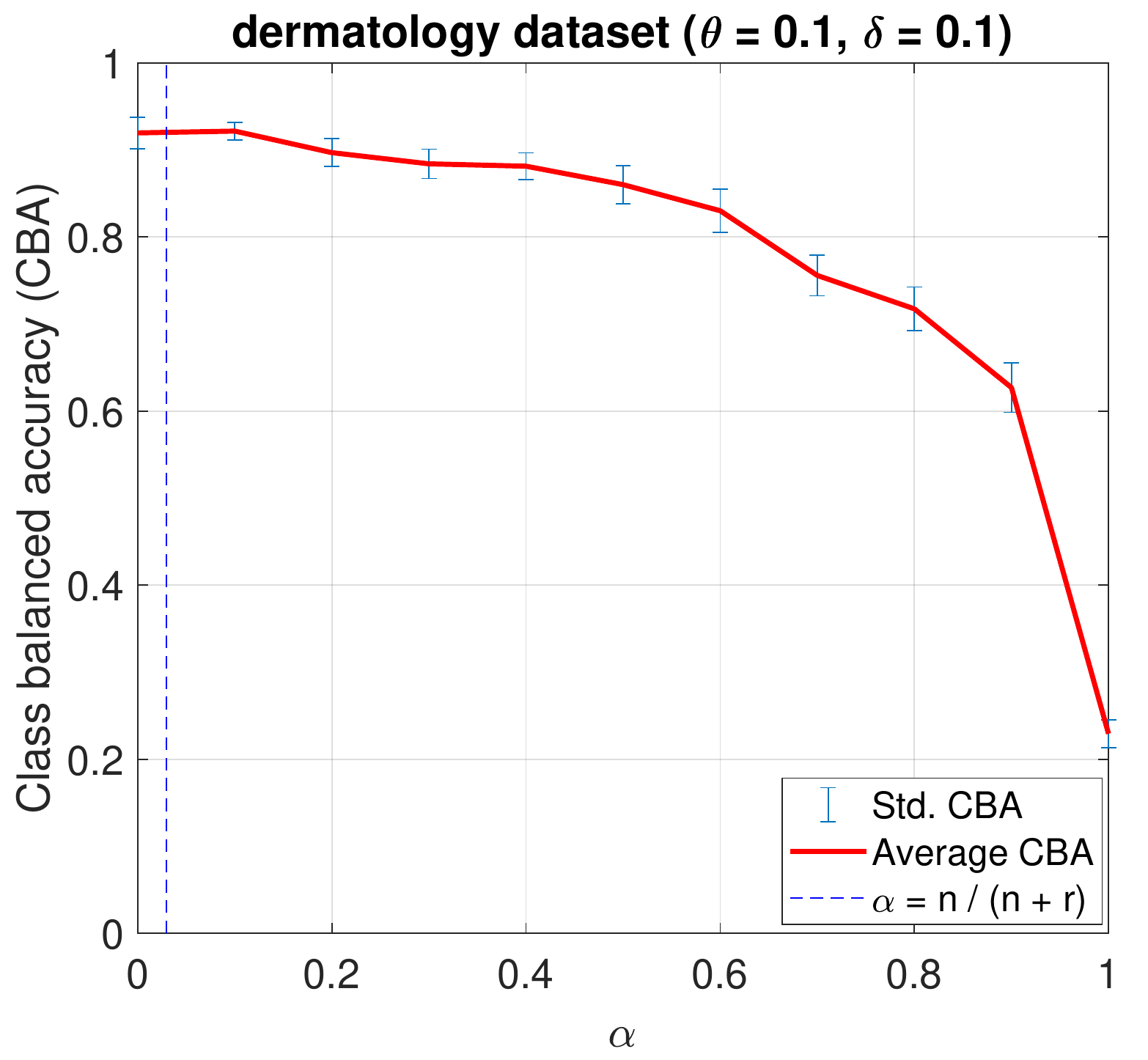}}
	\end{subfloat}
	\hfill%
	\begin{subfloat}[flag]{
			\includegraphics[width=0.32\textwidth,height=0.18\textheight]{flag_changed_alpha_0101_v2.pdf}}
	\end{subfloat}
	\hfill%
	\begin{subfloat}[german]{
			\includegraphics[width=0.32\textwidth,height=0.18\textheight]{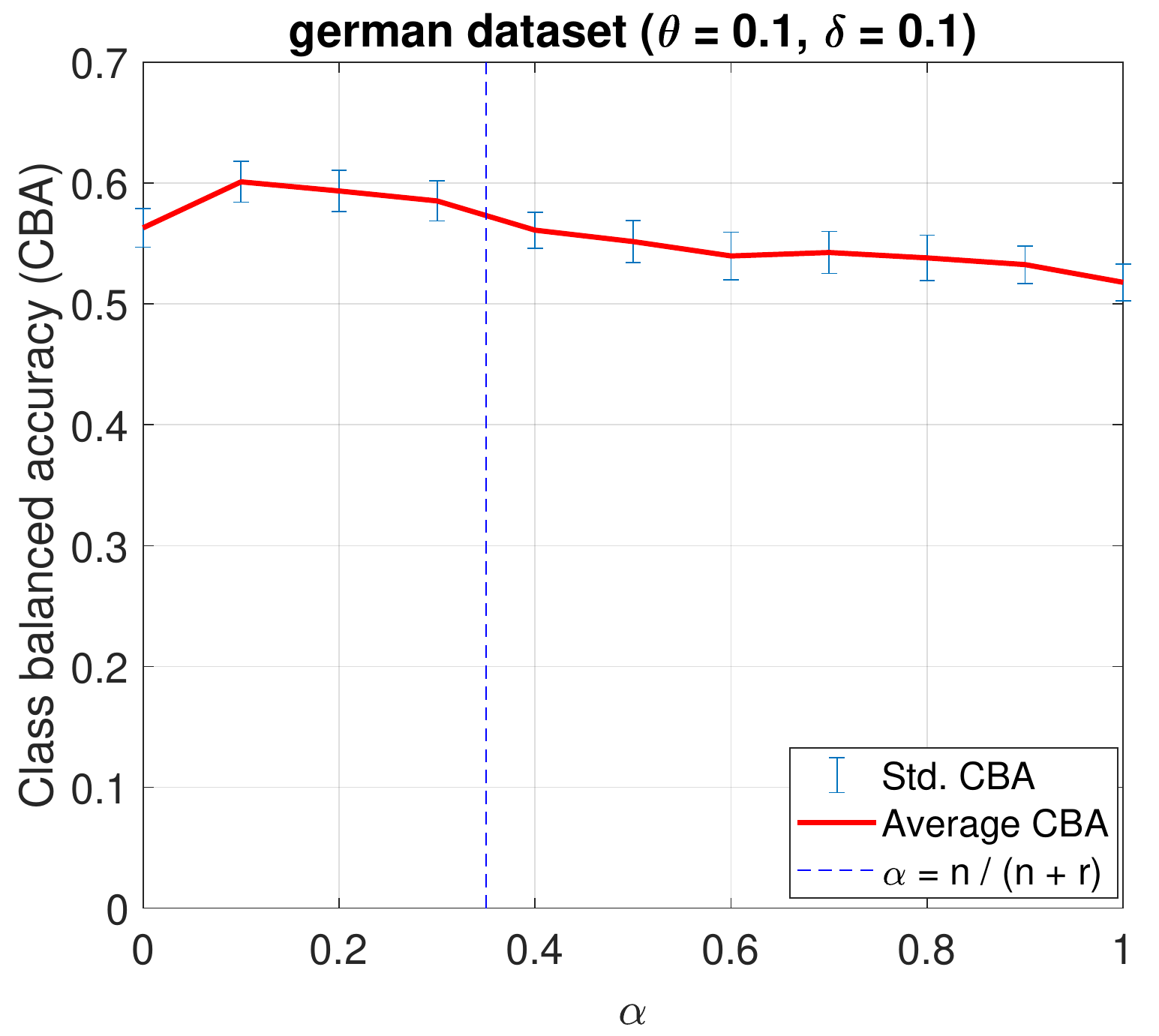}}
	\end{subfloat}
	\hfill
	\begin{subfloat}[heart]{
			\includegraphics[width=0.32\textwidth,height=0.18\textheight]{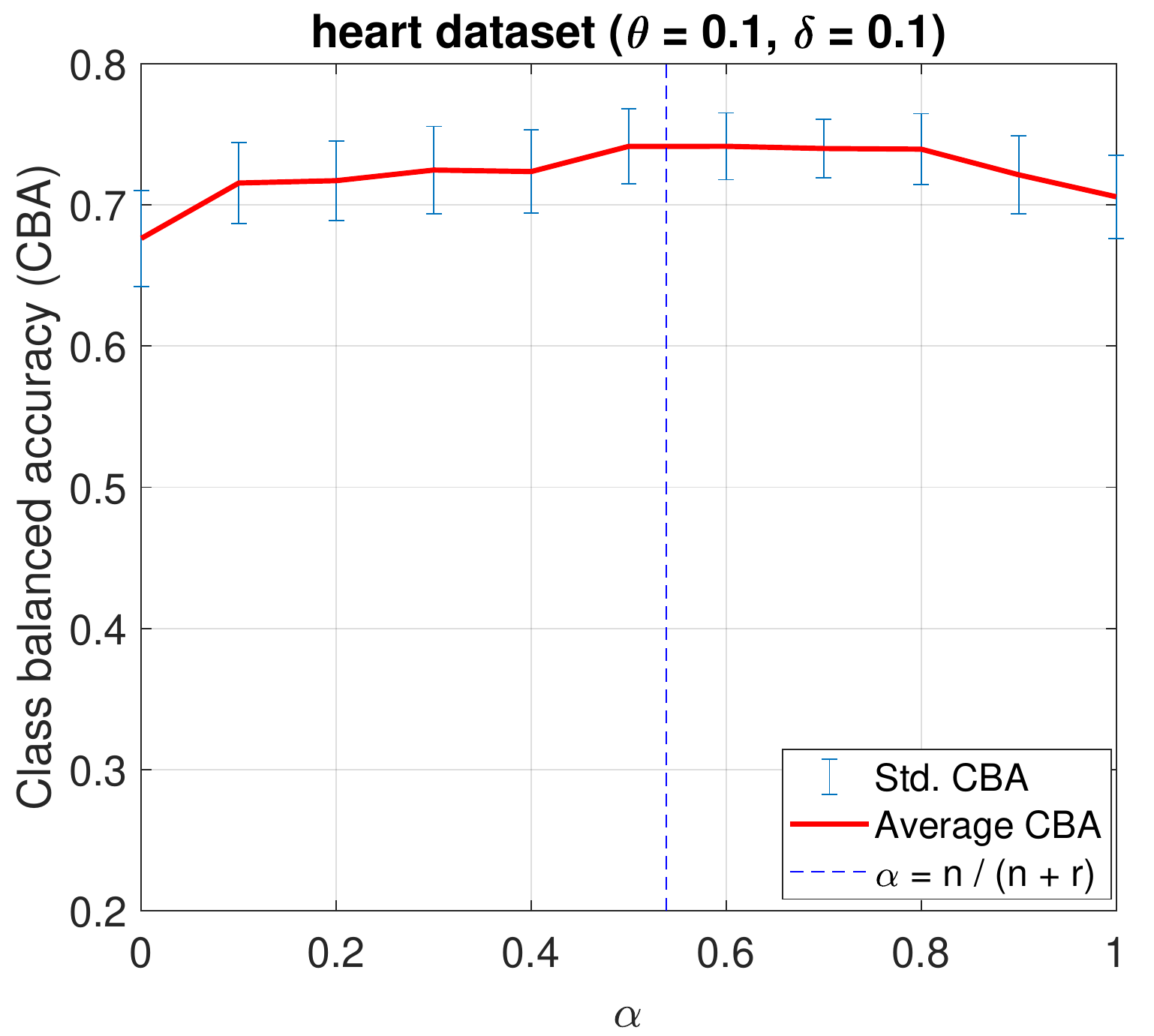}}
	\end{subfloat}
	\hfill
	\begin{subfloat}[japanese credit]{
			\includegraphics[width=0.32\textwidth,height=0.18\textheight]{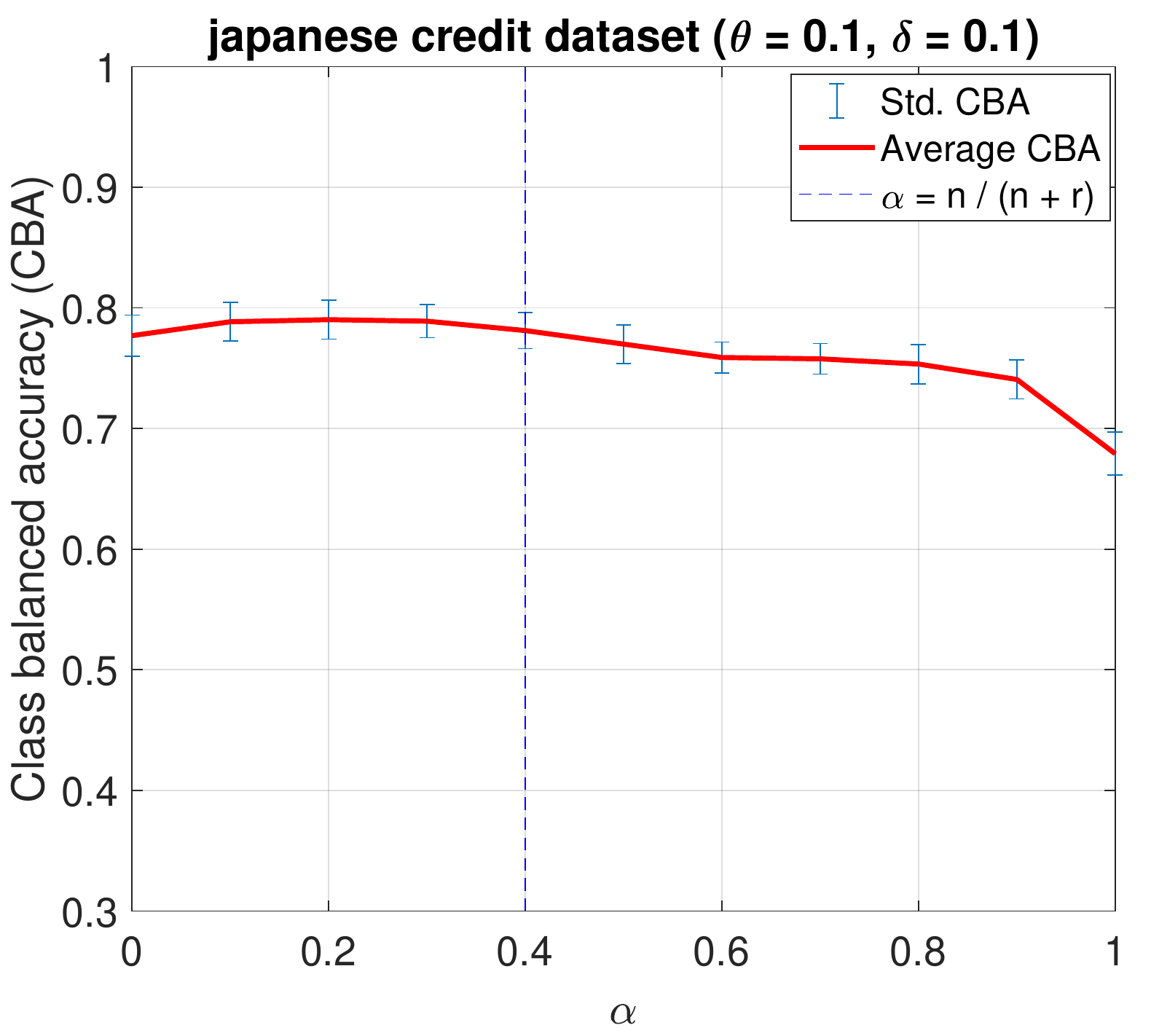}}
	\end{subfloat}
	\hfill
	\begin{subfloat}[post operative]{
			\includegraphics[width=0.32\textwidth,height=0.18\textheight]{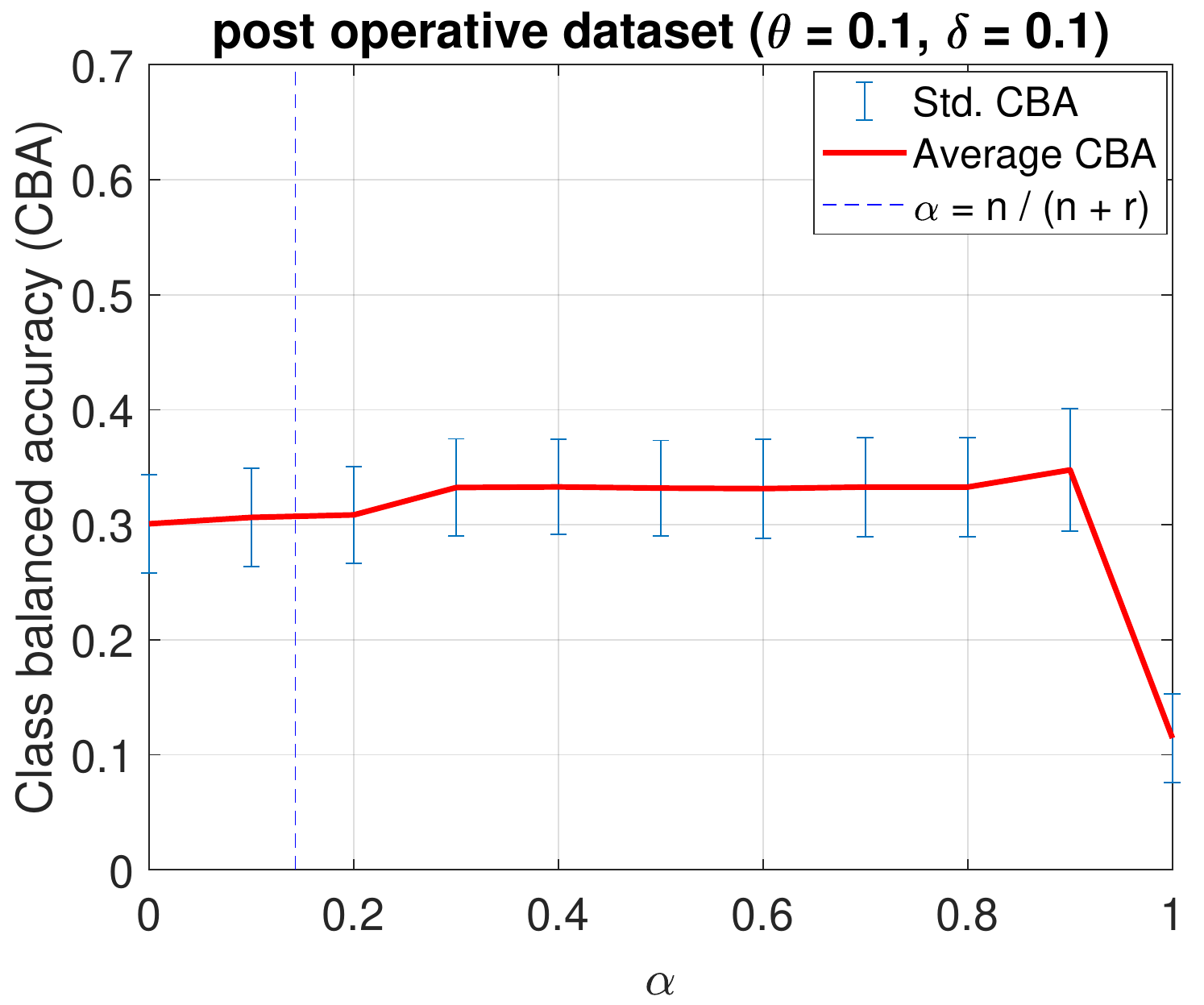}}
	\end{subfloat}
	\hfill
	\begin{subfloat}[tae]{
			\includegraphics[width=0.32\textwidth,height=0.18\textheight]{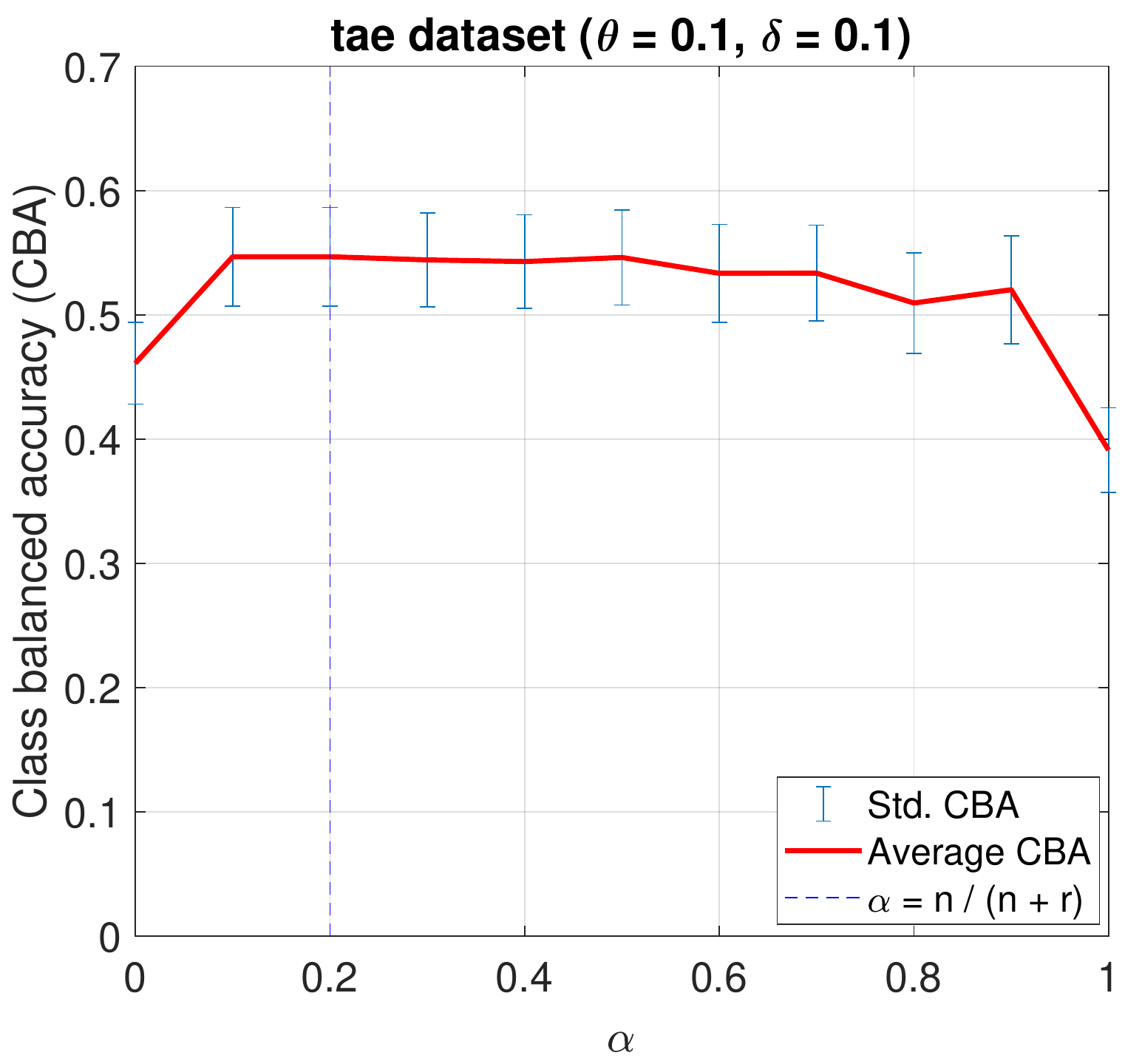}}
	\end{subfloat}
	\begin{subfloat}[zoo]{
			\includegraphics[width=0.32\textwidth,height=0.18\textheight]{zoo_changed_alpha_0101_v1.pdf}}
	\end{subfloat}
	\caption{Class balanced accuracy values for different values of $\alpha$ using the IOL-GFMM-v2 algorithm ($\theta = 0.1, \delta = 0.1$).}
	\label{Fig_S3}
\end{figure}

\begin{figure}[!ht]
	\begin{subfloat}[abalone]{
			\includegraphics[width=0.32\textwidth, height=0.18\textheight]{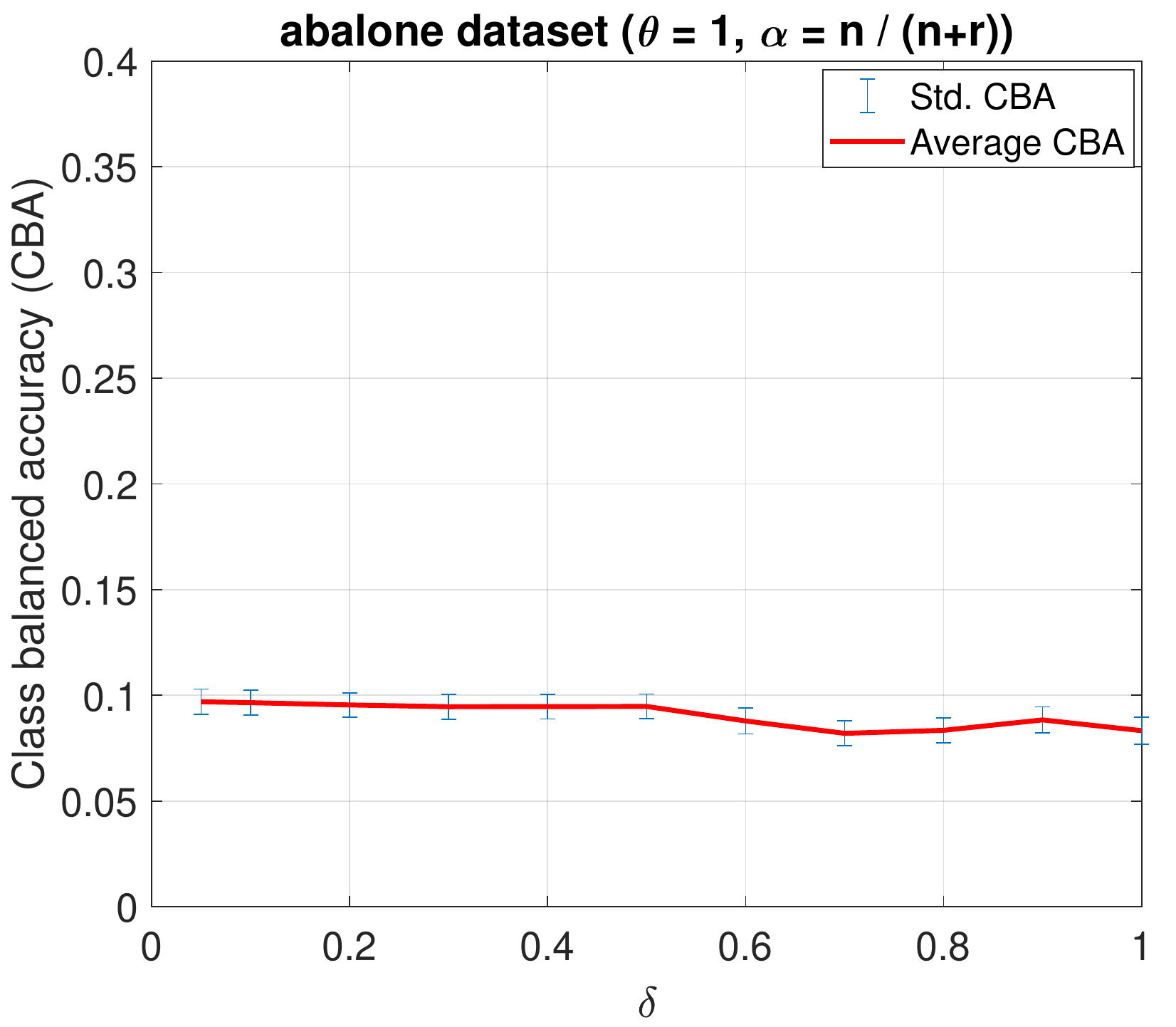}}
	\end{subfloat}
	\hfill%
	\begin{subfloat}[australian]{
			\includegraphics[width=0.32\textwidth,height=0.18\textheight]{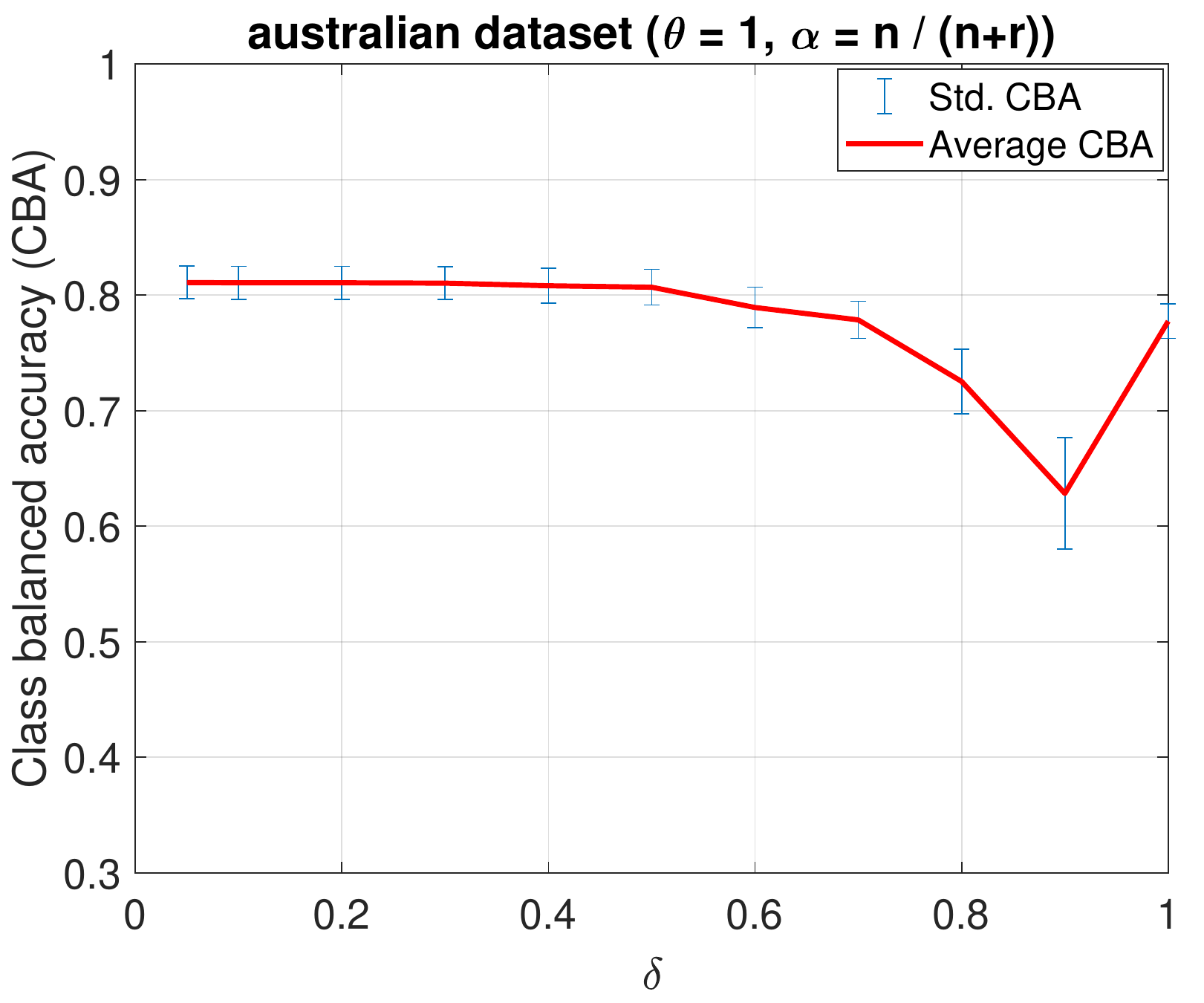}}
	\end{subfloat}
	\hfill%
	\begin{subfloat}[flag]{
			\includegraphics[width=0.32\textwidth,height=0.18\textheight]{flag_changed_delta_v1.pdf}}
	\end{subfloat}
	\hfill%
	\begin{subfloat}[german]{
			\includegraphics[width=0.32\textwidth,height=0.18\textheight]{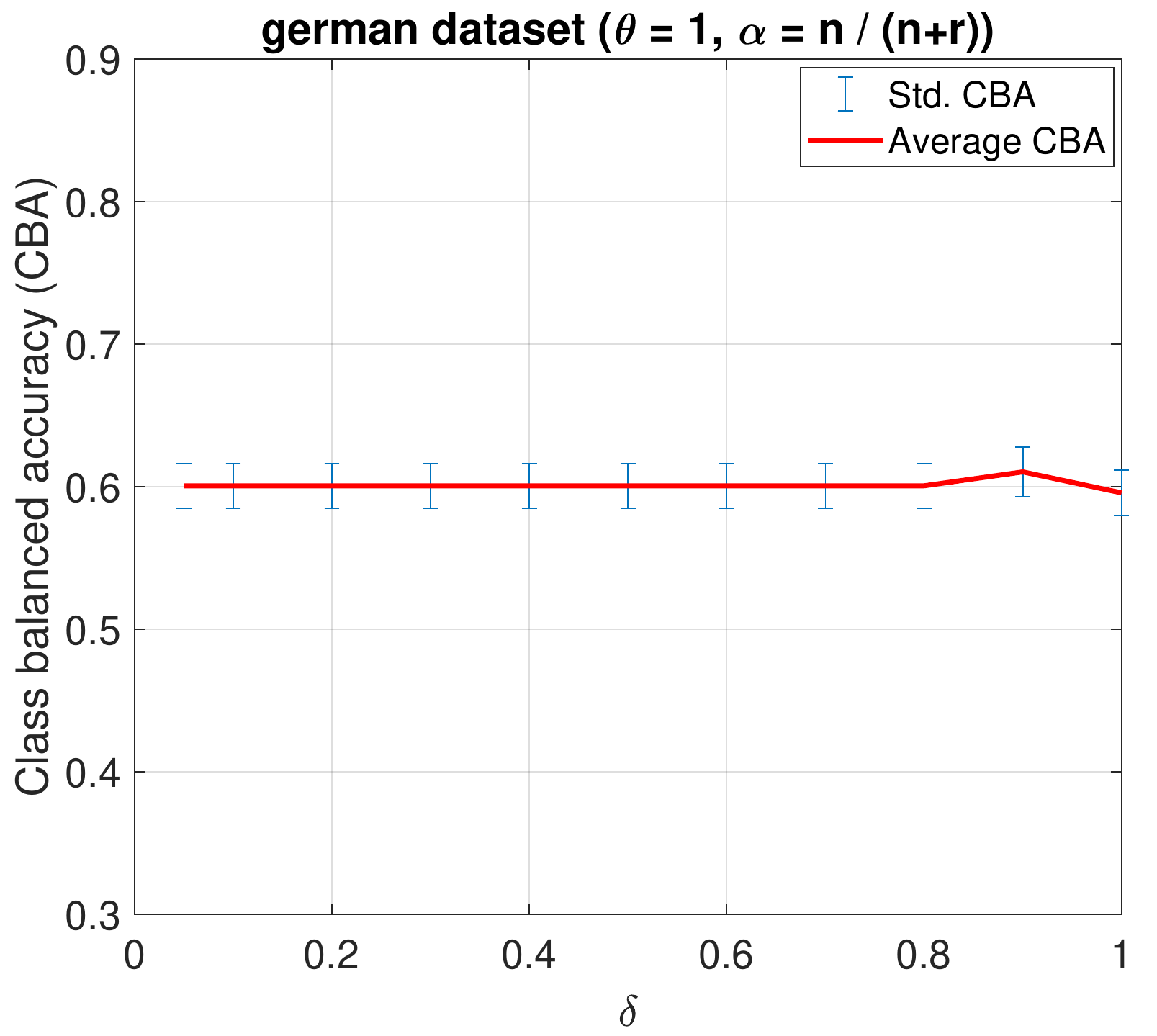}}
	\end{subfloat}
	\hfill
	\begin{subfloat}[heart]{
			\includegraphics[width=0.32\textwidth,height=0.18\textheight]{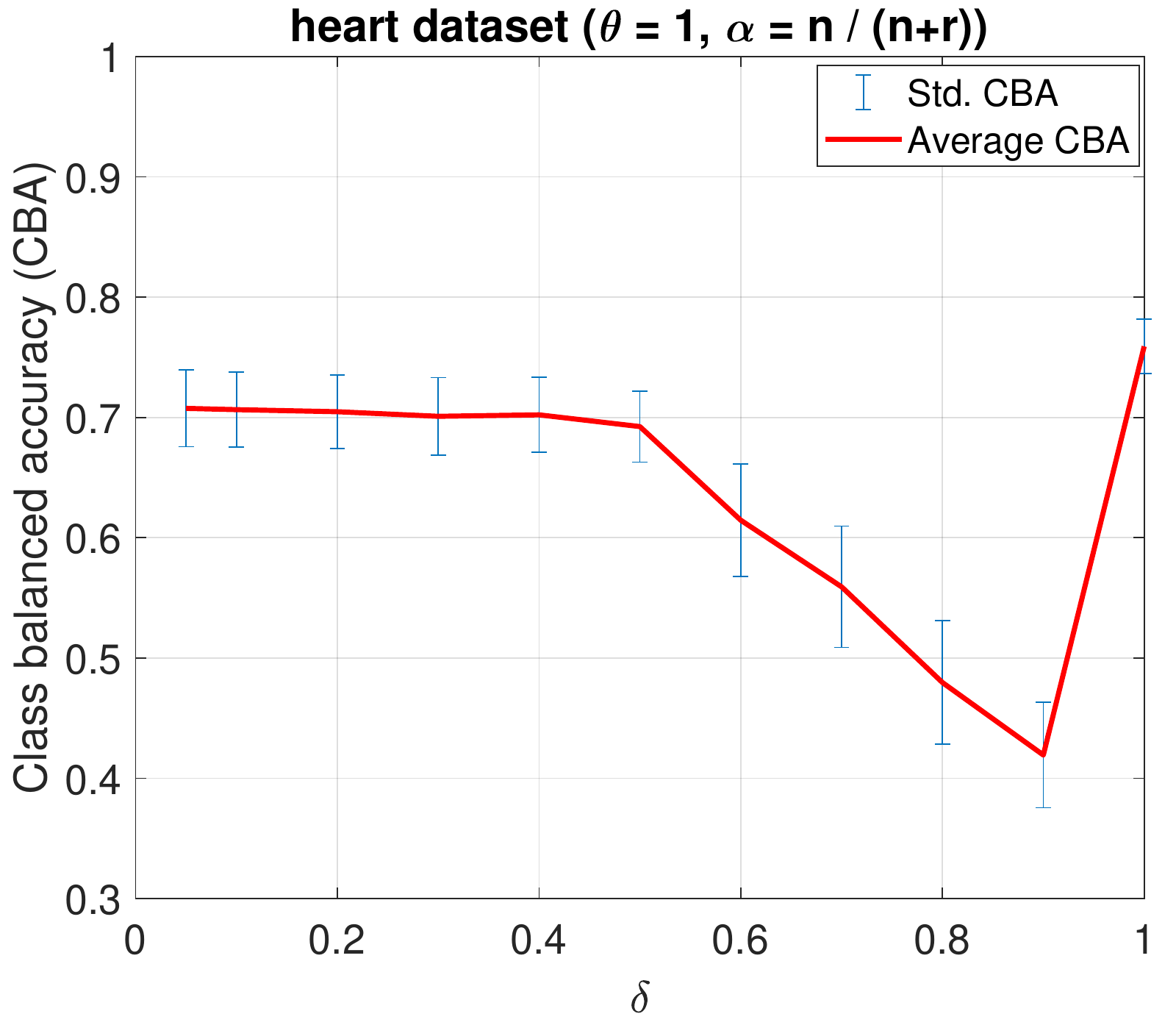}}
	\end{subfloat}
	\hfill
	\begin{subfloat}[post operative]{
			\includegraphics[width=0.32\textwidth,height=0.18\textheight]{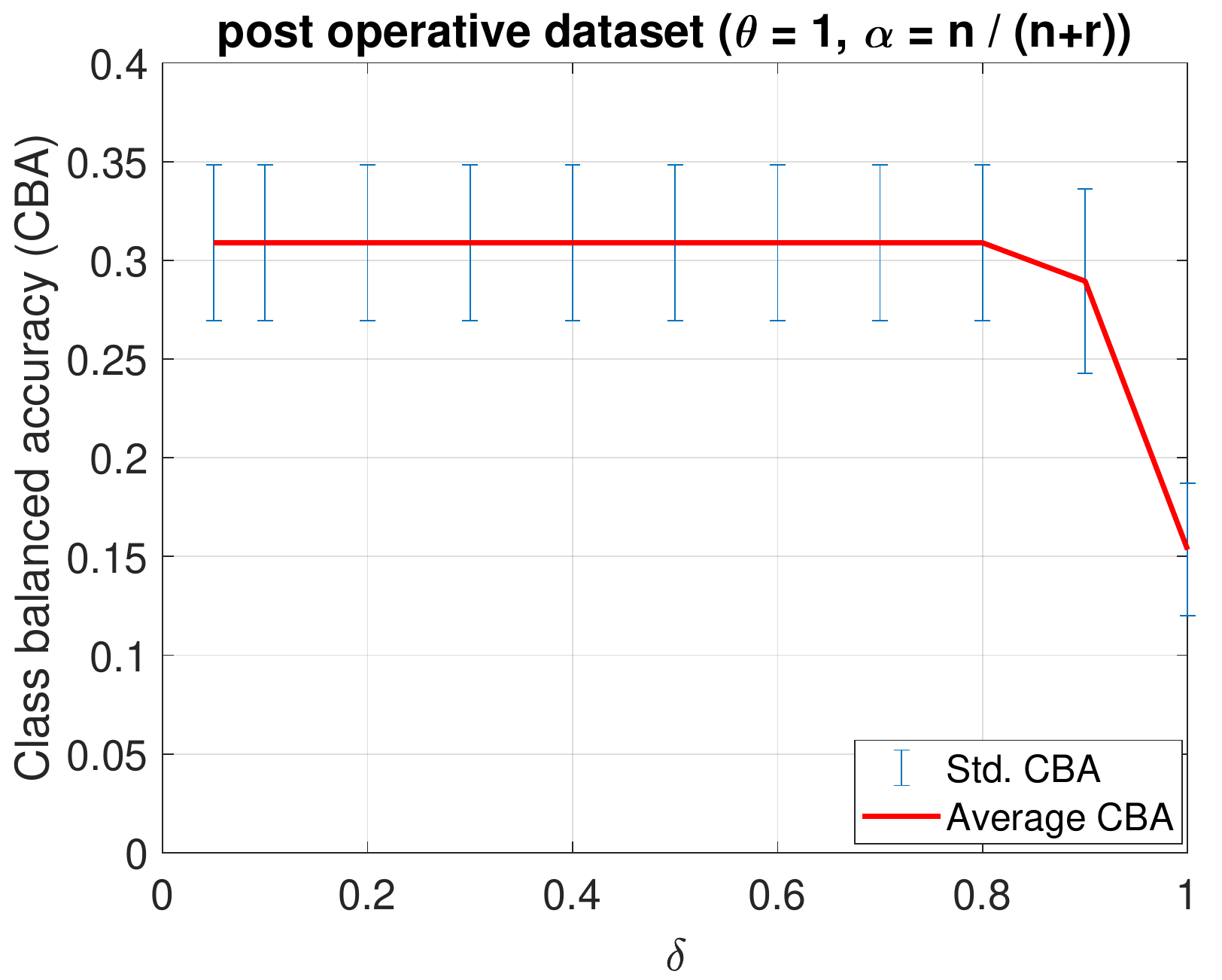}}
	\end{subfloat}
	\hfill
	\begin{subfloat}[tic tac toe]{
			\includegraphics[width=0.32\textwidth,height=0.18\textheight]{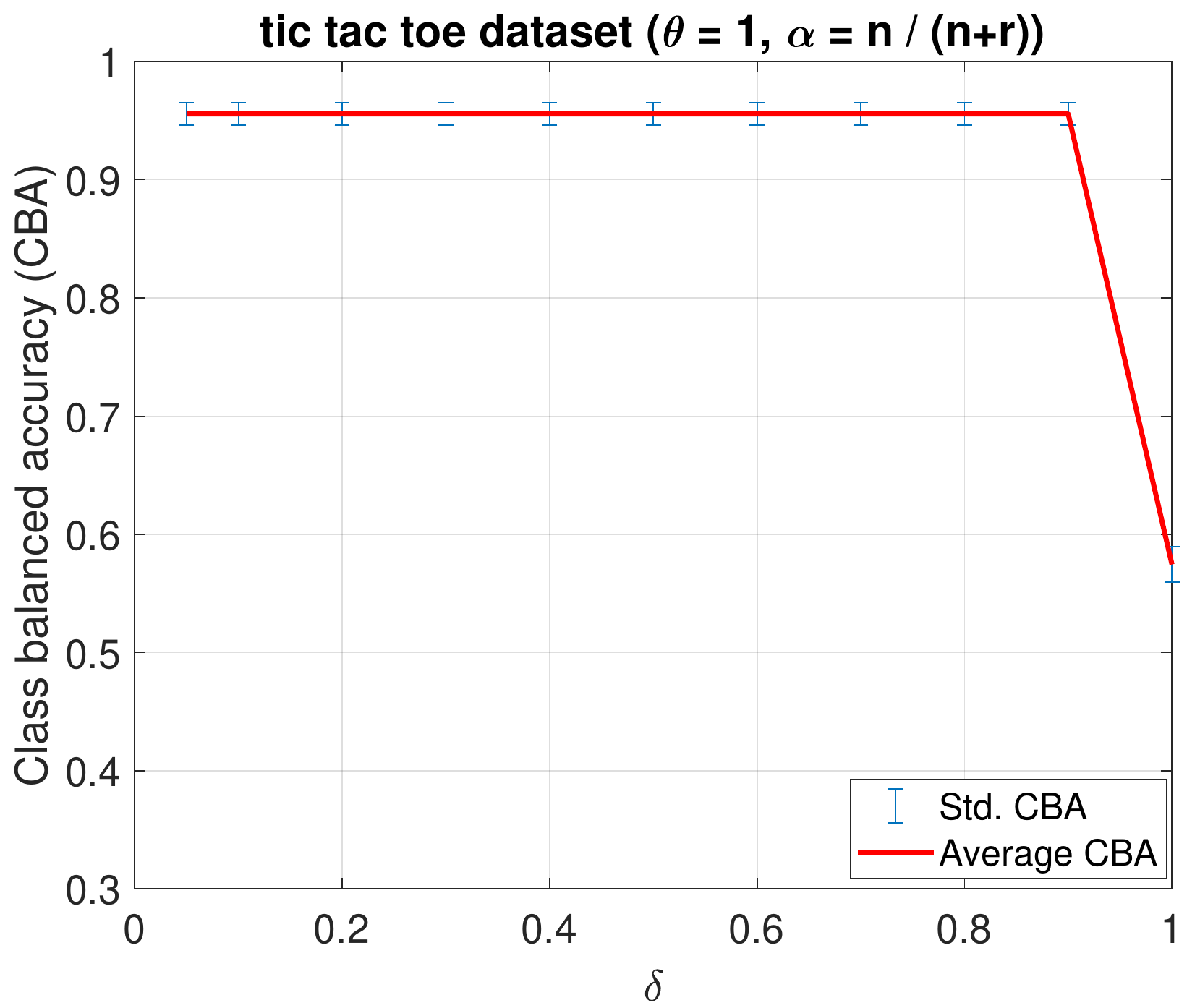}}
	\end{subfloat}
	\begin{subfloat}[zoo]{
			\includegraphics[width=0.32\textwidth,height=0.18\textheight]{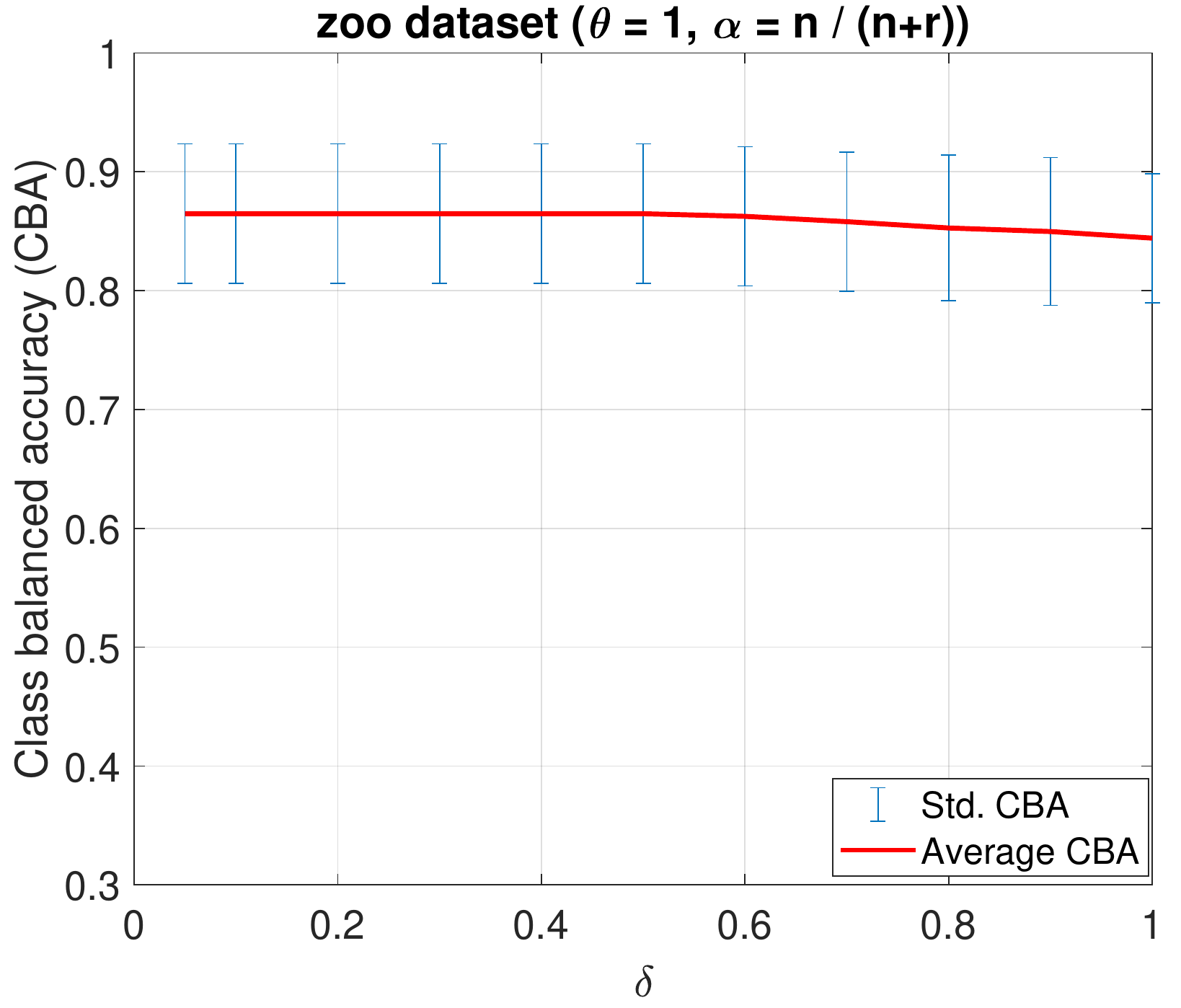}}
	\end{subfloat}
	\caption{Class balanced accuracy values for different values of $\delta$ using the IOL-GFMM-v1 algorithm ($\theta = 1, \alpha = n/(n+r)$).}
	\label{Fig_S4}
\end{figure}

\begin{figure}[!ht]
	\begin{subfloat}[abalone]{
			\includegraphics[width=0.32\textwidth, height=0.18\textheight]{abalone_changed_delta.pdf}}
	\end{subfloat}
	\hfill%
	\begin{subfloat}[australian]{
			\includegraphics[width=0.32\textwidth,height=0.18\textheight]{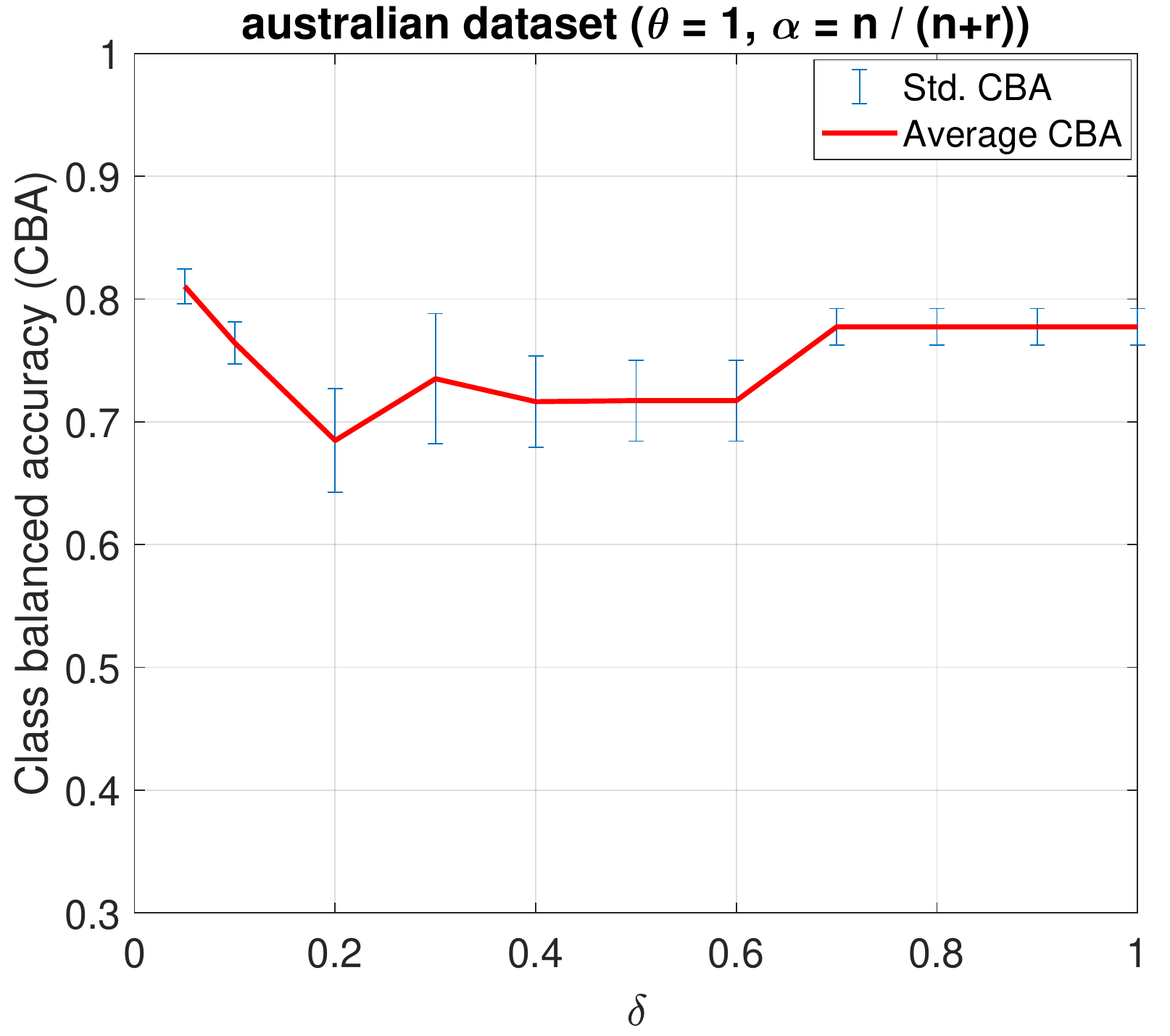}}
	\end{subfloat}
	\hfill%
	\begin{subfloat}[flag]{
			\includegraphics[width=0.32\textwidth,height=0.18\textheight]{flag_changed_delta_v2.pdf}}
	\end{subfloat}
	\hfill%
	\begin{subfloat}[german]{
			\includegraphics[width=0.32\textwidth,height=0.18\textheight]{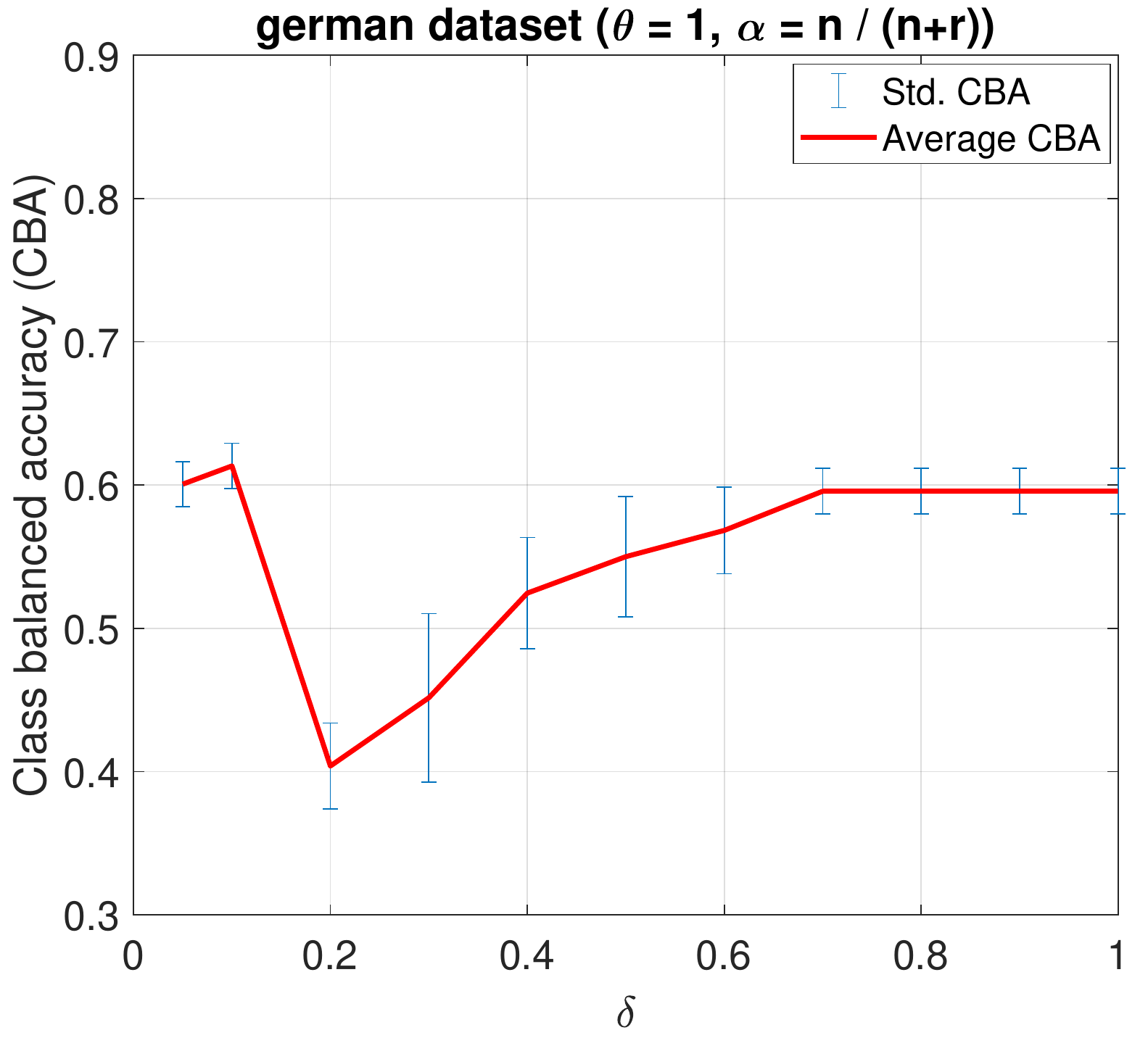}}
	\end{subfloat}
	\hfill
	\begin{subfloat}[heart]{
			\includegraphics[width=0.32\textwidth,height=0.18\textheight]{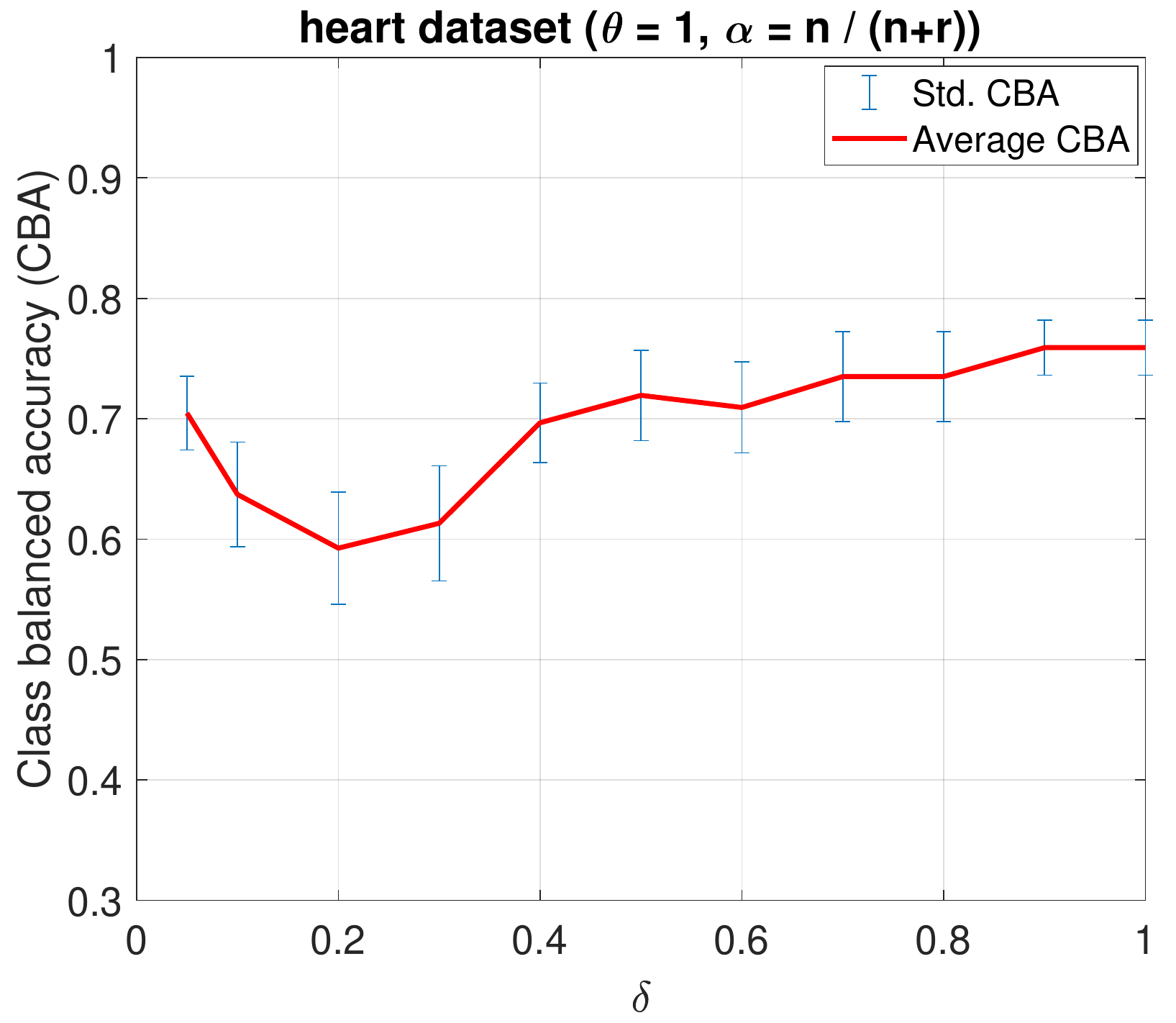}}
	\end{subfloat}
	\hfill
	\begin{subfloat}[post operative]{
			\includegraphics[width=0.32\textwidth,height=0.18\textheight]{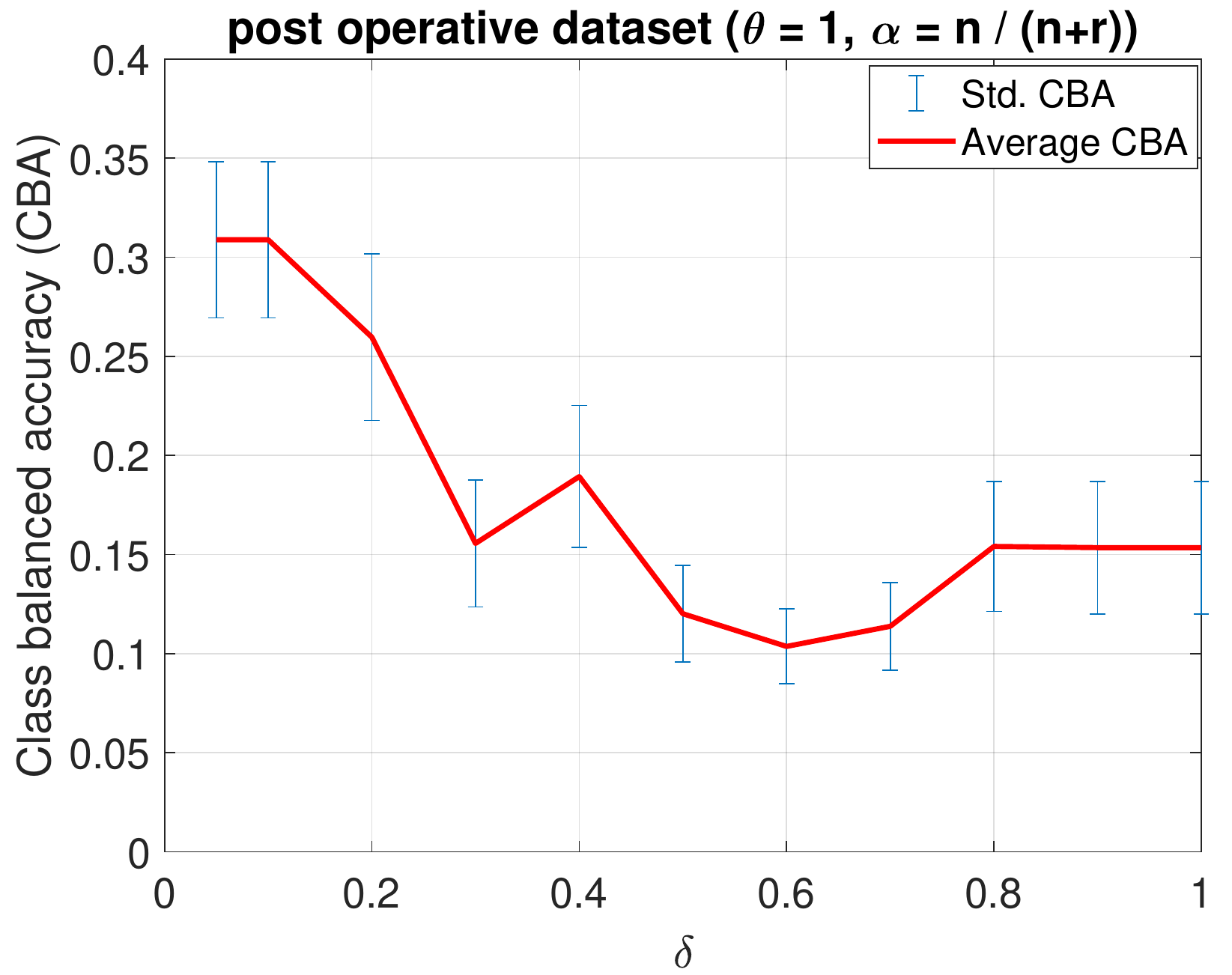}}
	\end{subfloat}
	\hfill
	\begin{subfloat}[tic tac toe]{
			\includegraphics[width=0.32\textwidth,height=0.18\textheight]{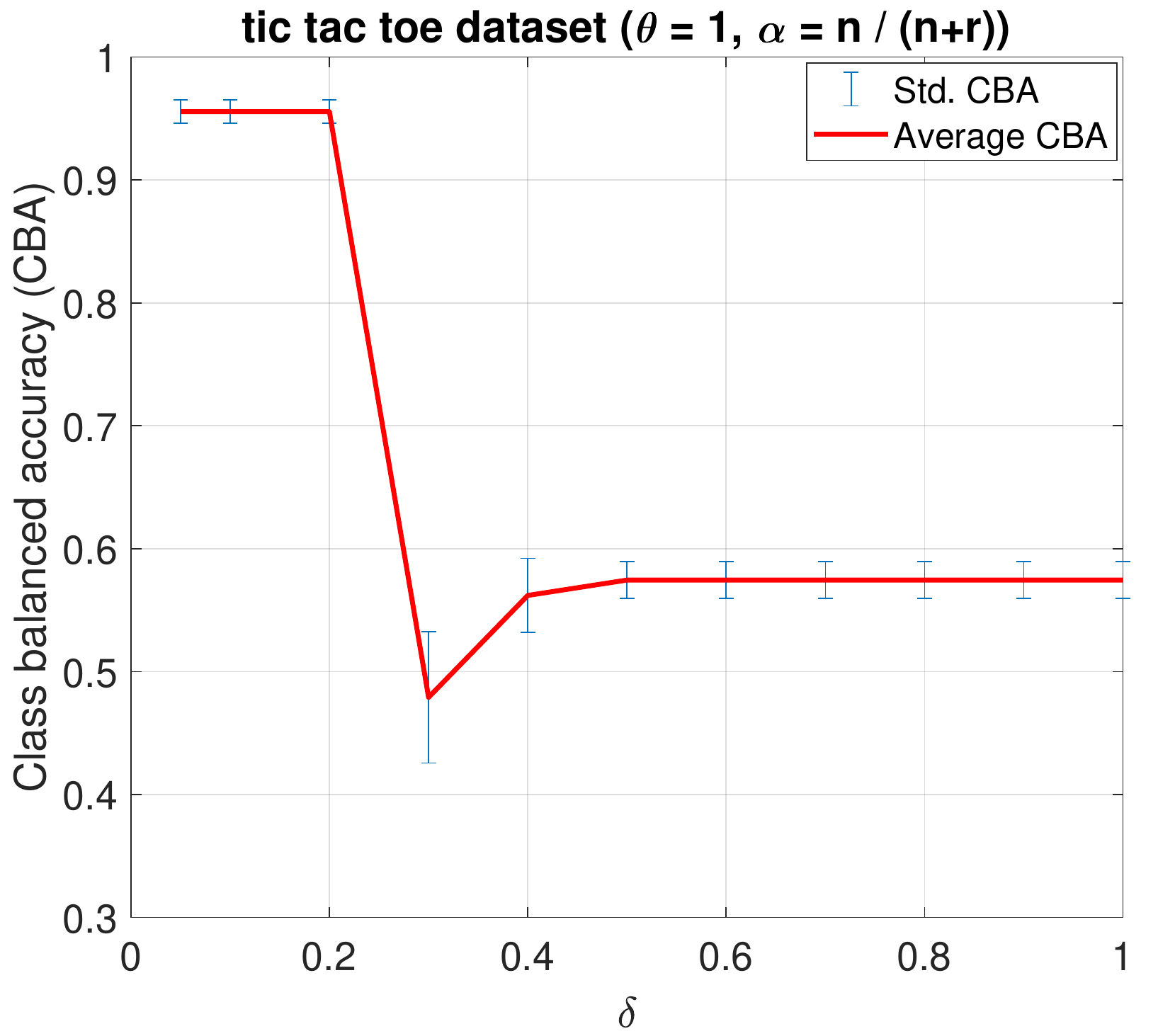}}
	\end{subfloat}
	\begin{subfloat}[zoo]{
			\includegraphics[width=0.32\textwidth,height=0.18\textheight]{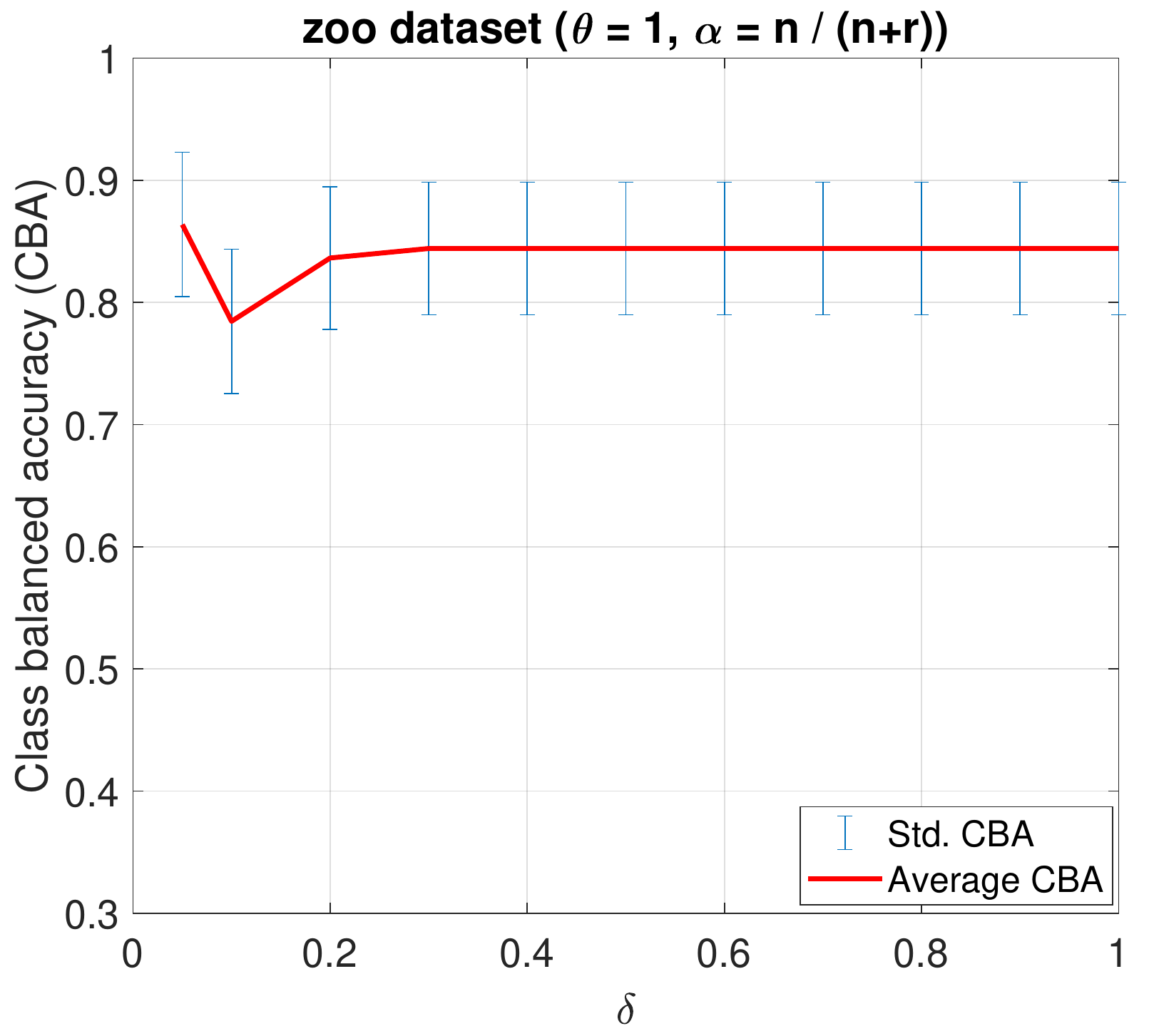}}
	\end{subfloat}
	\caption{Class balanced accuracy values for different values of $\delta$ using the IOL-GFMM-v2 algorithm ($\theta = 1, \alpha = n/(n+r)$).}
	\label{Fig_S5}
\end{figure}

\begin{figure}[!ht]
    \centering
    \includegraphics[width=0.9\textwidth]{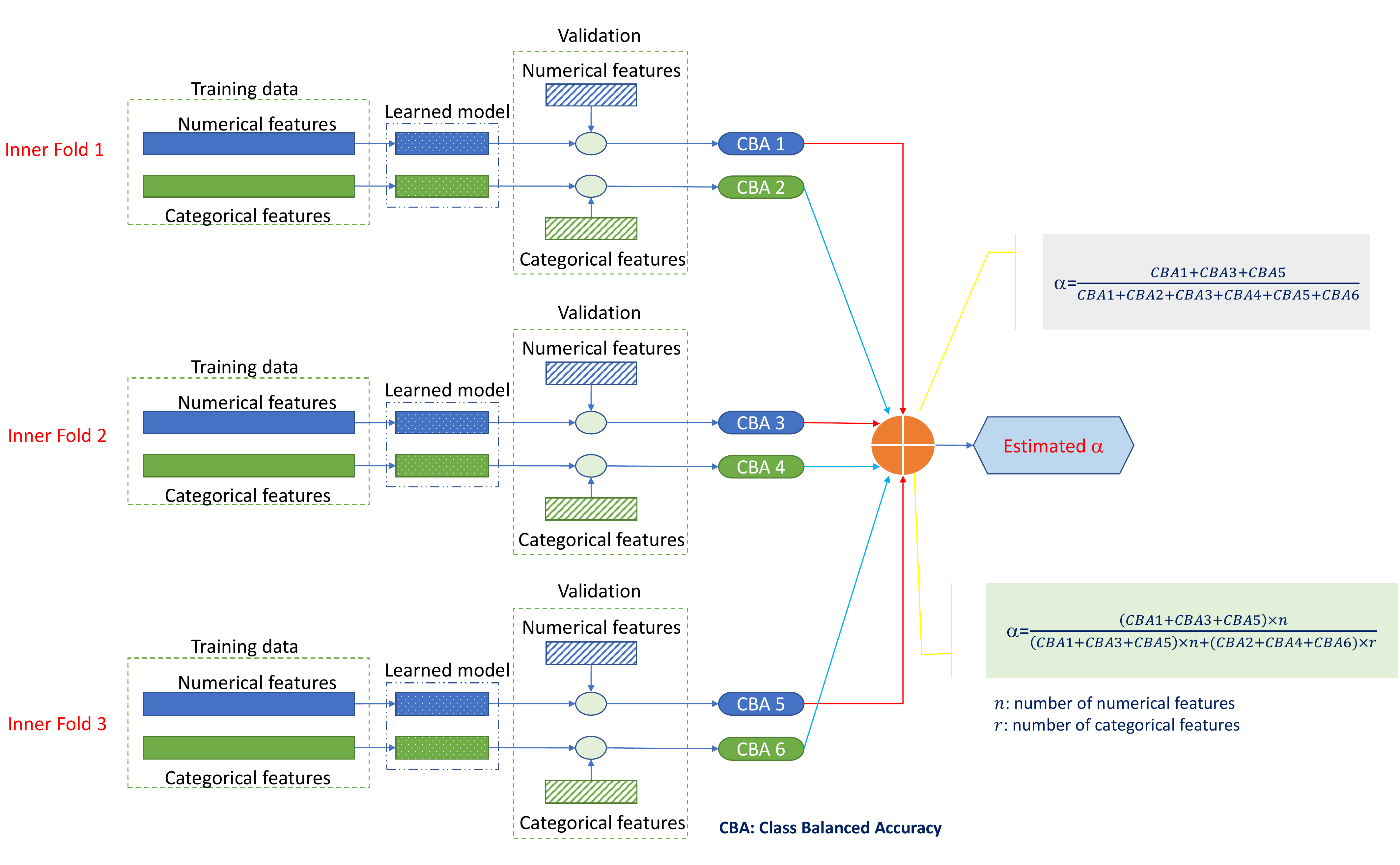}
    \caption{A demonstration for the methods used to estimate the values for parameter $\alpha$.}
    \label{Fig_S6}
\end{figure}

Fig. \ref{Fig_S1} to Fig. \ref{Fig_S3} show the change in the average class balanced accuracy (CBA) scores for different values of $\alpha$. These results support the contents in subsection IV.A.1 from the main paper. Fig \ref{Fig_S4} presents the change in the CBA values for different values of $\delta$ using the IOL-GFMM-v1 algorithm. Meanwhile, Fig. \ref{Fig_S5} shows similar results for the IOL-GFMM-v2 algorithm. These figures consolidate the comments in subsection IV.A.2 from the main paper. Fig \ref{Fig_S6} provides a graphical illustration for two ways of estimating the value of $\alpha$ shown in subsection IV.C.1 from the main paper.

\begin{figure}[!ht]
	\begin{subfloat}[abalone]{
			\includegraphics[width=0.32\textwidth, height=0.18\textheight]{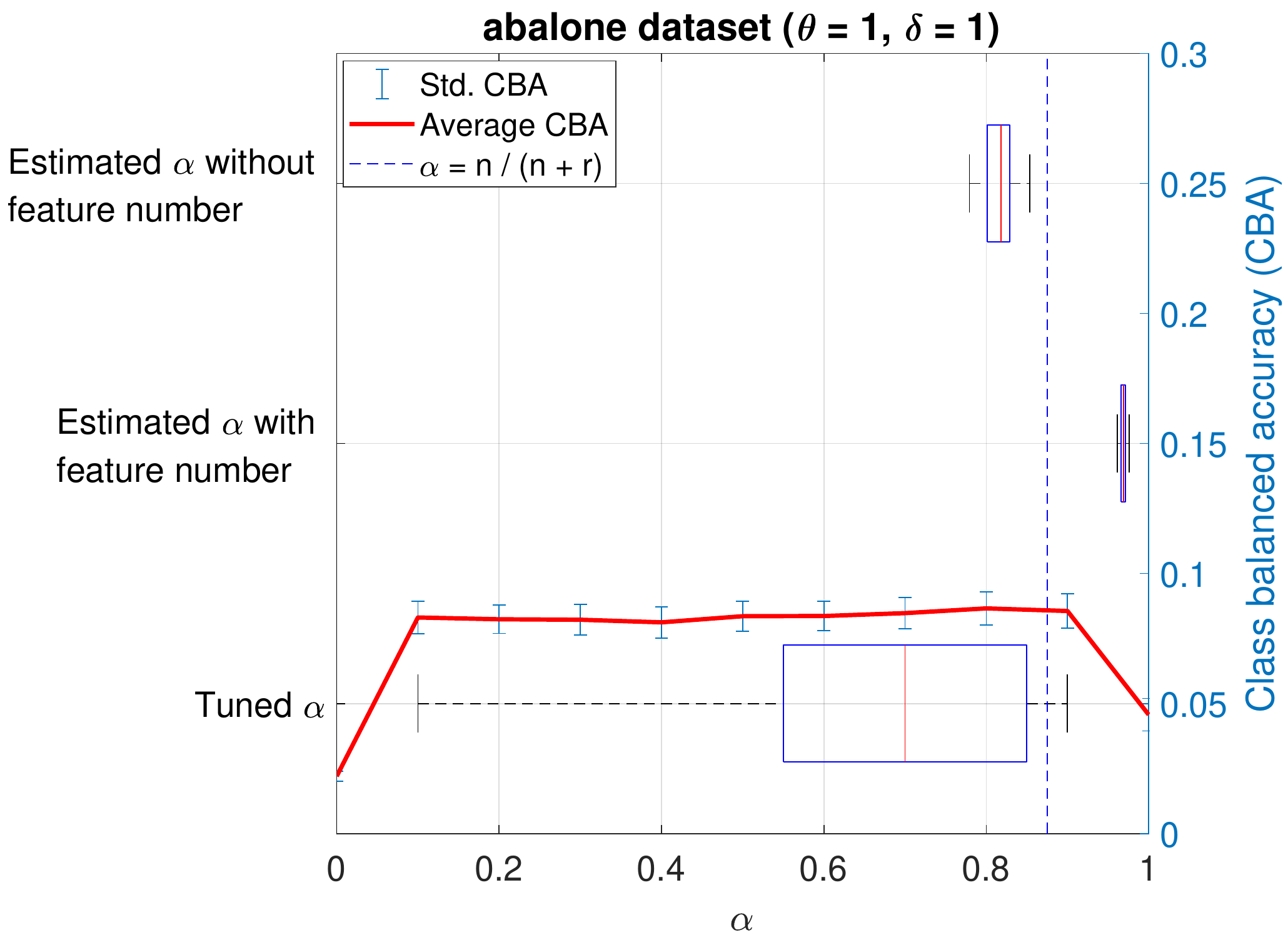}}
	\end{subfloat}
	\hfill%
	\begin{subfloat}[australian]{
			\includegraphics[width=0.32\textwidth,height=0.18\textheight]{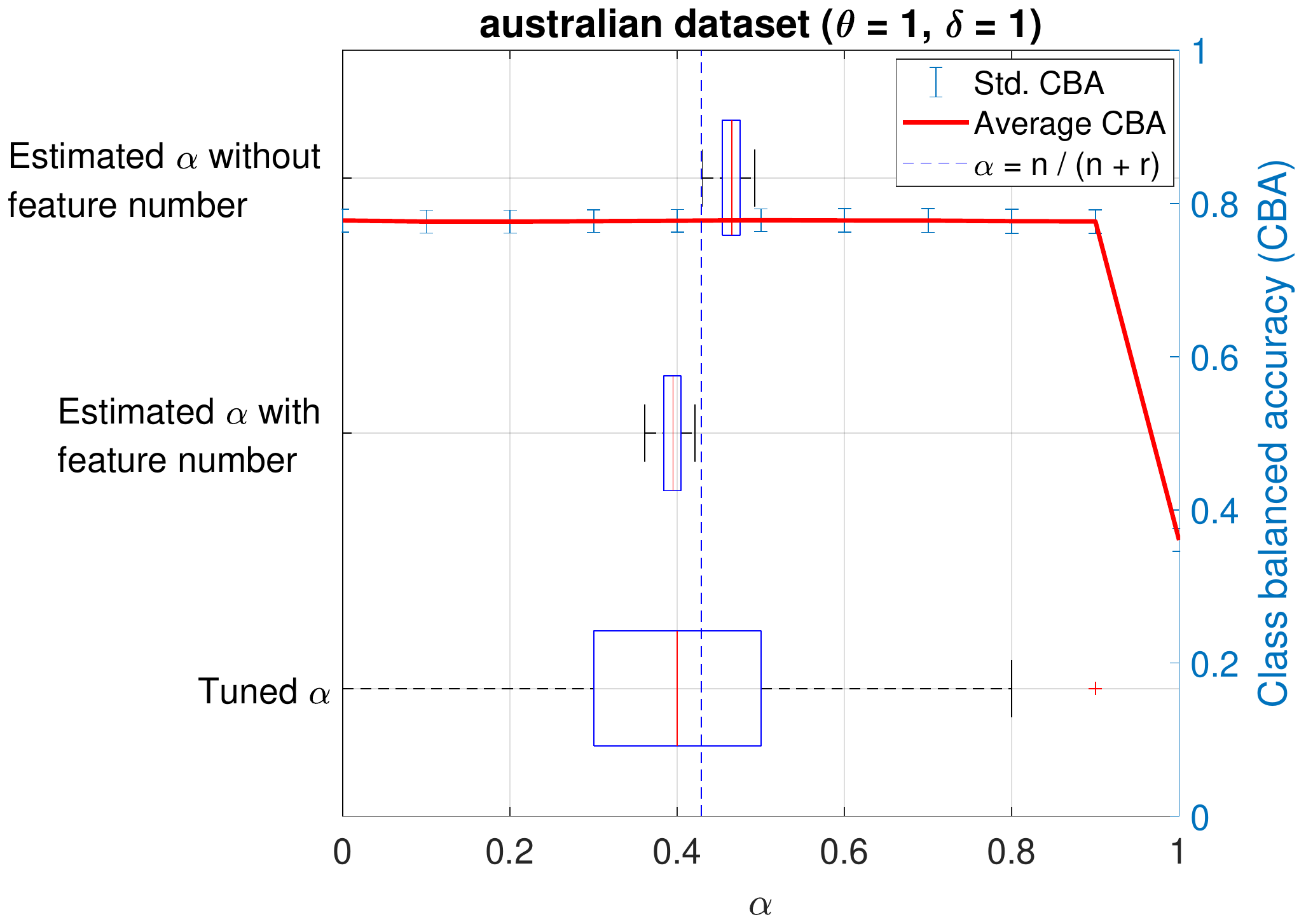}}
	\end{subfloat}
	\hfill%
	\begin{subfloat}[cmc]{
			\includegraphics[width=0.32\textwidth,height=0.18\textheight]{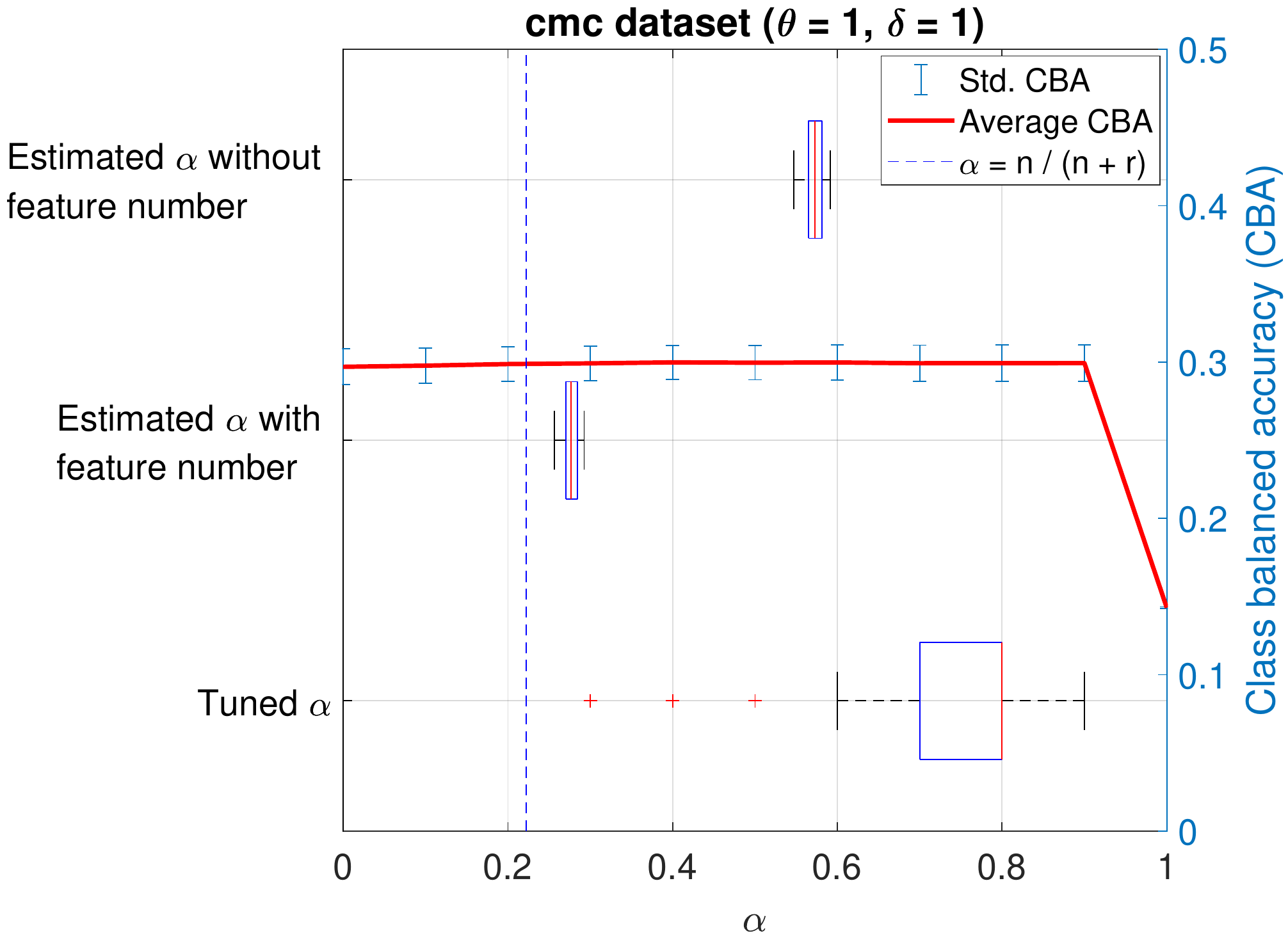}}
	\end{subfloat}
	\hfill%
	\begin{subfloat}[dermatology]{
			\includegraphics[width=0.32\textwidth,height=0.18\textheight]{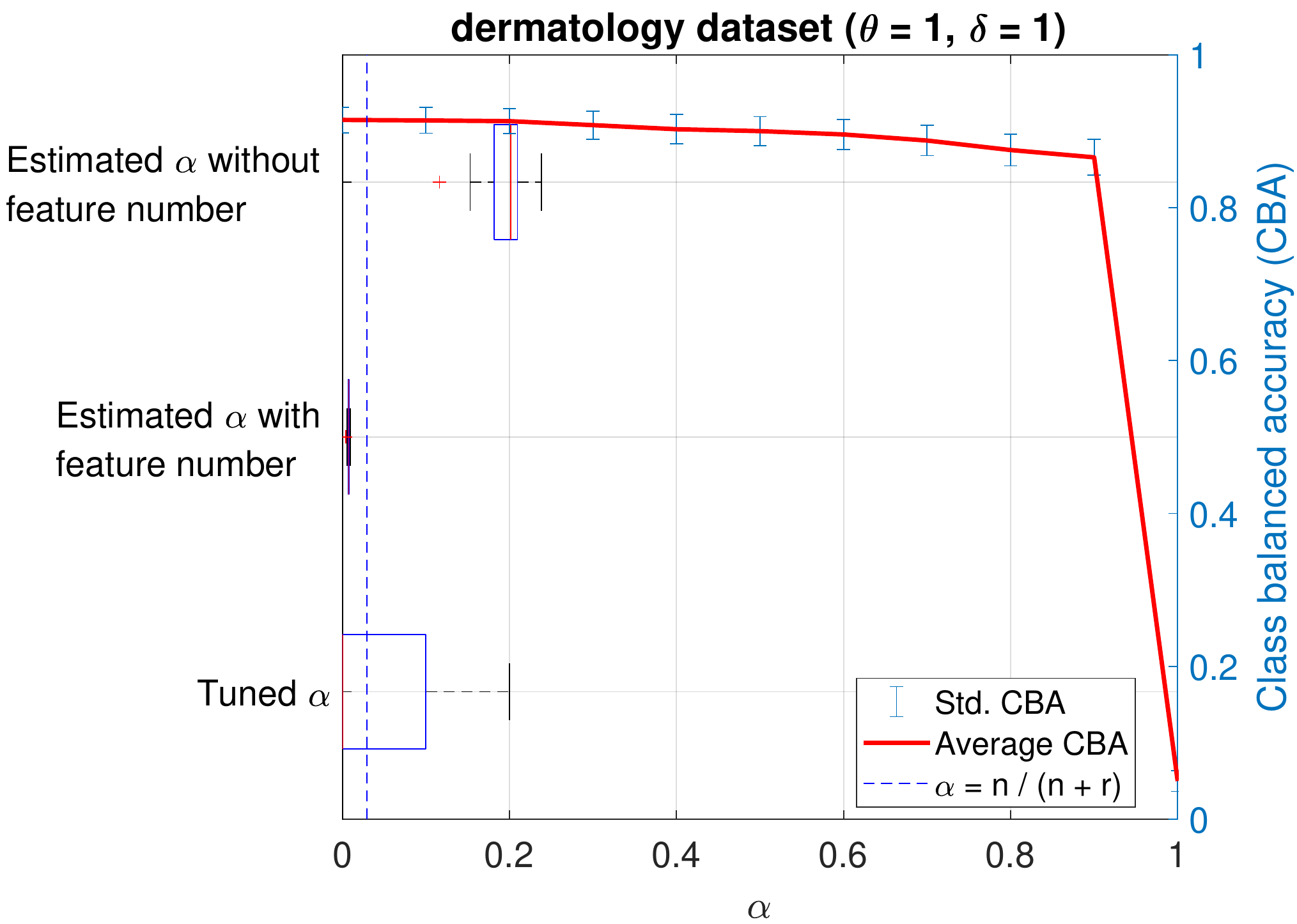}}
	\end{subfloat}
	\hfill%
	\begin{subfloat}[flag]{
			\includegraphics[width=0.32\textwidth,height=0.18\textheight]{flag_11.pdf}}
	\end{subfloat}
	\hfill%
	\begin{subfloat}[german]{
			\includegraphics[width=0.32\textwidth,height=0.18\textheight]{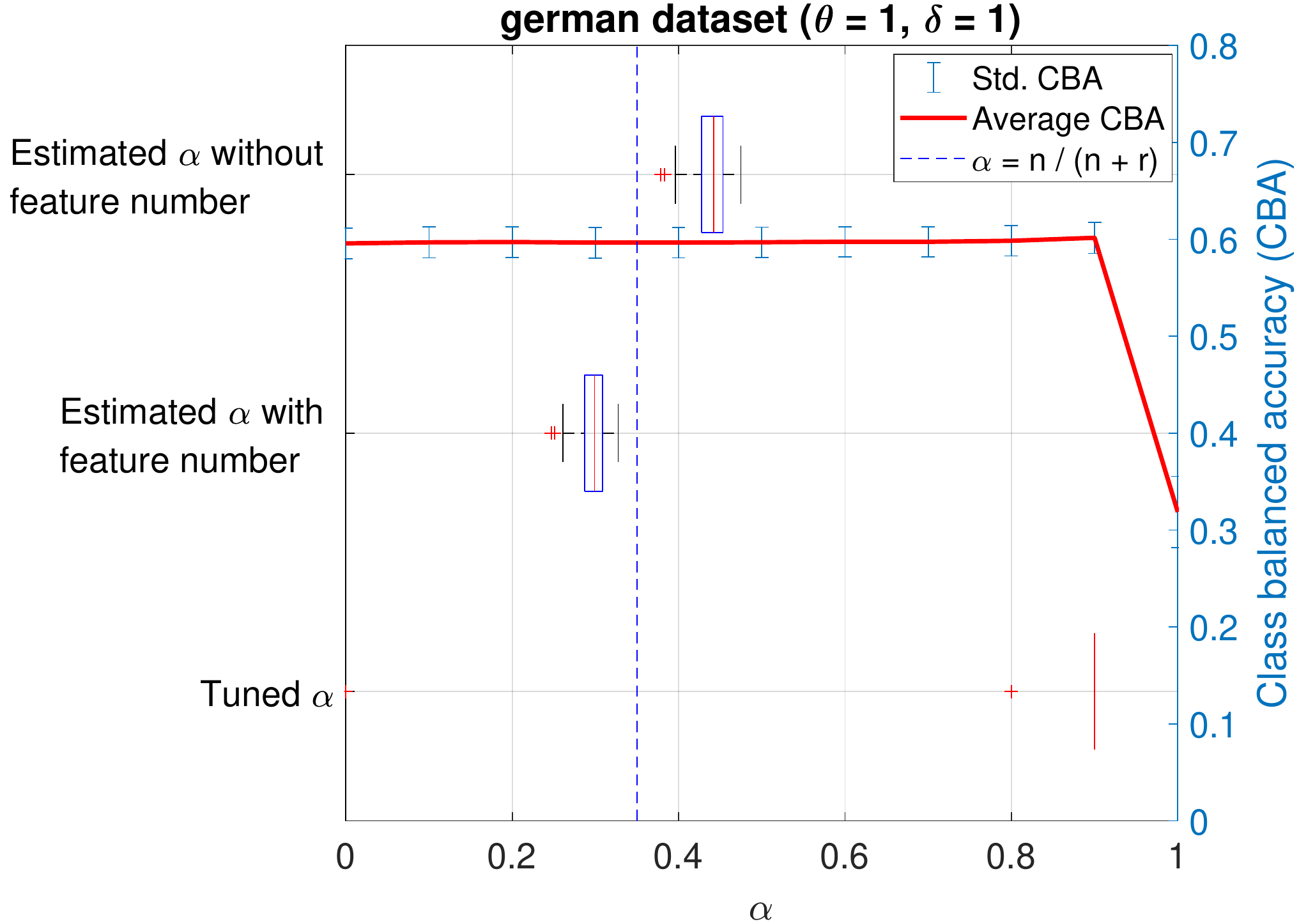}}
	\end{subfloat}
	\hfill
	\begin{subfloat}[heart]{
			\includegraphics[width=0.32\textwidth,height=0.18\textheight]{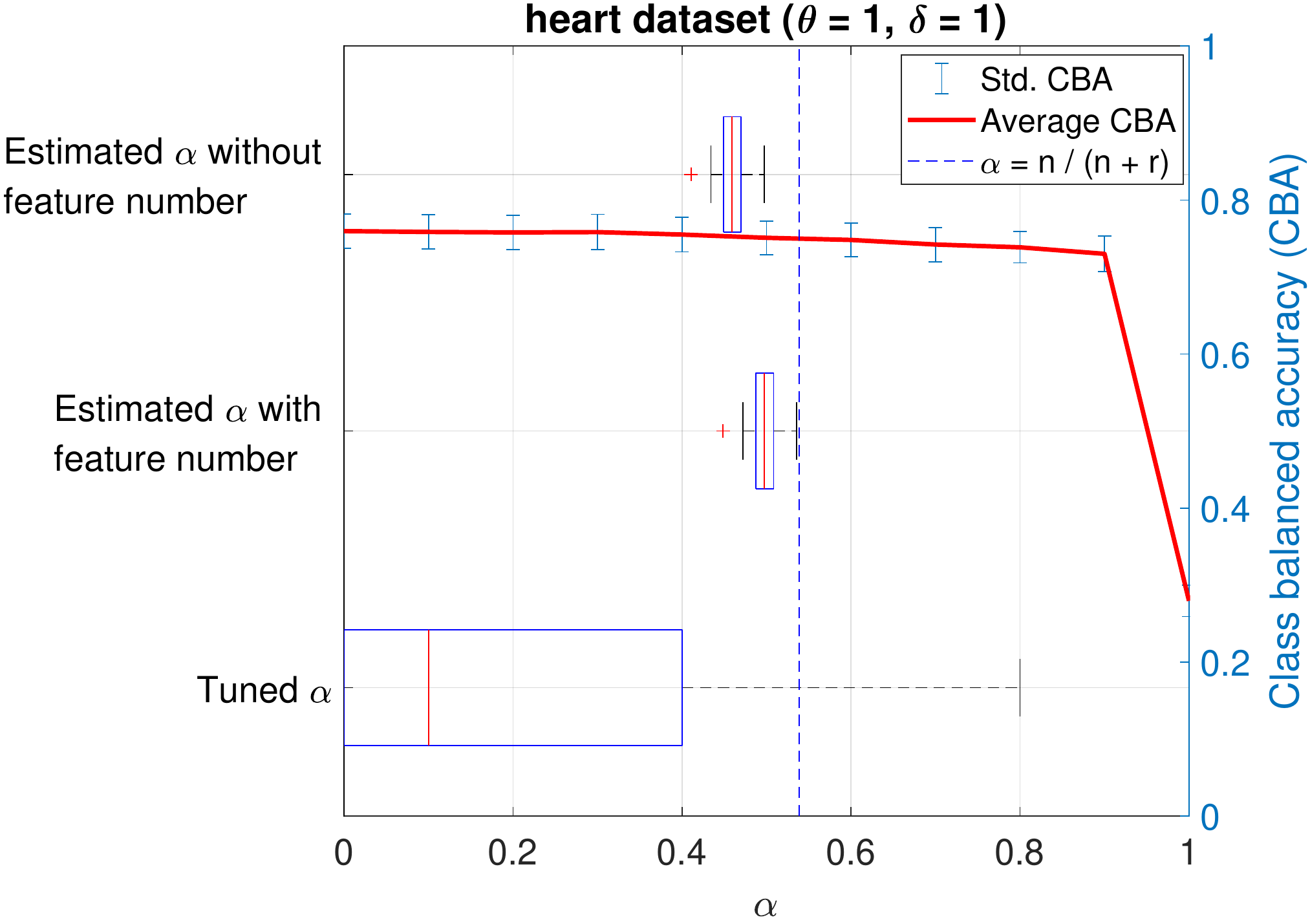}}
	\end{subfloat}
	\hfill
	\begin{subfloat}[japanese credit]{
			\includegraphics[width=0.32\textwidth,height=0.18\textheight]{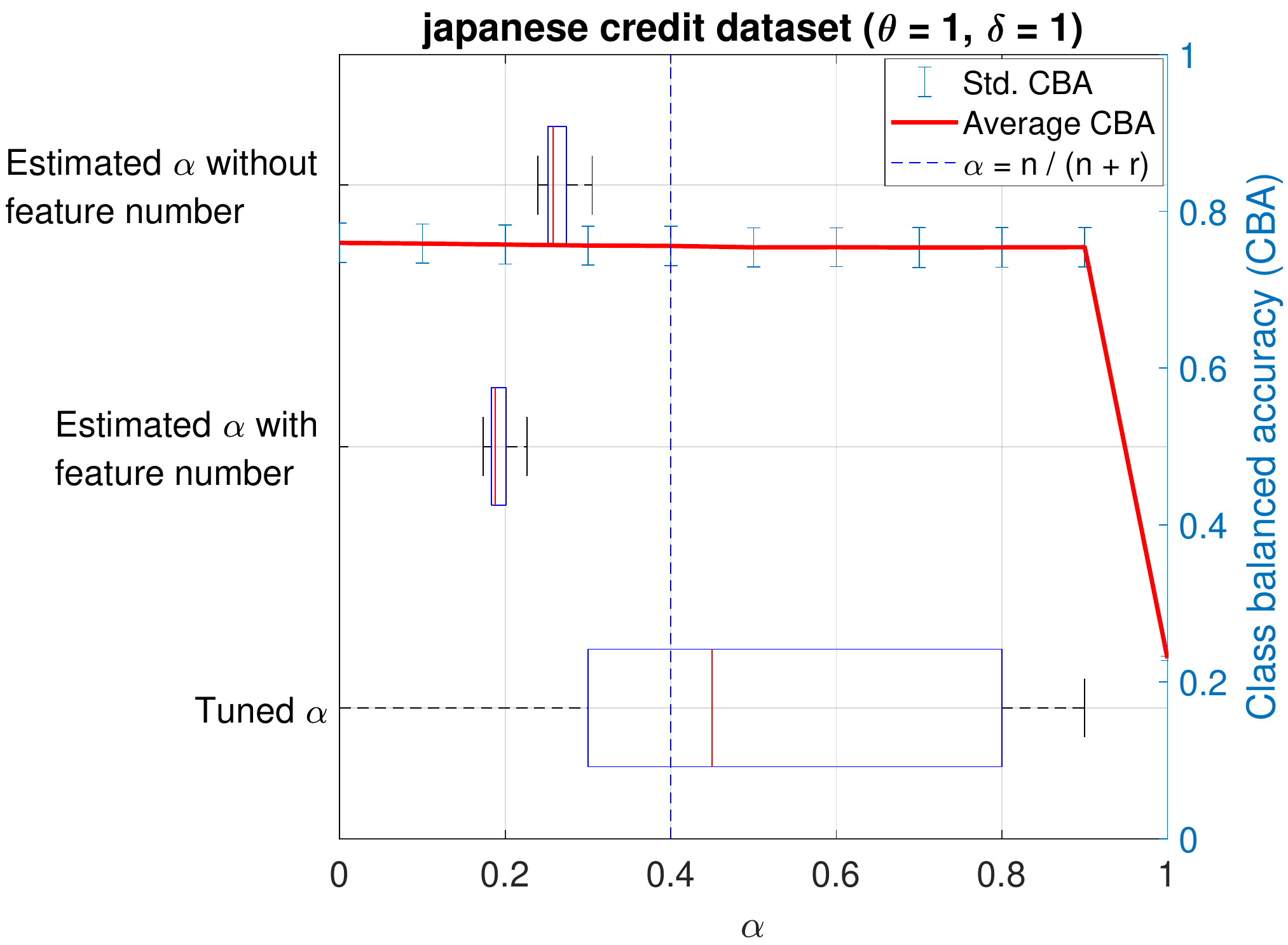}}
	\end{subfloat}
	\hfill
	\begin{subfloat}[post operative]{
			\includegraphics[width=0.32\textwidth,height=0.18\textheight]{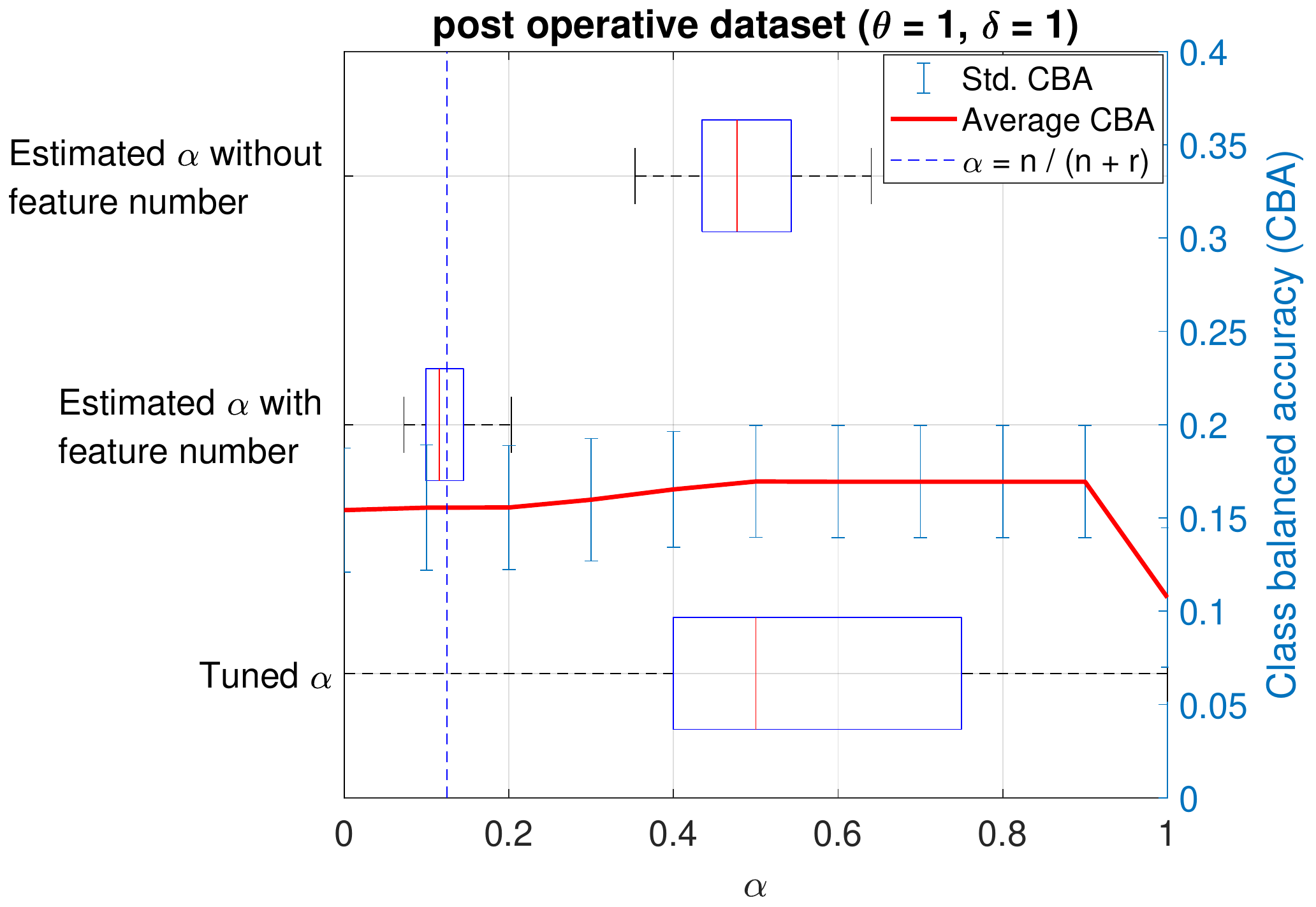}}
	\end{subfloat}
	\hfill
	\begin{subfloat}[tae]{
			\includegraphics[width=0.32\textwidth,height=0.18\textheight]{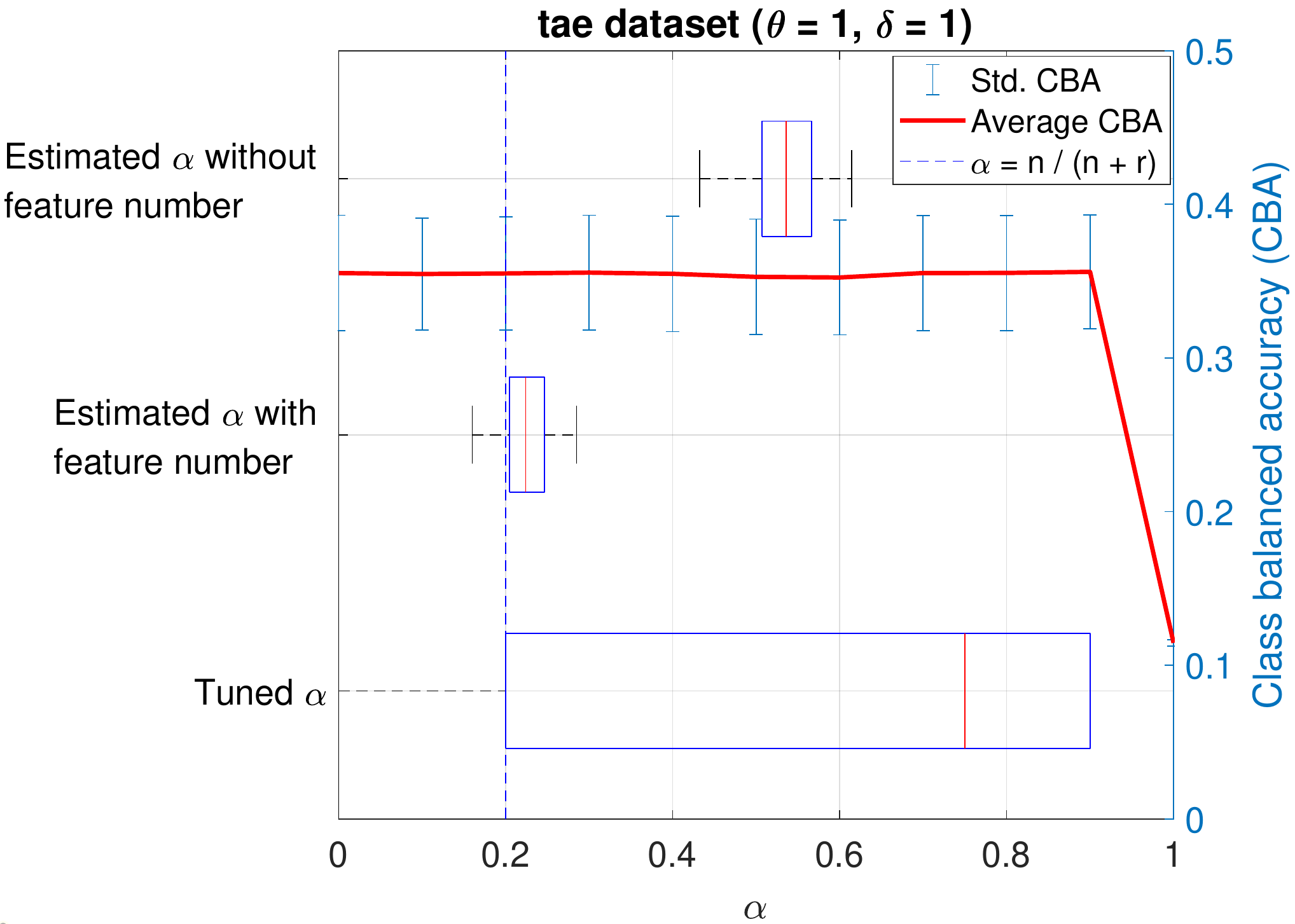}}
	\end{subfloat}
	\begin{subfloat}[zoo]{
			\includegraphics[width=0.32\textwidth,height=0.18\textheight]{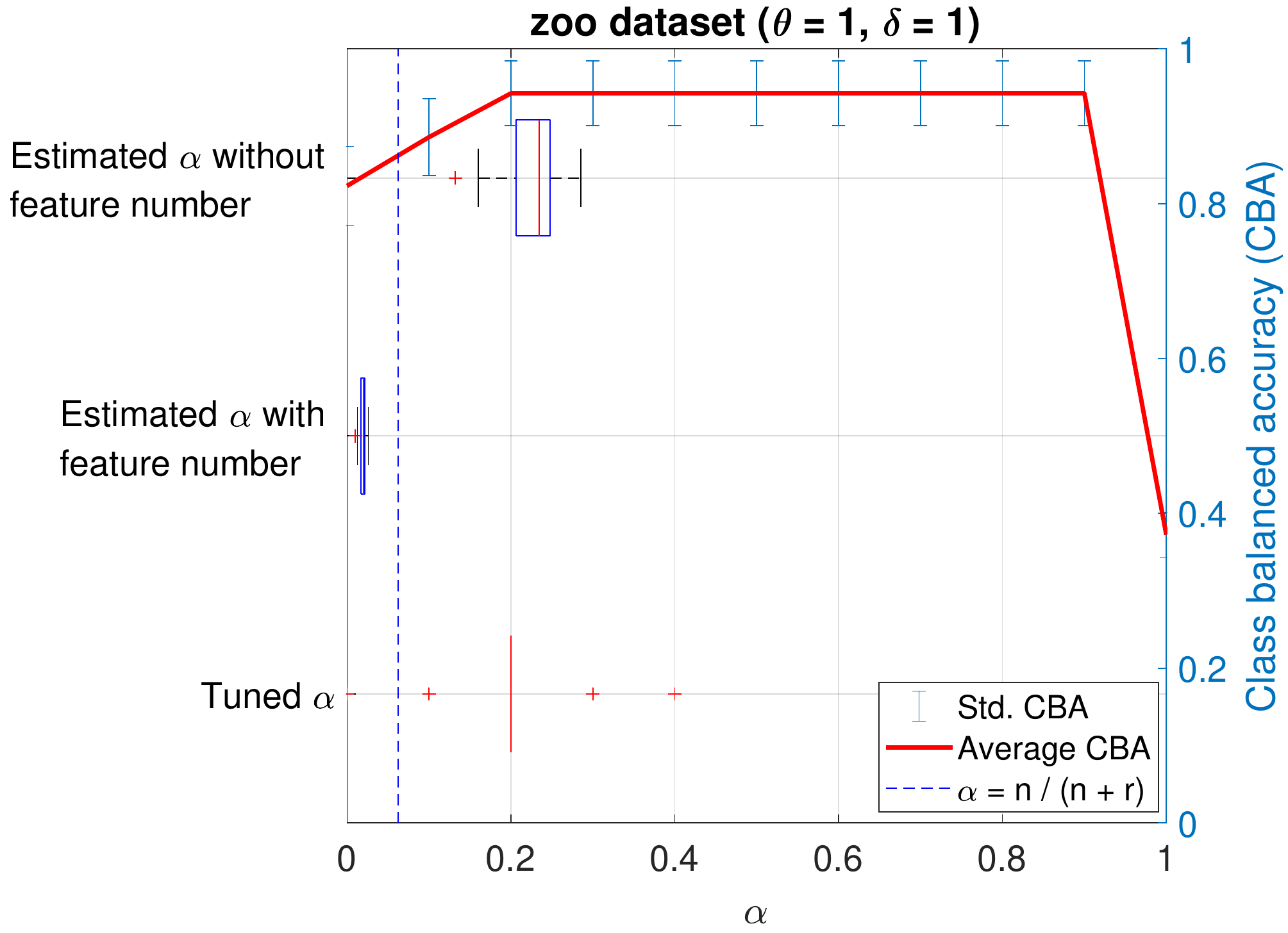}}
	\end{subfloat}
	\caption{Class balanced accuracy values and range of obtained $\alpha$ for different ways of estimating $\alpha$ ($\theta = 1, \delta = 1$).}
	\label{Fig_S7}
\end{figure}

\begin{figure}[!ht]
	\begin{subfloat}[abalone]{
			\includegraphics[width=0.32\textwidth, height=0.18\textheight]{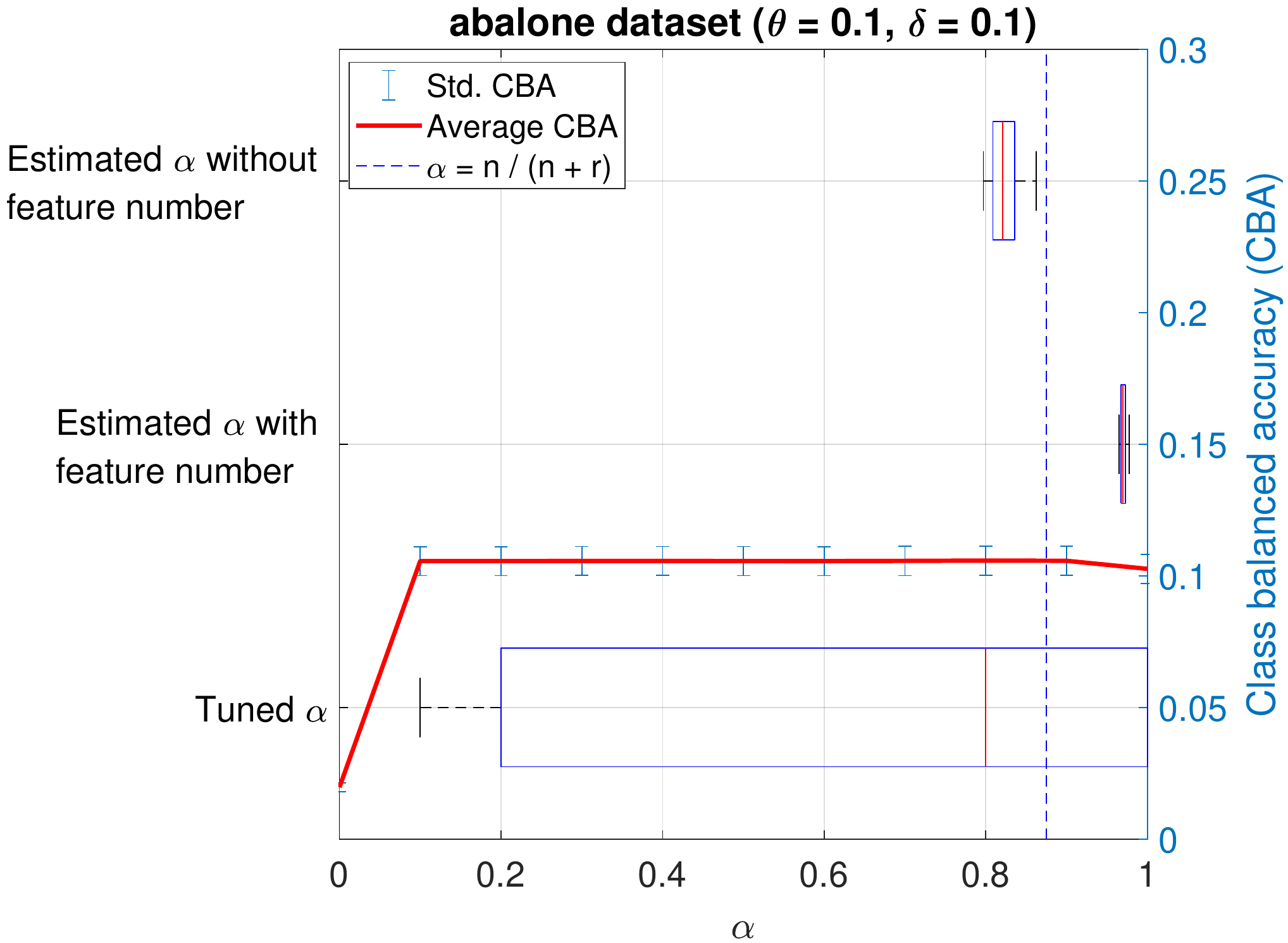}}
	\end{subfloat}
	\hfill%
	\begin{subfloat}[australian]{
			\includegraphics[width=0.32\textwidth,height=0.18\textheight]{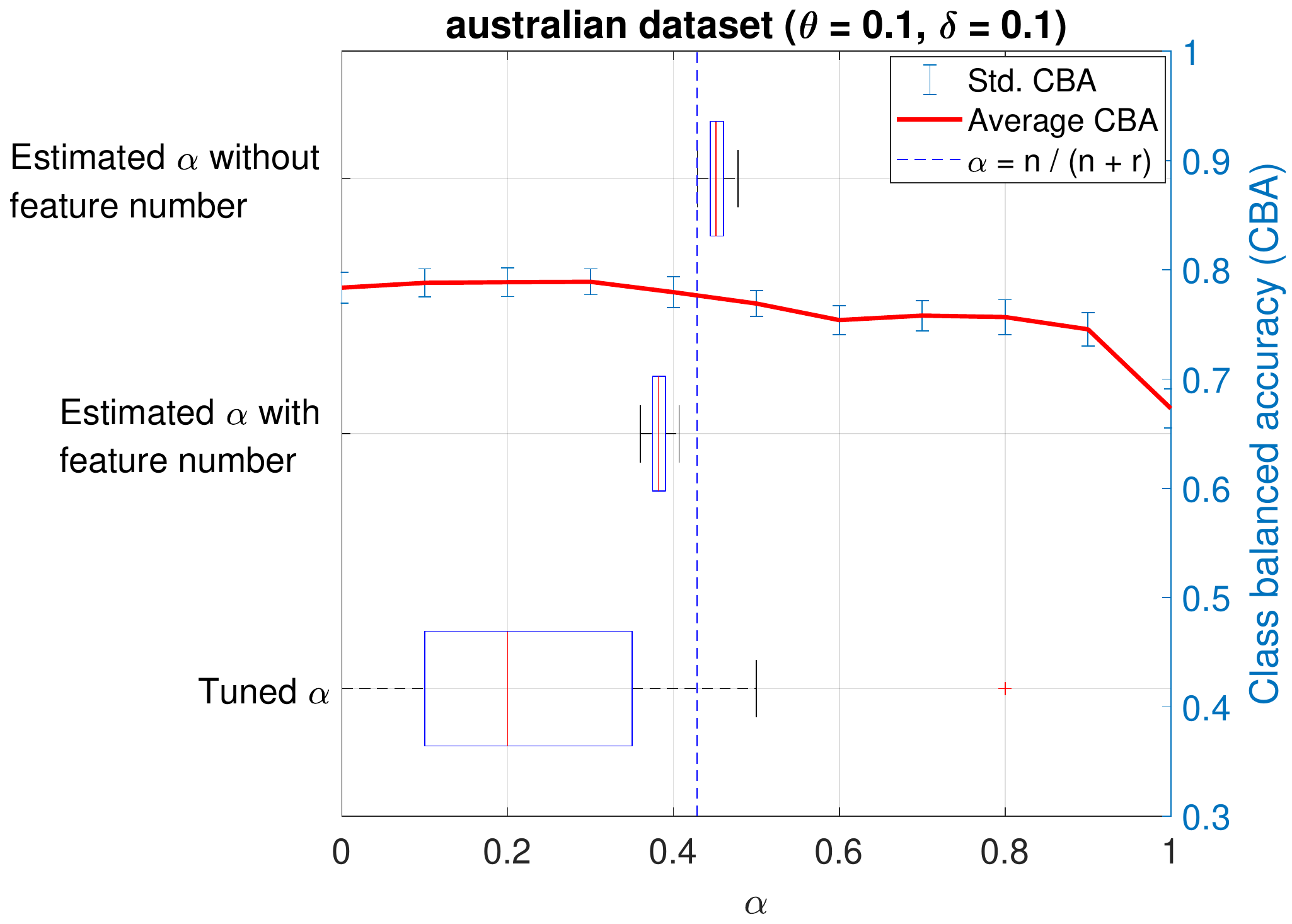}}
	\end{subfloat}
	\hfill%
	\begin{subfloat}[cmc]{
			\includegraphics[width=0.32\textwidth,height=0.18\textheight]{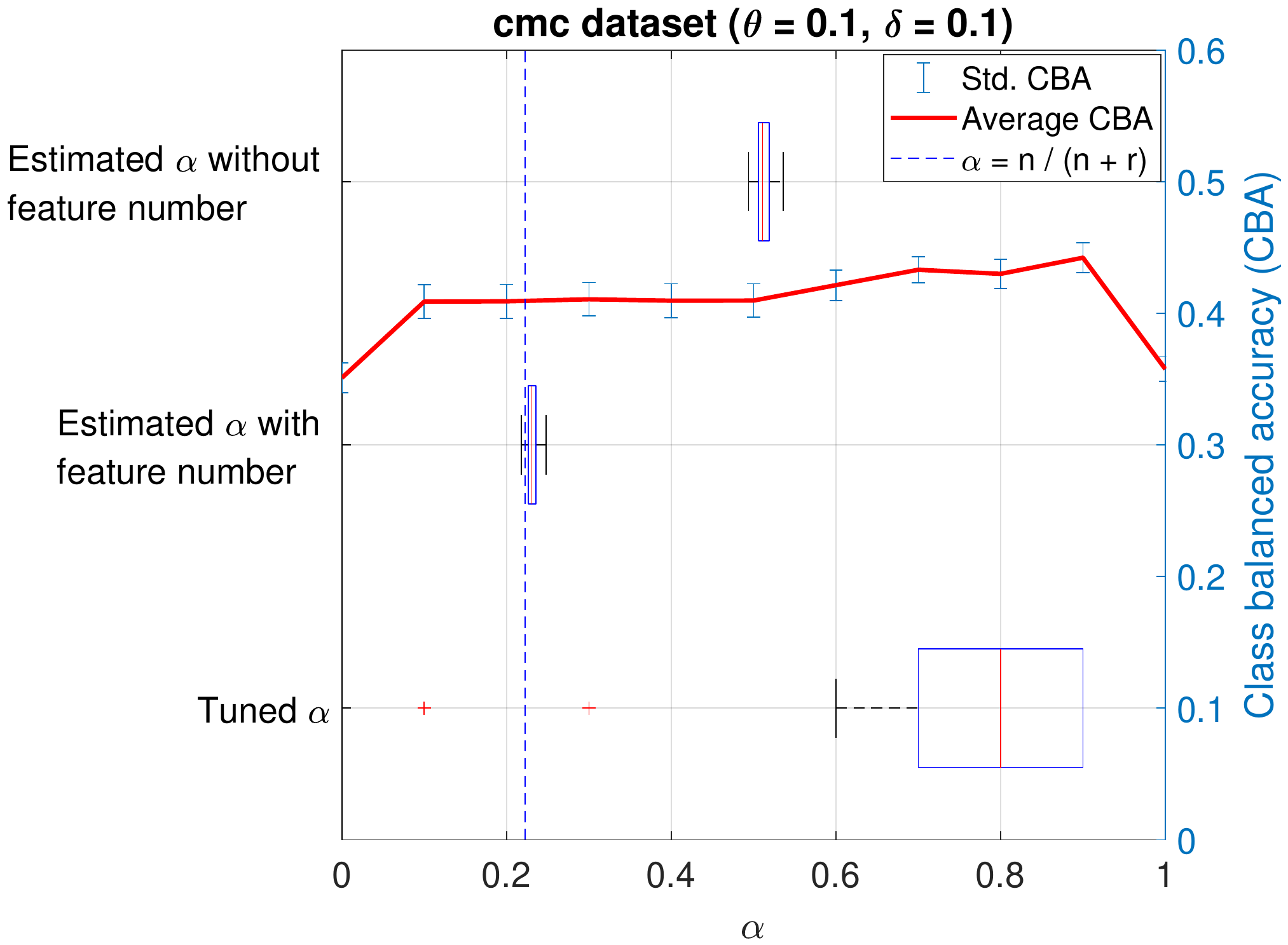}}
	\end{subfloat}
	\hfill%
	\begin{subfloat}[dermatology]{
			\includegraphics[width=0.32\textwidth,height=0.18\textheight]{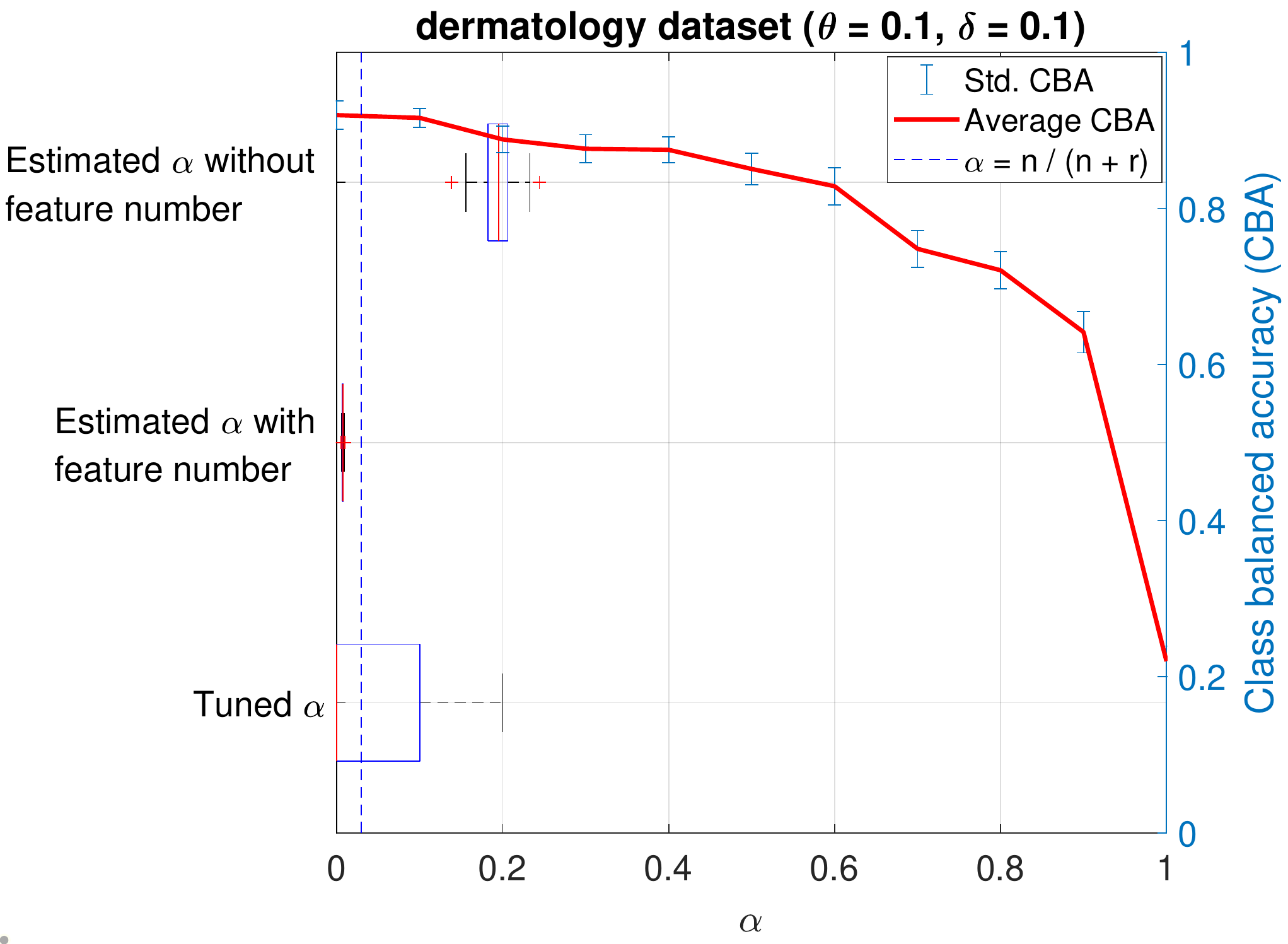}}
	\end{subfloat}
	\hfill%
	\begin{subfloat}[flag]{
			\includegraphics[width=0.32\textwidth,height=0.18\textheight]{flag_v1_0101.pdf}}
	\end{subfloat}
	\hfill%
	\begin{subfloat}[german]{
			\includegraphics[width=0.32\textwidth,height=0.18\textheight]{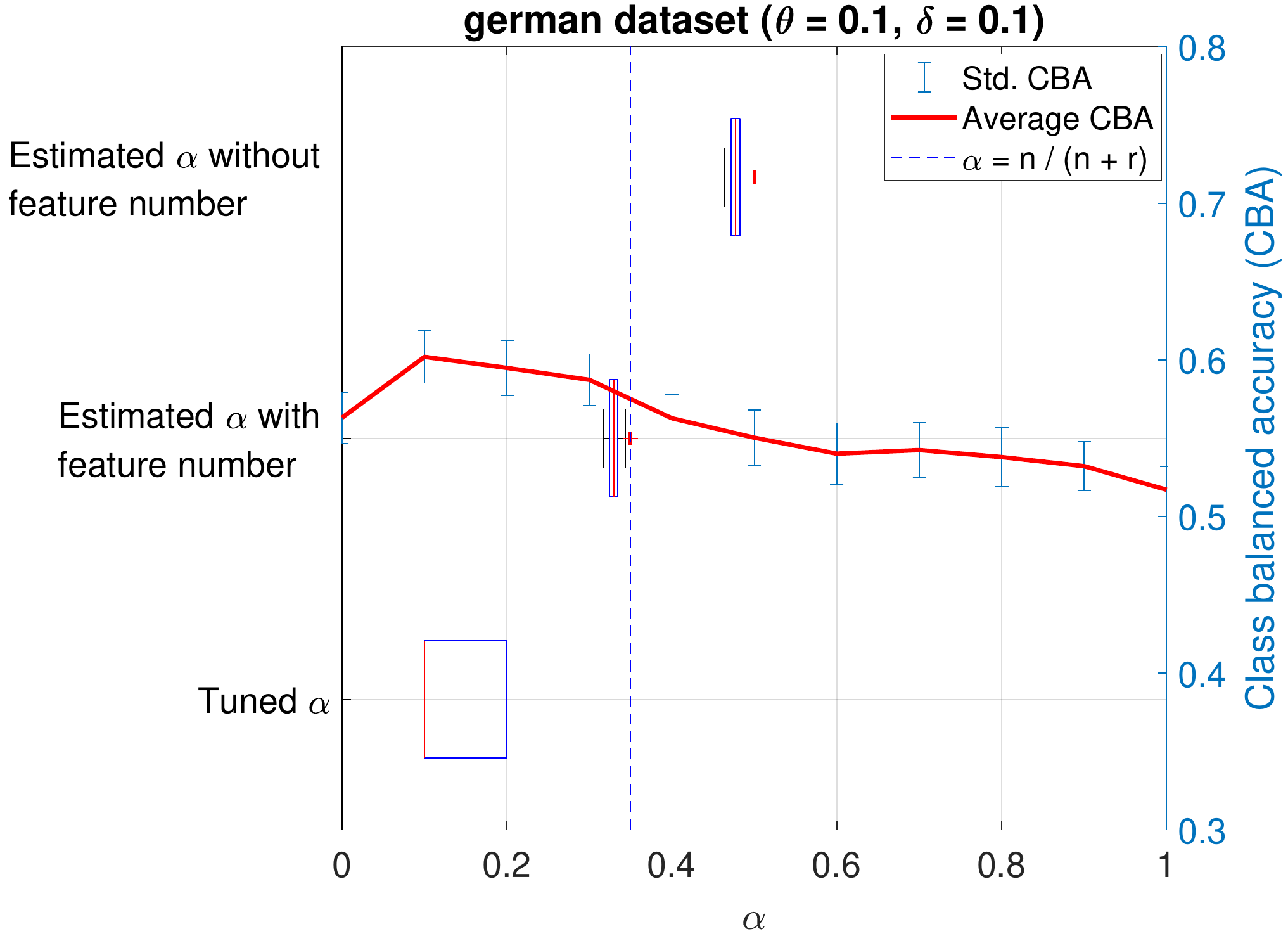}}
	\end{subfloat}
	\hfill
	\begin{subfloat}[heart]{
			\includegraphics[width=0.32\textwidth,height=0.18\textheight]{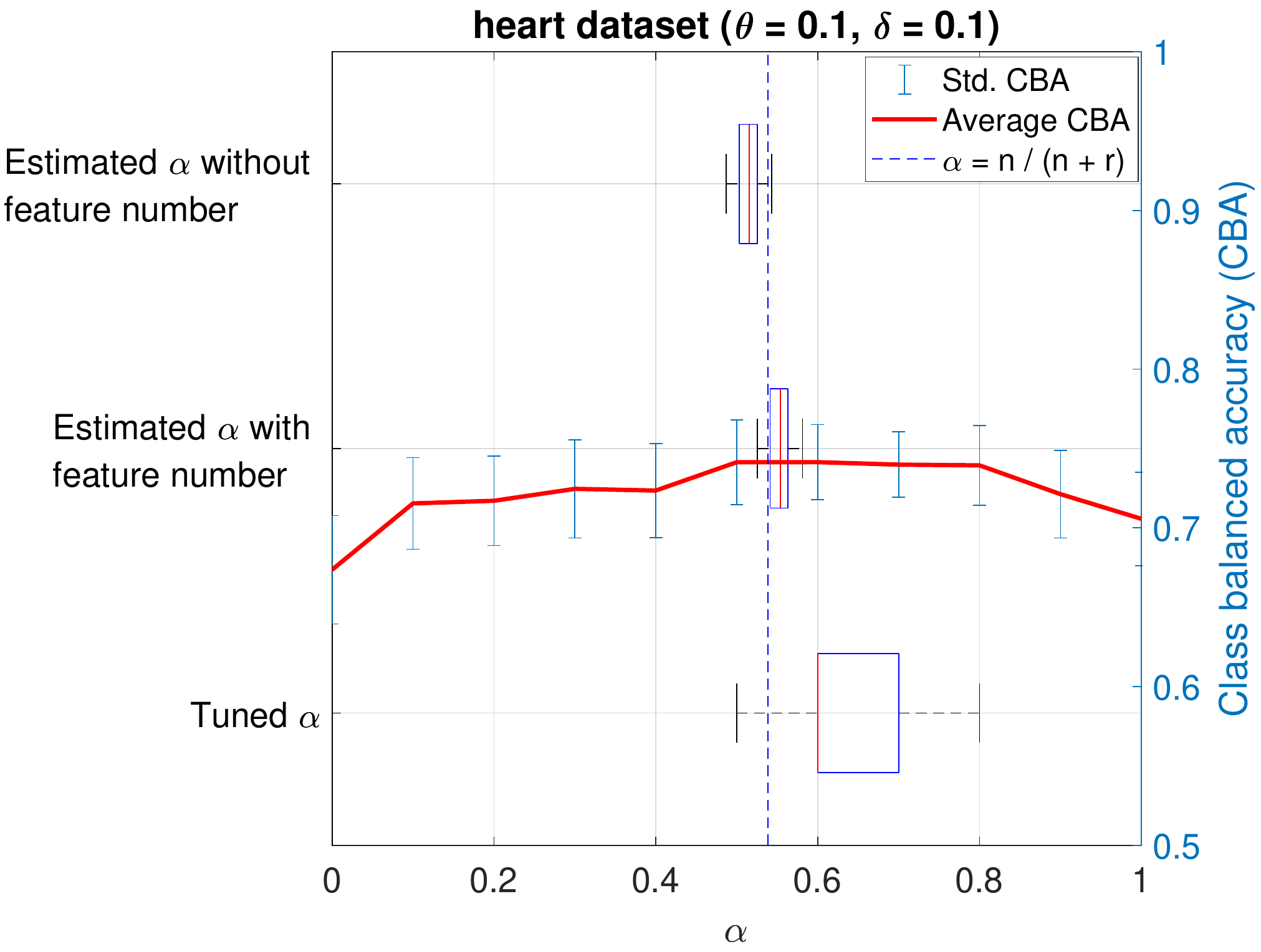}}
	\end{subfloat}
	\hfill
	\begin{subfloat}[japanese credit]{
			\includegraphics[width=0.32\textwidth,height=0.18\textheight]{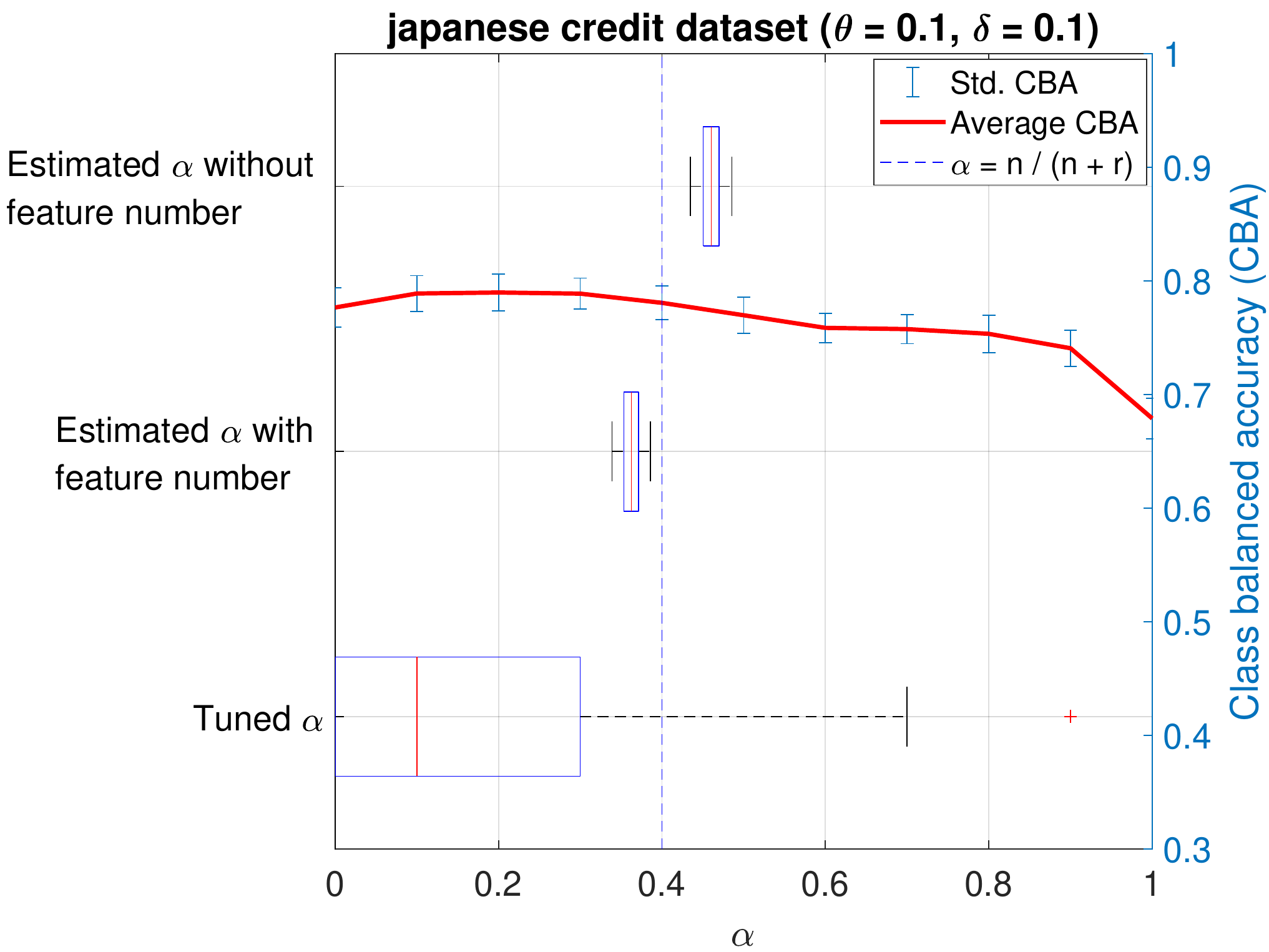}}
	\end{subfloat}
	\hfill
	\begin{subfloat}[post operative]{
			\includegraphics[width=0.32\textwidth,height=0.18\textheight]{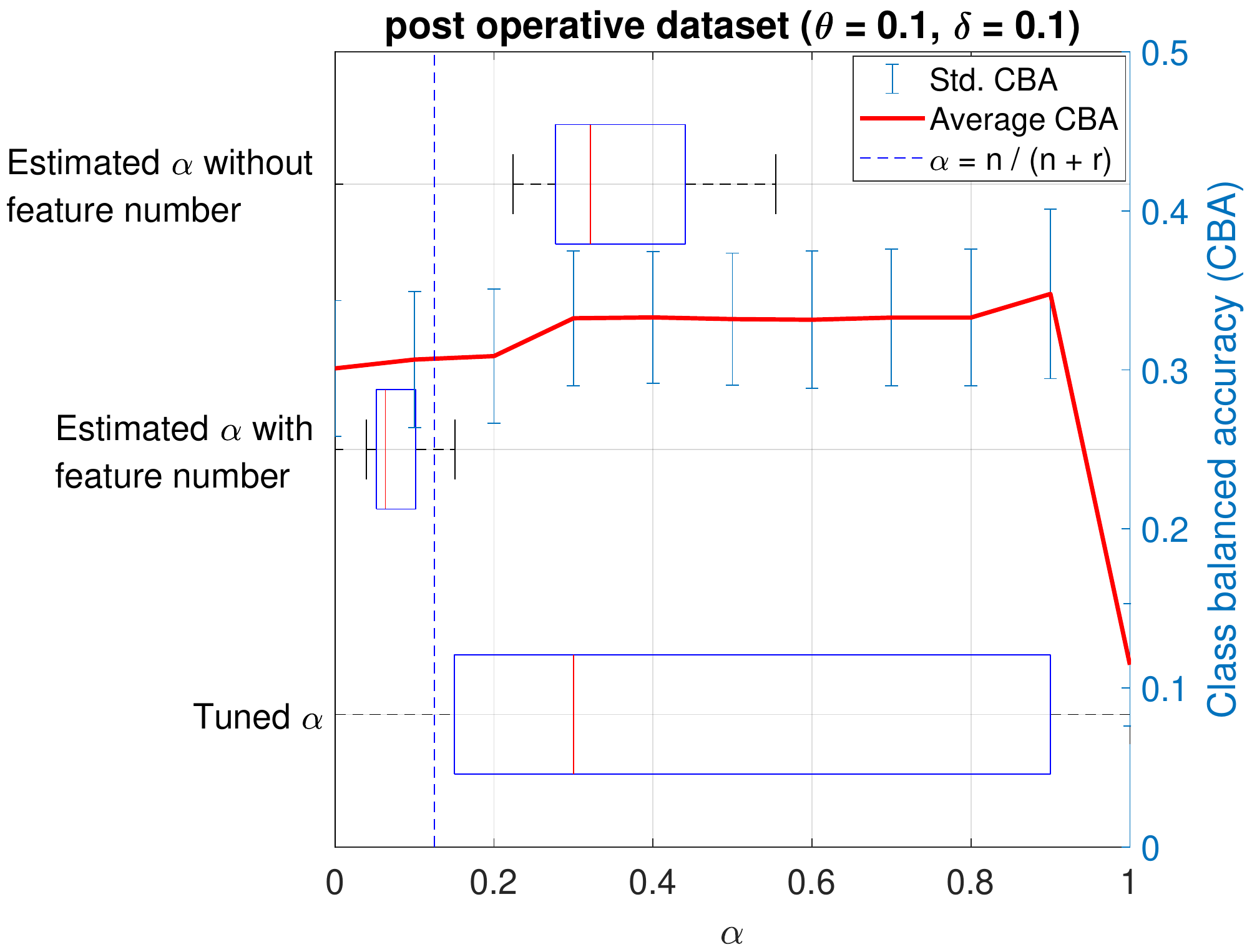}}
	\end{subfloat}
	\hfill
	\begin{subfloat}[tae]{
			\includegraphics[width=0.32\textwidth,height=0.18\textheight]{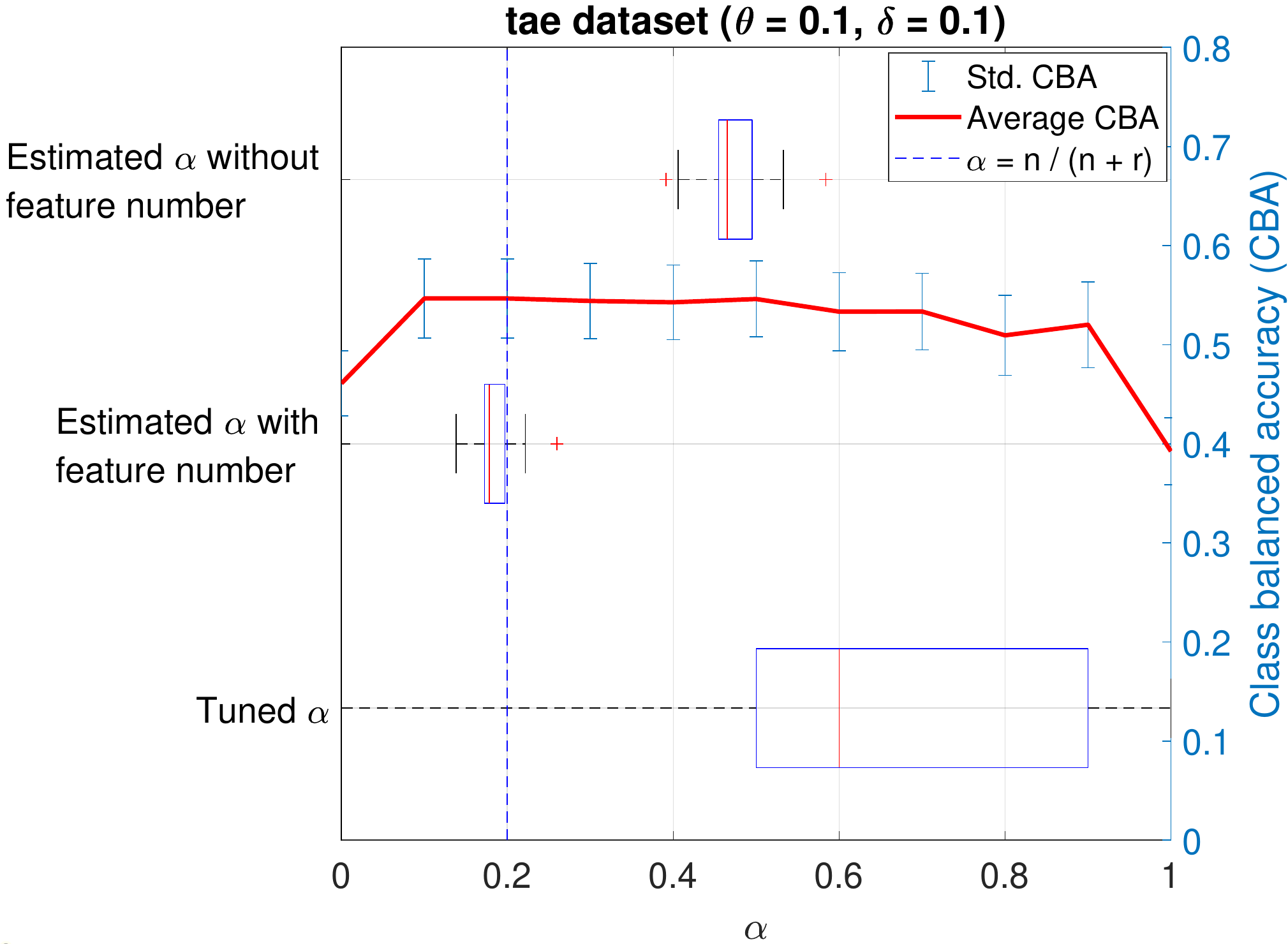}}
	\end{subfloat}
	\begin{subfloat}[zoo]{
			\includegraphics[width=0.32\textwidth,height=0.18\textheight]{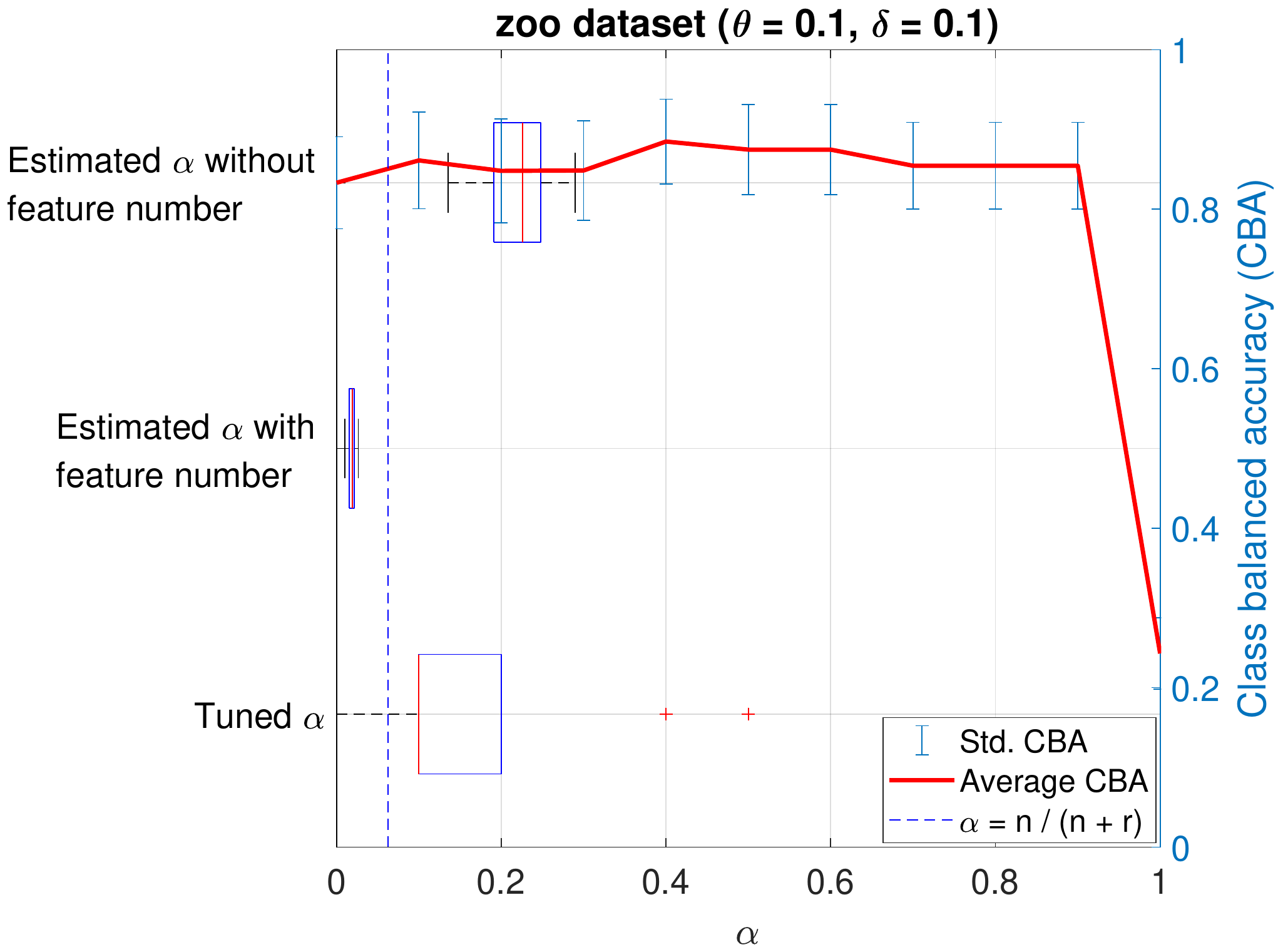}}
	\end{subfloat}
	\hfill
	\caption{Class balanced accuracy values and range of obtained $\alpha$ for different ways of estimating $\alpha$ (using EIOL-GFMM-v1 with $\theta = 0.1, \delta = 0.1$).}
	\label{Fig_S8}
\end{figure}

\begin{figure}[!ht]
	\begin{subfloat}[abalone]{
			\includegraphics[width=0.32\textwidth, height=0.18\textheight]{abalone_v1_0101.pdf}}
	\end{subfloat}
	\hfill%
	\begin{subfloat}[australian]{
			\includegraphics[width=0.32\textwidth,height=0.18\textheight]{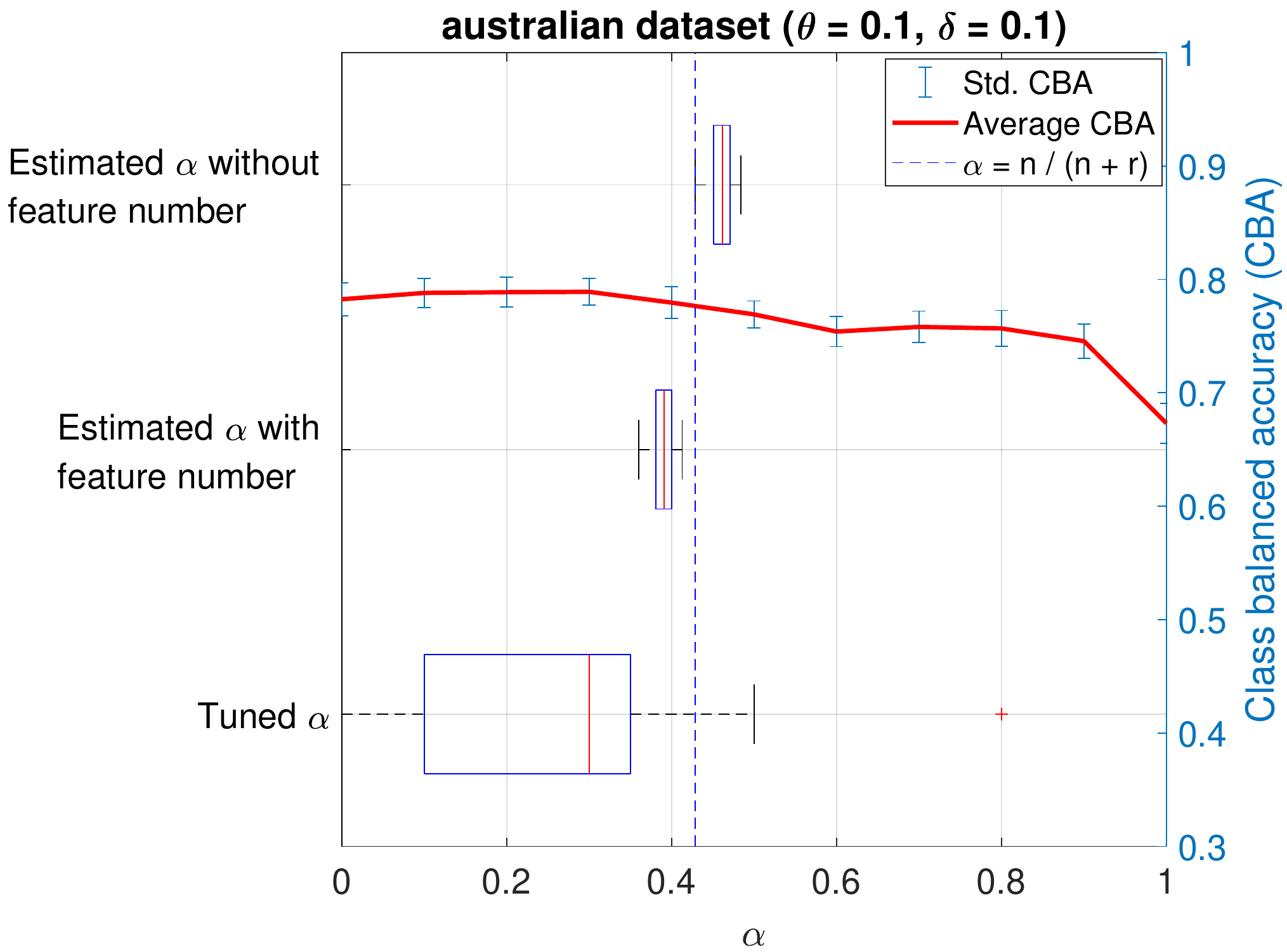}}
	\end{subfloat}
	\hfill%
	\begin{subfloat}[cmc]{
			\includegraphics[width=0.32\textwidth,height=0.18\textheight]{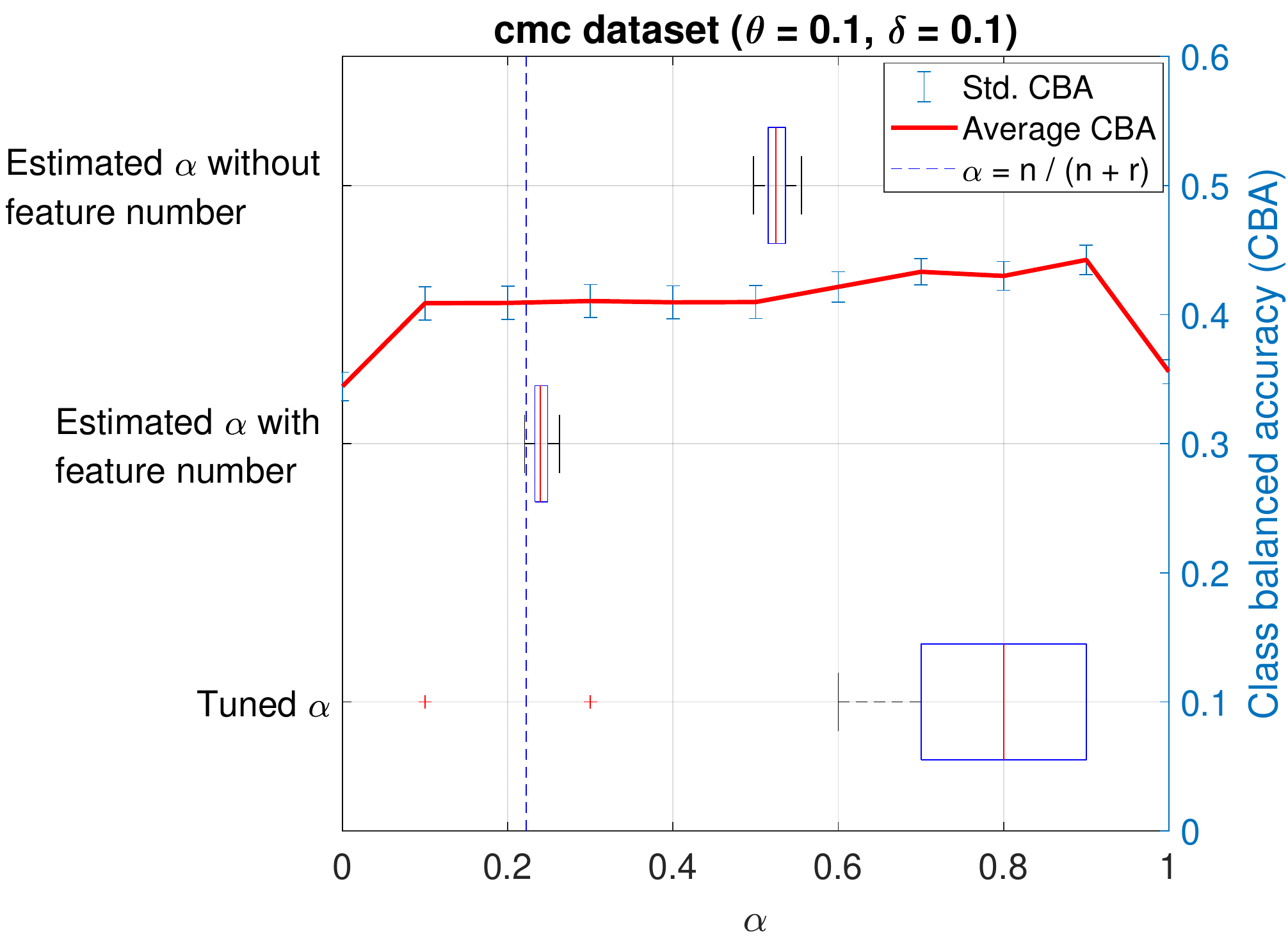}}
	\end{subfloat}
	\hfill%
	\begin{subfloat}[dermatology]{
			\includegraphics[width=0.32\textwidth,height=0.18\textheight]{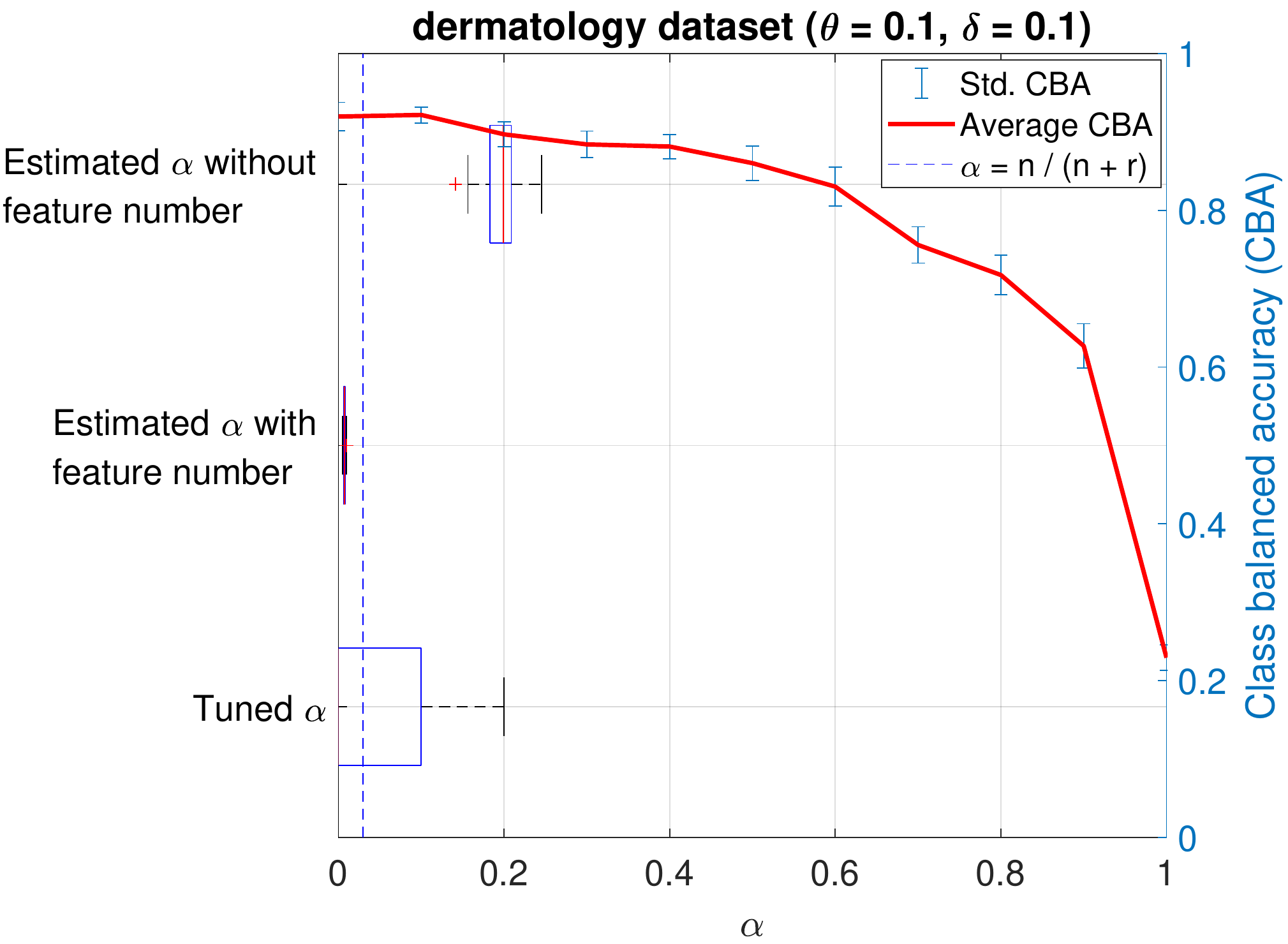}}
	\end{subfloat}
	\hfill%
	\begin{subfloat}[flag]{
			\includegraphics[width=0.32\textwidth,height=0.18\textheight]{flag_v2_0101.pdf}}
	\end{subfloat}
	\hfill%
	\begin{subfloat}[german]{
			\includegraphics[width=0.32\textwidth,height=0.18\textheight]{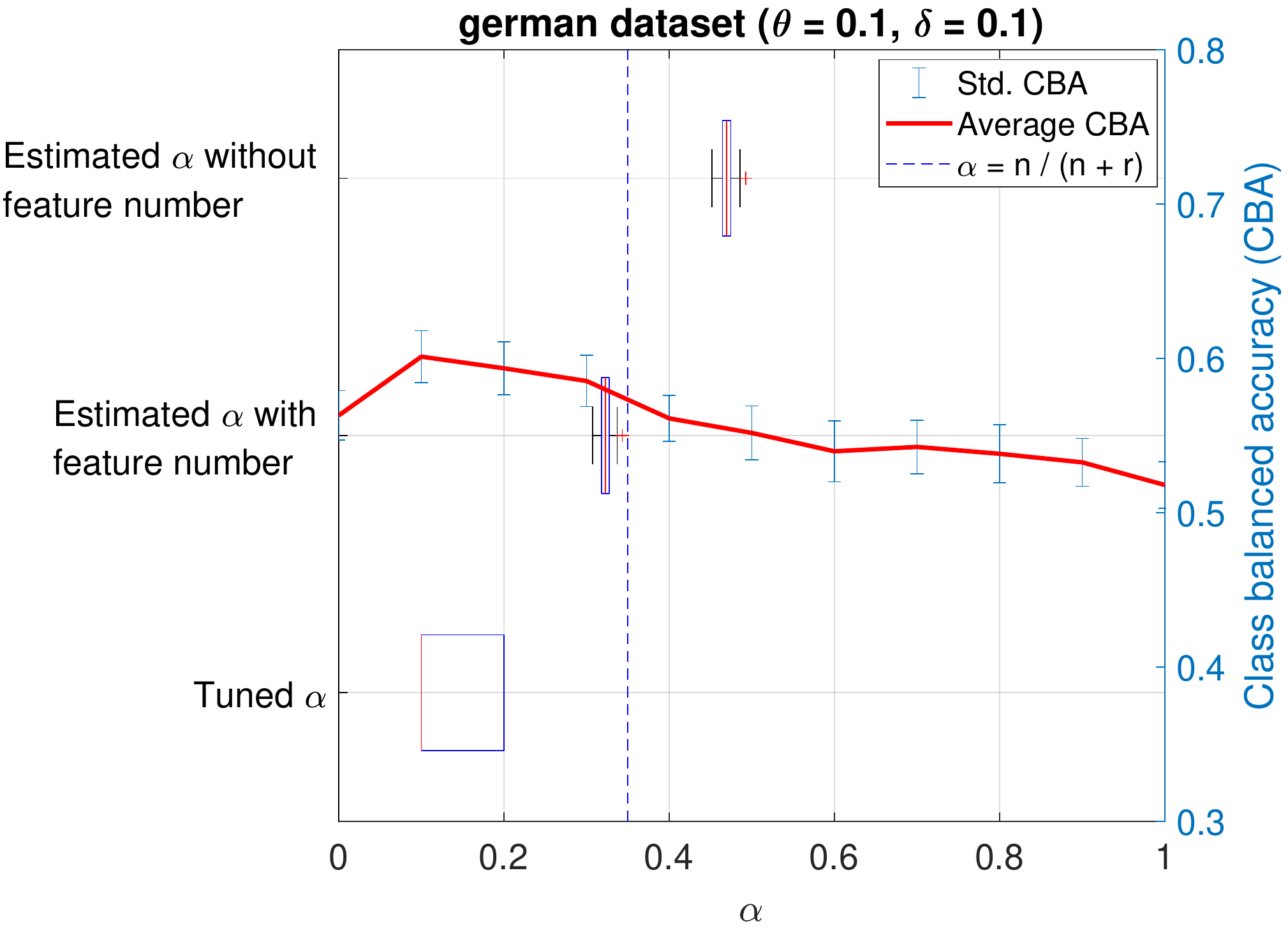}}
	\end{subfloat}
	\hfill
	\begin{subfloat}[heart]{
			\includegraphics[width=0.32\textwidth,height=0.18\textheight]{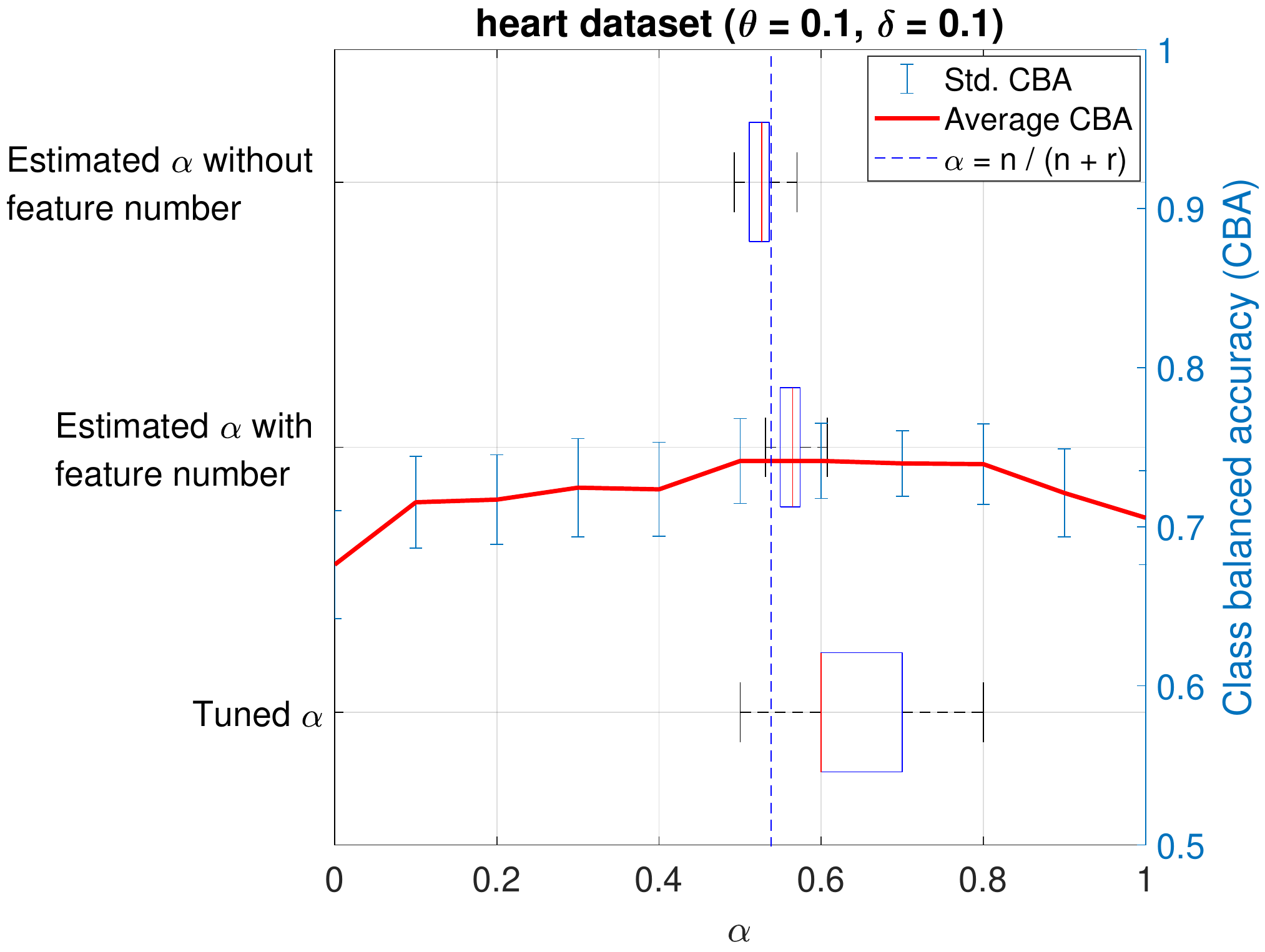}}
	\end{subfloat}
	\hfill
	\begin{subfloat}[japanese credit]{
			\includegraphics[width=0.32\textwidth,height=0.18\textheight]{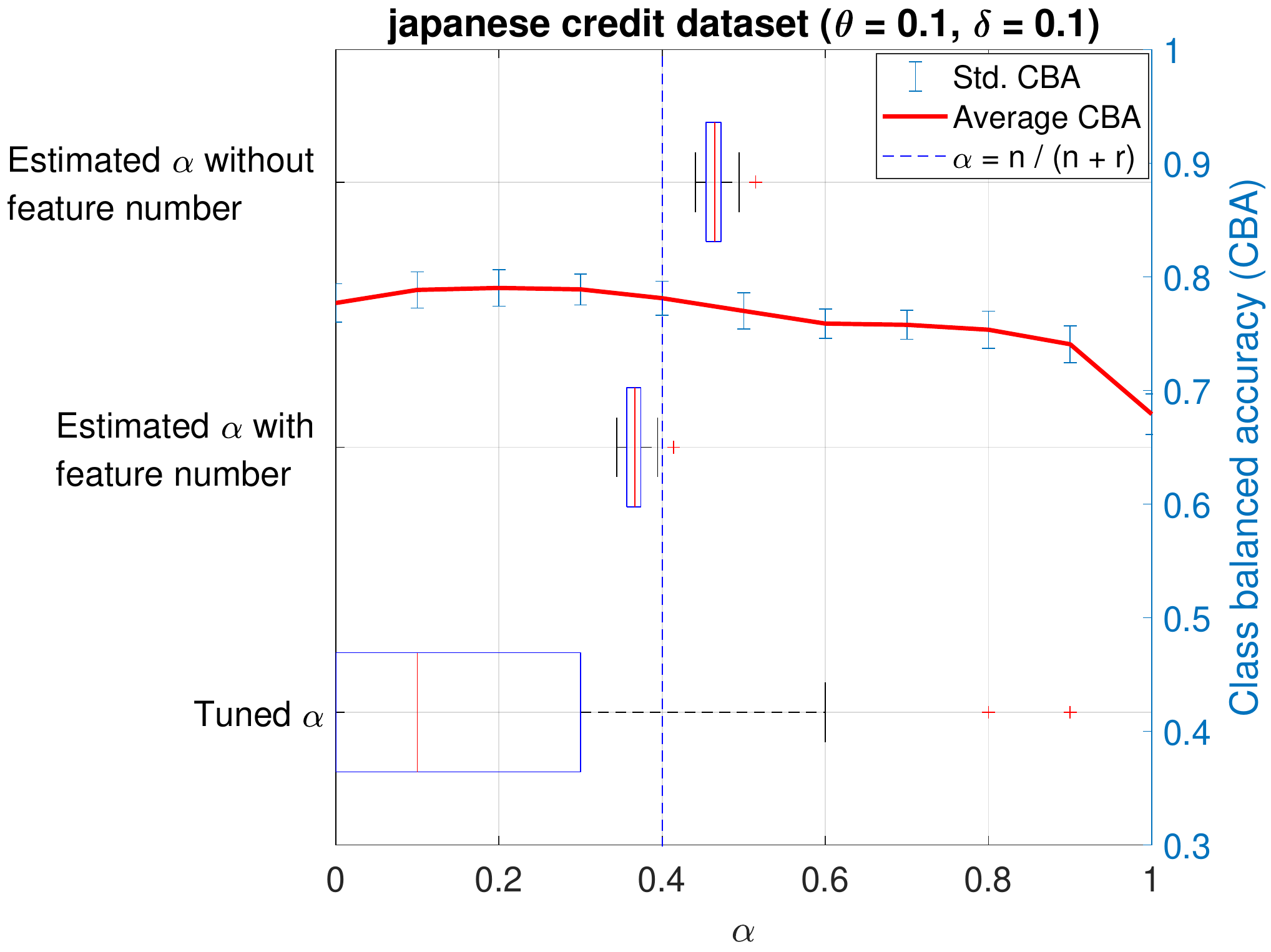}}
	\end{subfloat}
	\hfill
	\begin{subfloat}[post operative]{
			\includegraphics[width=0.32\textwidth,height=0.18\textheight]{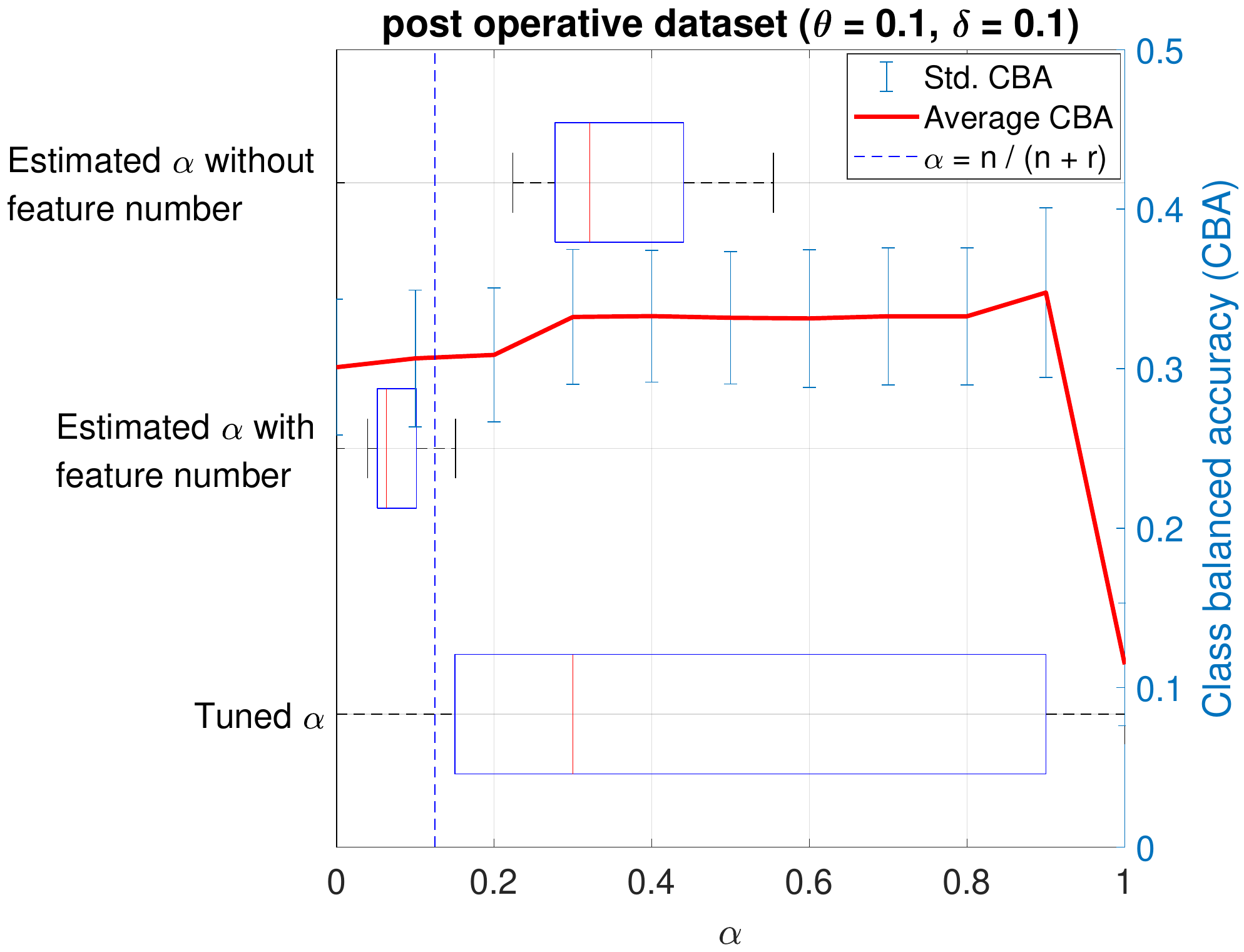}}
	\end{subfloat}
	\hfill
	\begin{subfloat}[tae]{
			\includegraphics[width=0.32\textwidth,height=0.18\textheight]{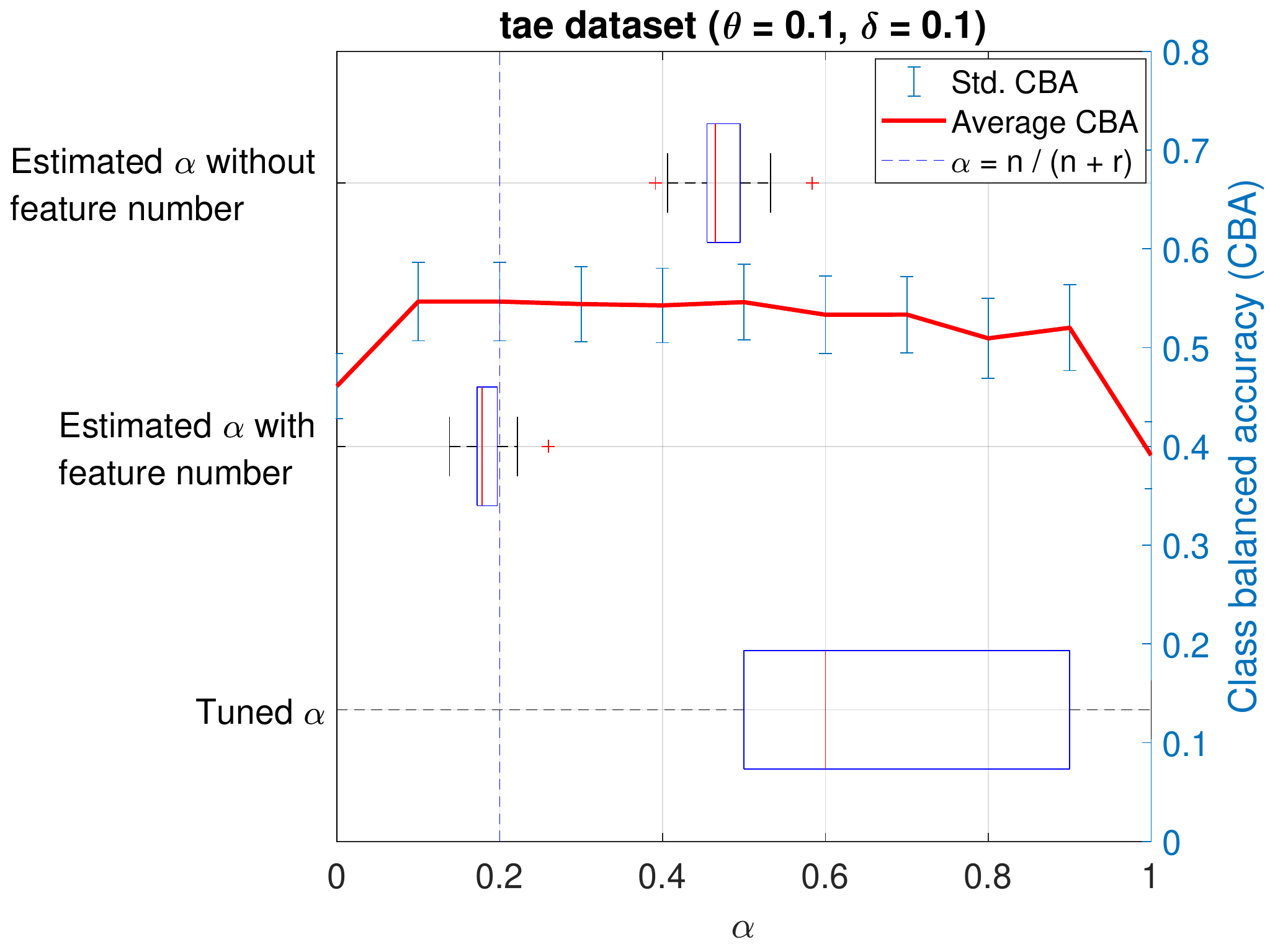}}
	\end{subfloat}
	\begin{subfloat}[zoo]{
			\includegraphics[width=0.32\textwidth,height=0.18\textheight]{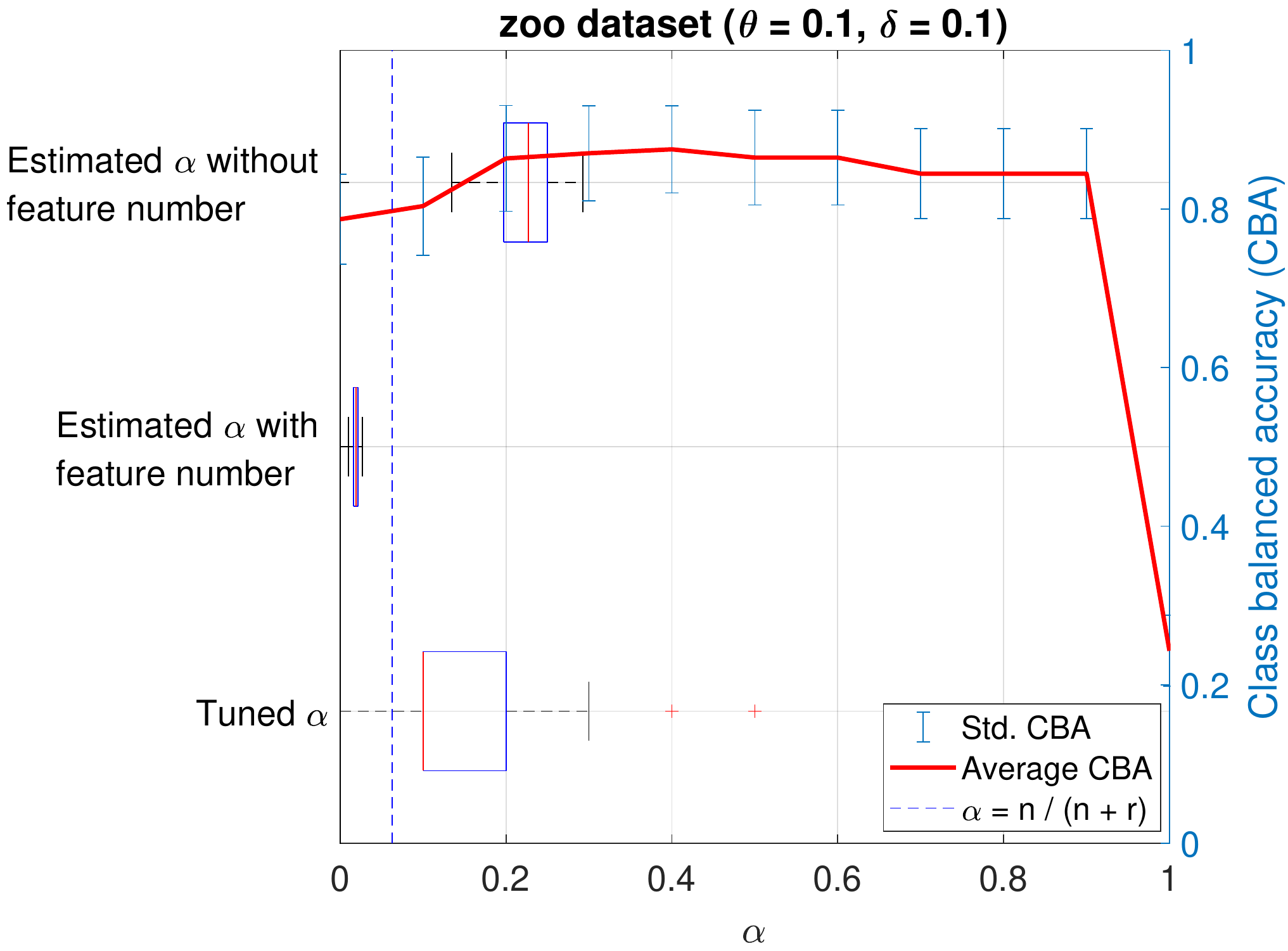}}
	\end{subfloat}
	\hfill
	\caption{Class balanced accuracy values and range of obtained $\alpha$ for different ways of estimating $\alpha$ (using EIOL-GFMM-v2 with $\theta = 0.1, \delta = 0.1$).}
	\label{Fig_S9}
\end{figure}

Figs. from \ref{Fig_S7} to \ref{Fig_S9} show the distribution of the obtained values of $\alpha$ using different methods to find the suitable values for $\alpha$. The best results in each row of the tables in this document are highlighted in bold. These figures support the explanation in subsection IV.C.1 from the main paper related to the effectiveness of each method of finding $\alpha$.

Table \ref{table_S2} presents the average class balanced accuracy for different learning algorithms with mixed-attribute handling ability over 40 iterations (10 times repeated stratified 4-fold cross-validation). The number of generated hyperboxes from these algorithms is shown in Table \ref{table_S3}. These are the results for the experiment mentioned in subsection IV.B.1 from the main paper.

Table \ref{table_S4} shows the average CBA for the proposed methods and the original improved online learning algorithm using different encoding techniques to transform the discrete features into numerical features. The ranks for these methods are presented in Table \ref{table_S5}. These results supplement to the claims in subsection IV.B.2 from the main paper.

Table \ref{table_S6} describes the average CBA values for the different methods to find the appropriate values for $\alpha$ using the proposed learning algorithms. Their ranks are shown in Table \ref{table_S7}. These are the detailed empirical results for the experiment presented in subsection IV.C.1 from the main paper.

Table \ref{table_S8} shows the average CBA values for four learning algorithms using the hyper-parameter tuning approach. The ranks for these results are presented in Table \ref{table_S9}. These results add more details to the content in subsection IV.C.2 from the main paper.

\begin{table}[!ht]
\centering
\caption{Average Class Balanced Accuracy for the Proposed EIOL-GFMM Algorithms and Two Existing Algorithms with the Mixed-Attribute Learning Ability} \label{table_S2}
\begin{scriptsize}
\begin{tabular}{lccccccccccccc}
\hline
\multirow{2}{*}{Dataset}          & \multirow{2}{*}{$\theta$} & \multicolumn{3}{c}{Onln-GFMM-M1}                & \multicolumn{3}{c}{Onln-GFMM-M2}                   & \multicolumn{3}{c}{EIOL-GFMM-v1} & \multicolumn{3}{c}{EIOL-GFMM-v2} \\ \cline{3-14} 
                                  &                           & $\eta = 0.1$     & $\eta   = 0.7$ & $\eta   = 1$ & $\beta = 0.25r$ & $\beta = 0.5r$ & $\beta = 0.75r$  & $\delta = 0.1$   & $\delta   = 0.7$ & $\delta   = 1$   & $\delta = 0.1$   & $\delta   = 0.7$ & $\delta   = 1$   \\ \hline
\multirow{3}{*}{abalone}          & 0.1                       & 0.10182          & 0.10156        & 0.1039       & 0.10195         & 0.10195        & 0.10195          & \textbf{0.10592} & 0.09893          & 0.0951           & \textbf{0.10592} & 0.09893          & 0.0951           \\ \cline{2-14} 
                                  & 0.7                       & 0.09182          & 0.08919        & 0.07923      & 0.08662         & 0.08662        & 0.08662          & \textbf{0.09384} & 0.07846          & 0.08748          & \textbf{0.09384} & 0.07846          & 0.08748          \\ \cline{2-14} 
                                  & 1                         & 0.08831          & 0.08357        & 0.0729       & 0.08455         & 0.08455        & 0.08455          & \textbf{0.094}   & 0.07766          & 0.08545          & \textbf{0.094}   & 0.07766          & 0.08545          \\ \hline
\multirow{3}{*}{australian}       & 0.1                       & 0.77345          & 0.76536        & 0.75532      & 0.62766         & 0.6276         & 0.6425           & \textbf{0.77871} & \textbf{0.77871} & 0.74137          & \textbf{0.77871} & 0.73931          & 0.74137          \\
\hline
\end{tabular}
\end{scriptsize}
\end{table}

\begin{table}[!ht]
\centering
\caption{The Average Number of Generated Hyperboxes from the Proposed Method and Other Existing Algorithms with Mixed-Attribute Learning Ability} \label{table_S3}
\begin{scriptsize}
\begin{tabular}{lccccccccccccc}
\hline
\multirow{2}{*}{Dataset}          & \multirow{2}{*}{$\theta$} & \multicolumn{3}{c}{Onln-GFMM-M1}                & \multicolumn{3}{c}{Onln-GFMM-M2}                   & \multicolumn{3}{c}{EIOL-GFMM-v1} & \multicolumn{3}{c}{EIOL-GFMM-v2} \\ \cline{3-14} 
                                  &                           & $\eta = 0.1$     & $\eta   = 0.7$ & $\eta   = 1$ & $\beta = 0.25r$ & $\beta = 0.5r$ & $\beta = 0.75r$  & $\delta = 0.1$   & $\delta   = 0.7$ & $\delta   = 1$   & $\delta = 0.1$   & $\delta   = 0.7$ & $\delta   = 1$   \\ \hline
\multirow{3}{*}{abalone}          & 0.1                       & 693.95       & 591.725        & 522.825      & 704.15          & 704.15         & 704.15          & 1675.2           & 1318.925          & 919.45          & 1675.2           & 1318.925          & 919.45          \\ \cline{2-14} 
                                  & 0.7                       & 77.225       & 56.875         & 46.75        & 79.375          & 79.375         & 79.375          & 1665.075         & 605.25            & 934.1           & 1665.075         & 605.25            & 934.1           \\ \cline{2-14} 
                                  & 1                         & 63.3         & 45.275         & 37.375       & 65.275          & 65.275         & 65.275          & 1662.125         & 599.225           & 932.375         & 1662.125         & 599.225           & 932.375         \\ \hline
\multirow{3}{*}{australian}       & 0.1                       & 472.15       & 348.225        & 222.6        & 198.775         & 206.05         & 279.025         & 493.225          & 493.225           & 203.2           & 493.225          & 204.7             & 203.2           \\
\hline
\end{tabular}
\end{scriptsize}
\end{table}

\begin{table}[!ht]
\centering
\caption{The Average Class Balanced Accuracy for the Proposed Method and the IOL-GFMM Algorithm using Encoding Techniques} \label{table_S4}
\begin{scriptsize}
\begin{tabular}{lccccccccccc}
\hline
Dataset                            & $\theta(= \delta)$ & CatBoost         & One-hot & LOO              & Label            & Target           & James-Stein      & Helmert          & Sum              & EIOL-GFMM-v1     & EIOL-GFMM-v2     \\ \hline
\multirow{3}{*}{abalone}             & 0.1                 & 0.1077           & 0.11216 & 0.08367          & \textbf{0.11217} & 0.11161          & 0.11161          & 0.11216          & 0.11215          & 0.10592          & 0.10592          \\ \cline{2-12} 
                                     & 0.7                 & \textbf{0.11117} & 0.04521 & 0.04599          & 0.07214          & 0.0544           & 0.0544           & 0.04521          & 0.03942          & 0.07846          & 0.07846          \\ \cline{2-12} 
                                     & 1                   & \textbf{0.11041} & 0.03651 & 0.05247          & 0.07843          & 0.07027          & 0.07027          & 0.02491          & 0.02491          & 0.08545          & 0.08545          \\ \hline
\multirow{3}{*}{australian}          & 0.1                 & 0.79148          & 0.6283  & 0.70876          & 0.77303          & 0.77283          & 0.77321          & 0.62855          & \textbf{0.79163} & 0.77871          & 0.77871          \\ \cline{2-12} 
                                     & 0.7                 & \textbf{0.81259} & 0.63919 & 0.60169          & 0.79844          & 0.80302          & 0.80712          & 0.63943          & 0.80088          & 0.78758          & 0.77846          \\ \cline{2-12} 
                                     & 1                   & 0.72686          & 0.36155 & 0.45375          & 0.38134          & 0.63626          & 0.64024          & 0.36384          & 0.36405          & \textbf{0.7771}  & \textbf{0.7771}  \\
\hline
\end{tabular}
\end{scriptsize}
\end{table}

\begin{table}[!ht]
\centering
\caption{The Ranks for the Proposed Method and the Original IOL-GFMM Algorithm using Encoding Techniques} \label{table_S5}
\begin{scriptsize}
\begin{tabular}{lccccccccccc}
\hline
Dataset                            & $\theta(= \delta)$ & CatBoost         & One-hot & LOO              & Label            & Target           & James-Stein      & Helmert          & Sum              & EIOL-GFMM-v1     & EIOL-GFMM-v2     \\ \hline
\multirow{3}{*}{abalone}             & 0.1                 & 7          & 2.5     & 10         & \textbf{1} & 5.5        & 5.5         & 2.5        & 4          & 8.5          & 8.5          \\ \cline{2-12} 
                                     & 0.7                 & \textbf{1} & 8.5     & 7          & 4          & 5.5        & 5.5         & 8.5        & 10         & 2.5          & 2.5          \\ \cline{2-12} 
                                     & 1                   & \textbf{1} & 8       & 7          & 4          & 5.5        & 5.5         & 9.5        & 9.5        & 2.5          & 2.5          \\ \hline
\multirow{3}{*}{australian}          & 0.1                 & 2          & 10      & 8          & 6          & 7          & 5           & 9          & \textbf{1} & 3.5          & 3.5          \\ \cline{2-12} 
                                     & 0.7                 & \textbf{1} & 9       & 10         & 5          & 3          & 2           & 8          & 4          & 6            & 7            \\ \cline{2-12} 
                                     & 1                   & 3          & 10      & 6          & 7          & 5          & 4           & 9          & 8          & \textbf{1.5} & \textbf{1.5} \\ \hline
\multirow{3}{*}{cmc}                 & 0.1                 & 10         & 7.5     & 9          & \textbf{1} & 6          & 5           & 7.5        & 2          & 3            & 4            \\
\hline
\end{tabular}
\end{scriptsize}
\end{table}

\begin{table}[!ht]
\centering
\caption{The Average Class Balanced Accuracy for Different Methods Used to Find the Values for Parameter $\alpha$} \label{table_S6}
\begin{scriptsize}
\begin{tabular}{llccccc}
\hline
Dataset                           & \multicolumn{1}{c}{Algorithm} & $\theta = \delta$ & Tuning $\alpha$ & Est-$\alpha$-v1 & Est-$\alpha$-v2 & $\alpha = n/(n+r)$ \\ \hline
\multirow{3}{*}{abalone}          & EIOL-GFMM-v1                   & 0.1               & 0.10452        & 0.10571         & 0.10589         & 0.10592           \\ \cline{2-7} 
                                  & EIOL-GFMM-v2                   & 0.1               & 0.10452        & 0.10571         & 0.10589         & 0.10592           \\ \cline{2-7} 
                                  & Both                           & 1                 & 0.08795        & 0.08498         & 0.08307         & 0.08545           \\ \hline
\multirow{3}{*}{australian}       & EIOL-GFMM-v1                   & 0.1               & 0.7806         & 0.77979         & 0.77416         & 0.77871           \\ \cline{2-7} 
                                  & EIOL-GFMM-v2                   & 0.1               & 0.78038        & 0.77991         & 0.77406         & 0.77871           \\ \cline{2-7} 
                                  & Both                           & 1                 & 0.77654        & 0.77734         & 0.77784         & 0.7771            \\ \hline
\multirow{3}{*}{cmc}              & EIOL-GFMM-v1                   & 0.1               & 0.43148        & 0.41052         & 0.41074         & 0.40938           \\
\hline
\end{tabular}
\end{scriptsize}
\end{table}

\begin{table}[!ht]
\centering
\caption{The Ranks for Different Methods Used to Find the Values for Parameter $\alpha$} \label{table_S7}
\begin{scriptsize}
\begin{tabular}{llccccc}
\hline
Dataset                           & \multicolumn{1}{c}{Algorithm} & $\theta = \delta$ & Tuning $\alpha$ & Est-$\alpha$-v1 & Est-$\alpha$-v2 & $\alpha = n/(n+r)$ \\ \hline
\multirow{3}{*}{abalone}          & EIOL-GFMM-v1                   & 0.1               & 4              & 3               & 2               & 1                  \\ \cline{2-7} 
                                  & EIOL-GFMM-v2                   & 0.1               & 4              & 3               & 2               & 1                  \\ \cline{2-7} 
                                  & Both                           & 1                 & 1              & 3               & 4               & 2                  \\ \hline
\multirow{3}{*}{australian}       & EIOL-GFMM-v1                   & 0.1               & 1              & 2               & 4               & 3                  \\ \cline{2-7} 
                                  & EIOL-GFMM-v2                   & 0.1               & 1              & 2               & 4               & 3                  \\ \cline{2-7} 
                                  & Both                           & 1                 & 4              & 2               & 1               & 3                  \\ \hline
\multirow{3}{*}{cmc}              & EIOL-GFMM-v1                   & 0.1               & 1              & 3               & 2               & 4                  \\
\hline
\end{tabular}
\end{scriptsize}
\end{table}

\begin{table}[!ht]
\centering
\caption{The Average Class Balanced Accuracy for the Learning Algorithms Using the Hyper-Parameter Tuning method} \label{table_S8}
\begin{tabular}{lcccc}
\hline
Dataset          & Onln-GFMM-M1 & Onln-GFMM-M2 & EIOL-GFMM-v1 & EIOL-GFMM-v2 \\ \hline
abalone          & 0.10072      & 0.09932      & \textbf{0.10431}      & \textbf{0.10431}      \\ 
australian       & 0.78501      & 0.79201      & \textbf{0.79484}      & 0.78588      \\ 
cmc              & 0.39265      & 0.40692      & \textbf{0.42634}      & 0.42522      \\ 
dermatology      & 0.87378      & 0.83522      & \textbf{0.91563}      & 0.91497      \\ 
flag             & 0.21735      & 0.27828      & \textbf{0.30107}      & 0.28806      \\
german           & 0.58233      & 0.55034      & \textbf{0.60345}      & 0.59929      \\ 
heart            & 0.72481      & \textbf{0.7772}       & 0.76922      & 0.75861      \\
japanese credit & 0.763        & 0.76685      & \textbf{0.79294}      & 0.79211      \\
post operative  & \textbf{0.32856}      & 0.29945      & 0.3582      & 0.31890      \\ 
tae              & \textbf{0.48682}     & 0.4853       & 0.47482      & 0.44618      \\
zoo              & 0.67941      & 0.8648       & \textbf{0.87179}      & 0.85685      \\ \hline
\end{tabular}
\end{table}

\begin{table}[!ht]
\centering
\caption{The Rank for the Learning Algorithms Using the Hyper-Parameter Tuning method} \label{table_S9}
\begin{tabular}{lcccc}
\hline
Dataset          & Onln-GFMM-M1 & Onln-GFMM-M2 & EIOL-GFMM-v1 & EIOL-GFMM-v2 \\ \hline
abalone          & 3            & 4            & \textbf{1.5}          & \textbf{1.5}          \\ 
australian       & 4            & 2            & \textbf{1}            & 3            \\ 
cmc              & 4            & 3            & \textbf{1}            & 2            \\ 
dermatology      & 3            & 4            & \textbf{1}            & 2            \\ 
flag             & 4            & 3            & \textbf{1}            & 2            \\ 
german           & 3            & 4            & \textbf{1}            & 2            \\ 
heart            & 4            & \textbf{1}            & 2            & 3            \\ 
japanese\_credit & 4            & 3            & \textbf{1}            & 2            \\ 
post\_operative  & \textbf{1}            & 4            & 3            & 2            \\ 
tae              & \textbf{1}            & 2            & 3            & 4            \\ 
zoo              & 4            & 2            & \textbf{1}            & 3            \\ \hline
\textbf{Average} & 3.182        & 2.909        & \textbf{1.5}        & 2.409        \\ \hline
\end{tabular}
\end{table}
\vfill

\ifCLASSOPTIONcaptionsoff
  \newpage
\fi

\bibliographystyle{IEEEtran}


%







